# Tensor Regression

Jiani Liu[1], Ce Zhu[1], Zhen Long[1] and Yipeng Liu[1]

[1] *School of Information and Communication Engineering, University of Electronic Science and Technology of China (UESTC), China,*
*jianiliu@std.uestc.edu.cn, eczhu@uestc.edu.cn,*
*zhenlong@std.uestc.edu.cn, yipengliu@uestc.edu.cn*

ABSTRACT

The presence of multidirectional correlations in emerging multidimensional data poses a challenge to traditional regression modeling methods. Traditional modeling methods based on matrix or vector, for example, not only overlook the data's multidimensional information and lower model performance, but also add additional computations and storage requirements. Driven by the recent advances in applied mathematics, tensor regression has been widely used and proven effective in many fields, such as sociology, climatology, geography, economics, computer vision, chemometrics, and neuroscience. Tensor regression can explore multidirectional relatedness, reduce the number of model parameters and improve model robustness and efficiency. It is timely and valuable to summarize the developments of tensor regression in recent years and discuss promising future directions, which will help accelerate the research process of tensor regression, broaden the research direction, and provide tutorials for researchers interested in high dimensional regression tasks.

The fundamentals, motivations, popular algorithms, related applications, available datasets, and software resources for tensor regression are all covered in this monograph. The first part focuses on the key concepts for tensor regression, mainly analyzing existing tensor regression algorithms from the perspective of regression families. Meanwhile, the adopted low rank tensor representations and optimization frameworks are also summarized. In addition, several extensions in online learning and sketching are described. The second part covers related applications, widely used public datasets and software resources, as well as some real-world examples, such as multitask learning, spatiotemporal learning, human motion analysis, facial image analysis, neuroimaging analysis (disease diagnosis, neuron decoding, brain activation, and connectivity analysis) and chemometrics. This survey can be used as a basic reference in tensor-regression-related fields and assist readers in efficiently dealing with high dimensional regression tasks.

# 1

# Introduction

Regression analysis is a key area of interest in the field of data analysis and machine learning. Regression analysis is devoted to exploring the dependencies between variables, thereby exploring the objective causal relationship of things and further predicting the possible state of things in the future.

Every day, a considerable amount of high dimensional data is collected and stored due to the rapid growth of digital imaging and sensing technologies. For example, numerous neuroimaging technologies, such as electroencephalography (EEG), electrocorticography (ECoG), and functional magnetic resonance imaging (fMRI), generate a lot of multidimensional data. Color images, hyperspectral images, or video sequences in computer vision can also be considered as three-dimensional or four-dimensional tensors. Data in climatology is usually organized in three modes: location, time, and variable type. Social network data also has many indexes, such as user, destination, and time. Fig. 1.1 provides some examples of high dimensional datasets, which are selected from some public datasets.

The emergence of high dimensional data has brought challenges to traditional data representation methods. Tensors, as high order ex-tensions of vectors, are considered as natural representations of high order data (Kolda and Bader, 2009; Cichocki *et al.*, 2015; Sidiropoulos *et al.*, 2017). Exploring the internal structure information of tensors can help people better represent high order data with fewer parameters, extract and interpret their attributes, and achieve data compression, storage, processing, and analysis. Tensor-based learning approaches have been proven to be more effective than classic vector-based methods, and have gotten a lot of attention in recent years, such as tensor completion (Long *et al.*, 2019), tensor principal component analysis (Lu *et al.*, 2016; Feng *et al.*, 2020), tensor sparse representation (Qi *et al.*, 2016; Xie *et al.*, 2017), tensor subspace learning (Lu *et al.*, 2011; Makantasis *et al.*, 2019), tensor clustering (Sun and Li, 2019; Poullis, 2019), tensor regression and classification (Tao *et al.*, 2005; Stoudenmire and Schwab, 2016; Wimalawarne *et al.*, 2016).

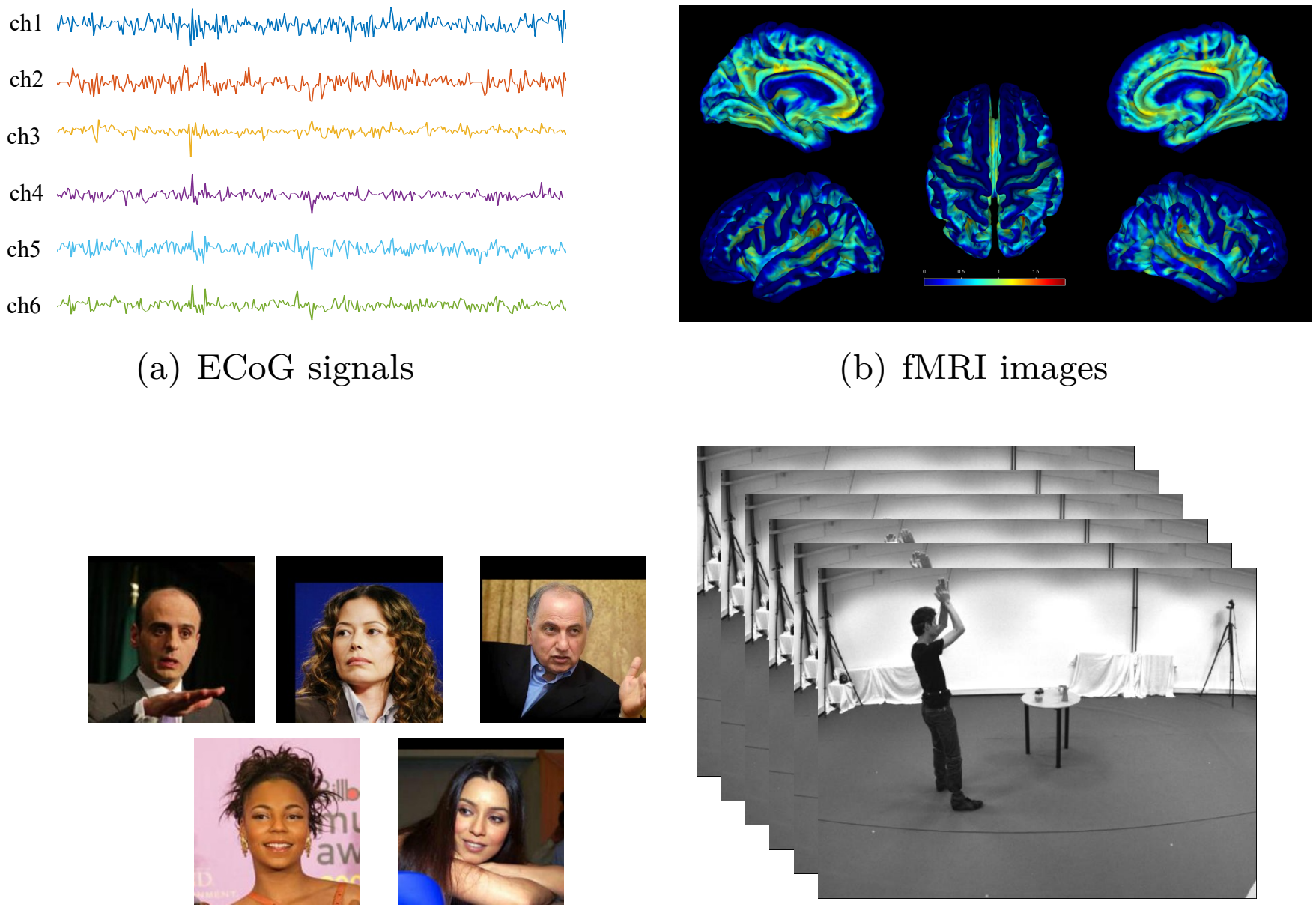

(a) ECoG signals

(b) fMRI images

(c) Facial images

(d) Video sequences

**Figure 1.1:** Some examples of high dimensional datasets.

For regression tasks with high dimensional predictors or responses, how to effectively explore the multidirectional correlation of these high dimensional data has become an important research topic in the field of image processing and machine learning (Zhou *et al.*, 2013). Most classical regression statistical models are modeled based on vectors as regression coefficients and are not suitable for high dimensional regression problems. There are two main disadvantages for simply employing the traditional regression methods. On the one hand, using traditional regression methods requires first performing vectorization operations on the multiway data, which will ignore the inherent multiway structural information contained in the high dimensional data, resulting in a degradation in terms of model performance. On the other hand, a huge vector-based model will require a large number of parameters, suffering from storage and computational burden and undesired numerical instability.

The objective of this review is to provide a systematic study and analysis of tensor-based regression models and their applications in recent years. Here we give some examples with multidimensional inputs or outputs that will be discussed in this review.

### Example 1: Weather forecasting

Weather forecasting is a common task to predict the state of weather at a given space based on historical recordings. The data for this example comes from the U.S. Historical Climatology Network (USHCN). It records the monthly measurements of four climate variables including maximum, minimum, and average temperatures over a month, and total monthly precipitations of approximately 1200 USHCN stations from 1915 to 2014. Table 1.1 lists the recordings for the four variables of some selected stations at a specific time. It records the temperature in degrees Fahrenheit and precipitation in hundreds of inches.

The objective is to design a regressor that could predict the possible value of these four variables at timepoint $t$ based on the previous recordings at $t-1, \cdots, t-K$. Let $\mathcal{X}(:, t, m) \in \mathbb{R}^P$ represent the recording of $m$-th variable at time $t$ of all the stations, then we aim to construct

**Table 1.1:** The monthly average of four climate variables for some selected stations in U.S. We have listed the recording for January 2014.

| Stations | Maximum temperature (F) | Minimum temperature (F) | Average temperature (F) | Total monthly precipitations |
|---|---|---|---|---|
| BREWTON 3 SSE | 56.3 | 32.0 | 44.2 | 3.15 |
| FAIRHOPE 2 NE | 55.5 | 30.0 | 42.7 | 2.34 |
| GAINESVILLE LOCK | 50.3 | 24.3 | 37.3 | 1.90 |
| GREENSBORO | 50.0 | 25.2 | 37.6 | 1.47 |
| HIGHLAND HOME | 50.9 | 23.0 | 36.9 | 3.35 |
| MUSCLE SHOALS AP | 47.8 | 22.9 | 35.3 | 1.71 |
| SAINT BERNARD | 44.3 | 20.1 | 32.2 | 2.99 |
| ··· | ··· | ··· | ··· | ··· |
| WORLAND | 31.1 | 5.1 | 18.1 | 0.38 |
| YELLOWSTONE PK MAMMOTH | 34.3 | 15.8 | 25.0 | 0.62 |

a data generation process as follows

$$\mathcal{X}(:,t,m) = \sum_{k=1}^{K} \mathcal{B}_k(:,:,m)\mathcal{X}(:,t-k,m) + \mathcal{E}(:,t,m) \tag{1.1}$$

where $\mathcal{B}_k \in \mathbb{R}^{P\times P\times M}$, $m = 1,\cdots,M$, $M = 4$ denotes the number of climate variables, $\mathcal{E}$ is the bias. In order to characterize both the spatial and temporal correlations of the climate data, the low rank constraint is commonly used over the coefficient tensor $\mathcal{B} \in \mathbb{R}^{P\times KP\times M}$ which is obtained by putting all the $\{\mathcal{B}_k\}_{k=1}^K$ together along the second mode. We have discussed a number of different methods for tackling this learning problem in this monograph.

### Example 2: Age estimation

Estimating age automatically from facial images or brain images has attracted a lot of attention in recent years. It is a challenging problem since the impact factors for the aging process are complex. Meanwhile, the facial images and brain images are indexed by multiple dimensions, commonly represented as a 3rd order tensor. Therefore, a lot of parameters are needed to predict the age as in the following model

$$y_n = \langle \mathcal{X}_n, \mathcal{B}\rangle + \epsilon_n, \tag{1.2}$$

where $\mathcal{X}$ is the employed brain images like MRI or fMRI images for $n$-th sample, $y_n$ is the corresponding sample age, the coefficient tensor $\mathcal{B}$ is in the same size as the predictor, $\epsilon_n$ is the bias for $n$-th sample.

Here we consider the UK biobank dataset for the age estimation task. The T1-weighted 3D structured MRI images and corresponding age labels are exploited to train the regression model. The MRI images are all sized at $182 \times 218 \times 182$. Thus, the parameters needed for model (1.2) will reach $7,000,000$, which makes the learning problem intractable.

In order to control model complexity and explore multidirectional relatedness, the low rank constraint is naturally considered, namely low rank tensor regression, which we will discuss later in this review.

### Example 3: A study for autism spectrum disorder

One major issue in neuroimaging analysis is to infer clinical assessments or human traits from neuroimages, such as the age estimation from the fMRI images described above. In addition to this, through modeling the data from a reverse perspective, we can obtain a tensor response regression task that aims to generate corresponding brain images in response to changes in disease state or age. In this way, it is easy to find out the activity pattern between different groups of subjects.

Specifically, here we consider the data from the Autism Brain Imaging Data Exchange (ABIDE) for autism spectrum disorder (ASD). ASD is a series of complex neurodevelopmental disorders that can affect children's social behavior and communication abilities. Studying the different patterns of information processing in the brains of people with ASD and normal children can better help the diagnosis and treatment of ASD. In other words, it is an important research topic to compare and analyze the neural images of normal children and children with ASD through existing data, and to find the brain regions or activity patterns that are most likely to distinguish the two groups. We could view this task as a tensor response regression problem, with the fMRI images being the response and three predictors including the ASD status, age, and sex.

Since there is a high dimensionality in the response, like the fractional amplitude of low-frequency fluctuations (fALFF) data sized at $91 \times 109 \times 91$ in ABIDE, the number of model parameters required will be very large, far exceeding the number of samples, which makes the problem difficult to solve. Therefore, how to explore the multidirectional correlation between data, and how to accurately reduce the dimensionality while ensuring that the model fitting error is small, become very important. This is also an issue that this review aims to deal with.

## Contributions

The main contributions of this review are summarized as follows:

- This is the first thorough overview of the fundamentals, motivations, popular algorithms, strategies for efficient implementation, related applications, available datasets, and software resources for tensor-based regression analysis. There is presently no comprehensive overview for tensor-based regression methods. Only some reviews for tensor-based analysis (Sun *et al.*, n.d.; Cichocki *et al.*, 2017) include a subsection for regression analysis, covering only a part of popular algorithms. And there is no classification and discussion of existing methods from multiple perspectives, nor a systematic summary of algorithms, applications, datasets, etc.

- This review groups and illustrates the existing tensor-based regression methods mainly from the perspective of different regression families, such as simple linear tensor regression, generalized tensor regression, penalized tensor regression, Bayesian tensor regression, quantile tensor regression, projection based tensor regression, kernelized tensor regression, tensor Gaussian process regression, tensor additive models, random forest based tensor regression and deep tensor regression. Meanwhile, a discussion of these methods is also given in terms of modeling (tensor-on-vector regression, vector-on-tensor regression, tensor-on-tensor regression), tensor representations (CP, Tucker, t-SVD, TT, TR), and optimization frameworks (ADMM, AM, GD).

- Another highlight of this overview is the use of some real cases in both Chapter 1 and 7 to help readers better understand the application background of tensor regression analysis, such as the forecasting task of spatio-temporal data, human motion trajectories reconstruction from corresponding image sequences, age estimation and disease diagnosis from brain images, a study of the difference between the neural images of children with ASD and normal children, and decoding electroencephalogram signals.

## Who should read this review

This review intends to give non-specialists and students interested in tensor regression analysis a decent starting point. The basics, core ideas, and theoretical characteristics of most tensor-based regression methods are covered, but it may be beneficial for readers to have some background in convex optimization, machine learning, and statistics.

We attempt to construct a more comprehensive overview covering most existing tensor regression methods. Moreover, we may pay more attention to the core ideas of the algorithms rather than the specific details and theoretical properties of the algorithms. This is our focus because we want to help readers understand the possible solutions to tensor regression tasks and the main ideas, advantages, and disadvantages of these solutions conceptually rather than mathematically.

In addition, those who are interested in starting related works can also read this review to learn how to use existing tensor-based regression methods to solve specific regression tasks with multiway data, what datasets can be selected, and what software packages are available to start related work as soon as possible.

## Organization of this review

The remaining part of the review proceeds as follows: Chapter 2 provides the notations, basic tensor-related operations, and some popular decomposition methods used in this review. The traditional regression methods are introduced in Chapter 3 and extended into the tensor field in Chapter 4 and Chapter 5. Meanwhile, Chapter 4 gives a comprehensive illustration, discussion, comparison, and summary of existing popular linear tensor regression methods. Chapter 5 illustrates the key concepts and algorithms of nonlinear tensor regression models. Then some strategies for efficient implementation, including sketching and online learning, are discussed in Chapter 6. Chapter 7 summarizes the common application fields and gives some practical examples in order to guide the research on tensor regression and validate the effectiveness of some tensor regression approaches. Finally, Chapter 8 concludes the open-source software, and Chapter 9 draws some conclusions and makes recommendations for possible future research works.

# 2

# Notations and preliminaries

In this section, we first give a brief overview of some notations, basic tensor operations and decomposition methods which are necessary for stating the tensor regression problems.

## 2.1 Notions

This survey expresses a scalar by a lowercase letter $a$, a vector by a lowercase boldface letter $\mathbf{a}$, a matrix by an uppercase boldface letter $\mathbf{A}$, and a tensor by a calligraphic letter $\mathcal{A}$. As shown in Fig. 2.1, a scalar is a 0th order tensor, a vector can be seen as a 1st order tensor, and a matrix can be seen as a 2nd order tensor. Some important notions are listed in Table 2.1.

## 2.2 Basic operations

In this section, some useful operations based on vectors or matrices are illustrated due to their contribution to tensor computations, as listed in Table 2.1.

**Definition 2.1 (Matrix trace).** The trace of matrix $\mathbf{A} \in \mathbb{R}^{I \times I}$ is calculated by $\mathrm{Tr}(\mathbf{A}) = \sum_{i=1}^{I}(\mathbf{A}(i,i))$.

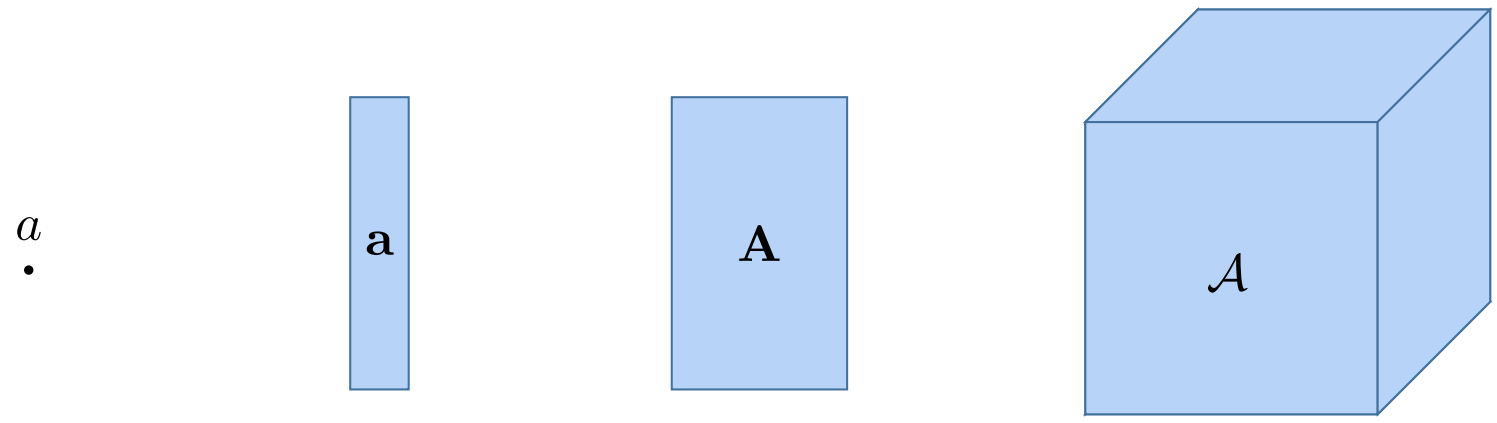

(a) scalar (b) vector (c) matrix (d) 3rd order tensor

**Figure 2.1:** The diagram representation of a scalar, vector, matrix and a 3rd order tensor.

**Table 2.1:** List of important notions

| | |
|---|---|
| $\mathbf{a}(i_1)$ | $i_1$-th element of vector $\mathbf{a} \in \mathbb{R}^{I_1}$ |
| $\mathbf{A}(i_1, i_2)$ | $i_1, i_2$-th element of matrix $\mathbf{A} \in \mathbb{R}^{I_1 \times I_2}$ |
| $\mathcal{A}(i_1, \cdots, i_N)$ | $i_1, \cdots, i_N$-th element of tensor $\mathcal{A} \in \mathbb{R}^{I_1 \times \cdots \times I_N}$ |
| $\mathrm{vec}(\cdot)$ | vectorization |
| $(\cdot)^{\mathrm{T}}$ | matrix transpose |
| $(\cdot)^{-1}$ | matrix inverse |
| $(\cdot)^{\dagger}$ | Moore-Penrose inverse |
| $\mathrm{Tr}(\cdot)$ | trace operation |
| $\mathbf{I}_N$ | identity matrix in size of $N \times N$ |
| $\mathbf{0}_N$ | zero matrix in size of $N \times N$ |
| $\mathrm{E}(\cdot)$ | expected value |
| $\mathrm{Corr}(\mathbf{a}, \mathbf{b})$ | correlation coefficient of variable $\mathbf{a}$ and $\mathbf{b}$ |
| $\mathrm{Var}(\mathbf{a})$ | variance of variable $\mathbf{a}$ |
| $\mathrm{Cov}(\mathbf{a}, \mathbf{b})$ | covariance of variable $\mathbf{a}$ and $\mathbf{b}$ |
| $\mathbf{A} \cdot \mathbf{B} \in \mathbb{R}$ | dot product of $\mathbf{A} \in \mathbb{R}^{I \times J}$ and $\mathbf{B} \in \mathbb{R}^{I \times J}$ |
| $\mathbf{a} \circ \mathbf{b} \in \mathbb{R}^{I \times J}$ | out product of $\mathbf{a} \in \mathbb{R}^{I}$ and $\mathbf{b} \in \mathbb{R}^{J}$ |
| $\mathbf{A} \otimes \mathbf{B} \in \mathbb{R}^{IK \times JL}$ | Kronecker product of $\mathbf{A} \in \mathbb{R}^{I \times J}$ and $\mathbf{B} \in \mathbb{R}^{K \times L}$ |
| $\mathbf{A} \odot \mathbf{B} \in \mathbb{R}^{IJ \times K}$ | Khatri-Rao product of $\mathbf{A} \in \mathbb{R}^{I \times K}$ and $\mathbf{B} \in \mathbb{R}^{J \times K}$ |
| $\Vert\mathbf{A}\Vert_{\mathrm{F}}$ | Frobenius norm of $\mathbf{A} \in \mathbb{R}^{I \times J}$ |
| $\Vert\mathbf{A}\Vert_{*}$ | nuclear norm of $\mathbf{A} \in \mathbb{R}^{I \times J}$ |
| $\Vert\mathbf{A}\Vert_{1}$ | $\ell_1$ norm of $\mathbf{A} \in \mathbb{R}^{I \times J}$ |
| $\Vert\mathbf{A}\Vert_{0}$ | $\ell_0$ norm of $\mathbf{A} \in \mathbb{R}^{I \times J}$ |
| $\overleftarrow{\mathcal{A}}^{n}$ | $n$-shifting of $\mathcal{A}$ |
| $\mathcal{A}^{\mathrm{T}} \in \mathbb{R}^{I_2 \times I_1 \times I_3}$ | the tensor transpose of $\mathcal{A} \in \mathbb{R}^{I_1 \times I_2 \times I_3}$ (Kilmer *et al.*, 2013) |
| $\overline{\mathcal{A}}$ | the discrete Fourier transform of $\mathcal{A}$ |

**Definition 2.2 (Dot product).** The dot product of matrices $\mathbf{A}$ and $\mathbf{B}$ in the same size yields

$$c = \mathbf{A} \cdot \mathbf{B} = \mathrm{vec}(\mathbf{A})^{\mathrm{T}} \mathrm{vec}(\mathbf{B}) \in \mathbb{R}. \tag{2.1}$$

Then the Frobenius norm of matrix $\mathbf{A}$ is defined as $\Vert\mathbf{A}\Vert_{\mathrm{F}} = \sqrt{\mathbf{A} \cdot \mathbf{A}} \in \mathbb{R}$.

**Definition 2.3 (Vector outer product).** The outer product of vectors $\mathbf{a} \in \mathbb{R}^I$ and $\mathbf{b} \in \mathbb{R}^J$ is defined as

$$\mathbf{C} = \mathbf{a} \circ \mathbf{b} \in \mathbb{R}^{I \times J} \tag{2.2}$$

with entries

$$\mathbf{C}(i, j) = \mathbf{a}(i)\mathbf{b}(j). \tag{2.3}$$

**Definition 2.4 (Kronecker product).** (Sidiropoulos *et al.*, 2017) The Kronecker product of matrices $\mathbf{A} \in \mathbb{R}^{I \times J}$ and $\mathbf{B} \in \mathbb{R}^{K \times L}$ can be defined by

$$\begin{aligned} \mathbf{C} &= \mathbf{A} \otimes \mathbf{B} \in \mathbb{R}^{IK \times JL} \qquad (2.4) \\ &= \begin{bmatrix} a_{1,1}\mathbf{B} & a_{1,2}\mathbf{B} & \dots & a_{1,J}\mathbf{B} \\ a_{2,1}\mathbf{B} & a_{2,2}\mathbf{B} & \dots & a_{2,J}\mathbf{B} \\ \vdots & \vdots & \ddots & \vdots \\ a_{I,1}\mathbf{B} & a_{I,2}\mathbf{B} & \dots & a_{I,J}\mathbf{B} \end{bmatrix}. \end{aligned}$$

**Definition 2.5 (Khatri-Rao product).** The Khatri-Rao product of matrices $\mathbf{A} \in \mathbb{R}^{I \times K}$ and $\mathbf{B} \in \mathbb{R}^{J \times K}$ is denoted as

$$\mathbf{A} \odot \mathbf{B} = [\mathbf{a}_1 \otimes \mathbf{b}_1, \mathbf{a}_2 \otimes \mathbf{b}_2, \cdots, \mathbf{a}_K \otimes \mathbf{b}_K] \in \mathbb{R}^{IJ \times K}, \tag{2.5}$$

where $\mathbf{a}_k$ and $\mathbf{b}_k$ are the $k$-th column of matrix $\mathbf{A}$ and $\mathbf{B}$.

**Definition 2.6 (Nuclear norm).** The nuclear norm (also known as the trace norm) of matrix $\mathbf{A} \in \mathbb{R}^{I \times J}$ is defined by

$$\|\mathbf{A}\|_* = \mathrm{Tr}(\sqrt{\mathbf{A}^{\mathrm{T}}\mathbf{A}}) = \sum_{k=1}^{\min(I,J)} \sigma_k, \tag{2.6}$$

where $\sigma_k$ is the $k$-th singular value of matrix $\mathbf{A}$.

**Definition 2.7 ($\ell_1$ norm).** The $\ell_1$ norm of matrix $\mathbf{A} \in \mathbb{R}^{I \times J}$ is defined by

$$\|\mathbf{A}\|_1 = \sum_{i=1}^{I} \sum_{j=1}^{J} |\mathbf{A}(i, j)|, \tag{2.7}$$

where $|\mathbf{A}(i, j)|$ is the absolute value of $\mathbf{A}(i, j)$.

**Definition 2.8 ($\ell_0$ norm).** The $\ell_0$ norm of matrix $\mathbf{A} \in \mathbb{R}^{I \times J}$ is defined by the number of non-zero elements in $\mathbf{A}$.

## 2.3 Graph networks

Before introducing tensor unfolding methods, we first give a definition for the graph networks. Graph networks can naturally represent the multiway tensor and its operations. It can help readers better understand the tensor operations and decompositions.

**Definition 2.9 (Graph networks).** Graph networks use the nodes with different edges to represent the multidimensional tensor (see examples in Fig. 2.2). For each node, the number of outgoing edges denotes the order of the tensor and the value of each outgoing line denotes the size of that specific mode. As in Fig. 2.2(a), a scalar is represented by a node with no edge, while a third order tensor $\mathcal{A}$ is represented by a node with three edges in Fig. 2.2(d).

In addition, we give an illustration of the matrix product in Fig. 2.2(e), where the edge connecting two nodes implies the contraction operation along that mode.

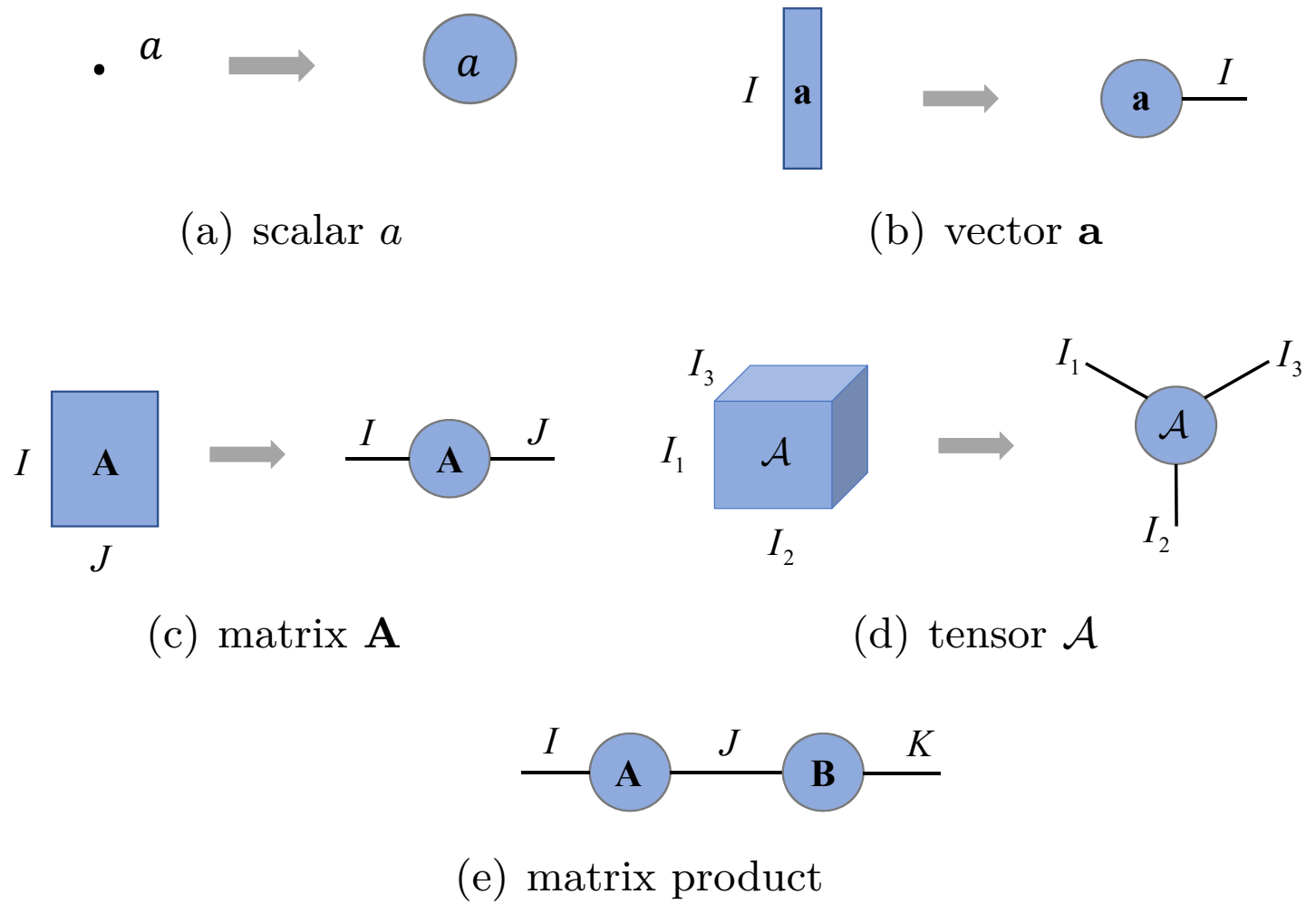

**Figure 2.2:** Graphical illustrations of (a) scalar $a$; (b) vector $\mathbf{a}$; (c) matrix $\mathbf{A}$; (d) third order tensor $\mathcal{A}$; (e) matrix product of $\mathbf{A}$ and $\mathbf{B}$.

## 2.4 Tensor unfolding

For tensor analysis, tensor unfolding methods are critical. Here we review three unfolding forms which are commonly used in classical tensor decomposition methods.

**Definition 2.10 (Mode-$n$ unfolding).** The mode-$n$ unfolding matrix of $\mathcal{A}$ is defined as $\mathbf{A}_{[n]} \in \mathbb{R}^{I_n \times I_{n+1} \cdots I_N I_1 \cdots I_{n-1}}$ or $\mathbf{A}_{(n)} \in \mathbb{R}^{I_n \times I_1 \cdots I_{n-1} I_{n+1} \cdots I_N}$, as shown in Fig. 2.3(b) and Fig. 2.3(c). The difference between these two kinds of mode-$n$ unfolding is the order of the remaining dimensions in the column of $\mathbf{A}_{[n]}$ and $\mathbf{A}_{(n)}$.

**Definition 2.11 ($n$-unfolding).** The $n$-unfolding matrix $\mathbf{A}_{\langle n \rangle} \in \mathbb{R}^{I_1 \cdots I_n \times I_{n+1} \cdots I_N}$ is obtained by simply reshaping $\mathcal{A}$ along its $n$-th dimension, as illustrated in Fig. 2.3(d). The first $n$ dimensions are rearranged as the row of $\mathbf{A}_{<n>}$ while the remaining dimensions are mapped as the column of $\mathbf{A}_{<n>}$.

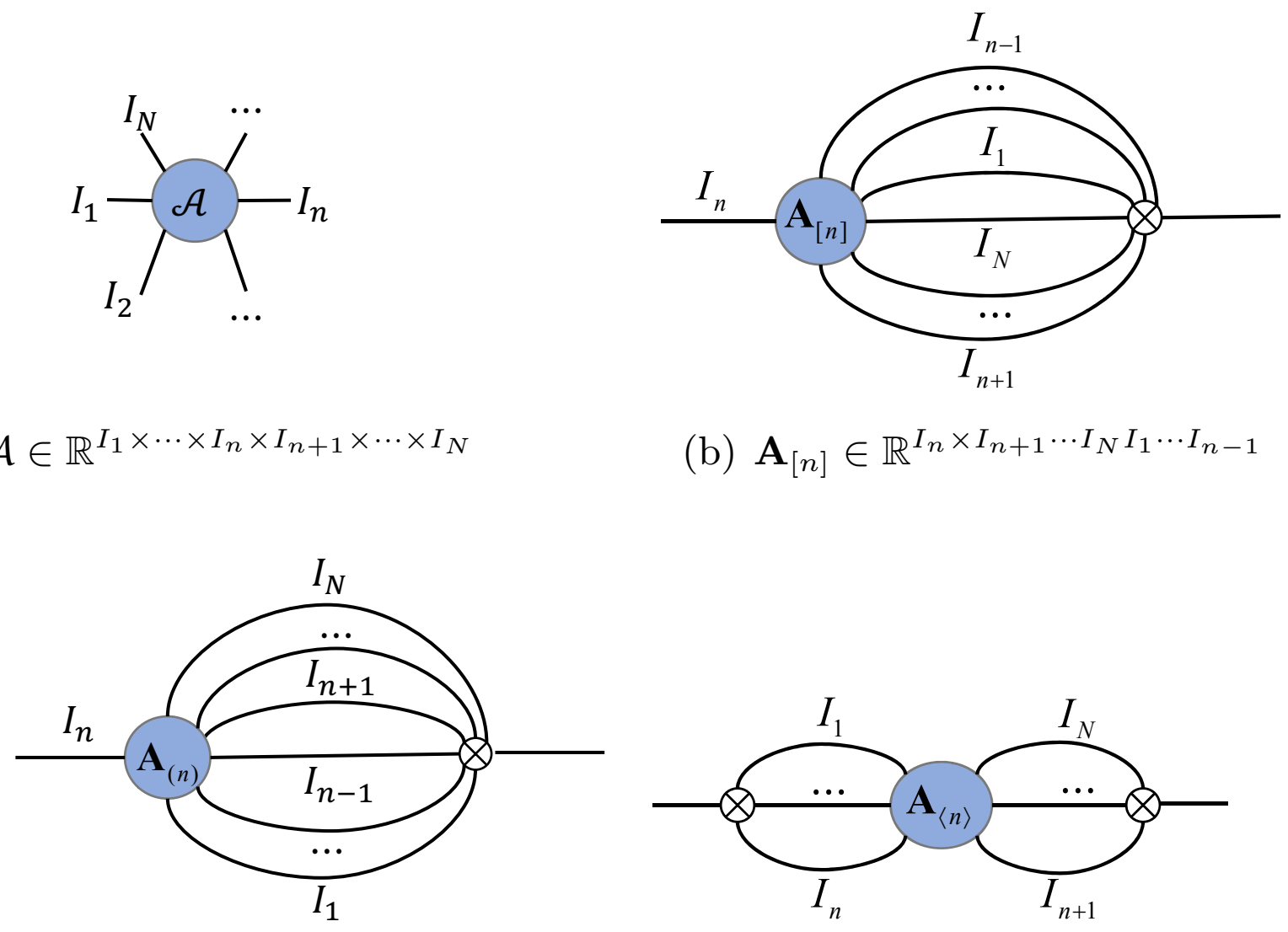

(a) $\mathcal{A} \in \mathbb{R}^{I_1 \times \cdots \times I_n \times I_{n+1} \times \cdots \times I_N}$ (b) $\mathbf{A}_{[n]} \in \mathbb{R}^{I_n \times I_{n+1} \cdots I_N I_1 \cdots I_{n-1}}$

(c) $\mathbf{A}_{(n)} \in \mathbb{R}^{I_n \times I_1 \cdots I_{n-1} I_{n+1} \cdots I_N}$ (d) $\mathbf{A}_{\langle n \rangle} \in \mathbb{R}^{I_1 \cdots I_n \times I_{n+1} \cdots I_N}$

**Figure 2.3:** Tensor diagram for different unfolding methods.

**Definition 2.12** (**$n$-shifting $k$-unfolding**). The $n$-shifting $k$-unfolding matrix of $\mathcal{A}$ is denoted as $\mathbf{A}_{\{n,k\}} \in \mathbb{R}^{I_{n+1}\cdots I_{n+k} \times I_{n+k+1}\cdots I_n}$, which first rotates $\mathcal{A}$ by $n$ and then rearranges the first $k$ dimensions of $\overleftarrow{\mathcal{A}}^n$ as the row of $\mathbf{A}_{\{n,k\}}$ and the remaining dimensions as the column.

For easy understanding, we provide a diagram to virtually illustrate the $n$-shifting $k$-unfolding method in Fig. 2.4.

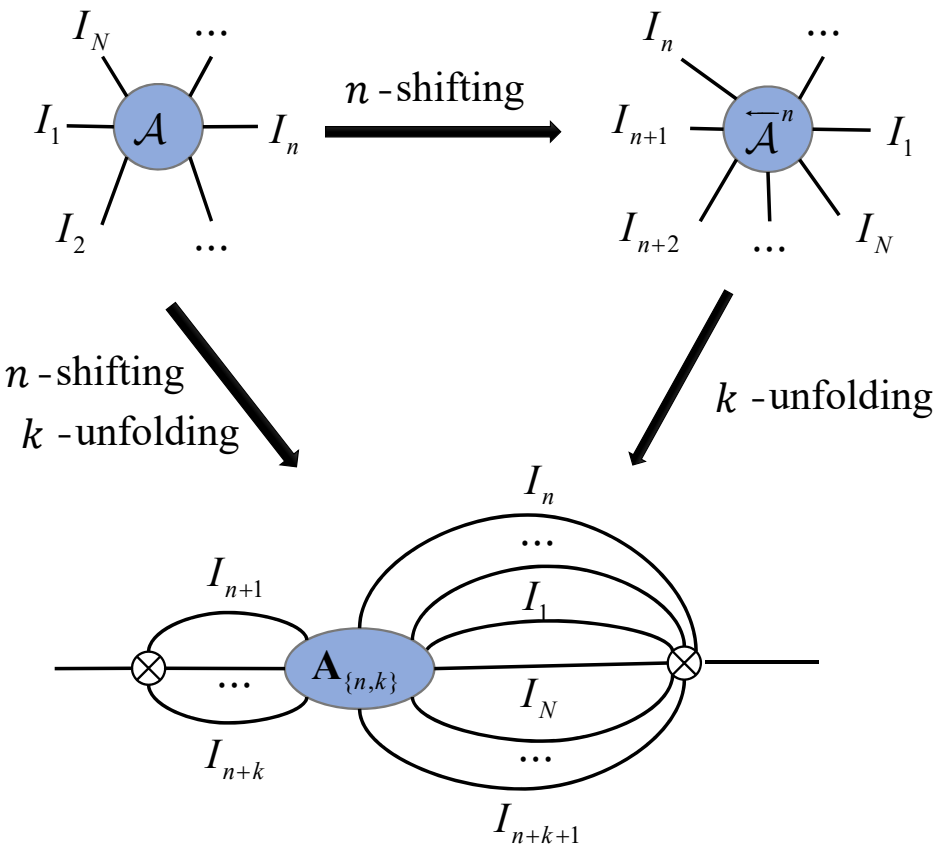

**Figure 2.4:** The diagram of $n$-shifting $k$-unfolding.

An example of $n$-shifting $k$-unfolding for a third order tensor is also given in Fig. 2.5.

It should be noted that in addition to the arrangement of dimensions, the ordering of elements also affects the specific form of the tensor unfolding matrices. In fact, there are two ordering methods for elements, namely column-major ordering (Fortron style) or row-major ordering (C style).

For example, as defined above, the mode-$n$ unfolding matrix $\mathbf{A}_{(n)}$ arranges the $n$-th mode of $\mathcal{A}$ as the row while the rest modes as the column of the mode-$n$ unfolding matrix in sequential order. The mathematical expression is as follows

$$\mathbf{A}_{(n)}(i_n, j) = \mathcal{A}(i_1, \cdots, i_n, \cdots, i_N), \tag{2.8}$$

where $j = \overline{i_1, \cdots, i_{n-1}, i_{n+1}, \cdots, i_N}$ is multi-index. The mathematical expression of multi-index $i = \overline{i_1, \cdots, i_N}$ can be either little-endian

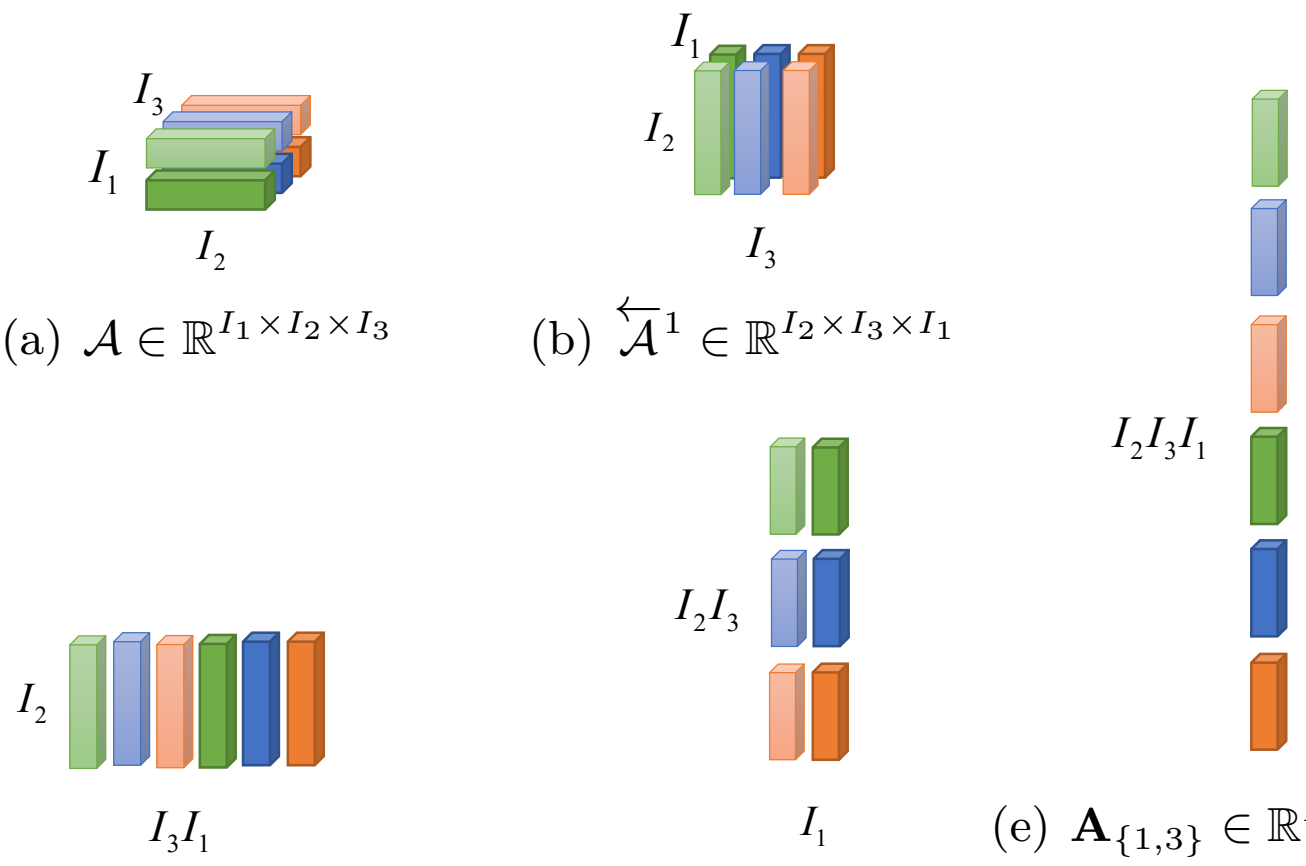

(a) $\mathcal{A} \in \mathbb{R}^{I_1 \times I_2 \times I_3}$ (b) $\overleftarrow{\mathcal{A}}^1 \in \mathbb{R}^{I_2 \times I_3 \times I_1}$ (c) $\mathbf{A}_{\{1,1\}} \in \mathbb{R}^{I_2 \times I_3 I_1}$ (d) $\mathbf{A}_{\{1,2\}} \in \mathbb{R}^{I_2 I_3 \times I_1}$ (e) $\mathbf{A}_{\{1,3\}} \in \mathbb{R}^{I_2 I_3 I_1 \times 1}$

**Figure 2.5:** An example of $n$-shifting $k$-unfolding matrices for a third order tensor.

convention

$$\overline{i_1, \cdots, i_N} = i_1 + (i_2 - 1)I_1 + (i_3 - 1)I_1 I_2 + \cdots + (i_N - 1)I_1 \cdots I_{N-1},$$

or big-endian

$$\overline{i_1, \cdots, i_N} = i_N + (i_{N-1} - 1)I_N + \cdots + (i_2 - 1)I_3 \cdots I_N + (i_1 - 1)I_2 \cdots I_N.$$

The little-endian notation is consistent with the Fortran style of indexing, while the big-endian notation corresponds to the C style. However, the employed order must be consistent during the whole algorithm. Typically, Matlab code uses the former Fortran style order, while Python code uses C style order.

## 2.5 Tensor product

**Definition 2.13 (Tensor inner product).** The inner product of given tensors $\mathcal{A} \in \mathbb{R}^{I_1 \times \cdots \times I_n \times \cdots \times I_N}$ and $\mathcal{B} \in \mathbb{R}^{I_1 \times \cdots \times I_n \times \cdots \times I_N}$ can be denoted as

$$c = \langle \mathcal{A}, \mathcal{B} \rangle = \operatorname{vec}(\mathcal{A})^{\mathrm{T}} \operatorname{vec}(\mathcal{B}) \in \mathbb{R}.$$

Based on the definition of tensor inner product, the Frobenius norm of tensor $\mathcal{A}$ can be expressed as $\|\mathcal{A}\|_{\mathrm{F}} = \sqrt{\langle \mathcal{A}, \mathcal{A} \rangle}$.

**Definition 2.14 (Hadamard product).** The Hadamard product of given tensors $\mathcal{A}$ and $\mathcal{B}$ in same size $I_1 \times \cdots \times I_n \times \cdots \times I_N$ can be denoted as

$$\mathcal{C} = \mathcal{A} \circledast \mathcal{B}$$

with entries

$$\mathcal{C}(i_1, \cdots, i_N) = \mathcal{A}(i_1, \cdots, i_N)\mathcal{B}(i_1, \cdots, i_N),$$

which is the element-wise product.

**Definition 2.15 (Mode-$n$ product).** The model-$n$ product of a given tensor $\mathcal{A} \in \mathbb{R}^{I_1 \times \cdots \times I_n \times \cdots \times I_N}$ and a matrix $\mathbf{B} \in \mathbb{R}^{J \times I_n}$ can be denoted as

$$\mathcal{C} = \mathcal{A} \times_n \mathbf{B} \in \mathbb{R}^{I_1 \times \cdots \times J \times \cdots \times I_N}$$

with entries

$$\mathcal{C}(i_1, \cdots, j, \cdots, i_N) = \sum_{i_n=1}^{I_n} \mathcal{A}(i_1, \cdots, i_n, \cdots, i_N)\mathbf{B}(i_n, j).$$

**Definition 2.16 (Tensor contracted product).** Tensor contracted product is a natural extension of matrix product. For two tensors $\mathcal{A} \in \mathbb{R}^{I_1 \times \cdots \times I_M \times J_1 \times \cdots \times J_N}$ and $\mathcal{B} \in \mathbb{R}^{J_1 \times \cdots \times J_N \times K_1 \times \cdots \times K_L}$, their contracted product along the common indices $\{J_1, \cdots, J_N\}$ can be expressed by

$$\mathcal{C} = \langle \mathcal{A}, \mathcal{B} \rangle_N \in \mathbb{R}^{I_1 \times \cdots \times I_M \times K_1 \times \cdots \times K_L}$$

with entries

$$\begin{aligned} \mathcal{C}(i_1, \cdots, i_M, k_1, \cdots, k_L) = \textstyle\sum_{j_1, \cdots, j_N} & \mathcal{A}(i_1, \cdots, i_M, j_1, \cdots, j_N) \\ & \mathcal{B}(j_1, \cdots, j_N, k_1, \cdots, k_L), \end{aligned}$$

where $N$ is the number of common modes $\{J_1, \cdots, J_N\}$.

Fig. 2.6 gives a graphical illustration of the tensor contracted product.

**Definition 2.17 (t-product).** For tensors $\mathcal{A} \in \mathbb{R}^{I_1 \times I_2 \times J}$ and $\mathcal{B} \in \mathbb{R}^{I_2 \times I_3 \times J}$, their t-product is formulated as

$$\mathcal{C} = \mathcal{A} * \mathcal{B} \in \mathbb{R}^{I_1 \times I_3 \times J}$$

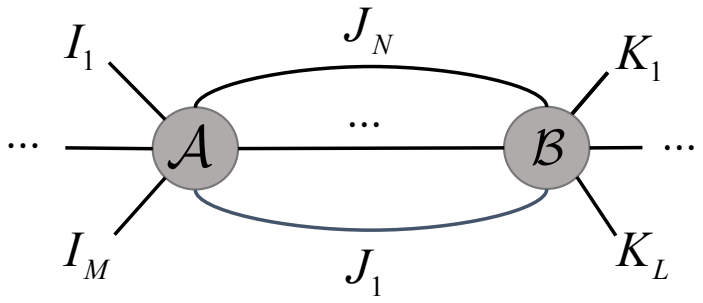

**Figure 2.6:** A graphical illustration of the tensor contracted product.

with entries

$$\mathcal{C}(i_1, i_3, :) = \sum_{i_2=1}^{I_2} \mathcal{A}(i_1, i_2, :) \bullet \mathcal{B}(i_2, i_3, :),$$

where $\bullet$ denotes the circular convolution of two tubes in the same size, it can be computed by the matrix multiplication in Fourier domain. Mathematically,

$$\mathcal{C} = \mathcal{A} * \mathcal{B} \Longleftrightarrow \overline{\mathcal{C}}(:, :, j) = \overline{\mathcal{A}}(:, :, j)\overline{\mathcal{B}}(:, :, j), j = 1, \cdots, J,$$

where $\overline{\mathcal{A}}$, $\overline{\mathcal{B}}$ and $\overline{\mathcal{C}}$ are the discrete Fourier transform of $\mathcal{A}$, $\mathcal{B}$ and $\mathcal{C}$ along the third mode, respectively.

The t-product can be implemented efficiently by FFT. However, the third dimension of $\mathcal{A}$ and $\mathcal{B}$ should be shared and the first mode of $\mathcal{B}$ has to match the second mode of $\mathcal{A}$.

## 2.6 Tensor decomposition

**Definition 2.18 (CP decomposition).** (Carroll and Chang, 1970; Harshman *et al.*, 1970) As a natural extension form of the matrix singular value decomposition (SVD), canonical polyadic decomposition (CPD) decomposes $\mathcal{A} \in \mathbb{R}^{I_1 \times \cdots \times I_N}$ into a sum of several rank-1 tensors, formulated as

$$\mathcal{A} = \sum_{r=1}^{R} \mathbf{b}_r^{(1)} \circ \cdots \circ \mathbf{b}_r^{(N)},$$

where $\mathbf{b}_r^{(n)} \in \mathbb{R}^{I_n}$ for $n = 1, \cdots, N$, $R$ is the CP rank.

**Lemma 2.1.** (Kolda, 2006) If a tensor $\mathcal{A}$ admits the CP decomposition $\mathcal{A} = \sum_{r=1}^{R} \mathbf{b}_r^{(1)} \circ \cdots \circ \mathbf{b}_r^{(N)}$, then

$$\begin{aligned} \text{vec}(\mathcal{A}) &= (\mathbf{B}^{(N)} \odot \cdots \odot \mathbf{B}^{(1)})\mathbf{1}_R, \\ \mathbf{A}_{(n)} &= \mathbf{B}^{(n)}(\mathbf{B}^{(N)} \odot \cdots \odot \mathbf{B}^{(n+1)} \odot \mathbf{B}^{(n-1)} \odot \cdots \odot \mathbf{B}^{(1)})^{\mathrm{T}}, \end{aligned}$$

where $\mathbf{B}^{(n)} = [\mathbf{b}_1^{(n)}, \cdots, \mathbf{b}_R^{(n)}] \in \mathbb{R}^{I_n \times R}$ for $n = 1, \cdots, N$, $R$ is the CP rank, $\mathbf{1}_R \in \mathbb{R}^{R \times 1}$ is a vector with all elements set to 1.

For simplicity, $\mathcal{A} = [\![\mathbf{B}^{(1)}, \cdots, \mathbf{B}^{(N)}]\!]$ is used to represent the CP decomposition. Fig. 2.7 provides an example for the CPD of a third order tensor.

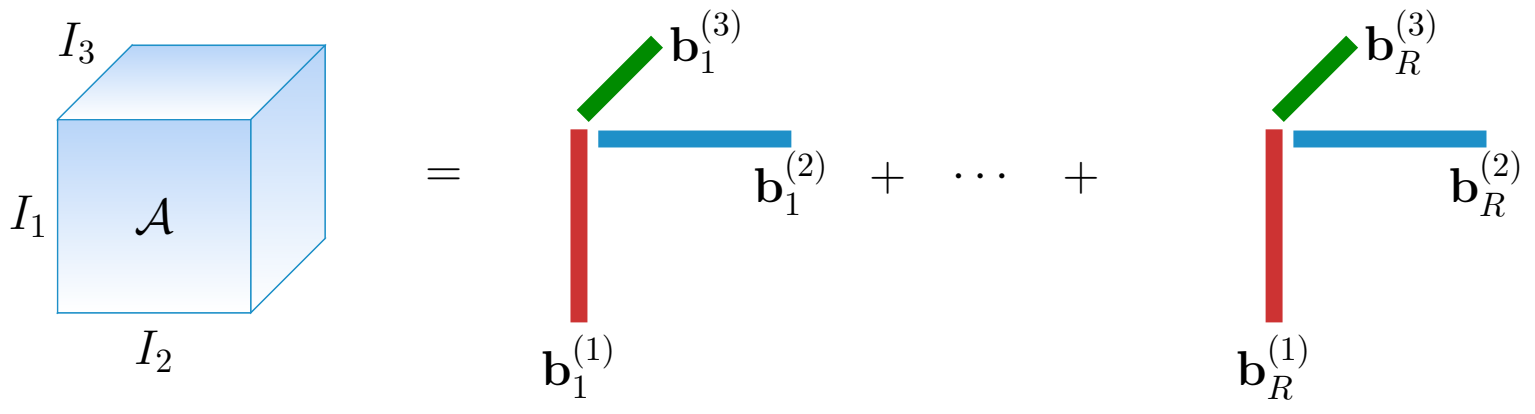

**Figure 2.7:** An example for CP decomposition.

**Definition 2.19 (Tucker decomposition).** (Tucker, 1963) The Tucker decomposition represents $\mathcal{A} \in \mathbb{R}^{I_1 \times \cdots \times I_N}$ as the mode-$n$ product of a core tensor $\mathcal{G}$ and several matrices. The mathematical expression is as follows

$$\mathcal{A} = \mathcal{G} \times_1 \mathbf{B}^{(1)} \times_2 \mathbf{B}^{(2)} \cdots \times_N \mathbf{B}^{(N)},$$

where $\mathcal{G} \in \mathbb{R}^{R_1 \times \cdots \times R_N}$ is the core tensor in the same order as $\mathcal{A}$, but the size of each dimension is reduced. $\mathbf{B}^{(n)} \in \mathbb{R}^{I_n \times R_n}$ for $n = 1, \cdots, N$, $\{R_1, \cdots, R_N\}$ is the Tucker rank. A simple representation of the Tucker decomposition is denoted as $\mathcal{A} = [\![\mathcal{G}; \mathbf{B}^{(1)}, \mathbf{B}^{(2)}, \cdots, \mathbf{B}^{(N)}]\!]$.

An example of the Tucker decomposition for a third order tensor is given in Fig. 2.8.

**Lemma 2.2.** (Kolda, 2006) If a tensor $\mathcal{A}$ admits the Tucker decomposition $\mathcal{A} = \mathcal{G} \times_1 \mathbf{B}^{(1)} \times_2 \mathbf{B}^{(2)} \cdots \times_N \mathbf{B}^{(N)}$, then

$$\begin{aligned} &\text{vec}(\mathcal{A}) = (\mathbf{B}^{(N)} \otimes \cdots \otimes \mathbf{B}^{(1)}) \, \text{vec}(\mathcal{G}), \\ &\mathbf{A}_{(n)} = \mathbf{B}^{(n)}\mathbf{G}_{(n)}(\mathbf{B}^{(N)} \otimes \cdots \otimes \mathbf{B}^{(n+1)} \otimes \mathbf{B}^{(n-1)} \otimes \cdots \otimes \mathbf{B}^{(1)})^{\mathrm{T}}, \end{aligned}$$

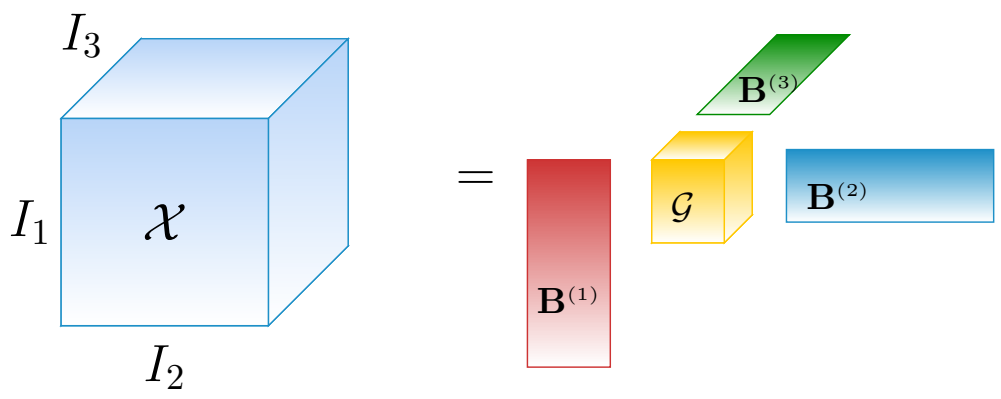

**Figure 2.8:** An example for a Tucker decomposition of a third order tensor.

where $\mathbf{A}_{(n)}$, $\mathbf{G}_{(n)}$ are the corresponding mode-$n$ unfolding matrices of $\mathcal{A}$ and $\mathcal{G}$, respectively.

**Definition 2.20 (Block term decomposition).** (De Lathauwer, 2008) For a tensor $\mathcal{A} \in \mathbb{R}^{I_1 \times \cdots \times I_N}$, block term decomposition decomposes it as a sum of $R$ low multilinear rank components. The mathematic expression is as follows

$$\mathcal{A} = \sum_{r=1}^{R} \mathcal{A}_r, \quad \mathcal{A}_r = \mathcal{G}_r \times \mathbf{B}_r^{(1)} \times \cdots \times \mathbf{B}_r^{(N)}, \tag{2.9}$$

where $\mathcal{A}_r$ is $r$-th low multilinear rank component of $\mathcal{A}$. Each $\mathcal{A}_r$ admits a Tucker decomposition with core tensor $\mathcal{G}_r \in \mathbb{R}^{J_1 \times \cdots \times J_N}$ and latent factor matrices $\{\mathbf{B}_r^{(n)} \in \mathbb{R}^{I_n \times J_n}\}_{n=1}^{N}$. $\{J_1, \cdots, J_N\}$ is the Tucker rank of $\mathcal{A}_r$. A graphical representation for block term decomposition of a third order tensor is given in Fig. 2.9.

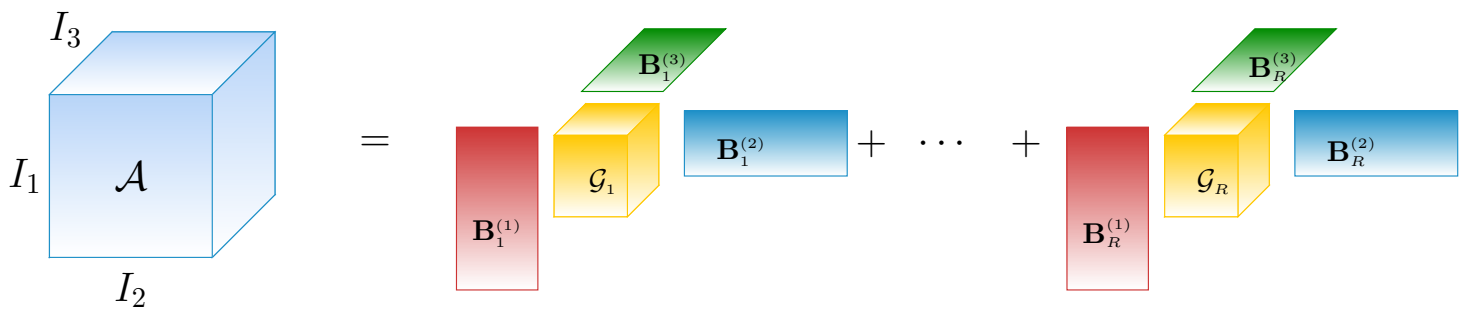

**Figure 2.9:** An example for block term decomposition of a third order tensor.

**Definition 2.21 (Hierarchical Tucker decomposition).** (Hackbusch and Kühn, 2009; Grasedyck, 2010) The hierarchical Tucker decomposition expresses tensor $\mathcal{A} \in \mathbb{R}^{I_1 \times \cdots \times I_N}$ using a binary tree. As shown in Fig.

2.10, for a parent node $q$, let $q_l$ and $q_r$ denote its left and right child nodes, then

$$\mathcal{G}_{\{\mathbb{C}_q\}} = \text{Merge}(\mathbf{G}_{\{\mathbb{C}_{q_l}\}}, \mathcal{B}_{\{\mathbb{C}_q\}}, \mathbf{G}_{\{\mathbb{C}_{q_r}\}}) \in \mathbb{R}^{I_{q_l} \times R_q \times I_{q_r}},$$

where $\mathbb{C}_q = \{\mathbb{C}_{q_l}, \mathbb{C}_{q_r}\}$, $\mathbb{C}_q$ indicate the set of dimension indices for node $q$. The operation Merge$(\cdot)$ can be seen as a series of tensor contraction operations. Specifically, as shown in Fig. 2.10, to obtain $\mathcal{G}_{\{\mathbb{C}_q\}}$, we need to first contract $\mathbf{G}_{\{\mathbb{C}_{q_l}\}}$ and $\mathcal{B}_{\{\mathbb{C}_q\}}$ along the edge $R_{q_l}$, then contract the resulting tensor with $\mathbf{G}_{\{\mathbb{C}_{q_r}\}}$ along the edge $R_{q_r}$.

After getting $\mathcal{G}_{\{\mathbb{C}_q\}}$, $\mathbf{G}_{\{\mathbb{C}_q\}} \in \mathbb{R}^{I_q \times R_q}$ can be derived directly by permuting and reshaping $\mathcal{G}_{\{\mathbb{C}_q\}}$. Merging all the nodes recursively, we can finally get the original tensor.

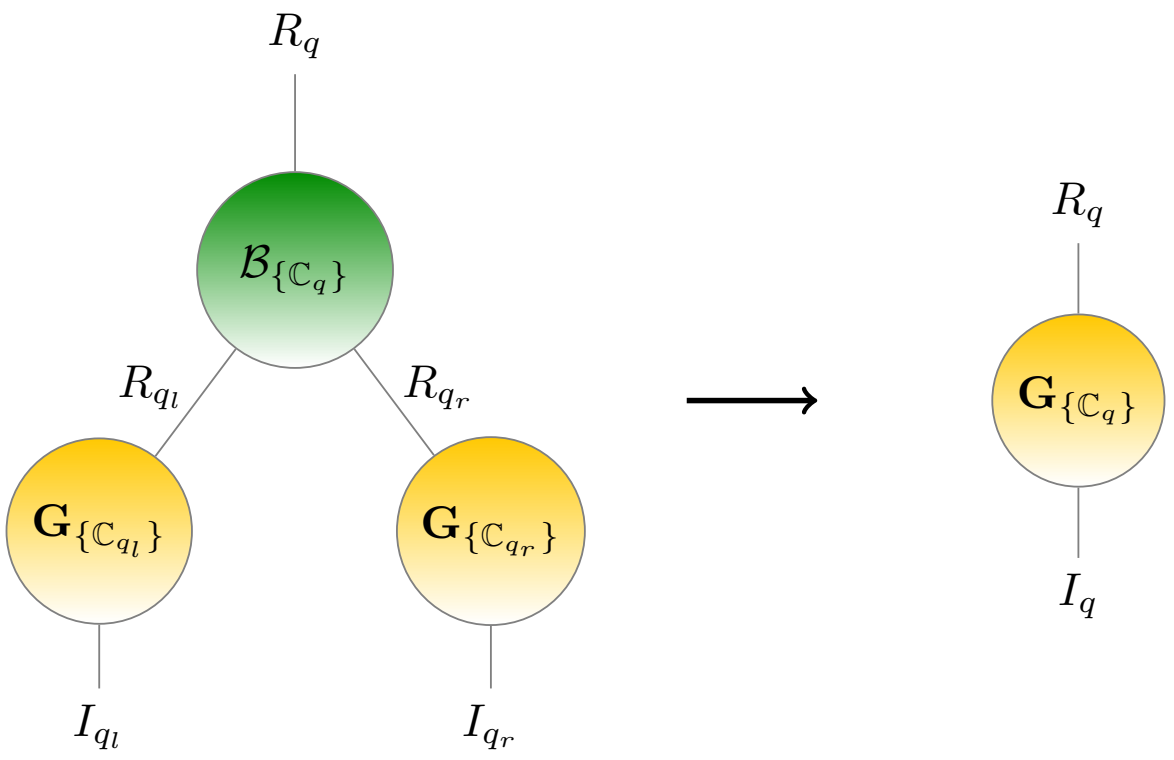

**Figure 2.10:** An illustration diagram for hierarchical Tucker decomposition.

For simplicity, we use term

$$\mathcal{A} = \mathfrak{H}(\mathbf{G}_{\{1\}}, \cdots, \mathbf{G}_{\{N\}}; \mathcal{B}_{\{1,2\}}, \cdots, \mathcal{B}_{\{N-1,N\}}; \mathbf{B}_{\{1,\cdots,N\}})$$

to represent the hierarchical Tucker decomposition hereafter. The matrix $\mathbf{G}_{\{n\}}$ is the $n$-th leaf node, $\mathcal{B}_{\{1,2\}}, \cdots, \mathcal{B}_{\{N-1,N\}}$ are the interior nodes and $\mathbf{B}_{\{1,\cdots,N\}}$ is the root node. Specifically, an example of the hierarchical Tucker decomposition form for a 5th order tensor is illustrated in Fig. 2.11.

Before introducing the t-SVD, we first define the concepts of tensor transpose, $f$-diagonal tensor and orthogonal tensor.

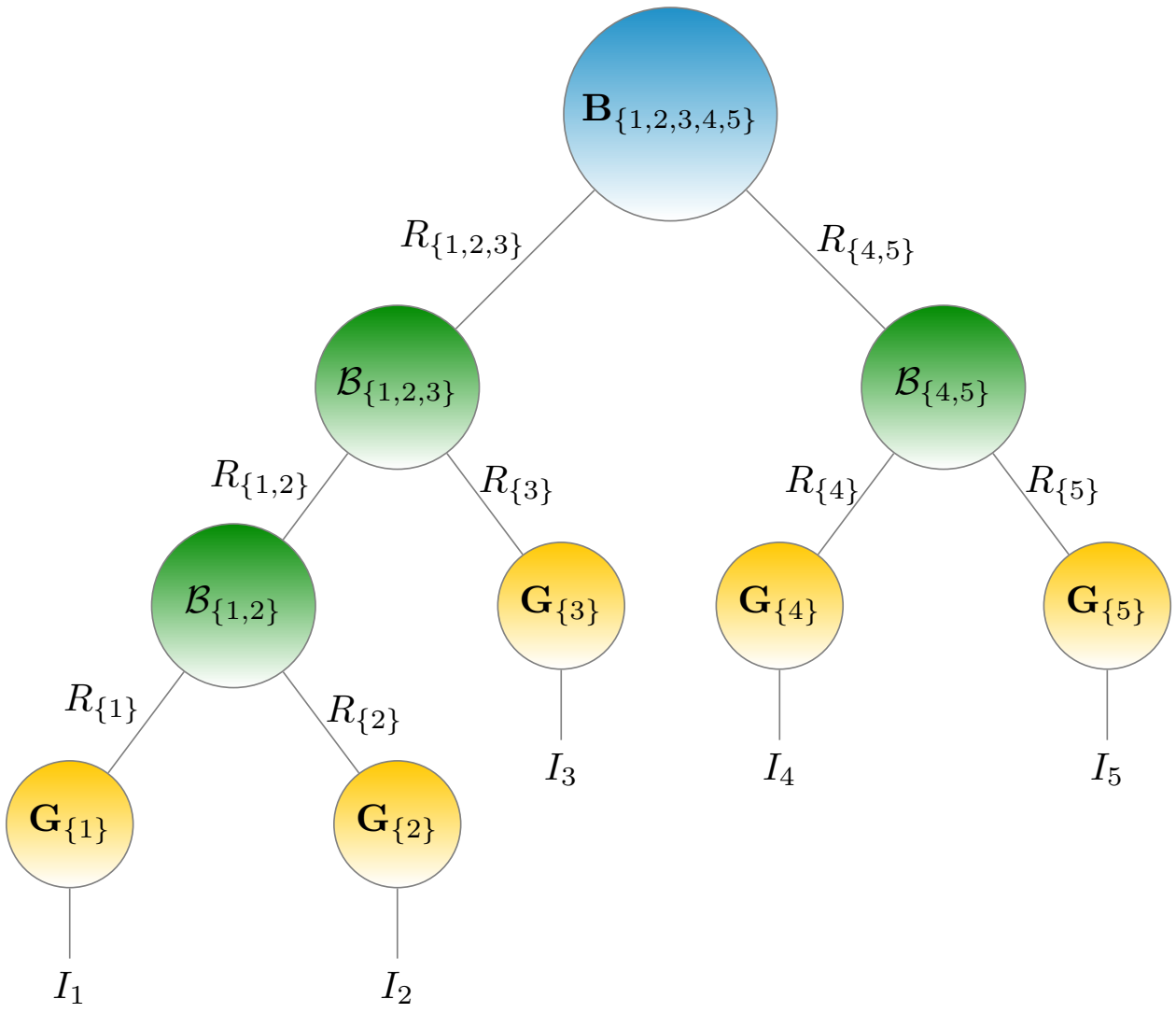

**Figure 2.11:** An example of the hierarchical Tucker decomposition form for a 5th order tensor.

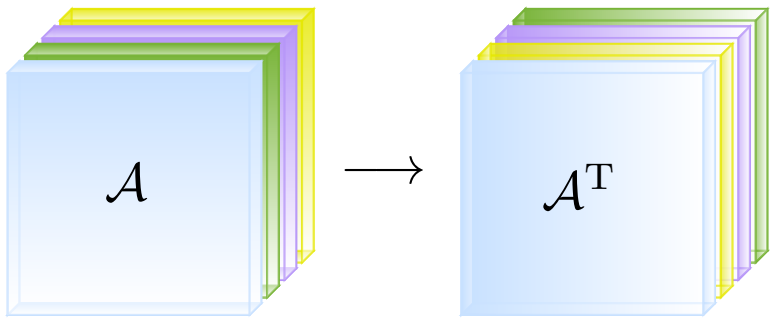

**Figure 2.12:** Tensor transpose for a third order tensor.

**Definition 2.22 (Tensor transpose).** For a third order tensor $\mathcal{A} \in \mathbb{R}^{I_1 \times I_2 \times I_3}$, its tensor transpose is expressed as $\mathcal{A}^{\mathrm{T}}$. It is obtained by transposing each frontal slice of $\mathcal{A}$ and reversing the order of transposed frontal slices through 2 to $I_3$, as shown in Fig. 2.12. Mathematically,

$$\begin{aligned} \mathcal{A}^{\mathrm{T}}(:,:,1) &= (\mathcal{A}(:,:,1))^{\mathrm{T}} \\ \mathcal{A}^{\mathrm{T}}(:,:,i_3) &= (\mathcal{A}(:,:,I_3+2-i_3))^{\mathrm{T}}, \end{aligned}$$

where $i_3 = 2, \cdots, I_3$.

**Definition 2.23 ($f$-diagonal tensor).** A tensor $\mathcal{A} \in \mathbb{R}^{I_1 \times I_2 \times I_3}$ is $f$-diagonal if each frontal slice of $\mathcal{A}$, namely $\mathcal{A}(:,:,i_3)$, $i_3 = 1, \cdots, I_3$, is a diagonal matrix, as shown in Fig. 2.13.

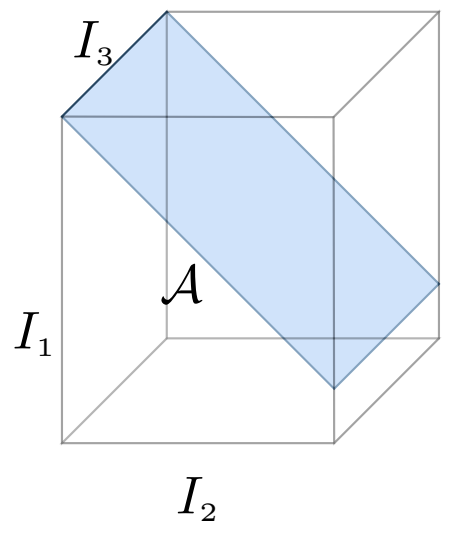

**Figure 2.13:** An example for a third order $f$-diagonal tensor.

**Definition 2.24 (Identity tensor).** The identify tensor, expressed by $\mathcal{I} \in \mathbb{R}^{I \times I \times J}$, is the tensor whose first frontal slice is an identity matrix while all other frontal tensors are all zeros. The mathematical expression is as follows

$$\mathcal{I}(:,:,1) = \mathbf{I};\ \mathcal{I}(:,:,2:J) = \mathbf{0}.$$

**Definition 2.25 (Orthogonal tensor).** Based on the definition of t-product, we say a tensor $\mathcal{Q}$ is orthogonal if it satisfies

$$\mathcal{Q}^{\mathrm{T}} * \mathcal{Q} = \mathcal{Q} * \mathcal{Q}^{\mathrm{T}} = \mathcal{I}.$$

**Definition 2.26 (t-SVD).** (Kilmer *et al.*, 2008; Kilmer and Martin, 2011) Based on the definition of t-product, the t-SVD of tensor $\mathcal{A} \in \mathbb{R}^{I_1 \times I_2 \times I_3}$ can be expressed as

$$\mathcal{A} = \mathcal{U} * \mathcal{S} * \mathcal{V}^{\mathrm{T}},$$

where $\mathcal{U} \in \mathbb{R}^{I_1 \times I_1 \times I_3}$ and $\mathcal{V} \in \mathbb{R}^{I_2 \times I_2 \times I_3}$ are orthogonal tensors, $\mathcal{S} \in \mathbb{R}^{I_1 \times I_2 \times I_3}$ is an $f$-diagonal tensor, as illustrated in Fig. 2.14. $\mathcal{V}^{\mathrm{T}} \in \mathbb{R}^{I_2 \times I_2 \times I_3}$ is the tensor transpose of $\mathcal{V}$. In Fourier domain, $\overline{\mathcal{A}}(:,:,i_3) = \overline{\mathcal{U}}(:,:,i_3)\overline{\mathcal{S}}(:,:,i_3)\overline{\mathcal{V}}(:,:,i_3)^{\mathrm{T}}$, $i_3 = 1, \cdots, I_3$. It means that t-SVD of $\mathcal{A}$ can be efficiently obtained by SVD along each frontal slice of $\overline{\mathcal{A}}$ in Fourier domain.

**Definition 2.27 (Tensor train decomposition).** (Perez-Garcia *et al.*, 2007; Oseledets, 2011; Imaizumi *et al.*, 2017) As shown in Fig. 2.15, the tensor train decomposition decomposes tensor $\mathcal{A} \in \mathbb{R}^{I_1 \times \cdots \times I_N}$ into a train structure. It can be expressed as follows

$$\mathcal{A} = [\![\mathcal{G}^{(1)}, \cdots, \mathcal{G}^{(N)}]\!]$$

with entries

$$\mathcal{A}(i_1, i_2, \cdots, i_N) = \mathcal{G}^{(1)}(:, i_1, :)\mathcal{G}^{(2)}(:, i_2, :) \cdots \mathcal{G}^{(N)}(:, i_N, :),$$

where $\mathcal{G}^{(n)} \in \mathbb{R}^{R_{n-1} \times I_n \times R_n}$ for $n = 1, \cdots, N$, $\{R_0, R_1, \cdots, R_N\}$ is the tensor train rank which meets the condition $R_0 = R_N = 1$.

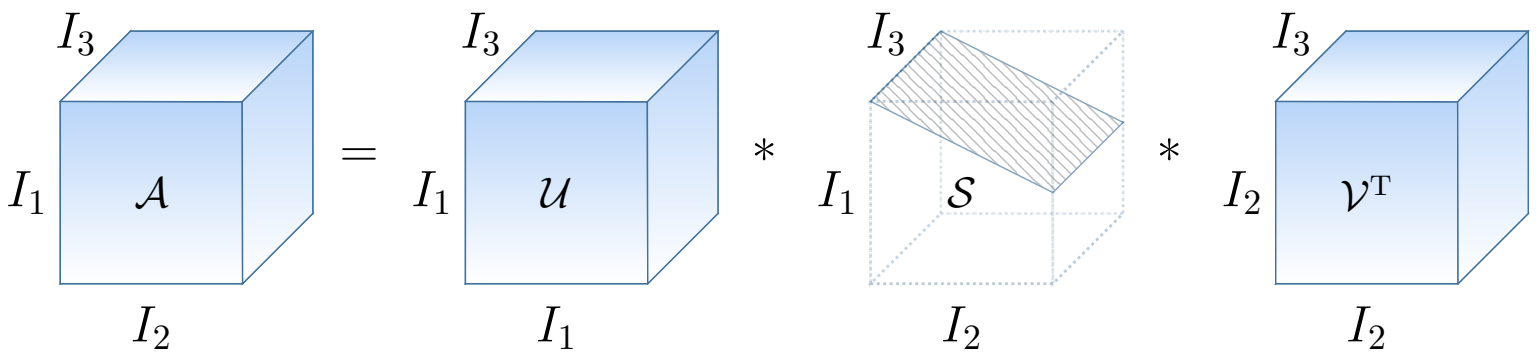

**Figure 2.14:** An illustration diagram for t-SVD decomposition.

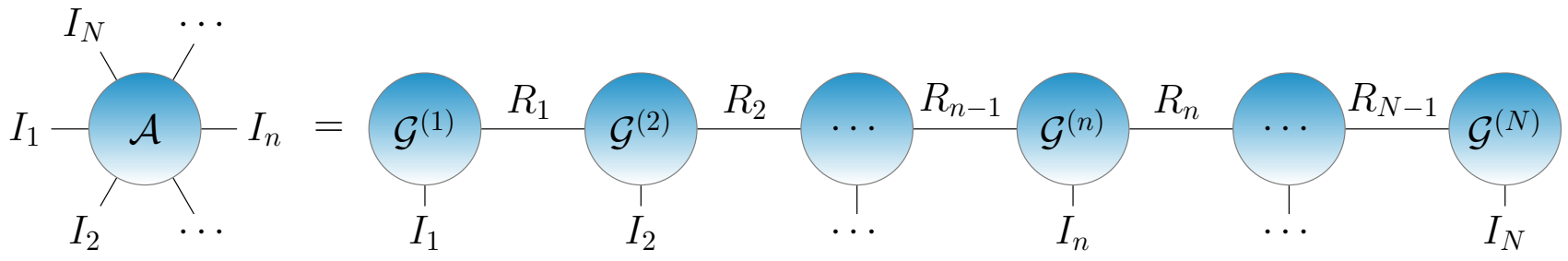

**Figure 2.15:** A graphical illustration for tensor train decomposition.

**Definition 2.28 (Tensor ring decomposition).** (Zhao *et al.*, 2018; Zhao *et al.*, 2016) Tensor ring decomposition uses several third order tensors to represent the original tensor $\mathcal{A} \in \mathbb{R}^{I_1 \times I_2 \times \cdots \times I_N}$ in a ring structure. The mathematical expression is as follows

$$\mathcal{A}(i_1, \cdots, i_N) = \mathrm{Tr}(\mathcal{G}^{(1)}(:, i_1, :) \cdots \mathcal{G}^{(N)}(:, i_N, :)),$$

where $\mathcal{G}^{(n)} \in \mathbb{R}^{R_{n-1} \times I_n \times R_n}$ for $n = 1, \cdots, N$, $\{R_0, R_1, \cdots, R_N\}$ is the tensor ring rank with $R_0 = R_N$, as shown in Fig. 2.16. For simplicity, we use $\mathcal{A} = \Re(\mathcal{G}^{(1)}, \cdots, \mathcal{G}^{(N)})$ hereafter to represent the tensor ring decomposition.

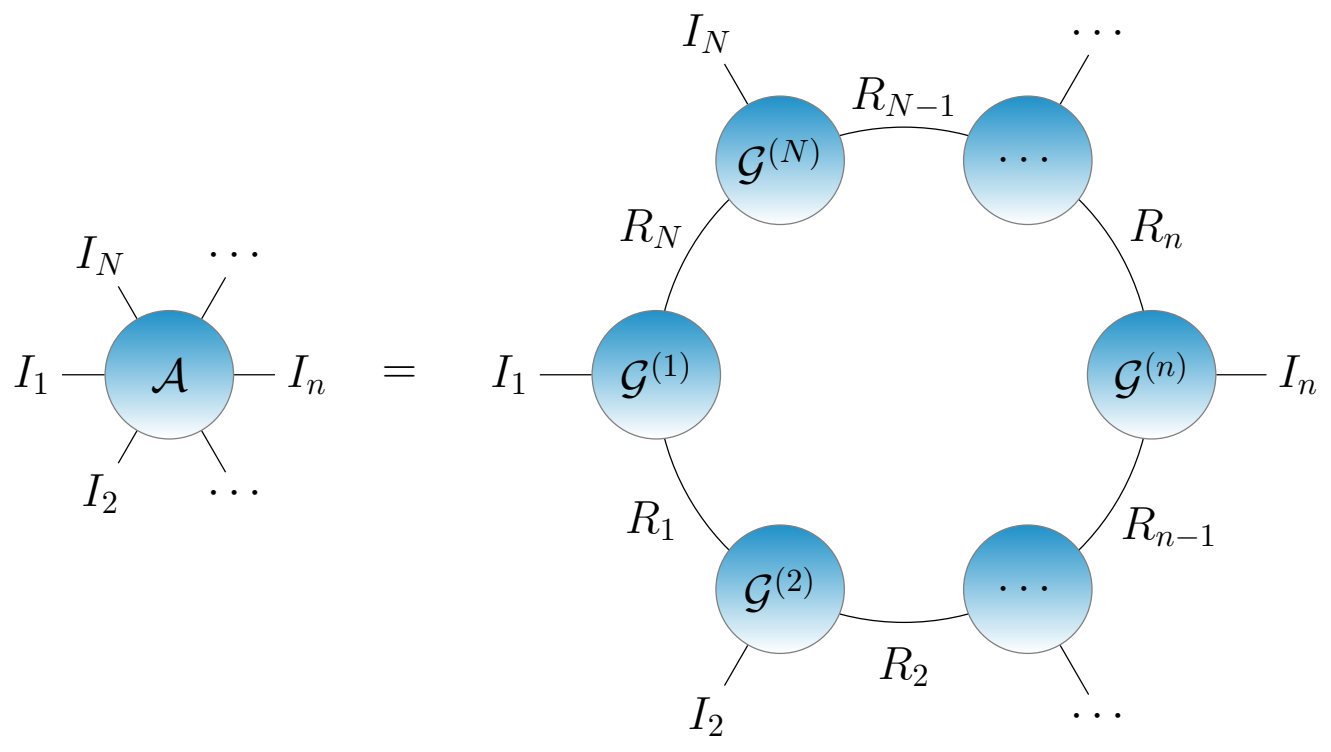

**Figure 2.16:** A graphical representation for tensor ring decomposition.

# 3

## Classical regression models

Regression methods are devoted to estimating the relationship between dependent variables and independent variables. Specifically, given $N$ samples $\{\mathbf{x}_n, y_n\}_{n=1}^{N}$, traditional regression analysis tends to explore the dependencies between the input variable $\mathbf{x}_n \in \mathbb{R}^P$ and the output $y_n \in \mathbb{R}$ through:

$$y_n = f(\mathbf{x}_n) + \varepsilon_n, \tag{3.1}$$

where $f(\cdot)$ is either a linear or nonlinear function, $\varepsilon_n$ is the bias of $n$-th sample. Regression analysis is commonly used for two purposes. The first is to forecast the most possible trends in the future or predict parameters of interest based on other information, such as climate forecasting, human age estimation, and human motion reconstruction based on videos. The other aims to identify the key factors which affect the response most. For example, disease diagnosis and brain connectivity analysis are dedicated to studying the causal relationship between different things. They focus on the areas of the brain that have a significant impact on a specific disease or behavior, or whether there is a causal relationship between different areas of the brain. Analyzing these unknown causal relationships can help doctors diagnose the disease and even enable early detection or prevention.

Regression analysis has been studied for hundreds of years and there are many popular approaches, such as generalized linear regression (Nelder and Wedderburn, 1972), partial least square (PLS) (Stone and Brooks, 1990), Gaussian process (Rasmussen, 2003), random forest (Liaw, Wiener, *et al.*, 2002), support vector regression (Smola and Schölkopf, 2004), and neural network (Specht *et al.*, 1991; Beck *et al.*, 2000).

## 3.1 Linear regression models

The most commonly used model is the linear regression model, which is very popular and significant in both scientific research and the industry field because of its simplicity. Considering the simple linear function, the linear regression model can be formulated as follows

$$y_n = \mathbf{x}_n^{\mathrm{T}} \beta + \varepsilon_n, \tag{3.2}$$

where $\beta \in \mathbb{R}^{P\times 1}$ is the coefficient matrix that needs to be trained, $\varepsilon_n$ is the bias. Letting $\mathbf{y} = [y_1, \cdots, y_N]^{\mathrm{T}} \in \mathbb{R}^{N\times 1}$, $\mathbf{X} = [\mathbf{x}_1, \cdots, \mathbf{x}_N]^{\mathrm{T}} \in \mathbb{R}^{N\times P}$, $\varepsilon = [\varepsilon_1, \cdots, \varepsilon_N]^{\mathrm{T}} \in \mathbb{R}^{N\times 1}$, we can get the matrix notation of linear regression as follows

$$\mathbf{y} = \mathbf{X}\beta + \varepsilon. \tag{3.3}$$

To tackle model (3.2), scientists have developed different methods using some prior knowledge, such as ridge regression (Hoerl and Kennard, 2000), least angle regression (Efron *et al.*, 2004), principal component regression (PCR) (Stone and Brooks, 1990), and PLS. These approaches can be cast into three classes: simple linear regression, shrinkage methods, and projection methods. They are based on different optimization frameworks listed in Table 3.1.

The simplest method is simple linear regression which estimates the regression coefficient $\beta$ by minimizing the empirical error. However, the simple linear regression model is too idealistic, and it is difficult to deal with many practical problems. For example, the noise term does not always obey a simple Gaussian distribution. There may exist an exponential relationship between input and output, or the output is discontinuous. Considering different data forms of the output, the generalized linear model (GLM) assumes that $y_n$ obeys any distribution in the exponential family with probability mass function as follows

$$p(y_n \mid \theta_n, \phi_n) = \exp\left\{ \frac{y_n\theta_n - b(\theta_n)}{a(\phi_n)} + c(y_n, \phi_n) \right\}, \tag{3.4}$$

where $\theta_n$ and $\phi_n$ are the natural and dispersion parameters, $a(\cdot)$, $b(\cdot)$ and $c(\cdot)$ are functions. Different settings for $\theta_n$, $\phi_n$, $a(\cdot)$, $b(\cdot)$ and $c(\cdot)$ will lead to different distribution forms, such as normal distribution or polynomial distribution for continuous data, Poisson distribution for count data, and Bernoulli distribution for binary output.

**Table 3.1:** Different optimization models for linear regression (Hastie *et al.*, 2009).

| Family | | Optimization models |
|---|---|---|
| Simple linear regression | | $\min_\beta \lVert \mathbf{y} - \mathbf{X}\beta \rVert_F^2$ |
| GLM | | $\max_\beta \sum_{n=1}^N \log p(y_n \vert \mathbf{x}_n, \beta)$ |
| Shrinkage methods | Ridge | $\min_\beta \lVert \mathbf{y} - \mathbf{X}\beta \rVert_F^2 + \lambda \lVert \beta \rVert_2^2$ |
| | Lasso | $\min_\beta \lVert \mathbf{y} - \mathbf{X}\beta \rVert_F^2 + \lambda \lVert \beta \rVert_1$ |
| Projection methods | PLS | $\max_{\varphi_m} \mathrm{Corr}^2(\mathbf{y}, \mathbf{X}\varphi_m)\,\mathrm{Var}(\mathbf{X}\varphi_m)$ s. t. $\lVert \varphi_m \rVert = 1, \varphi_m^{\mathrm{T}} \mathbf{S} \varphi_l = 0, l = 1, \cdots, m-1.$ |
| | PCR | $\max_{\mathbf{v}_m} \mathrm{Var}(\mathbf{X}\mathbf{v}_m)$ s. t. $\lVert \mathbf{v}_m \rVert = 1, \mathbf{v}_m^{\mathrm{T}} \mathbf{S} \mathbf{v}_l = 0, l = 1, \cdots, m-1.$ |

Then the GLM links the mean value of the output under specific distribution with the input $\mathbf{x}$ as follows

$$\begin{aligned} g(u_n) &= \mathbf{x}_n^{\mathrm{T}}\beta + \varepsilon_n, \\ u_n &= \mathrm{E}(y_n \vert \mathbf{x}_n). \end{aligned} \tag{3.5}$$

In GLM, there are three main components that need to be determined, including the linear predictor $\mathbf{x}_n^{\mathrm{T}}\beta + \varepsilon_n$, the link function $g(u_n)$, and the probability distribution of $y_n$. The linear predictor is the linear combinations of the input space. The link function $g(u_n)$ links the linear predictor with the mean value $u_n$. For example, in logistic regression, $g(u_n)$ is the logit function and the probability distribution of $y_n$ is the binomial distribution since the output is binary. The GLM is normally solved by the maximum likelihood estimation method. The corresponding optimization problems are given in Table 3.1.

Shrinkage methods are common methods for feature selection and model complexity control. This class of approaches is mainly motivated by the case when the sample number is less than the number of features. Since the ordinary least squares estimations in that case will have high variability, lead to overfitting, and fail to give an accurate prediction, this solution would also not be unique. Imposing different constraints on model parameters can not only reduce the size of the solution space but also enhance the interpretability of the model. This is because in practice, not all features are related to output. Through feature reduction technology, irrelevant features or noises can be deleted. The optimization problems of commonly used shrinkage methods, including ridge regression and Lasso regression, are given in Table 3.1.

Projection methods, like PCR and PLS, are dedicated to modeling collinear data. When the input data are correlated, the traditional least squares method will fail to give an accurate prediction. In order to make this problem tractable, projection methods aim to extract a small set of linear combinations of the input data which are uncorrelated for subsequent regression analysis instead of using original input features. So the key problem of this class of approaches is how to find the important feature combinations.

Two notable examples, PLS and PCR, both construct the uncorrelated linear combinations of input space sequentially. For example, PLS seeks one PLS direction $\varphi_m$ at a time and then deflates to find further updating directions which are uncorrelated with previous ones. Specifically, as shown in Table 3.1, PLS obtains the $m$-th PLS direction $\varphi_m$ at $m$-th iteration by maximizing both the correlation between selected feature combinations $\mathbf{X}\varphi_m$ and response $\mathbf{y}$ and the variance between $\mathbf{X}$ and the direction $\varphi_m$. The condition $\varphi_m^{\mathrm{T}}\mathbf{S}\varphi_l = 0, l = 1, \cdots, m-1$ ensures that all the selected feature combinations are uncorrelated. $\mathbf{S}$ is the sample covariance matrix of the $\mathbf{X}(:, p)$, $p = 1, \cdots, P$. Solving these sub-problems sequentially can finally result in important feature combinations. The optimization process of PLR and PLS is similar. The only difference between these two methods is that PCR only uses the input data for construction, whereas PLS also considers the correlation with the response. In fact, PCR first extracts the principal components of the input, and then uses the obtained principal components for subsequent regression analysis. The output is not considered in the feature extraction process, as shown in Table 3.1.

## 3.2 Nonlinear regression models

For real data analysis, the relationship between the explanatory variables and the dependent variables is usually nonlinear. In this case, using linear models can not provide accurate modeling and predictions. The nonlinear regression models find an appropriate curved function $f$ to fit the observed data and to predict, namely $y_n = f(\mathbf{x}_n) + \varepsilon_n$. Commonly used methods include the kernel methods (Hainmueller and Hazlett, 2014; De Jong and Ter Braak, 1994), Gaussian process (Rasmussen, 2003), additive models (Wood, 2011), regression trees or forests (Loh, 2011; Liaw, Wiener, *et al.*, 2002) and Neural networks (Beck *et al.*, 2000).

One simple idea is to transform the nonlinear models into linear ones through some transformations over the explanatory variable or the dependent variables. For example, kernel methods map the input variables $\mathbf{x}_n$ into a higher dimensional space through a function $\Phi(\mathbf{x}_n)$, and the newly derived datasets $\{\Phi(\mathbf{x}_n), y_n\}_{n=1}^{N}$ can be modeled by the linear models

$$y_n = \Phi(\mathbf{x}_n)^{\mathrm{T}}\beta + \varepsilon_n. \tag{3.6}$$

The least squares solution of the corresponding ridge regression is $\beta = (\Phi^{\mathrm{T}}\Phi)^{-1}\Phi^{\mathrm{T}}\mathbf{y} = \Phi^{\mathrm{T}}(\Phi\Phi^{\mathrm{T}} + \lambda\mathbf{I}_N)^{-1}\mathbf{y}$ with $\Phi = [\Phi(\mathbf{x}_1)^{\mathrm{T}}; \cdots; \Phi(\mathbf{x}_N)^{\mathrm{T}}]$, $\mathbf{y} = [y_1; \cdots; y_N]$, $\lambda$ is the weighting factor of the regularization term. Defining the kernel function $k(\mathbf{x}_n, \mathbf{x}_{n'}) = \Phi(\mathbf{x}_n)^{\mathrm{T}}\Phi(\mathbf{x}_{n'})$, we can get the prediction for new input $\mathbf{x}'$ as

$$y' = \Phi(\mathbf{x}')^{\mathrm{T}}\beta = \mathbf{k}'^{\mathrm{T}}(\mathbf{K} + \lambda\mathbf{I}_N)^{-1}\mathbf{y} \tag{3.7}$$

with no need to determine the specific expression of $\Phi$, where $\mathbf{k}'(n) = k(\mathbf{x}', \mathbf{x}_n)$, $\mathbf{K} = \Phi\Phi^{\mathrm{T}}$. It can be seen that the prediction of new samples is represented as a linear combination of the labels of observed samples. The weight of each sample is characterized by the similarity of the new sample and the observed one, which is measured by the kernel function.

Gaussian process assumes that the nonlinear function needed to estimate admits a Gaussian process distribution, namely

$$f(\mathbf{x}_n) \sim \mathrm{GP}(0, k(\mathbf{x}_n, \mathbf{x}_{n'})), \tag{3.8}$$

where $k(\mathbf{x}_n, \mathbf{x}_{n'})$ is the kernel function for training samples. In this way, if the noise $\varepsilon_n \sim \mathcal{N}(0, \tau^2)$, then

$$\mathbf{y} \sim \mathcal{N}(\mathbf{0}, \mathbf{K} + \tau^2 \mathbf{I}). \tag{3.9}$$

Based on Bayesian inference, the prediction of newly arrived sample $\mathbf{x}'$ is

$$y' \sim \mathcal{N}(m', \tau')$$

with $m' = \mathbf{k}'^{\mathrm{T}}(\mathbf{K} + \lambda \mathbf{I}_N)^{-1}\mathbf{y}$ and $\tau' = k'' - \mathbf{k}'^{\mathrm{T}}(\mathbf{K} + \lambda \mathbf{I}_N)^{-1}\mathbf{k}'$, $k'' = k(\mathbf{x}', \mathbf{x}')$. Then through estimating the hyperparameters of the Gaussian process, we can get both the expectation value and the confidence interval of the predictions. Comparing the results obtained by the kernel ridge regression and the Gaussian process, we can find that the Gaussian process is closely related to the kernel ridge regression. This is because both the Gaussian process and kernel ridge regression assume that all parameters and variables obey Gaussian distributions. The only difference is that kernel ridge regression only considers the mean value of the prediction, but does not consider the variance. In the Gaussian process, we have to predict the distribution of the prediction.

Additive models are nonparametric models which simplify the fully nonparametric model $f(\mathbf{x}_n) = f(x_{n,1}, \cdots, x_{n,P})$ through additivity hypothesis, where $\mathbf{x}_n \in \mathbb{R}^P$,$x_{n,p}$ is the $p$-th feature of $\mathbf{x}_n$. The mathematical expression is

$$f(\mathbf{x}_n) = \sum_{p=1}^{P} f_p(x_{n,p}). \tag{3.10}$$

If $f_p(x_{n,p}) = x_{n,p}\beta_p$, then it boils down to the simple linear models. The additive models are more flexible and general than parametric models because they do not need to assume a certain functional form. The functions $\{f_p\}_{p=1}^P$ can be pretty arbitrary smooth nonlinear functions. Meanwhile, additive models are simpler than the fully nonparametric model since the influence of the predictor variable on the response variable is assumed to be independent. It means that the interactions between different predictors are ignored. The fitting of the additive models is to perform spline smoothing on each predictor variable through an iterative process (backfitting algorithm), which iteratively estimates

$f_p$ while others are fixed until convergence

$$\min_{f_p} \sum_{n=1}^{N} \|y_n - \sum_{j \neq p} f_j(x_{n,j}) - f_p(x_{n,p})\|_{\mathrm{F}}^2, \tag{3.11}$$

where $\{y_n - \sum_{j \neq p} f_j(x_{n,j})\}_{n=1}^N$ are seen as the response when estimating $f_p$. The algorithm balances the fitting error and the degree of freedom and finally reaches the optimum.

Decision tree is a basic machine learning method for classification and regression tasks. One classical algorithm is the classification and regression tree (CART) algorithm (Loh, 2011). Decision tree splits the feature space into several units along different feature dimensions. For each unit, there will be a corresponding response, which can be simply assigned as the average of the samples divided into this unit or fitted by simple linear models. The main issue in the construction of a decision tree is the selection of the features and the splitting point. For given datasets $\{\mathbf{x}_n \in \mathbb{R}^P, y_n\}_{n=1}^N$, the decision tree is constructed by selecting an important feature and splitting the dataset by threshold $s$

$$\mathbb{R}_{\text{left}}(p, s) = \{\mathbf{x}_n | x_{n,p} \leq s\}, \mathbb{R}_{\text{right}}(p, s) = \{\mathbf{x}_n | x_{n,p} > s\}, \tag{3.12}$$

where $p = 1, \cdots, P$. To figure out the key features and corresponding best splitting point, one can minimize the following optimization problem

$$\min_{p,s}[\min_{c_1} \sum_{\mathbf{x}_n \in \mathbb{R}_{\text{left}}(p,s)} (y_n - c_1)^2 + \min_{c_2} \sum_{\mathbf{x}_n \in \mathbb{R}_{\text{right}}(p,s)} (y_n - c_2)^2], \tag{3.13}$$

where $c_1$ and $c_2$ are the estimated response of the left node and right node, respectively. When the predefined depth of the decision tree is reached or the number of patches in a node is less than a threshold, the splitting process will stop. When a new test sample arrives, the prediction will be given by the response of the unit it belongs to.

Random forest (Breiman, 2001) is a collection of multiple decision trees. It is an ensemble learning method that integrates multiple weak classifiers or regressors. It randomly selects subsets of the training sets to train different classifiers or regressors and makes predictions independently. And these predictions are finally aggregated into one single better prediction. In fact, decision tree and random forest are very robust and easy to interpret since in the regression tree, we can easily see which features or variables have the greatest influence on the response and their depth in the tree directly reflects their importance.

Neural networks aim to establish an end-to-end mapping function from the input variables into the corresponding output variables. It is commonly composed of an input layer, an output layer, and at least a hidden layer, as shown in Fig. 3.1. It is a kind of data-driven method whose performance highly relies on the training samples observed. Neural networks have shown powerful representation capability of any nonlinear functions and work well especially with large and complex datasets. Different types of neural networks are proposed with different network designs, such as convolutional neural network (CNN) and recurrent neural network (RNN). Benefitting from the convolutional operation in each layer, CNN is very effective in feature extraction and performs well in computer vision tasks, while RNN is developed to deal with sequential data analysis, which is commonly encountered in natural language processing and speech recognition.

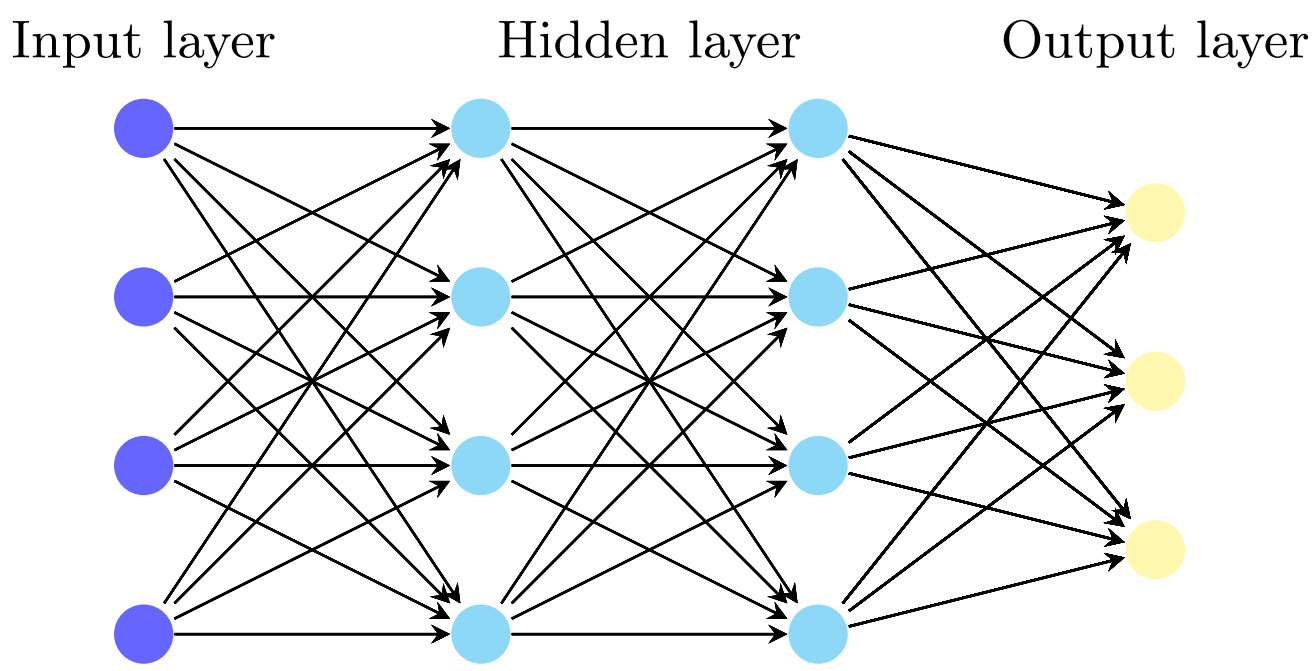

**Figure 3.1:** A graphical illustration for a basic neural network.

## 3.3 Multioutput regression

In Sections 3.1 and 3.2, we only consider single-output regression. But in practice, there exist many multioutput regression problems, such as multitask learning.

When it comes to multi-output regression problems, a simple and straightforward strategy is to fit a regression model and do estimation and prediction for each of the responses respectively. However, for multi-output regression, there may be correlations between different outputs, and the features that affect these outputs may be similar or even the same. Learning each task individually will ignore this prior information. Then reduced rank regression (Izenman, 1975) is proposed to fully explore the shared coefficient parameters between different responses. Given $N$ samples $\{\mathbf{x}_n, \mathbf{y}_n\}_{n=1}^{N}$, the reduced rank regression tries to find the solution of the following optimization problem:

$$\begin{aligned} &\min_{\mathbf{B}} \|\mathbf{Y} - f(\mathbf{X}, \mathbf{B})\|_{\mathrm{F}}^2 \\ &\text{s.t.} \quad \operatorname{rank}(\mathbf{B}) < R, \end{aligned} \tag{3.14}$$

where $\mathbf{X} = [\mathbf{x}_1, \cdots, \mathbf{x}_N]^{\mathrm{T}} \in \mathbb{R}^{N \times P}$, $\mathbf{Y} = [\mathbf{y}_1, \cdots, \mathbf{y}_N]^{\mathrm{T}} \in \mathbb{R}^{N \times Q}$, $\mathbf{B} \in \mathbb{R}^{P \times Q}$ is the coefficient matrix. The reduced rank regression assumes the coefficient matrix $\mathbf{B}$ to be low rank due to the consideration of the possible common parameters between different responses. This is a fundamental base of the subsequent multidimensional regression analysis.

## 3.4 Summary

In this chapter, we have given a concise introduction of some popular regression models for matrix-formed or vector-formed data in order to provide readers with a basic concept of regression analysis and commonly used methods, including both linear and nonlinear ones. In particular, we talk about multiple output regression tasks and reduced rank regression. This is naturally driven by the consideration that important features correlated to similar responses, like different metrics for a specific disease, are very likely to be the same. In fact, this is a basic idea that explains why the low-rank tensor approximation of the multidimensional coefficient tensor is effective when the output indices become multiple. Meanwhile, the methods we listed in this chapter are the basis of the tensor learning methods we will mention later. Most tensor regression methods are high-level extensions and innovations of existing regression methods. Understanding the traditional way of regression analysis is crucial to the understanding of the following chapters.

The regression tasks are very complicated in practice and affected by many factors, such as different kinds of noise, outliers, variables with heavy-tailed distributions, corrupted samples, etc. Therefore, besides the regression models for multidimensional data, how to model different kinds of noises, what loss functions to use, and how to improve the robustness of regression algorithms are still current research hotspots (Yi and Faloutsos, 2000; Liu *et al.*, 2017a; Diakonikolas *et al.*, 2019; Luo *et al.*, 2015; Sun *et al.*, 2019; Rousseeuw and Hubert, 2018; Zhang and Zhou, 2018; Takada *et al.*, 2019). In addition, with the emergence of some new application fields, the applicability of traditional regression methods for practical problems has also become a promising research direction, such as food safety control (Lim *et al.*, 2016; Cheng and Sun, 2017), smart transportation (Xu and Huang, 2015; Mei *et al.*, 2019), medical analysis (Zhu *et al.*, 2016; Zhen *et al.*, 2015), model performance assessment (Hewing *et al.*, 2019), and environmental risk assessment (Kang *et al.*, 2019; Arbaiy *et al.*, 2018). We refer interested readers to these books (Hastie *et al.*, 2009; Lawrence, 2019; Bates and Watts, 1988) and review paper (Fernández-Delgado *et al.*, 2018) for more details of traditional regression analysis.

# 4

# Linear tensor regression models

Tensor regression is proposed to tackle regression tasks related to high dimensional data. Like traditional regression analysis, tensor regression models can also be divided into linear models and nonlinear models. We will introduce the theories and algorithms of linear tensor regression models in this chapter and that of nonlinear ones in the next chapter.

Specifically, given $N$ samples $\{\mathcal{X}_n, \mathcal{Y}_n\}_{n=1}^N$, the linear tensor regression model can be formulated as

$$\mathcal{Y}_n = f(\mathcal{X}_n, \mathcal{B}) + \mathcal{E}_n, \tag{4.1}$$

where $\mathcal{X}_n \in \mathbb{R}^{P_1 \times \cdots \times P_L}$ is $n$-th predictor (input, independent variable), $\mathcal{Y}_n \in \mathbb{R}^{Q_1 \times \cdots \times Q_M}$ is $n$-th response (output, dependent variable), $\mathcal{B}$ is the coefficient array, $\mathcal{E}_n \in \mathbb{R}^{Q_1 \times \cdots \times Q_M}$ is the bias of $n$-th sample, $f(\mathcal{X}_n, \mathcal{B})$ indicates one tensor product form which can transform the input space into the output space using the coefficient array $\mathcal{B}$. Concatenating $N$ samples, model (4.1) can be rewritten as

$$\mathcal{Y} = f(\mathcal{X}, \mathcal{B}) + \mathcal{E}, \tag{4.2}$$

where $\mathcal{X} \in \mathbb{R}^{N \times P_1 \times \cdots \times P_L}$ with $\mathcal{X}(n, :, \cdots, :) = \mathcal{X}_n$, $\mathcal{Y} \in \mathbb{R}^{N \times Q_1 \times \cdots \times Q_M}$ with $\mathcal{Y}(n, :, \cdots, :) = \mathcal{Y}_n$, $\mathcal{E} \in \mathbb{R}^{N \times Q_1 \times \cdots \times Q_M}$ with $\mathcal{E}(n, :, \cdots, :) = \mathcal{E}_n$.

For high dimensional data regression analysis, the regression coefficient array $\mathcal{B}$ usually has a much higher order than the predictor $\mathcal{X}$ or response $\mathcal{Y}$. The simplest least squares estimation not only fails to achieve accurate estimates, but also brings huge storage and computational complexity. Therefore, some prior knowledge must be considered to simplify this regression problem. Driven by the low rank assumption of the regression coefficient matrix considered in the reduced rank regression for multi-output problems, a straightforward method is to consider low rankness in tensor form. In this way, not only the model parameters can be reduced, but also the multidirectional relatedness between the multidimensional predictor and response can be explored to improve the model performance. Some commonly used low tensor rank assumptions are based on the CP, Tucker, HT, t-SVD, TT, and TR decomposition methods we listed in Chapter 2.

In fact, compared with traditional regression models, the low tensor rank assumption is a key concept in tensor regression models. Reasonable assumptions on tensor rank determine the subspace exploration ability, parameters compression capability, and model predictive capability of the regression model. Besides, the approaches for linear tensor regression also vary under consideration of different tensor products, optimization methods, and prior knowledge (sparsity, smoothness).

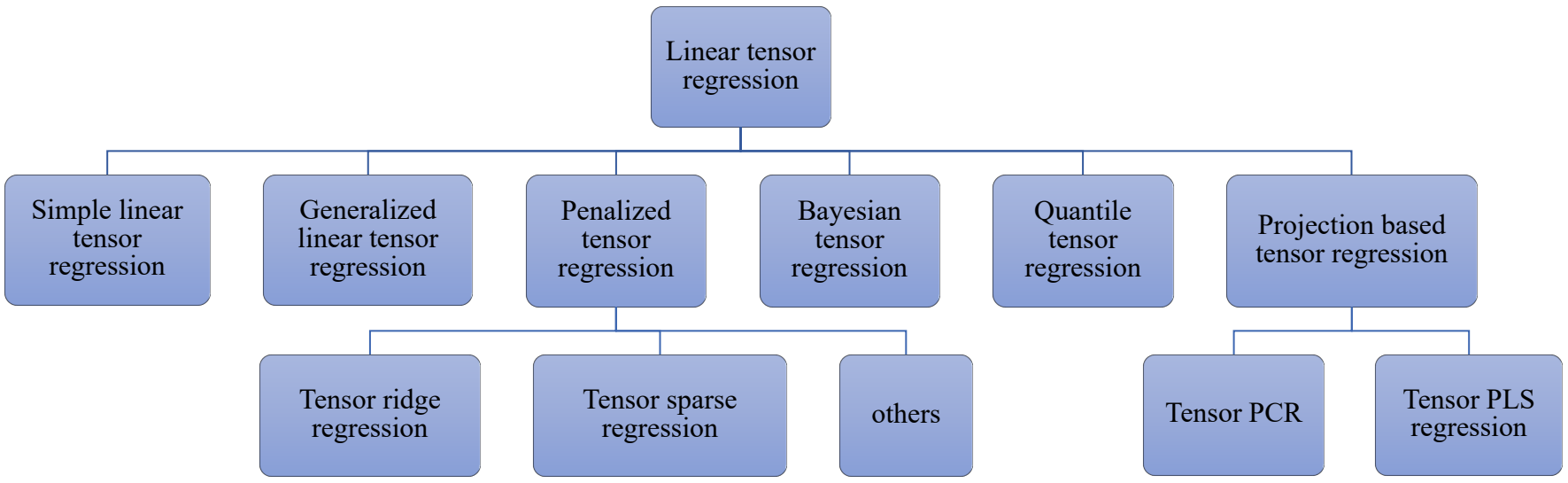

**Figure 4.1:** A taxonomy of linear tensor regression models.

Fig. 4.1 provides a taxonomy of tensor linear regression models by families. We will introduce these models and corresponding optimization methods in detail in the following. The difference and pros and cons of these approaches will also be compared and discussed.

## 4.1 Simple linear tensor regression

### 4.1.1 Tensor regression based on rank minimization

#### Multilinear multitask learning (MLMTL)

Multilinear multitask learning is a classical multi-output regression problem. Considering the parameters sharing between different tasks is the heart of multitask learning (Argyriou *et al.*, 2008; Romera-Paredes *et al.*, 2013). This strategy can not only reduce the model parameters but also achieve more accurate predictions through exploring the complex correlations of different tasks. For example, considering the classical restaurant rating problem, the objective is to predict $T_1$ customers' ratings for $T_2$ aspects (food, service, overall quality, etc) of this restaurant based on $d$ features, such as the restaurant's location, decoration, dining environment, serving speed, etc. There may exist common trends for different raters and personal preferences, thus they may share a small set of features for different tasks.

Generally speaking, we consider a regression problem of $T$ tasks. For each task $t$, the number of training samples is $m_t$. Given training dataset $\mathcal{D}_t = \{\mathbf{x}_i^t \in \mathbb{R}^{d_t}, y_i^t\}_{i=1}^{m_t}$, we can denote the coefficients of the $t$-th task as $\mathbf{b}_t \in \mathbb{R}^{d_t}$, which characterizes the relationship between the inputs $\{\mathbf{x}_i^t\}_{i=1}^{m_t}$ and outputs $\{y_i^t\}_{i=1}^{m_t}$ of $t$-th task. In this review, we only consider the homogeneous feature MTL, which implies that $d_t = d$ for $t = 1, \cdots, T$. In this setting, combining all the coefficients of $T$ tasks together, we can get the overall coefficient matrix as $\mathbf{B} = [\mathbf{b}_1, \cdots, \mathbf{b}_T] \in \mathbb{R}^{d \times T}$.

But for $T$ tasks, they may be indexed by several indices, such as the $T_1 \times T_2$ tasks indexed by customer and aspect in the restaurant rating system. So, taking account of the correlations along different dimensions, Romera-Paredes *et al.* (2013) reshape the regression coefficient matrix $\mathbf{B}$ into a high dimensional tensor $\mathcal{B} \in \mathbb{R}^{d \times T_1 \times \cdots \times T_N}$ and enforce it to be low Tucker rank. The optimization model is formulated as follows

$$\min_{\mathcal{B}} \mathfrak{L}(\mathcal{B}) + \lambda \mathfrak{F}(\mathcal{B}), \tag{4.3}$$

where the first term $\mathfrak{L}(\mathcal{B})$ is the data fitting term as follows

$$\mathfrak{L}(\mathcal{B}) = \sum_{t=1}^{T}\sum_{i=1}^{m_t} L\left(\left\langle \mathbf{x}_i^t, \mathbf{b}_t \right\rangle, y_i^t\right), \tag{4.4}$$

where $\{\mathbf{x}_i^t \in \mathbb{R}^d, y_i^t \in \mathbb{R}\}$ is $i$-th sample of $t$-th task, $m_t$ is the number of samples of $t$-th task, $L(\cdot,\cdot)$ here corresponds the loss function, which is commonly evaluated by the squared loss or the absolute loss. The regularization term $\mathfrak{F}(\mathcal{B})$ is used to minimize the Tucker rank of $\mathcal{B}$ as expressed in the following

$$\mathfrak{F}(\mathcal{B}) = \frac{1}{N+1}\sum_{j=1}^{N+1} \operatorname{rank}(\mathbf{B}_{(j)}), \tag{4.5}$$

where $\mathbf{B}_{(j)}$ is the mode-$j$ unfolding matrix for $j = 1, \cdots, N+1$, and $\lambda$ is the regularization parameter.

The Tucker rank minimization problem is generally NP-hard since the multilinear rank function is non-convex. As shown in (4.5), the Tucker rank is expressed as the sum of the rank of multiple unfolding matrices. As we know, if the nonzero singular value of matrix $\mathbf{X}$ is $\sigma_1 \geq \cdots \geq \sigma_R$, then the rank of $\mathbf{X}$ is $R$. Let $\Sigma = \operatorname{diag}(\sigma_1, \cdots, \sigma_R)$, then the matrix rank can be expressed as $\operatorname{rank}(\mathbf{X}) = \|\Sigma\|_0$.

Since the $\ell_0$ norm is non-convex, which makes the rank minimization problem an NP-hard problem, its convex envelope $\ell_1$ norm is commonly used to get a more computationally tractable problem (Tibshirani, 1996). In this way, replacing the $\ell_0$ norm in the definition of matrix rank function can yield the matrix nuclear norm, namely $\|\mathbf{X}\|_*= \|\Sigma\|_1= \sum_r \sigma_r(\mathbf{X})$, which is widely used in low rank matrix approximation (Recht *et al.*, 2010).

Extending into the tensor field, discovering an appropriate convex relaxation of the non-convex multilinear rank function in (4.5) has attracted a lot of interest. For example, simply replacing the rank function in (4.5) with its corresponding matrix nuclear norm can result in the overlapped trace norm (Romera-Paredes *et al.*, 2013), which is defined as the sum of the trace norm of the mode-$j$ unfolding matrices as follows

$$\|\mathcal{B}\|_{\text{overlapped}} = \frac{1}{N+1}\sum_{j=1}^{N+1} \|\mathbf{B}_{(j)}\|_*. \tag{4.6}$$

However, this convex relaxation is not optimal if each dimension of $\mathcal{B}$ is not equally sized or there exist different information sharing degrees with respect to different dimensions (Tomioka and Suzuki, 2013). Therefore, a new tensor latent trace norm is proposed (Tomioka and Suzuki, 2013) as follows:

$$\|\mathcal{B}\|_{\text{latent}} = \inf_{\mathcal{B}^{(1)}+\cdots+\mathcal{B}^{(N+1)}=\mathcal{B}} \sum_{j=1}^{N+1} \left\|\mathbf{B}_{(j)}^{(j)}\right\|_*, \tag{4.7}$$

where $\mathbf{B}_{(j)}^{(j)}$ is the mode-$j$ unfolding matrix of the tensor $\mathcal{B}^{(j)}$. Tensor latent trace norm takes a mixture of $N+1$ latent tensors, namely $\mathcal{B}^{(1)} + \cdots + \mathcal{B}^{(N+1)} = \mathcal{B}$. Each latent tensor is regularized to be low rank along a specific mode. For simplicity, for the $j$-th latent tensor $\mathcal{B}^{(j)}$, its mode-$j$ rank is considered, namely $\text{rank}(\mathbf{B}_{(j)}^{(j)})$ with convex surrogate $\left\|\mathbf{B}_{(j)}^{(j)}\right\|_*$. In this way, the latent tensor with a larger rank (along a specific mode) will be penalized more and the one with a lower rank will become prominent. It regularizes the rank of each latent tensor respectively instead of regularizing the different unfolding matrices of $\mathcal{B}$ directly as in a tensor overlapped trace norm. A scaled version is derived as follows (Wimalawarne *et al.*, 2014):

$$\|\mathcal{B}\|_{\text{scaled}} = \inf_{\mathcal{B}^{(1)}+\cdots+\mathcal{B}^{(N+1)}=\mathcal{B}} \sum_{j=1}^{N+1} \frac{1}{\sqrt{I_j}} \left\|\mathbf{B}_{(j)}^{(j)}\right\|_*, \tag{4.8}$$

where $I_j = d$ for $j = 1$ and $I_j = T_{j-1}$ for $j = 2, \cdots, N+1$. The tensor scaled latent trace norm has been shown to have the ability to recognize the mode with the lowest rank relative to its dimension, and further provides a better approximation for the Tucker rank function. This is especially the case for the coefficient array with heterogeneous dimensions or ranks. Besides, in Wimalawarne *et al.* (2014), the excess risk of these three tensor norms for different tasks, including tensor completion, multitask learning, and MLMTL, are all studied and verified experimentally.

### Tensor regression based on convex optimization and spectral regularization

In Signoretto *et al.* (2014) and Wimalawarne *et al.* (2016), a simple tensor based learning model with training dataset $\{\mathcal{X}_n, y_n\}_{n=1}^N$ is considered as

$$\min_{\mathcal{B}} \sum_{n=1}^{N} \mathfrak{L}(\mathcal{X}_n, y_n, \mathcal{B}) + \lambda \|\mathcal{B}\|_*, \tag{4.9}$$

where $\mathfrak{L}(\mathcal{X}_n, y_n, \mathcal{B})$ is the loss function, which is the squared loss for regression tasks

$$\mathfrak{L}(\mathcal{X}_n, y_n, \mathcal{B}) = (y_n - \langle \mathcal{X}_n, \mathcal{B} \rangle)^2$$

and logistic loss

$$\mathfrak{L}(\mathcal{X}_n, y_n, \mathcal{B}) = \log(1 + \exp(-y_n \langle \mathcal{X}_n, \mathcal{B} \rangle))$$

for classification. $\|\mathcal{B}\|_*$ is the tensor norm, $\lambda$ is the regularization parameter.

Under this learning framework, Signoretto *et al.* (2014) used the overlapped trace norm and proved that it relates to the Tucker rank. While in Wimalawarne *et al.* (2016), a comprehensive comparison of regularization with various tensor norms, including the overlapped trace norm, the latent trace norm, and the scaled latent trace norm, is conducted both theoretically and experimentally. Theoretically, given the empirical risk and expected risk

$$\hat{\mathscr{R}}(\mathcal{B}) = \frac{1}{N} \sum_n \mathfrak{L}(y_n, \langle \mathcal{B}, \mathcal{X}_n \rangle), \tag{4.10}$$

$$\mathscr{R}(\mathcal{B}) = \mathrm{E}_{(\mathcal{X}, y) \sim \mu} \mathfrak{L}(y, \langle \mathcal{B}, \mathcal{X} \rangle), \tag{4.11}$$

the optimal solution $\mathcal{B}^*$ and the estimated $\hat{\mathcal{B}}$ can be expressed as

$$\mathcal{B}^* = \arg\min_{\mathcal{B}} \mathscr{R}(\mathcal{B}) \quad \text{s.t.} \|\mathcal{B}\|_* \leq C_0, \tag{4.12}$$

$$\hat{\mathcal{B}} = \arg\min_{\mathcal{B}} \hat{\mathscr{R}}(\mathcal{B}) \quad \text{s.t.} \|\mathcal{B}\|_* \leq C_0, \tag{4.13}$$

where $\mu$ is the probability distribution from which $(\mathcal{X}_n, y_n)$ are sampled. Then the upper bounds of the excess risk for Eq. (4.9) with different tensor norms are given by following lemmas.

**Theorem 4.1.** (Wimalawarne *et al.*, 2016) With probability at least $1-\sigma$, for a given $\Lambda$-Lipchitz continuous loss function $\mathfrak{L}$, the excess risk for Eq. (4.9) with overlapped trace norm for any $\mathcal{B}^* \in \mathbb{R}^{P_1 \times \cdots \times P_L}$, $L \geq 3$, is bounded as

$$\mathscr{R}(\hat{\mathcal{B}}) - \mathscr{R}(\mathcal{B}^*) \leq 2\Lambda \frac{C}{\sqrt{N}} (\sum_l \sqrt{R_l}) \min_l (\sqrt{P_l} + \sqrt{P_{\neq l}}) + \sqrt{\frac{\log(\frac{2}{\sigma})}{2N}} \quad (4.14)$$

with $\|\mathcal{B}^*\|_{\mathrm{F}} \leq C$, $C_0 \leq C \sum_l \sqrt{R_l}$, multilinear ranks $(R_1, \cdots, R_L)$, and $P_{\neq l} = \prod_{k \neq l} P_k$.

**Theorem 4.2.** (Wimalawarne *et al.*, 2016) With probability at least $1-\sigma$, for a given $\Lambda$-Lipchitz continuous loss function $\mathfrak{L}$, the excess risk for Eq. (4.9) with latent trace norm for any $\mathcal{B}^* \in \mathbb{R}^{P_1 \times \cdots \times P_L}$, $L \geq 3$, is bounded as

$$\begin{aligned} \mathscr{R}(\hat{\mathcal{B}}) - \mathscr{R}(\mathcal{B}^*) \leq\ & 2\Lambda \frac{C}{\sqrt{N}} \sqrt{\min_l R_l} (\max_l (\sqrt{P_l} + \sqrt{P_{\neq l}}) \\ & + 1.5\sqrt{2\log(L)}) + \sqrt{\frac{\log(\frac{2}{\sigma})}{2N}} \end{aligned} \quad (4.15)$$

with $\|\mathcal{B}^*\|_{\mathrm{F}} \leq C$, $C_0 \leq C \sqrt{\min_l R_l}$, multilinear ranks $(R_1, \cdots, R_L)$, and $P_{\neq l} = \prod_{k \neq l} P_k$.

**Theorem 4.3.** (Wimalawarne *et al.*, 2016) With probability at least $1-\sigma$, for a given $\Lambda$-Lipchitz continuous loss function $\mathfrak{L}$, the excess risk for Eq. (4.9) with scaled latent trace norm for any $\mathcal{B}^* \in \mathbb{R}^{P_1 \times \cdots \times P_L}$, $L \geq 3$, is bounded as

$$\begin{aligned} \mathscr{R}(\hat{\mathcal{B}}) - \mathscr{R}(\mathcal{B}^*) \leq\ & 2\Lambda \frac{C}{\sqrt{N}} \sqrt{\min_l \frac{R_l}{P_l}} (\max_l P_l + \sqrt{P} \\ & + 1.5\sqrt{2\log(L)}) + \sqrt{\frac{\log(\frac{2}{\sigma})}{2N}} \end{aligned} \quad (4.16)$$

with $\|\mathcal{B}^*\|_{\mathrm{F}} \leq C$, $C_0 \leq C \sqrt{\min_l \frac{R_l}{P_l}}$, multilinear ranks $(R_1, \cdots, R_L)$, and $P_{\neq l} = \prod_{k \neq l} P_k$.

Comparing the upper bounds of the excess risk given above, we can see that the upper bound of the excess risk for the overlapped trace norm is related to all the multilinear ranks, that of the latent trace norm is only related to the minimum one of the multilinear ranks, and that of the scaled latent norm is related to the minimum ratio of the multilinear ranks and corresponding dimensions, namely $\min_l R_l/P_l$.

Rather than minimizing the tensor trace norms directly using the proximity operator, Wimalawarne *et al.* (2016) employed their dual formulations and optimized the dual problems based on ADMM. The experimental results in both regression and classification tasks illustrate the superiority of tensor-based learning methods over traditional learning methods and verify the robustness of the scaled latent norm.

However, for this rank-minimization-based optimization framework, although they enjoy the global convergence property, they often suffer from high computational complexity and huge memory print for large-scale regression tasks. This is mainly because this group of approaches requires performing fully SVD of all the mode-$j$ unfolding matrices at each iteration. So this group of methods is not applicable for large-scale applications.

### 4.1.2 Greedy based low rank tensor learning

For multivariate spatio-temporal analysis, how to appropriately exploit the spatial and temporal information is a critical problem. Focusing on two important types of spatial-temporal analysis, i.e., cokriging and forecasting, Bahadori *et al.* (2014) proposed a unified low tensor rank learning framework as follows

$$\begin{aligned}\hat{\mathcal{B}} &= \arg\min_{\mathcal{B}} \mathfrak{L}(\mathcal{B}, \mathcal{X}, \mathcal{Y}) \\ &= \arg\min_{\mathcal{B}} \{\sum_{m=1}^{M} \|\mathcal{B}(:,:,m)\mathcal{X}(:,:,m) - \mathcal{Y}(:,:,m)\|_{\mathrm{F}}^2\} \\ \text{s.t.} &\quad \operatorname{rank}(\mathcal{B}) \leq R,\end{aligned} \tag{4.17}$$

where $\mathcal{X} \in \mathbb{R}^{KP\times T\times M}$, $\mathcal{Y} \in \mathbb{R}^{P\times T\times M}$, $\mathcal{B} \in \mathbb{R}^{P\times KP\times M}$ are the tensor input, output and coefficient tensor, respectively. Here the coefficient tensor $\mathcal{B}$ is assumed to be low Tucker rank. $T$ is the number of timestamps, $P$ is the number of features like locations, $K$ is the number of lags, and $M$ is the number of variables.

Instead of directly minimizing the rank function through convex relaxation, Bahadori *et al.* (2014) employed a greedy algorithm with a convergence guarantee, namely Greedy. At each iteration, this algorithm searches for the mode which provides the largest decrease in objective function by adding a rank-1 approximation of the unfolding matrix of the coefficient tensor along that mode into the model. Mathematically,

$$\max_{l} \mathfrak{L}(\mathcal{B}, \mathcal{X}, \mathcal{Y}) - \mathfrak{L}(\text{refold}(\mathbf{B}_{(l)} + \mathbf{W}_l), \mathcal{X}, \mathcal{Y}), \tag{4.18}$$

where $\mathbf{W}_l$ is obtained by

$$\min_{\mathbf{W}_l} \mathfrak{L}(\text{refold}(\mathbf{B}_{(l)} + \mathbf{W}_l), \mathcal{X}, \mathcal{Y}) \quad \text{s.t. } \text{rank}(\mathbf{W}_l) = 1, \tag{4.19}$$

where refold$(\cdot)$ represents the reverse operation of the unfolding operation which satisfies refold$(\mathbf{B}_{(l)}) = \mathcal{B}$, $l = 1, \cdots, 3$. For each mode, the rank-1 estimation is obtained by the orthogonal matching pursuit (Pati *et al.*, 1993). An orthogonal projection procedure is also integrated after each iteration for performance enhancement. Meanwhile, Bahadori *et al.* (2014) proved that the difference between the loss function evaluated at $k$-th iteration and the optimal solution is bounded.

**Theorem 4.4.** (Bahadori *et al.*, 2014) Let $\mathcal{B}^*$ be the optimal solution of Eq. (4.17), $\mathcal{B}_k$ is the $k$-th iterate of Eq. (4.17) by greedy method, then

$$\mathfrak{L}(\mathcal{B}_k, \mathcal{X}, \mathcal{Y}) - \mathfrak{L}(\mathcal{B}^*, \mathcal{X}, \mathcal{Y}) \leq \frac{\sigma_y^2 \|\mathbf{B}^*_{(1)}\|_*^2}{k+1} \tag{4.20}$$

if $\mathbf{Y}_m^{\mathrm{T}}\mathbf{Y}_m$ are positive definite with $\mathbf{Y}_m = \mathcal{Y}(:, :, m)$ for $m = 1, \cdots, M$, where $\sigma_y$ is the largest singular value of the block diagonal matrix diag$(\mathbf{Y}_1, \cdots, \mathbf{Y}_M)$.

Through experiments on synthetic data and real applications as cokriging and forecasting, the Greedy algorithm is proven to be superior to corresponding rank minimization-based methods in both estimation error and computational complexity.

### 4.1.3 Tensor projected gradient (TPG)

Considering the regression tasks in the context of both multilinear multitask learning and spatio-temporal forecasting, given an input $\mathcal{X}$ and an output $\mathcal{Y}$, the following optimization problem is considered in Yu and Liu (2016)

$$\begin{aligned} \min_{\mathcal{B}} \quad & \mathfrak{L}(\mathcal{B}, \mathcal{X}, \mathcal{Y}) \\ \text{s.t.} \quad & \operatorname{rank}(\mathcal{B}) \leq R, \end{aligned} \tag{4.21}$$

where $\mathcal{X} \in \mathbb{R}^{T \times D_1 \times D_3}$ is the predictor, $\mathcal{Y} \in \mathbb{R}^{T \times D_2 \times D_3}$ is the response, $\mathcal{B} \in \mathbb{R}^{D_1 \times D_2 \times D_3}$ is the regression coefficient tensor whose Tucker rank is assumed to not exceed $R$. The empirical loss $\mathfrak{L}(\mathcal{B}, \mathcal{X}, \mathcal{Y})$ is calculated under a simple linear regression model

$$\mathcal{Y} = \ll \mathcal{X}, \mathcal{B} \gg + \mathcal{E}, \tag{4.22}$$

where the product $\ll \mathcal{X}, \mathcal{B} \gg$ is defined as the independent matrix multiplication on each slice in Yu and Liu (2016), the mathematical expression is $<< \mathcal{X}, \mathcal{B} >> (:,:,d_3) = \mathcal{X}(:,:,d_3)\mathcal{B}(:,:,d_3), d_3 = 1, \cdots, D_3$.

To tackle the optimization problem (4.22), instead of using the general solutions based on ADMM or alternating least squares (ALS), Yu and Liu (2016) proposed a simple and efficient method dubbed subsampled tensor projected gradient (TPG), which combines a general gradient descent step with a proximal point projection step. Specifically, at iteration $k$, the gradient descent step finds the gradient descent direction of loss function $\mathfrak{L}$ with respect to $\mathcal{B}^{k-1}$

$$\nabla \mathfrak{L}(\mathcal{B}^{k-1}, \mathcal{X}, \mathcal{Y}) = \ll \mathcal{X}^{\mathrm{T}}, \mathcal{Y} - \ll \mathcal{X}, \mathcal{B}^{k-1} \gg \gg, \tag{4.23}$$

and then updates $\mathcal{B}$ at iteration $k$ as follows

$$\mathcal{B}^k = \mathcal{B}^{k-1} - \eta \nabla \mathfrak{L}(\mathcal{B}^{k-1}; \mathcal{X}, \mathcal{Y}), \tag{4.24}$$

where $\eta$ is the step size. Then it is followed by a projection step. In the sense of Tucker decomposition, the objective for a projection step is as follows

$$\begin{aligned} \min_{\mathcal{B}} & \left\| \mathcal{B}^k - \mathcal{B} \right\|_{\mathrm{F}}^2 \\ \text{s.t. } & \operatorname{rank}(\mathcal{B}) \leq R, \end{aligned} \tag{4.25}$$

which finds a low Tucker rank approximation for $\mathcal{B}^k$. In general, it can be seen as an extension of projected gradient descent or iterative hard thresholding algorithm into the tensor field.

In order to further speed up the algorithm and reduce the memory cost, TPG uses two strategies. First, it employs the power iteration algorithm to find one leading singular vector at a time rather than performing SVD on the original tensor with optimality (local) guarantees, which makes it efficient especially for low rank coefficient tensors. The other is randomized sketching, which is proved to be effective for regression tasks with more samples or extremely sparse data (Clarkson and Woodruff, 2017). (See Section 6 for more details about the sketching methods).

Moreover, with the assumption over predictor $\mathcal{X}$ in Definition 4.1 and optimality (local) guarantees of the projection step, the theoretical properties of TPG are analyzed and stated as Theorem 4.5.

**Definition 4.1.** (**Restricted Isometry Property** (RIP)) For predictor $\mathcal{X}$, the following holds for all $\mathcal{B}$ with Tucker rank at most $R$

$$(1-\sigma_R)\|\mathcal{B}\|_{\mathrm{F}}^2 \leq \|\ll \mathcal{X}, \mathcal{B} \gg\|_{\mathrm{F}}^2 \leq (1+\sigma_R)\|\mathcal{B}\|_{\mathrm{F}}^2, \tag{4.26}$$

where the isometry constant is the smallest number $\sigma_R$.

**Theorem 4.5.** (Yu and Liu, 2016) Suppose predictor $\mathcal{X}$ satisfies the RIP condition with isometry constant $\sigma_{2R} \leq 1/3$, the TPG algorithm with step-size $\eta = 1/(1+\sigma_R)$ can achieve a feasible solution $\mathcal{B}$ for Eq. (4.22) such that

$$\|\mathcal{B} - \mathcal{B}^*\|_{\mathrm{F}}^2 \leq \frac{1}{1-\sigma_{2R}}\|\mathcal{E}\|_{\mathrm{F}}^2 \tag{4.27}$$

in at least $\frac{1}{\log(1/\alpha)} \log(\frac{\|\mathcal{Y}\|_{\mathrm{F}}^2}{\|\mathcal{E}\|_{\mathrm{F}}^2})$ iterations for a constant $\alpha$, where $\mathcal{B}^*$ is the optimal solution of Eq. (4.22).

### 4.1.4 Multilinear tensor regression for longitudinal relational data (MLTR)

Inspired by the Tucker decomposition form, Hoff (2015) extended the bilinear regression model into a more general model which is capable of exploring the relationship between multiway arrays. Stacking all the observations, the multilinear regression model can be formulated as

follows

$$\begin{aligned} \mathcal{Y} &= \mathcal{X} \times \{\mathbf{B}^{(1)}, \cdots, \mathbf{B}^{(M)}, \mathbf{I}_N\} + \mathcal{E}, \\ &= \mathcal{X} \times_1 \mathbf{B}^{(1)} \cdots \times_M \mathbf{B}^{(M)} \times_{M+1} \mathbf{I}_N + \mathcal{E}, \end{aligned} \tag{4.28}$$

where $\mathcal{Y} \in \mathbb{R}^{Q_1 \times \cdots \times Q_M \times N}$, $\mathcal{X} \in \mathbb{R}^{P_1 \times \cdots \times P_M \times N}$, $\mathbf{I}_N$ is a $N \times N$ diagonal matrix, $\mathbf{B}^{(m)} \in \mathbb{R}^{P_m \times Q_m}$ for $m = 1, \cdots, M$ is a series of coefficient matrices.

To get the model parameters $\mathbf{B}^{(m)}$, $m = 1, \cdots, M$, the corresponding optimization model can be given as follows

$$\min_{\mathbf{B}^{(1)}, \cdots, \mathbf{B}^{(M)}} \|\mathcal{Y} - \mathcal{X} \times_1 \mathbf{B}^{(1)} \cdots \times_M \mathbf{B}^{(M)} \times_{M+1} \mathbf{I}_N\|_{\mathrm{F}}^2. \tag{4.29}$$

But unlike the standard linear regression, there does not exist a closed-form solution due to the mode-$m$ product of $\mathbf{B}^{(1)}, \cdots, \mathbf{B}^{(M)}$. In order to tackle this problem over several variables, we resort to the idea of alternating minimization, which optimizes over individual coordinates. In other words, it updates each of the parameters $\mathbf{B}^{(1)}, \cdots, \mathbf{B}^{(M)}$ iteratively while the others are fixed.

Then based on Lemma 2.2, the optimization model (4.29) boils down to a simple multivariate linear regression model as follows

$$\min_{\mathbf{B}^{(m)}} \|\mathbf{Y}_{(m)} - \mathbf{B}^{(m)} \hat{\mathbf{X}}^{(m)}\|_{\mathrm{F}}^2, \tag{4.30}$$

where $\hat{\mathbf{X}}^{(m)} = \mathbf{X}_{(m)}(\mathbf{B}^{(M)} \otimes \cdots \otimes \mathbf{B}^{(m+1)} \otimes \mathbf{B}^{(m-1)} \otimes \cdots \otimes \mathbf{B}^{(1)})^{\mathrm{T}}$. This is the classical alternating least squares (ALS) algorithm.

Although this model constructs a map between the multidimensional input and multidimensional output, the order of the tensor predictor and response must be the same, which limits the application of this model.

## 4.2 Generalized linear tensor regression

### 4.2.1 Generalized linear tensor regression based on alternating minimization

Inspired by the demand of medical imaging analysis, Zhou *et al.* (2013) first extended the GLM into the tensor field. Considering the single-response regression problem (when $M = 1$ and $Q_1 = 1$), the systematic part of the generalized tensor linear model with CP decomposition (CP-GLTR) can be formulated as

$$\begin{aligned} g(\mu) &= \alpha + \gamma^{\mathrm{T}} \mathbf{z} + \langle \mathcal{B}, \mathcal{X} \rangle \\ &= \alpha + \gamma^{\mathrm{T}} \mathbf{z} + \left\langle \sum_{r=1}^{R} \mathbf{u}_r^{(1)} \circ \cdots \circ \mathbf{u}_r^{(L)}, \mathcal{X} \right\rangle, \end{aligned} \tag{4.31}$$

where $\mathcal{X} \in \mathbb{R}^{N \times P_1 \times \cdots \times P_L}$ is the tensor input, $\mathbf{z}$ is the vector input, $g(\mu)$ indicates a link function of the mean value $\mu = \mathrm{E}(y|\mathcal{X}, \mathbf{z})$, $\gamma$ is the coefficient vector, $\mathcal{B}$ is the coefficient tensor which is decomposed as $\mathcal{B} = \sum_{r=1}^{R} \mathbf{u}_r^{(1)} \circ \cdots \circ \mathbf{u}_r^{(L)}$, $\alpha$ is the intercept.

It can be seen from model (4.31) that modeling the regression coefficient array $\mathcal{B}$ using the CP decomposition can reduce the model parameters from $O(\prod_l P_l)$ to $O(\sum_l P_l R)$. Meanwhile, the relationship between input $\mathcal{X}$ and output $\mathcal{Y}$ can be preserved by fewer parameters through fully exploring the multidirectional structure information of $\mathcal{X}$ and $\mathcal{Y}$. In fact, based on Lemma 2.1, model (4.31) can be transformed into a model as follows

$$\begin{aligned} g(\mu) = \ & \alpha + \gamma^{\mathrm{T}} \mathbf{z} + \langle \mathbf{U}^{(l)}, \mathbf{X}_{(l)} (\mathbf{U}^{(L)} \odot \cdots \odot \mathbf{U}^{(l+1)} \\ & \odot \mathbf{U}^{(l-1)} \odot \cdots \odot \mathbf{U}^{(1)}) \rangle, \end{aligned} \tag{4.32}$$

where $\mathbf{U}^{(l)} = [\mathbf{u}_1^{(l)}, \cdots, \mathbf{u}_R^{(l)}]$. The model (4.32) can be treated as a traditional GLM with respect to each factor $\mathbf{U}^{(l)}$, then the trainable parameters can be updated easily by solving several traditional GLMs alternatively. It boils down to the ALS algorithm if the response satisfies the Gaussian distribution.

Meanwhile, in order to study the large sample number asymptotic for tensor regression in Eq. (4.31), the identifiability of CP factorization is first analyzed. To deal with the indeterminacy due to scaling and permutation, a specific constrained parameterization is adopted as follows

$$\begin{aligned} \mathscr{B} = \{ & \mathbf{U}^{(1)}, \cdots, \mathbf{U}^{(L)}; \mathbf{U}^{(l)}(1, r) = 1 \text{ for } l = 1, \cdots, L-1, r = 1, \cdots, R \\ & \text{and } \mathbf{u}_1^{(l)}(1) > \cdots > \mathbf{u}_R^{(l)}(1) \}, \end{aligned} \tag{4.33}$$

where the first row of $\mathbf{U}^{(1)}, \cdots, \mathbf{U}^{(L-1)}$ are all scaled to ones, and the permutation indeterminacy is fixed by arranging the first row of $\mathbf{U}^{(L)}$ in descending order. Based on this, a sufficient and necessary condition for local identifiability of parametrization in Eq. (4.31) is given with a nonsingular Fisher information matrix.

**Theorem 4.6.** (Zhou *et al.*, 2013) Suppose $\mathcal{B}^* = [\![\mathbf{U}^{*(1)}, \cdots, \mathbf{U}^{*(L)}]\!] \in \mathscr{B}$ is (globally) identifiable up to permutation and $\mathcal{X}_n$ are independently identically distributed from a bound distribution, then

- **Consistency** The Maximum likelihood estimation is consistent. $\hat{\mathcal{B}}$ converges to $\mathcal{B}^*$ in probability in (1) normal tensor regression with a compact space $\mathcal{B}^* \in \mathscr{B}$; (2) binary tensor regression; and (3) passion tensor regression with a compact space $\mathcal{B}^* \in \mathscr{B}$.

- **Asymptotic Normality** For $\mathcal{B}^* \in \mathscr{B}$ with nonsingular information matrix $I(\mathcal{B}^*)$ and consistent $\hat{\mathcal{B}}$, then $\text{vec}(\hat{\mathcal{B}})$ is asymptotically distributed as

$$\text{vec}(\hat{\mathcal{B}}) \rightarrow \mathcal{N}(\text{vec}(\mathcal{B}^*), NI^{-1}(\mathcal{B}^*)).$$

Under an identifiability condition, $\hat{\mathcal{B}}$ is shown to be root-$N$ consistent and converges to the best approximation of $\mathcal{B}$ in CP format with a large sample size.

However, determining the CP rank is an NP hard problem and it is also difficult to find the optimal solution with a given CP rank. Moreover, under the CP decomposition, the rank is fixed along each dimension of $\mathcal{B}$, which ignores the possibility that there may exist different parameter sharing degrees along different dimensions.

Driven by the drawbacks of low CP rank assumption presented before, it is natural to consider the popular Tucker decomposition with multilinear rank. Letting the coefficient tensor admit the Tucker decomposition $\mathcal{B} = \mathcal{G} \times_1 \mathbf{U}^{(1)} \times_2 \mathbf{U}^{(2)} \cdots \times_L \mathbf{U}^{(L)}$, the generalized tensor linear model with the Tucker decomposition (Tucker-GLTR) (Li *et al.*, 2013a) can be formulated as

$$\begin{aligned} g(\mu) &= \alpha + \gamma^{\mathrm{T}}\mathbf{z} + \langle \mathcal{B}, \mathcal{X} \rangle \\ &= \alpha + \gamma^{\mathrm{T}}\mathbf{z} + \left\langle \mathcal{G} \times_1 \mathbf{U}^{(1)} \times_2 \mathbf{U}^{(2)} \cdots \times_L \mathbf{U}^{(L)}, \mathcal{X} \right\rangle, \end{aligned} \tag{4.34}$$

where $\mathcal{G} \in \mathbb{R}^{R_1 \times \cdots \times R_L}$ is the core tensor factor, $\{R_1, \cdots, R_L\}$ is the Tucker rank.

Compared with the CP-based generalized linear regression model, the Tucker-based one has a more flexible representation capability and better compression capability. With the Tucker decomposition, the model parameter number is reduced to $O(\sum_l P_l R_l + \prod_l R_l)$. If the $R_l$ is all the same as the CP rank, then it is obvious that the Tucker-based model requires more parameters than the CP-based one. But in practice, the Tucker rank is much smaller than the CP rank, especially when tensor $\mathcal{B}$ is skewed in dimensions or ranks (Cichocki *et al.*, 2017). This is mainly because the Tucker decomposition can explore the multilinear rank, which allows the different rank values for each dimension.

In fact, CP decomposition can be seen as a special case of Tucker decomposition with the core tensor $\mathcal{G}$ restricted to be a superdiagonal tensor with only nonzero elements in its superdiagonal line. In that sense, the core tensor $\mathcal{G}$ for the Tucker decomposition can better exploit the interactions between different basis vectors in all modes instead of the restricted superdiagonal core tensor of CP. However, the same high order of core tensor $\mathcal{G}$ as the original tensor $\mathcal{B}$ really limits the application of the Tucker decomposition for high-order tensors.

Similar to the CP-based GLTR, the asymptotic properties of the Tucker-GLTR model are also studied. The only difference comes from the different fisher information matrices and constrained parameterization of CP decomposition and Tucker decomposition. In order to deal with the nonsingular transformation indeterminacy of the Tucker decomposition, the constrained parameterization adopted is as follows

$$\mathscr{B} = \{\mathcal{G}; \mathbf{U}^{(1)}, \cdots, \mathbf{U}^{(L)}; \quad \mathbf{U}^{(l)}(p_l, r_l) = 1 \text{ for } p_l = 1, \cdots, P_l, \\ r_l = 1, \cdots, R_l, l = 1, \cdots, L\}, \quad (4.35)$$

which means that the first $R_l$ rows of $\mathbf{U}^{(l)}$ are all set to be ones. With this space, the identifiability of the Tucker decomposition is ensured when the fisher information matrix is nonsingular under some mild assumptions.

To avoid the dimensionality curse, the hierarchical Tucker decomposition is employed to the generalized tensor linear model. The regression model with HT (HT-GLTR) (Hou and Chaib-Draa, 2015) can be expressed as follows

$$
\begin{aligned}
g(\mu) = \; & \alpha + \gamma^{\mathrm{T}}\mathbf{z} + \langle \mathcal{B}, \mathcal{X} \rangle & (4.36)\\
= \; & \alpha + \gamma^{\mathrm{T}}\mathbf{z} & (4.37)\\
& + \left\langle \mathfrak{H}(\mathbf{G}_{\{1\}}, \cdots, \mathbf{G}_{\{L\}}; \mathcal{T}_{\{1,2\}}, \cdots, \mathcal{T}_{\{L-1,L\}}; \mathbf{T}_{\{1,\cdots,L\}}), \mathcal{X} \right\rangle .
\end{aligned}
$$

Following the same generalized regression model, these three methods based on CP, Tucker, and HT are all solved using the maximum likelihood estimation method. Unlike Tucker decomposition, HT decomposes the original tensor $\mathcal{B}$ into a binary tree structure. It can maintain all the good properties of the Tucker decomposition but further decomposes the core tensor into several low-order tensor factors. In that case, the parameter number will not increase exponentially with the order of the original coefficient tensor $\mathcal{B}$. The experiments on medical image analysis have also verified the flexibility and scalability of HT decomposition especially for the high dimensional datasets with small sample sizes.

### 4.2.2 Generalized linear tensor regression based on projected gradient descent

In addition to the alternating minimization framework, a general framework for generalized tensor regression models under different low-rank structures using the projected gradient descent (GLTR-PGD) algorithm is proposed in Chen *et al.* (2019). Considering the generalized linear tensor regression model $g(\mu) = \langle \mathcal{B}, \mathcal{X} \rangle$ without additional vector input and bias, the negative log-likelihood risk objective for given training samples $\{\mathcal{X}_n, y_n\}_{n=1}^{N}$ is

$$\mathfrak{L}(\mathcal{B}) = \frac{1}{N} \sum_{n=1}^{N} \frac{b(\langle \mathcal{B}, \mathcal{X}_n \rangle) - y_n \langle \mathcal{B}, \mathcal{X}_n \rangle}{a(\phi_n)} - c(y_n, \phi_n) \qquad (4.38)$$

when the response $y_n$ admits the exponential family of distributions. In order to exploit the low rank structure of the coefficient tensor $\mathcal{B}$, three forms of low rank constraint are listed, of which two are based on matrix slices, the last one is to ensure the Tucker rank is upper bounded. Taking a third order tensor $\mathcal{B} \in \mathbb{R}^{P_1 \times P_2 \times P_3}$ as an example, we can get the mathematical expression of the three low rank structural assumptions as follows.

$$\mathscr{B}_1(R) = \{\mathcal{B} : \sum_{p_3=1}^{P_3} \text{rank}(\mathcal{B}(:,:,p_3)) \leq R\}$$

$$\mathscr{B}_2(R) = \{\mathcal{B} : \sum_{p_3=1}^{P_3} \text{rank}(\mathcal{B}(:,:,p_3)) \leq R, \sum_{p_3=1}^{P_3} \|\mathcal{B}(:,:,p_3)\|_0 \leq S\}$$

$$\mathscr{B}_3(R_1, R_2, R_3) = \{\mathcal{B} : \text{rank}(\mathbf{B}_{(l)}) \leq R_l \text{ for } l = 1, \cdots, 3\}.$$

In this way, the optimization objective of generalized tensor regression models under different low-rank structures can be expressed as

$$\min_{\mathcal{B}} \mathfrak{L}(\mathcal{B}) \text{ s.t. } \mathcal{B} \in \mathscr{B}, \tag{4.39}$$

where $\mathscr{B}$ can be either of the low rank structural spaces $\mathscr{B}_1(R)$, $\mathscr{B}_2(R)$ or $\mathscr{B}_3(R_1, \cdots, R_L)$. It can be tackled by the projected gradient descent method through a gradient descent step $g_k = \hat{\mathcal{B}}_k - \eta\nabla\mathfrak{L}(\hat{\mathcal{B}}_k)$ and a projection step which projects the updated $g_k$ into the low rank structural space $\hat{\mathcal{B}}_{k+1} = P_{\mathscr{B}}(g_k)$ iteratively until convergence. Besides, under the condition of ensuring that the projection operator satisfies a contractive property, the general risk bounds of the method based on the projection gradient descent applied to the Gaussian linear tensor regression model under different low-rank structure assumptions are given.

## 4.3 Penalized tensor regression

For linear tensor regression tasks, different models can be obtained through different learning methods. According to the principle of Occam's razor, for the same problem, the simplest model among the feasible models is the best. Therefore, in order to improve the generalization ability of the model and avoid overfitting, the model complexity usually needs to be restricted, and one classical method is regularization. From the perspective of Bayesian estimation, the regularization term corresponds to the prior probability of the model. It is generally assumed that a model with a large complexity has a small prior probability and a simple model has a large prior probability.

Based on different prior knowledge, shrinkage methods can be classified into several groups, such as tensor ridge regression with $\ell_2$ norm constraint on $\mathcal{B}$, tensor sparse regression with $\ell_1$ norm penalty on $\mathcal{B}$ and others. We will review the related works on shrinkage methods in detail in the following from these three categories.

### 4.3.1 Tensor ridge regression

#### Higher rank tensor ridge regression (hrTRR)

Considering the single response regression, given a set of labeled training set $\{\mathcal{X}_n, y_n\}$, the higher rank tensor ridge regression (hrTRR) based on CP decomposition, Guo *et al.* (2012) aims to minimize the following objective function

$$\begin{aligned} & \mathfrak{L}\left(\left\{\mathbf{U}^{(1)}, \mathbf{U}^{(2)}, \cdots, \mathbf{U}^{(L)}\right\}, \gamma\right) \\ = \quad & \frac{1}{2}\sum_{n=1}^{N}\left(y_n - \left\langle \mathcal{X}_n, \sum_{r=1}^{R} \mathbf{u}_r^{(1)} \circ \cdots \circ \mathbf{u}_r^{(L)} \right\rangle - \gamma\right)^2 \\ & + \frac{\lambda}{2}\left\|\sum_{r=1}^{R} \mathbf{u}_r^{(1)} \circ \cdots \circ \mathbf{u}_r^{(L)}\right\|_{\mathrm{F}}^2, \end{aligned} \tag{4.40}$$

where the first term is the empirical risk, the second term is the penalty term on the model parameters $\{\mathbf{U}^{(1)}, \mathbf{U}^{(2)}, \cdots, \mathbf{U}^{(L)}\}$, $\mathbf{U}^{(l)} = [\mathbf{u}_1^{(l)}, \cdots, \mathbf{u}_R^{(l)}]$ for $l = 1, \cdots, L$, $\gamma$ is the intercept.

The model (4.40) is an extension of standard ridge regression. As deduced in Guo *et al.* (2012), based on alternating minimization, each model parameter $\mathbf{U}^{(l)}$ can be obtained by gradient-descent-style methods. In addition, for each subproblem of $\mathbf{U}^{(l)}$, a closed-form solution can be obtained by relaxing the penalty constraint on the whole coefficient tensor $\mathcal{B}$ into a separable regularization term on each factor $\mathbf{U}^{(l)}$ as follows

$$\mathfrak{J}(\mathcal{B}) = \sum_{l=1}^{L} \|\mathbf{U}^{(l)}\|_{\mathrm{F}}^2. \tag{4.41}$$

In this paper, a solution based on $\epsilon$-insensitive loss function is also studied and has shown better performance in most experiments.

### Tensor-on-tensor regression

To construct the linear relationship map between both the high dimensional predictor and response, Lock (2018) proposed to use the tensor contracted product to predict the response $\mathcal{Y}$ from a tensor $\mathcal{X}$. The tensor-on-tensor regression model is as follows

$$\mathcal{Y} = \langle \mathcal{X}, \mathcal{B} \rangle_L + \mathcal{E}, \tag{4.42}$$

where $\mathcal{B} \in \mathbb{R}^{P_1 \times \cdots \times P_L \times Q_1 \times \cdots \times Q_M}$ is the coefficient array which is in much higher order than $\mathcal{X}$ and $\mathcal{Y}$.

By assuming the low rank property of coefficient array $\mathcal{B}$ under CP decomposition, the optimization problem with a Frobenius norm regularization term on $\mathcal{B}$ can be expressed as follows

$$\begin{aligned} &\min_{\mathcal{B}} \|\mathcal{Y} - \langle \mathcal{X}, \mathcal{B} \rangle_L\|_{\mathrm{F}}^2 + \lambda \|\mathcal{B}\|_{\mathrm{F}}^2 \\ &\text{s.t.} \quad \mathrm{rank}_{\mathrm{CP}}(\mathcal{B}) \leq R. \end{aligned} \tag{4.43}$$

In order to handle this problem, the objective (4.43) is first transformed into an unregularized version with modified input and output, and then tackled by the ALS algorithm through optimizing over the factor matrices of $\mathcal{B}$ under CP decomposition.

Actually, tensor-on-tensor regression based on CP decomposition (TTR-CP) is exactly the same as hrTRR if the response is a single value for each sample. But TTR-CP drives a closed-form solution for the tensor ridge regression based on a low CP rank assumption with no need to relax the penalty term. Meanwhile, it generalizes the tensor regression analysis to apply for both tensor predictors and responses of any order.

Variants of tensor-on-tensor regression based on tensor train and tensor ring, called TTR-TT/TR, can refer to Liu *et al.* (2020c) and Liu *et al.* (2021). Compared with the traditional CP decomposition, the coefficient tensor admits the tensor train decomposition in Liu *et al.* (2020c)

$$\mathcal{B} = [\![ \mathcal{U}^{(1)}, \cdots, \mathcal{U}^{(L)}, \mathcal{V}^{(1)}, \cdots, \mathcal{V}^{(M)} ]\!], \tag{4.44}$$

as shown in Fig. 4.2, and admits the tensor ring decomposition in Liu *et al.* (2021)

$$\mathcal{B} = \mathfrak{R}(\mathcal{U}^{(1)}, \cdots, \mathcal{U}^{(L)}, \mathcal{V}^{(1)}, \cdots, \mathcal{V}^{(M)}), \tag{4.45}$$

as shown in Fig. 4.3.

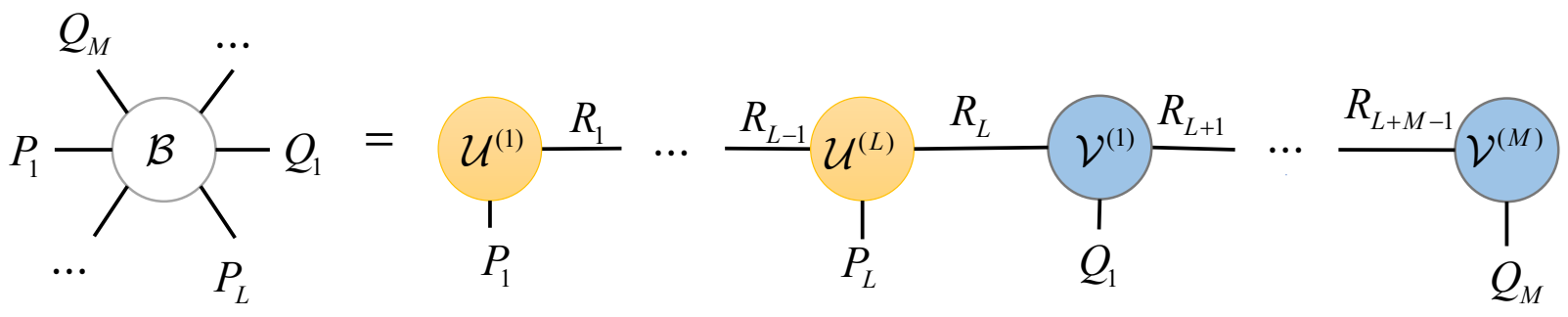

**Figure 4.2:** The diagram description of the tensor train network of the coefficient tensor.

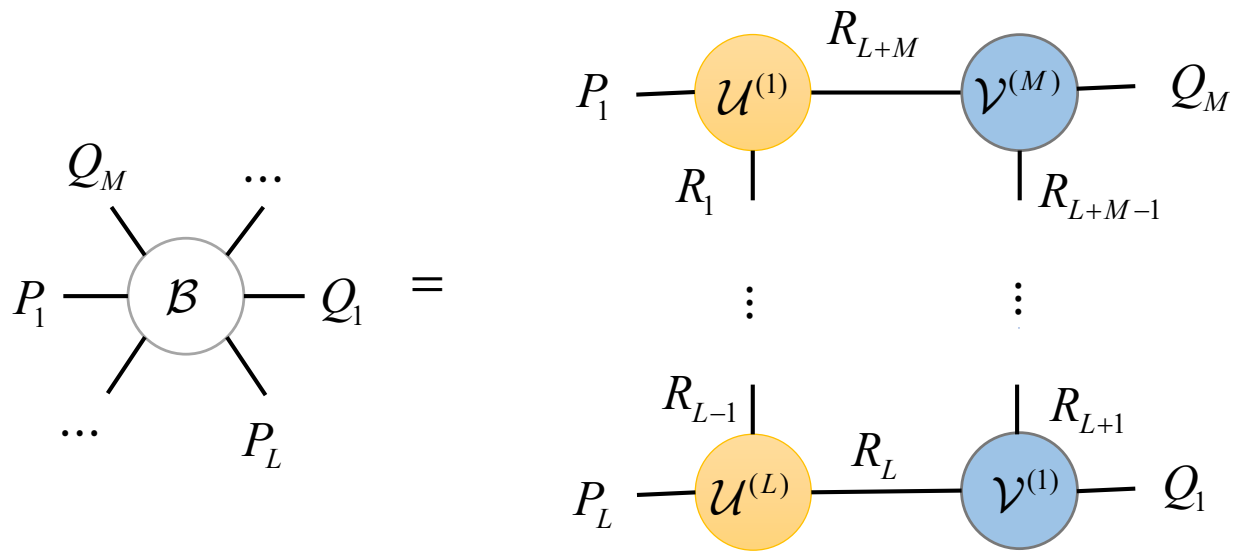

**Figure 4.3:** The diagram description of the tensor ring network of the coefficient tensor.

Benefiting from tensor networks, the TT/TR based tensor-on-tensor regression is expected to obtain better results, especially efficient for high dimensional tensors with a small size in each mode. Additionally, in order to determine the appropriate rank for specific tasks, Liu *et al.* (2020c) adopted a tensor train rounding procedure to obtain adaptive ranks (Oseledets, 2011). Specifically, tensor train rounding aims to approximate the given tensor in TT format with optimal ranks, which can reduce the tensor train rank at each iteration with required accuracy.

### High order low-rank regression (HOLRR)

Unlike the vector-on-tensor and tensor-on-tensor regression models mentioned before, Rabusseau and Kadri (2016) proposed a closed-form solution for the tensor-on-vector regression model based on a low Tucker rank assumption over the regression coefficient tensor. Taking consideration of complexity control by the Frobenius norm penalty term, the optimization model is given by

$$\begin{aligned} \min_{\mathcal{B}} \; & \|\mathcal{B} \times_1 \mathbf{X} - \mathcal{Y}\|_{\mathrm{F}}^2 + \lambda \|\mathcal{B}\|_{\mathrm{F}}^2 \\ \text{s.t.} \; & \operatorname{rank}_{\text{Tucker}}(\mathcal{B}) \leq (R_0, R_1, \cdots, R_M)\,, \end{aligned} \tag{4.46}$$

where $\mathcal{B} \in \mathbb{R}^{P_1 \times Q_1 \times \cdots \times Q_M}$, $\mathbf{X} \in \mathbb{R}^{N \times P_1}$. For each sample, the predictor is a vector.

Based on the tensor factorization, problem (4.46) can be transformed into the following problem

$$\begin{aligned} \min_{\mathcal{B}} \; & \|\mathcal{G} \times_1 \hat{\mathbf{X}}\mathbf{U}^{(0)} \cdots \times_{M+1} \mathbf{U}^{(M)} - \hat{\mathcal{Y}}\|_{\mathrm{F}}^2 \\ \text{s.t.} \; & (\mathbf{U}^{(m)})^{\mathrm{T}} \mathbf{U}^{(m)} = \mathbf{I}_{R_m}, \forall m \in \{0, \cdots, M\}, \end{aligned} \tag{4.47}$$

where $\mathbf{U}^{(m)} \in \mathbb{R}^{Q_m \times R_m}$ for $m = 1, \cdots, M$, $\mathbf{U}^{(0)} \in \mathbb{R}^{P_1 \times R_0}$, $\hat{\mathbf{X}} = [\mathbf{X}^{\mathrm{T}}, \lambda \mathbf{I}_{P_1}]^{\mathrm{T}} \in \mathbb{R}^{(N+P_1) \times P_1}$, $\hat{\mathcal{Y}}$ with its mode-1 unfolding matrix obtained by $\hat{\mathbf{Y}}_{(1)} = [\mathbf{Y}_{(1)}^{\mathrm{T}}, \mathbf{0}_{P_1}]^{\mathrm{T}} \in \mathbb{R}^{(N+P_1) \times Q_1 \cdots Q_M}$.

Then the problem (4.47) can be tackled by an efficient closed-form solution, namely high order low-rank regression (HOLRR). It solves the approximation problem by simply finding the $M+1$ projections of $\hat{\mathcal{Y}}$ onto $\hat{\mathbf{X}}\mathbf{U}^{(0)}$,$\cdots$ and $\mathbf{U}^{(M)}$ along different modes independently without iterations. Compared with other methods, HOLRR is computationally efficient and enjoys approximation guarantees.

**Theorem 4.7.** (Rabusseau and Kadri, 2016) With $\mathcal{B}^*$ being the optimal solution of Eq. (4.46) and $\mathcal{B}$ being the solution obtained by the HOLRR algorithm, then

$$\mathfrak{L}(\mathcal{B}) \leq (p+1)\mathfrak{L}(\mathcal{B}^*),$$

where $\mathfrak{L}(\mathcal{B}) = \|\mathcal{B} \times_1 \mathbf{X} - \mathcal{Y}\|_{\mathrm{F}}^2 + \lambda \|\mathcal{B}\|_{\mathrm{F}}^2$.

Moreover, based on a similar framework, an application in structured point cloud data analysis has been recently proposed (Yan *et al.*, 2019). But instead of the orthogonality constraint in (4.47), Yan *et al.* (2019) employed a weighted orthogonality regularizer as follows

$$(\mathbf{U}^{(m)})^{\mathrm{T}} \Sigma_m^{-1} \mathbf{U}^{(m)} = \mathbf{I}_{R_m}, \forall m \in \{1, \cdots, M\},$$

where $\Sigma_m$ represents the spatial correlation of the noise. Employing this weighted orthogonality regularizer can further take the spatial covariance structure information into account to get better estimation for the basis $\mathbf{U}^{(m)}$.

However, the correlation within the input space is ignored, which makes it infeasible for some tasks, such as image-based clinical diagnosis and video-based motion reconstruction. In addition, a kernelized version of HOLRR is also proposed and has shown better performance. But even the kernelized vision can not exploit the high dimensional structure information of the input.

### 4.3.2 Tensor sparse regression

In recent years, neuroimaging analysis has attracted a lot of attention for its possible use in early diagnosis and computer-assisted systems of diagnosis and treatment. It also provides a way to understand brain connectivity and key regions that affect specific diseases or behaviors. However, the collection of task-related medical images is usually difficult due to the long time cost, the limited sample size, and complex acquisition environments. Therefore, for neuroimaging analysis, the sample size is commonly smaller than the parameter number, making it intractable. Even when imposing a low rank structure on the regression coefficient tensor, the parameter number can also exceed the sample size in some cases. Thus, it is necessary to employ the regularization term to control the model complexity and improve the generalization ability of the model. As mentioned earlier, in addition to controlling the complexity of the model, the regularization term can actually take account of prior knowledge. Specifically, for neuroimaging studies, the sparsity constraint is commonly used due to the consideration that the specific disease or behavior is only related to subregions of the brain, as assumed in biology that each brain region has a specific function.

For neuroimaging data analysis, there are two main studies: one is prediction or classification tasks based on high dimensional predictors, the other is causal analysis or comparison tasks with a tensor response. The specific applications will be discussed later in Chapter 7 in detail. Here we review the popular methods for these two studies and a few sparse models developed for spatio-temporal data analysis and image or video processing.

### Sparse generalized linear tensor regression (SGLTR)

For example, the generalized linear tensor regression is extended to a penalized version with a sparsity constraint on the latent factors of tensor $\mathcal{B}$ under different decomposition methods (Zhou *et al.*, 2013; Li *et al.*, 2013a). First, with the CP decomposition, the regularized version of the optimization model based on maximum likelihood estimation (MLE) can be expressed as

$$\max_{\alpha,\gamma,\mathbf{U}^{(1)},\ldots,\mathbf{U}^{(L)}} \mathfrak{L}\left(\alpha,\gamma,\mathbf{U}^{(1)},\ldots,\mathbf{U}^{(L)}\right) - \lambda\sum_{l=1}^{L}\|\mathbf{U}^{(l)}\|_1, \tag{4.48}$$

where $\mathfrak{L}(\alpha,\gamma,\mathbf{U}^{(1)},\ldots,\mathbf{U}^{(L)})$ is the log-likelihood function, the second term enforces all the latent factors $\mathbf{U}^{(l)}$, $l = 1,\cdots,L$ of $\mathcal{B}$ under CP decomposition to be sparse. With the sparsity constraint, the regularized tensor regression model will be able to identify the relevant regions associated with the response and achieve better predictions.

Variants mentioned in Section 4.2.1 can also be easily extended into a regularized version and solved by existing software libraries for penalized GLM regression problems. But it is necessary to note that the sparsity constraint can vary with different considerations. For example, for Tucker-based sparse generalized linear tensor regression, the sparsity constraint can be imposed on the core tensor only, or on all the factors jointly.

### Sparse multilinear regression based on rank minimization

The success of rank minimization-based frameworks in MLMTL and tensor completion (Liu *et al.*, 2012; Long *et al.*, 2019) also sheds light on new directions for sparse multilinear linear regression. In Song and Lu (2017), a new method, dubbed Regularized multilinear regression and selection (Remurs), is proposed based on the tensor nuclear norm. Given $N$ samples $\{\mathcal{X}_n, y_n\}_{n=1}^{N}$, the corresponding optimization model can be formulated as

$$\min_{\mathcal{B}} \frac{1}{2}\sum_{n=1}^{N}\left(y_n - \langle\mathcal{X}_n,\mathcal{B}\rangle\right)^2 + \tau\|\mathcal{B}\|_{\text{overlapped}} + \gamma\|\mathcal{B}\|_1, \tag{4.49}$$

where $\mathcal{X}_n \in \mathbb{R}^{P_1 \times P_2 \times P_3}$, $\|\mathcal{B}\|_{\text{overlapped}}$ denotes the tensor overlapped trace norm, which is a type of tensor nuclear norm. Compared with the sparse generalized linear tensor regression models which enforce the latent factors to be sparse, Remurs imposes the sparsity constraint on the coefficient array $\mathcal{B}$ directly.[1] It can improve the interpretability of regression models to some extent. The classical ADMM algorithm is used to tackle the proposed model and the convergence property is studied in terms of the objective function.

As illustrated before, for this kind of rank minimization-based regression framework, the computational burden caused by the requirement of full SVD at each iteration is a main challenge. This is why more recent attention has focused on the provision of tensor factorization-based optimization methods. However, the newly proposed t-SVD enables the optimization problem based on rank minimization to be effectively implemented. Derived from this, Li *et al.* (2019b) introduced the tubal tensor nuclear norm based on t-SVD to replace the original tensor nuclear norm, namely sparse tubal-regularized multilinear regression (Sturm). The optimization objective can then be transferred as

$$\min_{\mathcal{B}} \frac{1}{2} \sum_{n=1}^{N} (y_n - \langle \mathcal{X}_n, \mathcal{B} \rangle)^2 + \tau \|\mathcal{B}\|_{\text{TNN}} + \gamma \|\mathcal{B}\|_1, \tag{4.50}$$

where $\|\mathcal{B}\|_{\text{TNN}}$ is denoted as

$$\|\mathcal{B}\|_{\text{TNN}} = \frac{1}{P_3} \sum_{p_3=1}^{P_3} \|\overline{\mathcal{B}}(:,:,p_3)\|_*, \tag{4.51}$$

where $\overline{\mathcal{B}}$ is the discrete Fourier transform of $\mathcal{B}$ along the third mode. As shown in (4.51), rather than performing full SVD on the unfolding matrices of $\mathcal{B}$, Sturm minimizes the nuclear norm of each frontal slice of $\overline{\mathcal{B}}$. It captures the coherence along the third mode through the Fourier transform, while the tubal tensor nuclear norm constraint explores the structural information in the first two modes. In this way, it helps to avoid the computational burden brought by the full SVD operations on the unfolding matrices (Lu *et al.*, 2019). However, it will be impractical if the size of each slice of $\mathcal{B}$ is large, and the permutation also affects the performance of t-SVD based algorithms (Hu *et al.*, 2016).

[1]The $\ell_1$ norm of tensor $\mathcal{B}$ refers to the number of non-zero elements in $\mathcal{B}$, which can be calculated by the $\ell_1$ norm of any one of its unfolding matrices.

### Sparse tensor regression based on projection gradient methods

Considering the efficiency of the projected gradient descent-based method, the TPG method for spatial-temporal data analysis described in Section 4.1.3 is also extended to a sparsity regularized version with an external sparse constraint on the coefficient tensor $\mathcal{B}$ (Ahmed *et al.*, 2019). Aiming at the neuroimaging analysis, the constrained optimization problem can be formulated as

$$\min_{\mathcal{B}\in\mathscr{C}} \frac{1}{2}\sum_{n=1}^{N}\|y_n - \langle \mathcal{X}_n, \mathcal{B}\rangle\|_2^2, \tag{4.52}$$

where $\|\cdot\|_2$ refers to the $\ell_2$ norm.[2] The coefficient tensor $\mathcal{B}$ is forced to be $\{R_1, \cdots, R_L\}$-rank and $S$-sparse under the Tucker decomposition, and $\mathscr{C}$ is used to represent a collection of such tensors

$$\mathscr{C} = \{\mathcal{G} \times_1 \mathbf{U}^{(1)} \times_2 \mathbf{U}^{(2)} \times_3 \cdots \times_L \mathbf{U}^{(L)} : \mathcal{G} \in \mathbb{R}^{R_1\times R_2\times\cdots\times R_L},$$
$$\mathbf{U}^{(l)} \in \mathbb{R}^{P_l\times R_l}, \|\mathbf{U}^{(l)}(:, r_l)\|_0 \leq S_l, l = 1, \cdots, L, r_l = 1, \cdots, R_l\}.$$

Similar to the TPG algorithm, the estimation of $\mathcal{B}$ can be achieved by an iterative algorithm, which updates $\mathcal{B}$ along its gradient descent direction over the objective function and then projects the updated $\mathcal{B}$ into a low-dimensional subspace at each iteration. The resulting algorithm is called sparse TPG (STPG). But rather than only forcing $\mathcal{B}$ to be a low Tucker rank in the TPG algorithm, STPG projects it into the set $\mathscr{C}$ which satisfies both low rank property and sparsity. In this way, the number of model parameters is further reduced and it takes account of the typical case that the response only depends on parts of the input variables. In addition, under the assumption of independently and identically distributed sub-Gaussian predictors, the sample complexity bound only poses a polylogarithmic dependence on $\bar{P} := \max P_l, l = 1, 2, \cdots, L$. The experiments on neuroimaging analysis verify the effectiveness of the introduction of sparsity constraint on $\mathcal{B}$, especially in a small sample size regime.

[2]The $\ell_2$ norm is also known as Frobenius norm, which is normally expressed as $\|\cdot\|_2$ for vectors, $\|\cdot\|_\mathrm{F}$ for matrices or tensors.

Moreover, collecting all the samples together, the regression model can be expressed as

$$\mathbf{y} = \langle \mathcal{X}, \mathcal{B} \rangle_L + \mathbf{e}$$

with $\mathbf{y} = [y_1, \cdots, y_N]$, $\mathcal{X}(n, :, \cdots, :) = \mathcal{X}_n$ for $n = 1, \cdots, N$, $\mathbf{e}$ is the bias. If the linear map $\langle \mathcal{X}, \mathcal{B} \rangle_L$ satisfies the RIP denoted in Definition 4.2, the theoretical properties of STPG are analyzed and stated as Theorem 4.8.

**Definition 4.2.** (($R, S, \tau, \sigma_{R,S,\tau}$)**-Restricted Isometry Property**) For predictor $\mathcal{X}$, the restricted isometry constant $\sigma_{R,S,\tau}$ of linear map $\mathcal{X}$ is the smallest quantity such that

$$(1 - \sigma_{R,S,\tau}) \|\mathcal{B}\|_{\mathrm{F}}^2 \leq \|\langle \mathcal{X}, \mathcal{B} \rangle_L\|_{\mathrm{F}}^2 \leq (1 + \sigma_{R,S,\tau}) \|\mathcal{B}\|_{\mathrm{F}}^2 \tag{4.53}$$

for all $\mathcal{B} \in \mathscr{C}_{R,S,\tau}$ with

$$\begin{aligned} \mathscr{C}_{R,S,\tau} = \{ \mathcal{G} \times_1 \mathbf{U}^{(1)} \times_2 \mathbf{U}^{(2)} \times_3 \cdots \times_L \mathbf{U}^{(L)} : \mathcal{G} \in \mathbb{R}^{R_1 \times \cdots \times R_L}, \|\mathcal{G}\|_1 \leq \tau, \\ \mathbf{U}^{(l)} \in \mathbb{R}^{P_l \times R_l}, \|\mathbf{U}^{(l)}(:, r_l)\|_0 \leq S_l, \|\mathbf{U}^{(l)}(:, r_l)\|_1 \leq 1, \\ l = 1, \cdots, L, r_l = 1, \cdots, R_l \}, \end{aligned}$$

which imposes additional constraints over the $\ell_1$ norm of the core tensor $\mathcal{G}$ and $\ell_2$ norm of the columns of each factor matrix, where $R$ and $S$ stand for $[R_1, \cdots, R_L]$ and $[S_1, \cdots, S_L]$ respectively.

**Theorem 4.8.** (Ahmed *et al.*, 2019) For some fixed $\alpha \in (0, 1)$, suppose predictor $\mathcal{X}$ satisfies the RIP condition with isometry constant $\sigma_{2R,S,2\tau} \leq \frac{\alpha}{4+\alpha}$, the STPG algorithm with step-size $\eta = 1/(1 + \sigma_{2R,S,2\tau})$ can achieve a feasible solution $\mathcal{B}$ for Eq. (4.22) such that

$$\|\mathcal{B} - \mathcal{B}^*\|_{\mathrm{F}}^2 \leq \frac{2}{1 - \sigma_{2R,S,2\tau}} (1 + C + \frac{b}{1 - \alpha}) \|\mathbf{e}\|_{\mathrm{F}}^2 \tag{4.54}$$

in at least $\frac{1}{\log(1/\alpha)} \log(\frac{\|\mathbf{y} - \langle \mathcal{X}, \mathcal{B}^0 \rangle_L\|_{\mathrm{F}}^2}{\|C\mathbf{e}\|_{\mathrm{F}}^2})$ iterations for a constant $C > 0$, where $b = \frac{1+3\sigma_{2R,S,2\tau}}{1-\sigma_{2R,S,2\tau}}$, $\mathcal{B}^0$ and $\mathcal{B}^*$ are the initialization tensor and the optimal solution when tacking Eq. (4.22) using STPG algorithm respectively.

### Sparse tensor regression based on stagewise algorithms

Based on the same regression model, a new method is proposed in He *et al.* (2018) based on the sequential extraction method (Phan *et al.*, 2015; Mishra *et al.*, 2017), namely fast stagewise unit-Rank tensor factorization (SURF). Rather than minimizing the rank of $\mathcal{B}$ or factorizing $\mathcal{B}$ with a given CP rank, SURF sequentially seeks the rank-1 component of $\mathcal{B}$ through solving the following optimization problem

$$\begin{aligned} &\min_{\mathcal{B}_r} \frac{1}{N} \sum_{n=1}^{N} \left(y_{n,r} - \langle \mathcal{X}_n, \mathcal{B}_r \rangle\right)^2 + \lambda_r \|\mathcal{B}_r\|_1 + \alpha \|\mathcal{B}_r\|_\text{F}^2 \\ &\text{s.t.} \quad \text{rank}_\text{CP}\left(\mathcal{B}_r\right) \leq 1, \end{aligned} \tag{4.55}$$

where $\mathcal{B}_r$ is the $r$-th rank-1 component of $\mathcal{B}$, $y_{n,r}$ is the current residue of response with

$$y_{n,r} := \begin{cases} y_n, & \text{if } r = 1, \\ y_{n,r-1} - \langle \mathcal{X}_n, \mathcal{B}_{r-1} \rangle, & \text{otherwise} . \end{cases}$$

Introducing the CP decomposition $\mathcal{B}_r = \lambda_r \mathbf{u}_r^{(1)} \circ \cdots \circ \mathbf{u}_r^{(L)}$, (4.55) can be reformulated as

$$\begin{aligned} &\min_{\sigma_r, \mathbf{u}_r^{(l)}} \frac{1}{N} \|\mathbf{y}_r - \lambda_r \mathcal{X} \times_1 \mathbf{u}_r^{(1)} \times_2 \cdots \times_L \mathbf{u}_r^{(L)}\|_2^2 \\ &\qquad + \alpha \lambda_r \prod_{l=1}^{L} \|\mathbf{u}_r^{(l)}\|_1 + \beta \lambda_r^2 \prod_{l=1}^{L} \|\mathbf{u}_r^{(l)}\|_2^2 \\ &\text{s.t.} \quad \lambda_r \geq 0, \|\mathbf{u}_r^{(l)}\|_1 = 1, l = 1, \cdots, L, \end{aligned} \tag{4.56}$$

which can be divided into several traditional elastic net penalized regression problems based on the alternating minimization. Then the classical stagewise regression algorithm can be employed to solve each subproblem. Compared with Lasso, when determining the features to be selected at each iteration, SURF not only needs to select the mode, but also the index under each mode. This is the main difference between SURF and stagewise Lasso. In fact, this method can be seen as an extension of stagewise Lasso to the tensor field or an application of tensor deflation (Phan *et al.*, 2015) in regression tasks. The solution paths obtained by SURF have been proven to approach the solution paths of corresponding regularized regression. Numerical experiments show that SURF is fast and effective in the case of large datasets or low rank coefficient arrays, but the convergence rate is slow.

### Sparse multi-response tensor regression (SMRTR)

Most studies first model disease diagnosis tasks as classification problems with a binary output that indicates whether the subject is a patient or a normal control. However, it is difficult to distinguish between the multiple states of disease through classification problems using a binary output. Therefore, some works consider adopting image scans and some clinical scores to train the model, in order to obtain richer information for further diagnosis.

Specifically, in Li *et al.* (2016), the authors associate the MRI images with multiple clinical outcomes jointly and apply it in an Alzheimer's disease study. Given $N$ observations $\{(\mathcal{X}_1, \mathbf{y}_1), \cdots, (\mathcal{X}_N, \mathbf{y}_N)\}$, the regression model can be expressed as

$$\mathbf{y}_n = \begin{bmatrix} \langle \mathcal{B}_1, \mathcal{X}_n \rangle \\ \vdots \\ \langle \mathcal{B}_J, \mathcal{X}_n \rangle \end{bmatrix} + \mathbf{e}_n = \begin{bmatrix} \left\langle \sum_{r=1}^R \mathbf{u}_{1r}^{(1)} \circ \cdots \circ \mathbf{u}_{1r}^{(L)}, \mathcal{X}_n \right\rangle \\ \vdots \\ \left\langle \sum_{r=1}^R \mathbf{u}_{Jr}^{(1)} \circ \cdots \circ \mathbf{u}_{Jr}^{(L)}, \mathcal{X}_n \right\rangle \end{bmatrix} + \mathbf{e}_n, \tag{4.57}$$

where $\mathcal{X}_n \in \mathbb{R}^{P_1 \times \cdots \times P_L}$ is $n$-th sample of the predictor, $\mathbf{y}_n \in \mathbb{R}^J$ stands for a vector of $J$ responses, $\mathbf{u}_{jr}^{(l)} \in \mathbb{R}^{P_l}$, $l = 1, \cdots, L$, $j = 1, \cdots, J$. It can be observed that the model for each response of (4.57) is just the same as the model of CP-GLTR and hrTRR.

If estimating the parameters $\{(\mathbf{u}_{1r}^{(1)}, \cdots, \mathbf{u}_{1r}^{(L)}), \cdots, (\mathbf{u}_{Jr}^{(1)}, \cdots, \mathbf{u}_{Jr}^{(L)})\}$ without any other constraints, the model estimated by SMRTR will have made no difference to solving these tasks separately by previous methods. So considering how to exploit the correlations in these responses is at the heart of this work. To tackle this issue and further reduce the number of model parameters, a group sparsity penalty term is constructed as follows

$$\mathfrak{J}(\mathcal{B}_1, \cdots, \mathcal{B}_J) = \lambda \sum_{l=1}^{L} \sum_{r=1}^{R} \sum_{k=1}^{P_l} \left( \sum_{j=1}^{J} (\mathbf{u}_{jrk}^{(l)})^2 \right)^{1/2}.$$

This penalty makes the parameters $\{\mathbf{u}_{jrk}^{(l)}\}_{j=1}^{J}$ as a group, which encourages that coefficients corresponding to the same region of the image scans but across different responses will either be selected into the model or discarded simultaneously. In this way, the correlations between different outcomes are implied. Specifically, we can consider that for a specific disease, although there may be many assessment scores, they should only be affected by some subregions of the image scan, and the relevant regions for these responses should be approximately the same.

#### Sparse tensor response regression

As we illustrated before, besides disease diagnosis or clinical assessments based on medical images, there is another type of study called tensor response regression, which attracts a lot of attention. In fact, these two studies are based on the same subjects but with different focuses. The former one aims to predict the clinical diagnosis results from the medical images, while the latter one focuses on analyzing the change of medical images when the disease state or other variables vary.

In fact, the simplest solution for this task is to regress each value in the tensor response respectively (Penny *et al.*, 2011). But the spatial or temporal correlatedness in the tensor response is directly ignored, making the corresponding methods difficult to get accurate predictions. Then some studies make efforts on smoothing the tensor response by exploiting the correlation information provided by the neighbors around each voxel (Li *et al.*, 2011; Skup *et al.*, 2012; Li *et al.*, 2013b). For example, in Li *et al.* (2011), the regression parameters are estimated by iteratively increasing neighbors around each voxel and smoothing the observations within the neighbors with weights. However, only modeling the local similarity is not sufficient.

As illustrated in Section 4.3.1, Rabusseau and Kadri (2016) proposed an efficient closed-form solution HOLRR to analyze the association between the tensor predictor and tensor response jointly. Compared with the neighbors based methods, regressing the response jointly can make full use of the non-local information and better explore multidirectional relatedness. But there is no sparsity constraint on the regression coefficient array, making HOLRR inapplicable for feature selection. Under the same regression model $\mathcal{Y}_n = \mathcal{B} \times_{M+1} \mathbf{x}_n + \mathcal{E}_n$, Sun and Li (2017) derived a new regression model called the sparse tensor response model (STORE), which assumes a distinct low-rank structure on $\mathcal{B}$ and introduces an additional sparsity constraint. Specifically, letting $\mathcal{B}$ admit the CP decomposition structure $\mathcal{B} = \sum_r^R \lambda_r \mathbf{u}_r^{(1)} \circ \cdots \circ \mathbf{u}_r^{(M)} \circ \mathbf{u}_r^{(M+1)}$, the penalized optimization model can be expressed as

$$\min_{\lambda_r, \mathbf{u}_r^{(j)}} \frac{1}{N} \sum_{n=1}^{N} \|\mathcal{Y}_n - \sum_r \lambda_r ((\mathbf{u}_r^{(M+1)})^{\mathrm{T}} \mathbf{x}_n) \mathbf{u}_r^{(1)} \circ \cdots \circ \mathbf{u}_r^{(M)}\|_{\mathrm{F}}^2 \quad (4.58)$$
$$\text{s.t. } \|\mathbf{u}_r^{(j)}\|_2 = 1, \|\mathbf{u}_r^{(j)}\|_0 \leq s_j, j = 1, \cdots, M+1,$$

which can be solved based on the alternating updating algorithm, $\mathcal{Y}_n \in \mathbb{R}^{Q_1 \times \cdots \times Q_M}$, $\mathbf{x}_n \in \mathbb{R}^P$, $\mathcal{B} \in \mathbb{R}^{Q_1 \times \cdots \times Q_M \times P}$. The added sparsity constraint on latent factors $\|\mathbf{u}_r^{(j)}\|_0 \leq s_j$ makes it possible to select the features affected by the response and that can be achieved by truncation. Meanwhile, a non-asymptotic error bound of the actual estimation derived by the proposed algorithm is provided under several assumptions, including the unique decomposition condition required to ensure the identifiability of tensor decomposition, error bound over the initialization parameters to keep it not far away from the true parameters, and an error assumption to ensure the observations are not too noisy. Besides, by introducing symmetric constraints on latent factors, the algorithm can also be applied to structural neuroimaging analysis with symmetric responses.

Following a more generalized sparsity principle, Li and Zhang (2017) proposed an envelop-based tensor response regression model (TETORE). Compared with common sparse constraints expressed by $\ell_0$ or $\ell_1$ regularization, it first seeks linear combinations of the response that are irrelevant to the regression and then focuses on the correlated parts in the subsequence regression modeling. Based on the same tensor response regression model, the envelop-based method assumes that there exists a series of subspaces $\mathscr{S}_m \in \mathbb{R}^{Q_m}$ satisfying

$$\begin{aligned} \mathcal{Y} \times_m \mathbf{Q}_m \mid \mathbf{x} \quad &\sim \quad \mathcal{Y} \times_m \mathbf{Q}_m, \\ \mathcal{Y} \times_m \mathbf{Q}_m \quad &\perp\!\!\!\perp \quad \mathcal{Y} \times_m \mathbf{P}_m \mid \mathbf{x}, \end{aligned} \quad (4.59)$$

where $\mathbf{P}_m \in \mathbb{R}^{Q_m \times Q_m}$ is the projection matrix onto $\mathscr{S}_m$, $\mathbf{Q}_m = \mathbf{I}_{r_m} - \mathbf{P}_m \in \mathbb{R}^{Q_m \times Q_m}$ is the projection onto the complement space $\mathscr{S}_m^{\perp}$, $m = 1, \cdots, M$. $\sim$ and $\perp\!\!\!\perp$ means identically distributed and statistical independence respectively.

Specifically, the first condition essentially means that the projection of $\mathcal{Y}$ on the $\{\mathbf{Q}_1, \cdots, \mathbf{Q}_M\}$ with known $\mathbf{x}$, namely $\mathcal{Y} \times_m \mathbf{Q}_m \mid \mathbf{x}$, is identically distributed as $\mathcal{Y} \times_m \mathbf{Q}_m$. In other words, the projection of $\mathcal{Y}$ on the $\{\mathbf{Q}_1, \cdots, \mathbf{Q}_M\}$ does not rely on the change of $\mathbf{x}$. The second condition illustrates that $\mathcal{Y} \times_m \mathbf{Q}_m$ also does not affect the other part of the response $\mathcal{Y} \times_m \mathbf{P}_m$. So (4.59) implies that only the projection of response $\mathcal{Y}$ on $\{\mathbf{P}_1, \cdots, \mathbf{P}_M\}$, $\mathcal{Y} \times_m \mathbf{P}_m$, varies with the predictor $\mathbf{x}$ and matters in the regression procedure.

**Proposition 4.1.** (Li and Zhang, 2017) For regression model $\mathcal{Y}_n = \mathcal{B} \times_{M+1} \mathbf{x}_n + \mathcal{E}$, the assumption in Eq. (4.59) is true if and only if

$$\mathcal{B} = [\![\Theta; \Gamma_1, \cdots, \Gamma_M, \mathbf{I}]\!] \tag{4.60}$$

$$\Sigma_m = \Gamma_m \Omega_m \Gamma_m^{\mathrm{T}} + \Gamma_{0m} \Omega_{0m} \Gamma_{0m}^{\mathrm{T}}, \tag{4.61}$$

where $\Theta \in \mathbb{R}^{R_1 \times \cdots \times R_M \times P}$, $m = 1, \cdots, M$, $\Gamma_m \in \mathbb{R}^{Q_m \times R_m}$ denotes a basis for $\mathscr{S}_m$, $\Gamma_{0m} \in \mathbb{R}^{Q_m \times (Q_m - R_m)}$ denotes the complement basis for $\mathscr{S}_m$, $\mathbf{P}_m = \Gamma_m \Gamma_m^{\mathrm{T}}$, $\mathbf{Q}_m = \Gamma_{0m} \Gamma_{0m}^{\mathrm{T}}$, $\Omega_k \in \mathbb{R}^{R_m \times R_m}$ and $\Omega_{0k} \in \mathbb{R}^{(Q_m - R_m) \times (Q_m - R_m)}$ are two symmetric postive definite matrices.

Given the result of Proposition 4.1, the tensor response regression model can be rewritten as

$$[\![\mathcal{Y}_n; \Gamma_1^{\mathrm{T}}, \cdots, \Gamma_M^{\mathrm{T}}]\!] = \Theta \times_{M+1} \mathbf{x}_n + [\![\mathcal{E}_n; \Gamma_1^{\mathrm{T}}, \cdots, \Gamma_M^{\mathrm{T}}]\!], \tag{4.62}$$

which implies that only parts of linear combinations of output variables, namely $[\![\mathcal{Y}_n; \Gamma_1^{\mathrm{T}}, \cdots, \Gamma_M^{\mathrm{T}}]\!]$ , are relevant to the regression analysis. In other words, $[\![\mathcal{Y}_n; \Gamma_{01}^{\mathrm{T}}, \cdots, \Gamma_{0M}^{\mathrm{T}}]\!] = [\![\mathcal{E}_n; \Gamma_{01}^{\mathrm{T}}, \cdots, \Gamma_{0M}^{\mathrm{T}}]\!]$, which has no dependency on the predictor $\mathbf{x}_n$. Therefore, the regression modeling can now focus on the $[\![\mathcal{Y}_n; \Gamma_1^{\mathrm{T}}, \cdots, \Gamma_M^{\mathrm{T}}]\!]$. It is more flexible but could lose some interpretability. In addition, identifying the irrelevant information in the response first brings a great reduction in the number of model parameters. Two estimators are proposed to obtain the required parameters $\{\Theta, \Gamma_m, \Omega_m, \Omega_{0m}, m = 1, \cdots, M\}$, one iterative estimator and a one-step estimator. The theoretical analysis in Li and Zhang (2017) shows that both the iterative and one-step envelop methods are consistent and the iterative envelop estimator achieves a smaller asymptotic variance than OLS. Meanwhile, the asymptotic normality of the envelope estimator and OLS estimator when the envelop basis is unknown is also discussed and compared.

Moreover, Raskutti and Yuan (2015) proposed a unified framework for sparse tensor regression models to tackle high dimensional tensor regression tasks when both or either of the response and the predictor are tensors. In the case of tensors, the considered standard convex regularization framework is as follows

$$\min_{\mathcal{B}} \frac{1}{2N} \sum_{n=1}^{N} \|\mathcal{Y}_n - \langle \mathcal{B}, \mathcal{X}_n \rangle_L\|_{\mathrm{F}}^2 + \lambda \mathfrak{J}(\mathcal{B}), \tag{4.63}$$

where $\mathcal{B} \in \mathbb{R}^{P_1 \times \cdots \times P_L \times Q_1 \times \cdots \times Q_M}$, $\mathfrak{J}(\mathcal{B})$ is a norm regularizer, and $\lambda > 0$ is a tuning parameter.

Based on this unified framework, this paper first discusses various forms of regularizers and the corresponding solutions, especially the sparsity constraints. Meanwhile, a general risk bound is obtained and proved under the assumption of Gaussian error while the regularizers must be convex and weakly decomposable. Extended analysis and experimental verification on several popular regression problems are also given, such as multiple response regression, multivariate sparse auto-regressive models and pairwise interaction models.

### Optimal rank tensor ridge regression (orTRR)

Besides the entry-wise sparsity constraints, component-wise sparsity constraints are also exploited to achieve rank estimation (Guo *et al.*, 2012). It achieves rank selection during the training process and leads to the optimal rank tensor ridge regression. Specifically, based on the CP decomposition, the original tensor can be seen as a sum of $R$ rank-1 tensors. Therefore, enforcing these rank-1 components to be group sparsity as follows

$$\mathfrak{J}(\mathcal{B}) = \sum_{r=1}^{R} \left( \sum_{l=1}^{L} \left\| \mathbf{u}_r^{(l)} \right\|_2^2 \right)^{\frac{1}{2}}, \tag{4.64}$$

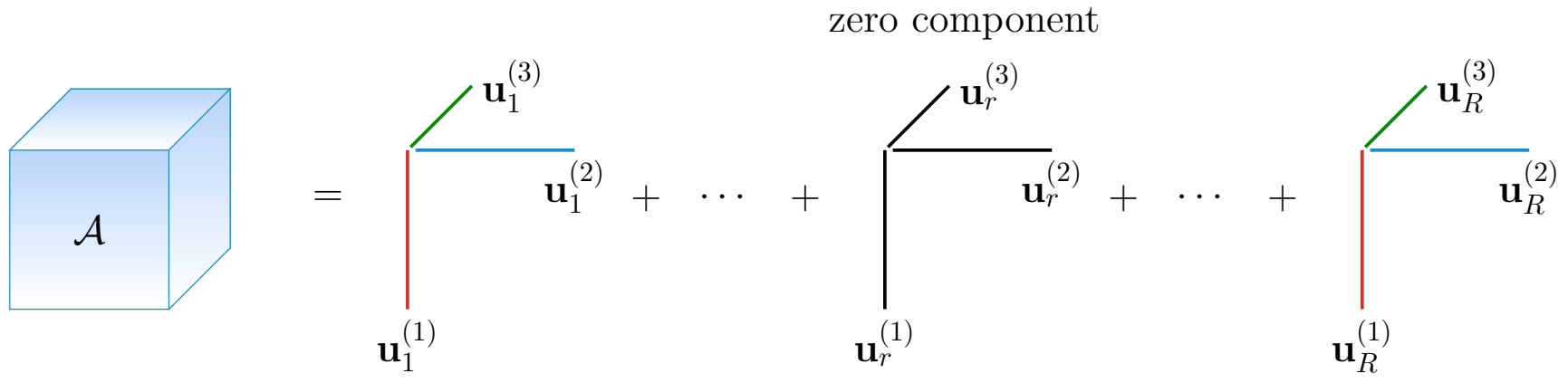

**Figure 4.4:** The diagram description of the group sparse rank decomposition.

the CP rank will be reduced if the algorithm detects unnecessary components. As shown in Fig. 4.4, each rank-1 component is treated as a group, and the sparsity is imposed over these groups. When there is one component enforced to be all zeros, the CP rank will be reduced by 1 and the classical iteratively reweighted least squares algorithm (Daubechies *et al.*, 2010) is employed to make the constrained problem efficiently tractable. Moreover, an extension with tensor ring decomposition is proposed in Liu *et al.* (2020a).

In fact, the equivalence between sparsity constrained matrix factorization problems and the rank minimization problems has been analyzed and discussed (Fan *et al.*, 2019; Zha *et al.*, 2020). Several tensor rank estimation methods have been proposed based on the consideration of sparsity constraint over the latent factors, such as sparse Tucker decomposition for rank estimation by enforcing the core tensor or the factor matrices to be sparse (Yokota and Cichocki, 2014), and the Bayesian based learning with sparsity constraint on latent factors (Zhao *et al.*, 2015). However, there are relatively few historical studies on rank estimation for tensor regression tasks currently, which may be a promising direction for future research.

### 4.3.3 Other variants

In addition to the low-dimensional structures as low-rankness and sparsity, some other constraints are also employed in specific applications to simulate complex regression systems, such as smoothness, nonnegativity, and orthogonality.

For example, a sparse Tucker regression model with orthonormality constraints over the factor matrices is proposed in Ou-Yang *et al.* (2020). Mathematically, the optimization problem is given as

$$\min_{\mathcal{G},\mathbf{U}^{(1)},\cdots,\mathbf{U}^{(L)}} \quad \frac{1}{2}\sum_{n}(y_n - \langle \mathcal{X}_n, [\![\mathcal{G}; \mathbf{U}^{(1)}, \cdots, \mathbf{U}^{(L)}]\!]\rangle)^2 \tag{4.65}$$

$$+\lambda \sum_{p_1,\cdots,p_L} |[\![\mathcal{G}; \mathbf{U}^{(1)}, \cdots, \mathbf{U}^{(L)}]\!]_{p_1,\cdots,p_L}|^{\gamma}$$

$$\text{s.t. } {\mathbf{U}^{(l)}}^{\mathrm{T}}\mathbf{U}^{(l)} = \mathbf{I}_{R_l}, l = 1, \cdots, L, \tag{4.66}$$

which is solved using the classical ADMM. In this case, the sparsity constraint is imposed directly over the coefficient tensor, $\gamma \in (0, 1)$.

In order to discover effective brain connectivity Valdes-Sosa *et al.* (2011) and Karahan *et al.* (2015) treated the classical Granger causality as a tensor autoregression model based on CP decomposition (GC-CP) and enhanced the model identifiability by taking account of sparsity, smoothness, nonnegativity and orthogonality over the latent factors of the connectivity tensor.

Specifically, the Granger causality (Vinck *et al.*, 2015; Granger, 1988) can be viewed as a multivariate autoregressive model (MAR) as follows

$$\mathbf{y}_t = \sum_{q=1}^{I_{\text{lag}}} \mathbf{B}_q \mathbf{y}_{t-q} + \varepsilon_t, \tag{4.67}$$

where $\mathbf{y}_t \in \mathbb{R}^{I_{Cx}}$ is the blood oxygen level dependent signal at time point $t$, $I_{\text{lag}}$ is the number of the past lags included in the model, $\varepsilon_t$ is the noise, $\{\mathbf{B}_1, \cdots, \mathbf{B}_{I_{\text{lag}}}\}$ represent the effect of signals at previous $I_{\text{lag}}$ time points on that of current time point. It is based on the assumption that a cause must appear before and reduce the uncertainty of its consequence.

Collecting all the time samples $t = I_{\text{lag}}, \cdots, I_{T\sigma} + I_{\text{lag}}$ and past lags $q = 1, \cdots, I_{\text{lag}}$, Karahan *et al.* (2015) represents the Granger casualty (inference) connectivity as the following tensor autoregression model

$$\mathbf{Y} = \langle \mathcal{B}, \mathcal{Y} \rangle_2 + \mathbf{E}, \tag{4.68}$$

where

$$\begin{aligned}
&\mathcal{B} \in \mathbb{R}^{I_{Cx} \times I_{Cx} \times I_{\text{lag}}}, \quad \mathcal{B}(q,:,:) = \mathbf{B}_q, \\
&\mathcal{Y} \in \mathbb{R}^{I_{\text{lag}} \times I_{Cx} \times I_{T\sigma}}, \quad \mathcal{Y}(q,:,:) = \mathbf{Y}_{t-q}, \\
&\mathbf{Y}_{t-q} = [\mathbf{y}_{I_{\text{lag}}+1-q}, \cdots, \mathbf{y}_{i_{T\sigma}+I_{\text{lag}}-q}]^{\mathrm{T}} \in \mathbb{R}^{I_{Cx} \times I_{T\sigma}}.
\end{aligned}$$

Under CP decomposition, the connectivity tensor $\mathcal{B}$ can be decomposed into spatial signatures and temporal signature for the lag, as shown in the following optimization problem

$$\begin{aligned}
\min_{\mathbf{M}_s, \mathbf{M}_r, \mathbf{T}} \{ & \frac{1}{2} \|\mathbf{Y} - \langle \mathcal{B}, \mathcal{Y} \rangle_2\|_{\mathrm{F}}^2 + \lambda_1 \|\mathbf{M}_s\|_1 + \frac{1}{2}\lambda_2 \|\mathbf{L}\mathbf{M}_s\|_{\mathrm{F}}^2 \\
& + \lambda_3 \|\mathbf{M}_r\|_1 + \frac{1}{2}\lambda_4 \|\mathbf{L}\mathbf{M}_r\|_{\mathrm{F}}^2 + \lambda_5 \|\mathbf{T}\|_1 + \frac{1}{2}\lambda_6 \|\mathbf{L}\mathbf{T}\|_{\mathrm{F}}^2 \} \\
\text{s.t. } & \mathcal{B} = [\![\mathbf{M}_s, \mathbf{M}_r, \mathbf{T}]\!], \quad \mathbf{M}_s \geq 0, \quad \mathbf{M}_r \geq 0, \\
& \mathbf{M}_s^{\mathrm{T}} \mathbf{M}_s = \mathbf{I}, \quad \mathbf{M}_r^{\mathrm{T}} \mathbf{M}_r = \mathbf{I},
\end{aligned} \tag{4.69}$$

where the $\mathbf{M}_s \in \mathbb{R}^{I_{Cx} \times R}$ and $\mathbf{M}_r \in \mathbb{R}^{I_{Cx} \times R}$ represent the spatial signature for the sender and receiver nodes, respectively, $\mathbf{T} \in \mathbb{R}^{I_{T\sigma} \times R}$ is the temporal signature, $\mathbf{L}$ is a Laplacian matrix.

Considering the specific conception of each factor, different constraints are imposed to enhance the model identifiability. Specifically, the spatial signatures $\mathbf{M}_s$ and $\mathbf{M}_r$ are enforced to be sparse and smooth, while the nonnegativity and orthogonality are taken into account to make there exist only one nonnegative element in each row of the spatial factors. In this way, there will be a subset of brain regions included with respect to the specific task and the connected spatial regions will be nonoverlapping local smooth patches. With the classical ADMM algorithm, all the constraints can be integrated into the objective function and tackled with respect to each variable.

Besides, to control local consistency, a spatial Laplacian regularization is imposed on the predicted climate values with similar location (Bahadori *et al.*, 2014) as follows

$$\mathfrak{J}(\mathcal{Y}) = \sum_{m=1}^{M} \mathrm{Tr}(\mathcal{Y}(:,:,m)\mathbf{L}\mathcal{Y}(:,:,m)), \tag{4.70}$$

where $\mathcal{Y}$ is the predicted value, $\mathbf{L}$ is the Laplacian matrix which represents the spatial similarity for $\mathcal{Y}$. Compared with the ordinary Frobenius norm, this Laplacian regularization term can make use of the spatial characteristics of the data itself and make the predicted values of adjacent positions close.

Moreover, another special form of the penalty term is employed to the tensor response regression model $\mathcal{Y} = \mathcal{B} \times_3 \mathbf{X} + \mathcal{E} = \mathcal{G} \times_1 \mathbf{U}^{(1)} \times_2 \mathbf{U}^{(2)} \times_3 \mathbf{X} + \mathcal{E}$ for smoothness in Yan *et al.* (2019).

$$\mathfrak{J}(\mathbf{g}) = \mathbf{g}^{\mathrm{T}}(\mathbf{X}^{\mathrm{T}}\Sigma_3^{-1}\mathbf{X}) \otimes (\lambda \mathbf{P}_2 \otimes \mathbf{U}^{(1)^{\mathrm{T}}}\Sigma_1^{-1}\mathbf{U}^{(1)} + \lambda \mathbf{U}^{(2)^{\mathrm{T}}}\Sigma_2^{-1}\mathbf{U}^{(2)} \otimes \mathbf{P}_1 + \lambda^2 \mathbf{P}_2 \otimes \mathbf{P}_1)\mathbf{g}, \tag{4.71}$$

where $\mathbf{g} = \mathrm{vec}(\mathcal{G})$, $\Sigma_1$ and $\Sigma_2$ denote the spatial correlation of the noise, $\Sigma_3$ represents the between-sample variation, $\mathbf{P}_m$ is the penalization matrix which enforces the mode-$m$ of the original tensor to be smooth with the consideration of the spatial structure information. Compared with directly enforcing the original coefficient tensor $\mathcal{B}$ to be smooth, (4.71) constrains the latent factors and thus the problem can be efficiently computed along each mode separately.

Furthermore, using the tensor representation, Wang *et al.* (2020) formulated the multitask survival analysis as the following problem

$$\min_{\mathcal{B}} \sum_{m=1}^{M}\sum_{t=1}^{T}(\frac{1}{2}\|\mathcal{S}(:,t,m) \odot (\mathcal{Y}(:,t,m) - \mathcal{X}(:,:,m)\mathcal{B}(:,t,m))\|_2^2) + \frac{1}{2}\alpha\|\mathcal{B}\|_{\mathrm{F}}^2 + \beta\|\mathcal{B}\|_{\mathrm{F},1} + \frac{1}{2}\gamma\|\mathcal{B}\|_{\mathrm{ts}}, \tag{4.72}$$

where $\mathcal{S}$ is used to handle the censored information (Reddy *et al.*, 2015), it indicates whether a sample contributes to the training process. $\alpha$, $\beta$, and $\gamma$ are weighting factors. Wang *et al.* (2020) incorporated the Frobenius norm for overfitting and two regularization terms for shared features and temporal smoothness as follows

$$\|\mathcal{B}\|_{\mathrm{F},1} = \sum_{p=1}^{P}\sqrt{\|\mathcal{B}(p,:,:)\|_{\mathrm{F}}^2}, \tag{4.73}$$

$$\|\mathcal{B}\|_{\mathrm{ts}} = \sum_{m=1}^{M}\sum_{t}^{T_1}\|\mathcal{B}(:,t+1,m) - \mathcal{B}(:,t,m)\|_2^2, \tag{4.74}$$

where $\|\mathcal{B}\|_{\mathrm{F},1}$ constraints that the features selected into the model are shared across all the $M$ tasks and $T$ timepoints, namely the inter-task

correlations. $\|\mathcal{B}\|_{\text{ts}}$ controls the intra-task temporal smoothness, which means the estimated parameters at adjacent time points are enforced to be as similar as possible.

As we see in model (4.72), instead of using the low-rankness constraints in almost all the previous models, Wang *et al.* (2020) identified the most important and shared parameters using the group sparsity constraints, which confirms that the group sparsity constrained problems can be transformed to rank minimization problems to some extent as we mentioned earlier. But here the shared parameters are considered jointly across different tasks and timestamps, and the difference between these two modes is ignored. From the perspective of rank minimization, only the rank along the first mode is minimized and correlations along other modes are not considered.

There are two ways to enforce local consistency, including $\ell_2$ norm based ones and $\ell_1$ norm based ones. Currently, only the $\ell_2$ norm based one is investigated in tensor regression, as presented before. However, $\ell_1$ norm based ones, like tensor total variation (Liu *et al.*, 2017b), can not only penalize the similarity of adjacent locations, but also allow for sharp transitions (Mirzaei *et al.*, 2019). But from the perspective of computational complexity, the non-differentiability of the $\ell_1$ norm makes it more complicated to deal with than the $\ell_2$ norm.

## 4.4 Bayesian tensor regression

As illustrated before, in order to balance the modeling error and model complexity, different low-dimensional structural assumptions are imposed on the coefficient tensor, such as low-rankness, sparsity, smoothness and orthogonality. But a clear disadvantage of such approaches is their heavy dependence on the selection of tuning parameters. For different regression tasks, users need to manually adjust some hyperparameters through experience or cross-validation. This makes the generalization and application of such approaches greatly limited. In addition, the penalized optimization is unsatisfactory for predictive uncertainties.

To tackle these issues, many references attempt to adjust the model complexity for specific tasks through Bayesian learning methods. For example, Suzuki (2015) proposed a Bayesian learning approach for low rank tensor regression problems and analyzed the convergence rate of the Bayesian tensor estimator. We refer to this algorithm as the Bayesian tensor regression model (BTRM). Under the simple linear tensor regression model $y_n = \langle \mathcal{B}, \mathcal{X}_n \rangle + \varepsilon_n$, BTRM adopts the Gaussian prior in the latent factors of the tensor coefficient under CP decomposition as follows

$$\pi(\mathbf{U}^{(1)}, \cdots, \mathbf{U}^{(L)} \mid R) \propto \exp\left\{-\frac{R}{2\sigma_p^2}\sum_{l=1}^{L} \mathrm{Tr}\left[\mathbf{U}^{(l)^{\mathrm{T}}}\mathbf{U}^{(l)}\right]\right\}, \tag{4.75}$$

where $\mathcal{B} = \sum_{r=1}^{R} \mathbf{u}_r^{(1)} \circ \cdots \circ \mathbf{u}_r^{(L)}$, $\mathbf{u}_r^{(l)}$ is $r$-th column of $\mathbf{U}^{(l)}$, $\sigma_p > 0$. And the prior distribution for rank $1 \leq R \leq R_{\max}$ is assumed as follows

$$\pi(R) = \frac{1}{N_\xi}\xi^R (P_1 + \cdots + P_L), \tag{4.76}$$

where $0 < \xi < 1$ is a positive real number, $R_{\max}$ is a larger number than the true rank of $\mathcal{B}$, $N_\xi$ is the normalizing constant.

Then the posterior distribution can be obtained under the assumption of Gaussian noise and can be computed by Markov Chain Monte Carlo method Gibbs sampling (Xiong *et al.*, 2010; Rai *et al.*, 2014). The Gaussian priors on the latent factors $\{\mathbf{U}^{(1)}, \cdots, \mathbf{U}^{(L)}\}$ and exponentially decaying prior on rank $R$ bring the model the ability to automatically adjust the rank with taking account of the predictive accuracy. In addition, a near-optimal convergence rate is achieved and proved without any strong convexity assumptions.

**Theorem 4.9.** (Suzuki, 2015) Denote the $\hat{\mathcal{B}}$ and $\mathcal{B}^*$ as the Bayes estimator and the true tensor, $R^*$ is the CP rank of $\mathcal{B}^*$, the (near) optimal convergence rate is given as

$$\|\hat{\mathcal{B}} - \mathcal{B}^*\|_N^2 = O\left(\frac{R^*(\sum_{l=1}^{L} P_l)\log(L\sqrt{N(\sum_{l=1}^{L} P_l)^L})}{N}\right), \tag{4.77}$$

where

$$\|\hat{\mathcal{B}} - \mathcal{B}^*\|_N^2 = \frac{1}{N}\sum_{n=1}^{N}\langle \mathcal{X}_n, \hat{\mathcal{B}} - \mathcal{B}^*\rangle^2 \tag{4.78}$$

is the in-sample predictive accuracy of the estimator $\hat{\mathcal{B}}$.

It can be seen that the convergence rate for the Bayes estimator relates to the actual degree of freedom and is rank adaptive. Moreover, the convergence rate of out-of-sample predictive accuracy is also analyzed. Let $p(\mathcal{X})$ represent the distribution of $\mathcal{X}$ that generates the observations $\{\mathcal{X}_n\}_{n=1}^N$, the out-of-sample predictive accuracy is defined as

$$\|\hat{\mathcal{B}} - \mathcal{B}^*\|^2_{\ell_2(p(\mathcal{X}))} = \mathrm{E}_{\mathcal{X}\sim p(\mathcal{X})}[\langle \mathcal{X}, \hat{\mathcal{B}} - \mathcal{B}^* \rangle^2]. \tag{4.79}$$

The minimax optimal learning rate for out-of-sample predictive accuracy is roughly given by

$$\inf_{\hat{\mathcal{B}}} \sup_{\mathcal{B}^* \in \mathbb{T}} \|\hat{\mathcal{B}} - \mathcal{B}^*\|^2_{\ell_2(p(\mathcal{X}))} \geq C \frac{R^*(\sum_{l=1}^L P_l)}{N}, \tag{4.80}$$

where $\hat{\mathcal{B}}$ is any estimator, $\mathbb{T}$ is a set of tensors, $C$ is a constant.

Focusing on a similar linear regression model $y_n = \alpha + \mathbf{z}_n^{\mathrm{T}}\gamma + \langle \mathcal{X}_n, \mathcal{B} \rangle + \varepsilon_n$, Guhaniyogi *et al.* (2017) introduced a novel class of multiway shrinkage priors for the coefficient tensor, called multiway Dirichlet generalized double Pareto (M-DGDP) prior. It imparts the shrinkage prior into three parts, resulting in a hierarchical margin-level prior as follows

$$\mathbf{u}_r^{(l)} \sim \mathcal{N}\left(0, (\phi_r \tau)\, \mathbf{W}_{lr}\right), \quad w_{lr,p_l} \sim \mathrm{Exp}\left(\lambda_{lr}^2/2\right), \quad \lambda_{lr} \sim \mathrm{Ga}\left(a_\lambda, b_\lambda\right), \tag{4.81}$$

where $\mathcal{B}$ also admits CP decomposition $\mathcal{B} = [\![\mathbf{U}^{(1)}, \cdots, \mathbf{U}^{(L)}]\!]$, $\mathbf{u}_r^{(l)}$ is $r$-th column of $\mathbf{U}^{(l)}$, $\mathbf{\Phi} = [\phi_1, \cdots, \phi_R] \sim \mathrm{Dirichlet}(\alpha_1, \cdots, \alpha_R)$ encourages the latent factors to be low-rank under the tensor decomposition, $\tau \sim \mathrm{Ga}(a_\tau, b_\tau)$ is a global scale which fits in each component by $\tau_r = \phi_r \tau$ for $r = 1, \cdots, R$, $\mathbf{W}_{lr} = \mathrm{diag}(w_{lr,1}, \cdots, w_{lr,P_l})$, $l = 1, \cdots, L$ are Margin-specific scale parameters of each component. A detailed expression of the statistical distributions mentioned in this review is referred to in Table 4.1.

Compared with Suzuki (2015), M-DGDP prior can not only achieve rank selection by the Dirichlet$(\alpha_1, \cdots, \alpha_R)$ prior on the $\mathbf{\Phi}$, but also consider shrinkage priors at global scale level and local scale level to

**Table 4.1:** A detailed expression of some statistical distributions. $\Gamma(\cdot)$ and $B(\cdot)$ stand for the gamma function and multinomial beta function, respectively.

| | | Probability distribution function |
|---|---|---|
| Continuous distributions | Normal Distribution $\mathcal{N}(\mu, \sigma)$ | $p(x) = \frac{1}{\sigma\sqrt{2\pi}} \exp(-(x-\mu)^2/(2\sigma^2))$ |
| | Exponential distribution $\text{Exp}(\lambda)$ | $p(x) = \lambda \exp(-\lambda x), x \in [0, \infty)$ |
| | Double exponential distribution $\text{DE}(\lambda)$ | $p(x) = \frac{1}{2\lambda} \exp(-\lvert x-\mu \rvert/\lambda)$ |
| | Gamma distribution $\text{Ga}(a, b)$ | $p(x) = \frac{x^{a-1}\exp(-x/b)}{\Gamma(a)b^a}, x \in [0, \infty)$ |
| | Inverse gamma distribution $\text{IG}(a, b)$ | $p(x) = \frac{x^{-(a+1)}\exp(-1/bx)}{\Gamma(a)b^a}, x \in [0, \infty)$ |
| Discrete distributions | Dirichlet distribution $\text{Dirichlet}(\alpha_1, \cdots, \alpha_K)$ | $p(\mathbf{x}) = \frac{1}{B(\alpha_1, \cdots, \alpha_K)} \prod_{k=1}^{K} x_k^{\alpha_k - 1}$ |

achieve more accurate estimation. Then the model can be expressed as

$$\begin{aligned} y|\gamma, \mathcal{B}, \sigma &\sim \mathcal{N}(\mathbf{z}^{\mathrm{T}}\boldsymbol{\gamma} + \langle \mathcal{X}, \mathcal{B} \rangle, \sigma^2), \\ \mathcal{B} &= \sum_{r=1}^{R} \mathcal{U}_r, \mathcal{U}_r = \mathbf{u}_r^{(1)} \circ \cdots \circ \mathbf{u}_r^{(L)}, \\ \sigma^2 &\sim \pi_\sigma, \boldsymbol{\gamma} \sim \pi_{\boldsymbol{\gamma}}, \mathbf{u}_r^{(l)} \sim \pi_{\mathbf{u}}, \end{aligned} \tag{4.82}$$

where the prior distribution over the noise variance $\pi_\sigma$ is a conjugate inverse-gamma prior, the regression coefficient $\boldsymbol{\gamma}$ is modeled by a normal distribution. Fig. 4.5 provides an intuitive illustration for this prior specification. An efficient Gibbs sampling scheme is employed to generate iterates for most promoters in (4.82) based on the proposed prior (4.81). Guhaniyogi *et al.* (2017) provided a strong posterior consistency for the Bayesian tensor regression model under a sufficient condition on the prior of tensor predictors.

Driven by the lack of study for time-varying networks in economics, Billio *et al.* (2018) extended Bayesian learning approaches to the dynamic tensor regression model, namely the Bayesian tensor dynamic model (BTDM), and employed the same hierarchical prior specification as in Guhaniyogi *et al.* (2017). Specifically, as a special example, the tensor autoregression model of lag-order 1 is considered as follows

$$\mathcal{Y}_t = \mathcal{B} \times_4 \text{vec}(\mathcal{Y}_{t-1}) + \mathcal{E}_t, \tag{4.83}$$

where $\mathcal{Y}_t \in \mathbb{R}^{I_1 \times I_2 \times I_3}$, $\mathcal{B} \in \mathbb{R}^{I_1 \times I_2 \times I_3 \times I_4}$ with $I_4 = I_1 I_2 I_3$ is the coefficient tensor. The hierarchical shrinkage prior enforced on the latent factors $\{\mathbf{u}_r^{(j)}, j = 1, \cdots, 4, r = 1, \cdots, R\}$ of $\mathcal{B}$ is the same as the prior distribution in (4.81).

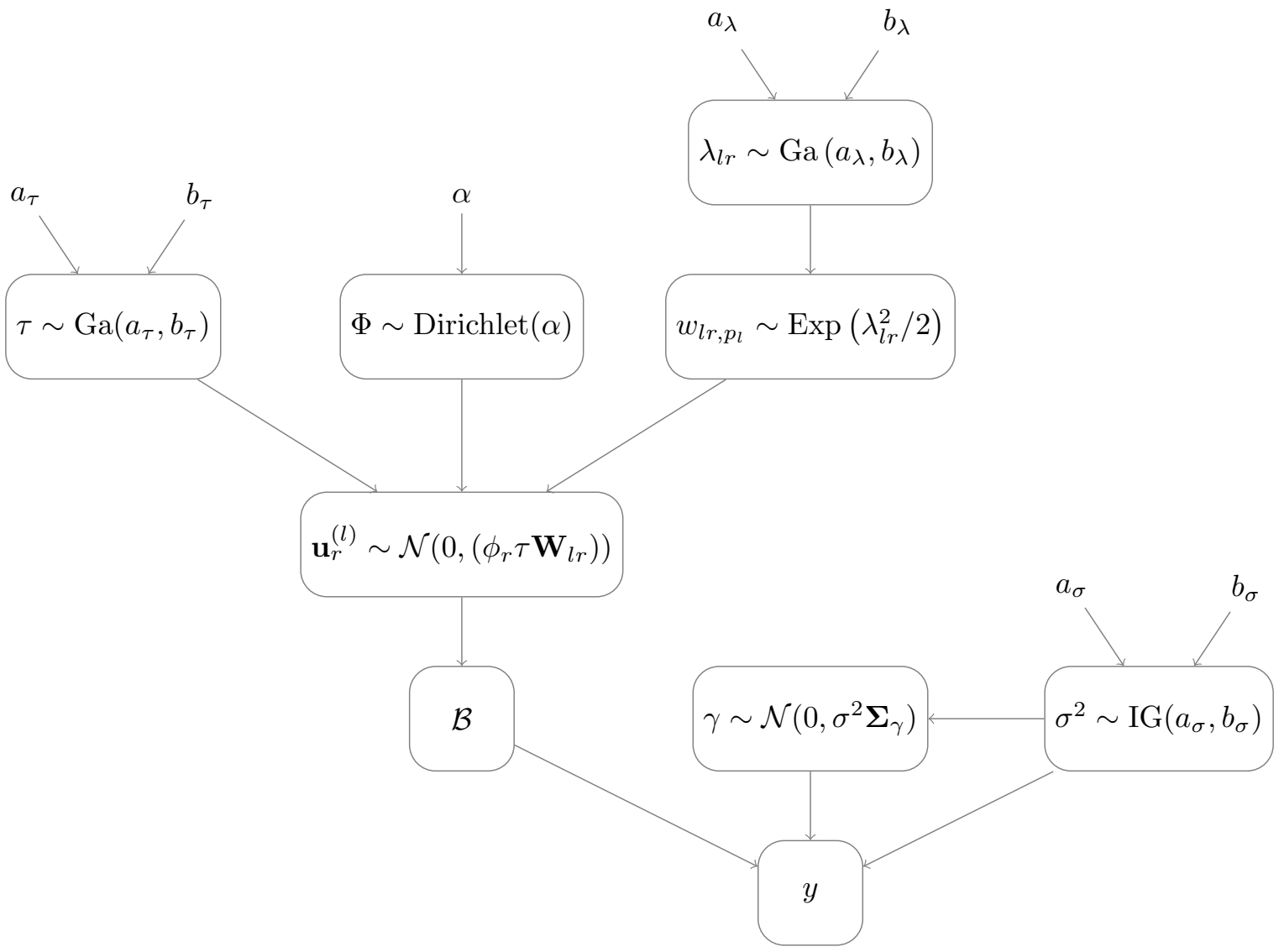

**Figure 4.5:** The diagram of the hierarchical shrinkage prior for Bayesian dynamic tensor regression (Guhaniyogi *et al.*, 2017).

Following the similar Bayesian learning approach, the soft tensor regression (Softer) (Papadogeorgou *et al.*, 2019) extends the classical CP decomposition into a soft version for the estimation of the coefficient tensor as follows

$$\mathcal{B} = \sum_{r=1}^{R} \mathcal{U}_r^{(1)} \circledast \cdots \circledast \mathcal{U}_r^{(L)}, \tag{4.84}$$

where $\mathcal{U}_r^{(l)}$ is in the same size as $\mathcal{B}$. The key idea for soft CP decomposition is to first transfer the high dimensional coefficient tensor $\mathcal{B}$ into even high dimensional latent factors $\mathcal{U}_r^{(l)}$ with no restricted assumption, and then introduce the sparsity on the latent factors through specifying prior distributions.

Specifically,

$$\mathcal{U}_r^{(l)}(p_1, \cdots, p_L) \sim \mathcal{N}(\gamma_{r,p_l}^{(l)}, \sigma_l^2 \zeta_r), p_l = 1, \cdots, P_l, l = 1, \cdots, L. \quad (4.85)$$

In this way, entries of the tensor $\mathcal{B}$ are assumed to be centered around a $\gamma$-based rank-$R$ CP factorization. Mathematically,

$$\mathrm{E}(\mathcal{B} \mid \gamma_r^{(l)}, \sigma_l^2, \zeta_r) = \sum_{r=1}^{R} \gamma_r^{(1)} \circ \cdots \circ \gamma_r^{(L)}, \quad (4.86)$$

where $\gamma_r^{(l)} = [\gamma_{r,1}^{(l)}, \cdots, \gamma_{r,P_l}^{(l)}]$. The prior distribution for $\gamma_r^{(l)}$ is the same as $\mathbf{u}_r^{(l)}$ in (4.81). From theoretical analysis and experimental verification, softening the CP decomposition can increase model flexibility and lead to more accurate estimation of regression coefficient tensors.

Moreover, a Bayesian mixed effect sparse tensor response regression model (BMESTRR) is proposed to simultaneously estimate the activated regions and connectivity patterns in task-based fMRI data analysis (Spencer *et al.*, 2019). In addition to the activation-only models identifying the activation of brain voxels due to a given stimulus or behavior (Sun and Li, 2017; Li and Zhang, 2017; Suzuki, 2015; Guhaniyogi *et al.*, 2017), exploring how the pattern of different spatially-adjacent groups of voxels in the brain connect with each other is also another area of interest. However, models incorporating brain activation and connectivity analysis are comparatively rare. Given observations $\{x_{n,t}, \mathcal{Y}_{n,g,t}\}$, the additive mixed effect model is as follows

$$\mathcal{Y}_{n,g,t} = \mathcal{B}_g x_{n,t} + d_{n,g} + \mathcal{E}_{n,g,t}, \quad (4.87)$$

where $\mathcal{Y}_{n,g,t} \in \mathbb{R}^{P_{1,g} \times \cdots \times P_{D,g}}$ is the observed fMRI data in brain region $g$ of the $n$-th subject at the $t$-th time point, $x_{n,t}$ is the activation related predictor, $n = 1, \cdots, N$, $t = 1, \cdots, T$, $g = 1, \cdots, G$, $\mathcal{E}_{n,g,t}$ is a bias.

The coefficient tensor $\mathcal{B}_g \in \mathbb{R}^{P_{1,g} \times \cdots \times P_{D,g}}$ represents the association between the predictor $x_{n,t}$ and the output $\mathcal{Y}_{n,g,t}$. In other words, whether the $\{p_{1,g}, \cdots, p_{D,g}\}$th element in $\mathcal{B}_g$ is non-zero determines whether the voxel is activated or not for a specific stimulus. To better assess brain activation, a similar prior distribution as in Fig. 4.5 is imposed on the latent factors of $\mathcal{B}_g$ under CP decomposition.

The term $d_{n,g}$ represents the random effects in $g$-th region for $n$-th subject. Investigating the conditional distributions $(d_{n,g}, d_{n,g'}) \mid \{d_{n,g''} :$

$g'' \neq g, g'\}$ can explore the connectivity patterns between G different regions. In Spencer *et al.* (2019), to favor most pairs $d_{n,g}$ and $d_{n,g'}$ conditional independent, $d_{n,g}$s are modeled jointly with a Gaussian graphical Lasso prior (Wang *et al.*, 2012) as follows

$$\begin{aligned}
&\mathbf{d}_n = [d_{n,1}, \cdots, d_{n,g}]^{\mathrm{T}} \sim \mathcal{N}(\mathbf{0}, \Sigma^{-1}),\\
&p(\boldsymbol{\sigma}|\zeta) = C^{-1} \prod_{k<l} [\mathrm{DE}\left(\sigma_{kl}|\zeta\right)] \prod_{k=1}^{G} \left[\mathrm{Exp}\left(\sigma_{kk}|\frac{\zeta}{2}\right)\right] \mathbf{1}_{\boldsymbol{\Sigma}\in\mathscr{P}^+}, \quad (4.88)\\
&\zeta \sim \mathrm{Ga}(a_\zeta, b_\zeta),
\end{aligned}$$

where $\boldsymbol{\sigma} = (\sigma_{k,l} : k \leq l)$ is a vector of upper triangular and diagonal entries of the precision matrix $\Sigma$, $C$ is a normalizing constant, $\mathrm{DE}(\cdot)$ represents the double exponential or Laplace density function, $\mathscr{P}^+$ is the space of symmetric positive definite matrices. In this way, only a few pairs of regions are assumed to be connected with the specific stimulus. As a result, the proposed model can accurately estimate the brain activation and connectivity patterns between $G$ different regions jointly with no need to tune the parameters in advance or set by experience.

Bayesian tensor regression is also used for change detection and analysis (Idé, 2019). The variational inference strategy is firstly employed for expectation maximization instead of the Gibbs sampling. It derives the expected value and variance of latent factors through variational inference. The resulting algorithm, namely Bayesian ALS, can be seen as an extension of the classical ALS algorithm. Although the derivation process of variational reasoning is more complicated, its calculation process is much faster than Gibbs sampling.

## 4.5 Quantile tensor regression

Minimizing the squared error loss pursues the expected mean value of the response and is not robust to outliers. In order to address this issue, some other cost functions are proposed to give different degrees of penalty for different residual values, such as Huber loss, absolute value loss and pinball loss. The least absolute derivations regression corresponds to the median value of the predictive distribution, and quantile regression goes further and fits several regression curves with respect to different quantiles, which is capable of giving a more comprehensive illustration to the predictive distribution. For a random variable $Y \in \mathbb{R}$, the $\tau$-th quantile of $Y$ is denoted as

$$Q_\tau = \inf\{y : \mathfrak{F}(y) \geq \tau\}, \tag{4.89}$$

where $\mathfrak{F}(y) = p(Y \leq y)$ is the cumulative distribution function of $Y$, $\tau \in (0, 1)$. In fact, using the quantile to describe the distribution benefits a lot more than the mean value, especially when the predictive distribution is skewed. The $\tau$-th quantile divides the data into $\tau$ below and $1 - \tau$ above this quantile. When $\tau = 0.5$, the corresponding quantile is the median value.

Lu *et al.* (2020) extends quantile regression into the tensor field and provides estimation methods and statistical analysis. Specifically, the $\tau$-th conditional quantile of response $y_n$ given tensor predictor $\mathcal{X}_n$ is used as the outcome of the linear regression model instead of the response $y_n$ itself.

$$Q_\tau(y_n|\mathcal{X}_n) = \epsilon + \langle \mathcal{B}, \mathcal{X}_n \rangle, \tag{4.90}$$

where $\mathcal{B}$ is the coefficient tensor, which admits the Tucker decomposition as

$$\mathcal{B} = \mathcal{G} \times_1 \mathbf{U}^{(1)} \times_2 \cdots \times_L \mathbf{U}^{(L)}, \tag{4.91}$$

which reduces the number of parameters in Eq. (4.90) from $O(\prod_{l=1}^L P_l)$ to $O(R^L + \sum_{l=1}^L P_l R)$ with all the Tucker ranks equal to $R$.

To obtain the desired parameters $\mathcal{G}, \mathbf{U}^{(1)}, \cdots, \mathbf{U}^{(L)}$ in this quantile tensor regression model, the pinball loss

$$\rho_\tau(r) = \begin{cases} \tau r, & \text{if } r \geq 0 \\ (\tau - 1)r, & \text{if } r < 0 \end{cases}$$

is employed and yields the following problem

$$\min_{\mathcal{G}, \mathbf{U}^{(1)}, \cdots, \mathbf{U}^{(L)}} \sum_n \rho_\tau(y_n - \langle \mathcal{G} \times_1 \mathbf{U}^{(1)} \times_2 \cdots \times_L \mathbf{U}^{(L)}, \mathcal{X}_n \rangle). \tag{4.92}$$

Similar to the Tucker tensor regression, the alternating minimization algorithm can tackle this problem. Besides, if $P_1, \cdots, P_L$ are large, the factor matrices $\mathbf{U}^{(1)}, \cdots, \mathbf{U}^{(L)}$ are assumed to be sparse and orthogonal for controlling model complexity. The corresponding optimization problem is expressed as

$$\begin{aligned} \min_{\mathcal{G}, \mathbf{U}^{(1)}, \cdots, \mathbf{U}^{(L)}} \quad & \sum_n \rho_\tau(y_n - \langle \mathcal{G} \times_1 \mathbf{U}^{(1)} \times_2 \cdots \times_L \mathbf{U}^{(L)}, \mathcal{X}_n \rangle) \\ & \qquad + \mathfrak{J}(\mathbf{U}^{(1)}, \cdots, \mathbf{U}^{(L)}) \qquad (4.93) \\ & \text{s.t. } \mathbf{U}^{(l)^{\mathrm{T}}} \mathbf{U}^{(l)} = \mathbf{I} \text{ for } l = 1, \cdots, L \end{aligned}$$

and tackled by the ADMM, where $\mathfrak{J}(\mathbf{U}^{(1)}, \cdots, \mathbf{U}^{(L)})$ constraints the sparsity of the latent factors, which can either take form of $\lambda \| \mathbf{U}^{(L)} \otimes \cdots \otimes \mathbf{U}^{(1)} \|_1 = \lambda \prod_l \| \mathbf{U}^{(l)} \|_1$ or $\sum_l \lambda_l \| \mathbf{U}^{(l)} \|_1$.

Meanwhile, the statistical analysis of the quantile tensor regression shows that the estimator $\hat{\theta}$ without sparse constraints on the factor matrices not only converges to the true parameter $\theta = \{\mathcal{G}, \mathbf{U}^{(1)}, \cdots, \mathbf{U}^{(L)}\}$, but also converges at a rate of $1/\sqrt{N}$ under some mild conditions. In addition, under some assumptions, the estimation error of the sparse Tucker quantile tensor regression model (4.94) is bounded as

$$\| \hat{\mathcal{B}} - \mathcal{B}^* \|_{\mathrm{F}} \leq C \sqrt{S \log(\max(P_1 \cdots P_L, N))/N}, \qquad (4.94)$$

where $S$ denotes the sparsity of the matrix $\mathbf{U} = \mathbf{U}^{(L)} \otimes \cdots \otimes \mathbf{U}^{(1)}$, namely $\|\mathbf{U}\|_0 \leq S$, $C$ is a constant. It can be seen that the upper bound for the estimation error depends on the number of nonzero parameters $S$, the logarithmic of the size of each predictor $P_1 \cdots P_L$ and sample number $N$.

## 4.6 Projection based tensor regression

### 4.6.1 Tensor principal component regression

As the high order extension of principal component regression, tensor principal component regression still first performs principal component analysis over the high dimensional predictor to extract the important features and then regress the corresponding response with the obtained latent features. But the difference here is that the principal component analysis methods used for multiway covariates regression are based on tensor analysis.

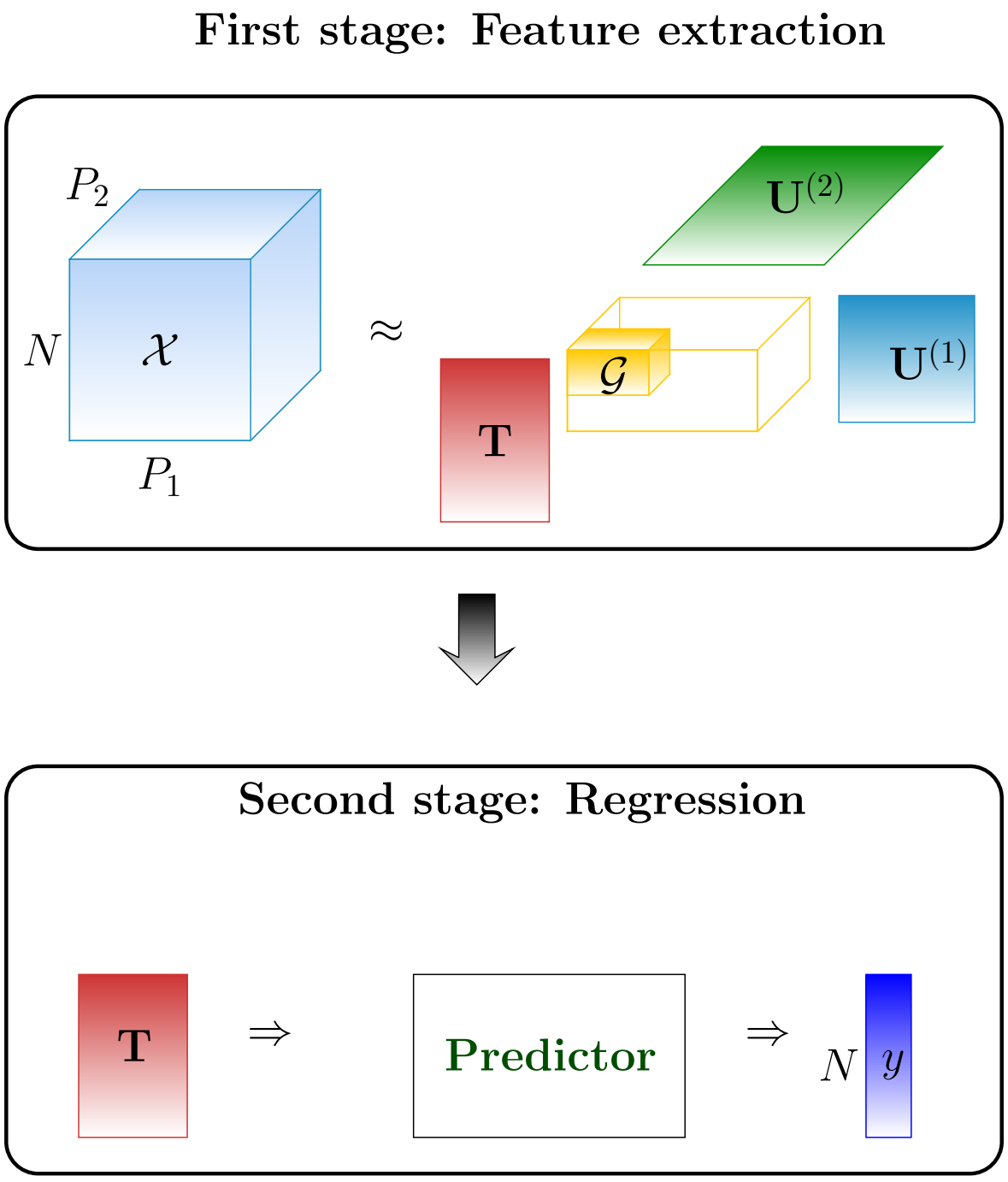

**Figure 4.6:** The two-stage framework for high dimensional regression frameworks.

For the two-stage framework, the first stage commonly aims to perform feature selection (dimension reduction), and the second stage tends to explore the relationship between the latent feature factors of the predictor and the response. As shown in Fig 4.6, one mostly used technology for high dimensional feature extraction is tensor decomposition like CP or Tucker decomposition. After the tensor decomposition, the latent factors can be used for subsequent regression analysis and the number of model parameters required will be reduced.

Given predictor $\mathcal{X} \in \mathbb{R}^{N \times P_1 \times P_2}$ and $\mathbf{y} \in \mathbb{R}^{N \times 1}$, multiway component regression first decomposes the predictor tensor using the CP decomposition (MCR-CP)(Bro, 1997; Geladi *et al.*, 1998) as follows

$$\mathcal{X} = [\![\mathbf{A}, \mathbf{U}^{(1)}, \mathbf{U}^{(2)}]\!], \tag{4.95}$$

which can be rewritten in matrix form as

$$\mathbf{X}_{(1)} = \mathbf{A}(\mathbf{U}^{(2)} \otimes \mathbf{U}^{(1)})^{\mathrm{T}}, \tag{4.96}$$

and then regress the response $\mathbf{y}$ over the latent factor $\mathbf{A}$ using the classical PLS model

$$\begin{aligned} \mathbf{A} &= \mathbf{T}\mathbf{H}^{\mathrm{T}} + \mathbf{E} && (4.97)\\ \mathbf{y} &= \mathbf{T}\mathbf{v} + \epsilon. && (4.98) \end{aligned}$$

The coefficient of determination, namely $R^2$, is used to measure how much of the variation in $\mathcal{X}$, $\mathbf{y}$ is predictable from the latent factor $\mathbf{T}$ as

$$\begin{aligned} R_x^2 &= 1 - \frac{\|\mathbf{X}_{(1)} - \mathbf{T}\mathbf{H}^{\mathrm{T}}(\mathbf{U}^{(2)} \otimes \mathbf{U}^{(1)})^{\mathrm{T}}\|^2}{\|\mathbf{X}_{(1)}\|^2}, \\ R_y^2 &= 1 - \frac{\|\mathbf{y} - \mathbf{T}\mathbf{v}\|^2}{\|\mathbf{y}\|^2}. \end{aligned} \tag{4.99}$$

Based on the same framework, employing different decompositions over the predictor $\mathcal{X}$ can yield different models, such as the Tucker decomposition (Smilde *et al.*, 1994; Kiers and Smilde, 1998).

In Li *et al.* (2013a), a dimension reduction procedure is performed on the original fMRI images and the downsized core factors are used for the subsequent disease diagnoses analysis. Moreover, an innovative individualized multilayer tensor learning model (Tang *et al.*, 2020) is proposed for multimodality imaging analysis, which improves the feature extraction step by decomposing the original multimodality images into modality-specific features shared by different subjects and individual-specific features shared by different modalities. In this way, the correlations along different modalities and different subjects can be well explored during the feature extraction stage and further provide more powerful predictions.

However, this strategy separates the entire regression framework into two independent steps, which makes the feature extraction process unsupervised and may discard some important features for prediction. Solutions to tackle this issue are either to optimize the objective function of these two steps jointly or to take account of the original multiway predictor and control the model complexity by regularization terms. For example, the optimization problem which maximizes the $R^2$ of predictor and response simultaneously is proposed in Smilde and Kiers (1999) as follows

$$\max \alpha R_x^2 + (1-\alpha) R_y^2 \tag{4.100}$$

for given $0 < \alpha < 1$, which balances the importance of fitting $\mathcal{X}$ and $\mathbf{y}$. Similarly, both the CP and Tucker models can be assumed over the predictor. It can also be extended into a generalized version where the predictor and response are both high-dimensional. It is to say, for given $\mathcal{X}$ and $\mathcal{Y}$, the multiway covariates regression (MCR) pursues the solution for the following problem

$$\begin{aligned} \mathbf{X}_{(1)} &= \mathbf{T}\mathbf{P}_x^{\mathrm{T}} \\ \mathbf{Y}_{(1)} &= \mathbf{T}\mathbf{P}_y^{\mathrm{T}} \\ \max \quad & \alpha R_x^2 + (1-\alpha) R_y^2 \\ R_x^2 &= 1 - \|\mathbf{X}_{(1)} - \mathbf{T}\mathbf{P}_x^{\mathrm{T}}\|^2 / \|\mathbf{X}_{(1)}\|^2 \\ R_y^2 &= 1 - \|\mathbf{Y}_{(1)} - \mathbf{T}\mathbf{P}_y^{\mathrm{T}}\|^2 / \|\mathbf{Y}_{(1)}\|^2 \end{aligned} \tag{4.101}$$

with

$$\mathbf{P}_x = \mathbf{U}_x^{(2)} \otimes \mathbf{U}_x^{(1)}$$
$$\mathbf{P}_y = \mathbf{U}_y^{(2)} \otimes \mathbf{U}_y^{(1)}$$

in CP model;

$$\mathbf{P}_x^{\mathrm{T}} = \mathbf{G}_{x(1)} {\mathbf{U}_x^{(2)}}^{\mathrm{T}} \otimes {\mathbf{U}_x^{(1)}}^{\mathrm{T}}$$
$$\mathbf{P}_y^{\mathrm{T}} = \mathbf{G}_{y(1)} {\mathbf{U}_y^{(2)}}^{\mathrm{T}} \otimes {\mathbf{U}_y^{(1)}}^{\mathrm{T}}$$

in Tucker model, where $\mathbf{G}_{x(1)}$ and $\mathbf{G}_{y(1)}$ are the mode-1 unfolding matrices of $\mathcal{G}_x$ and $\mathcal{G}_y$ respectively with

$$\begin{aligned} \mathcal{X} &= [\![\mathcal{G}_x; \mathbf{T}_x, \mathbf{U}_x^{(1)}, \mathbf{U}_x^{(2)}]\!] \\ \mathcal{Y} &= [\![\mathcal{G}_y; \mathbf{T}_y, \mathbf{U}_y^{(1)}, \mathbf{U}_y^{(2)}]\!]. \end{aligned}$$

Moreover, it is generalized to incorporate multiway blocks in either the input side or output side (Smilde *et al.*, 2000). The resulting algorithm is named multiway multiblock covariates regression models.

Another example for ensemble learning for the two-stage framework is presented in Xu *et al.* (2019). It first decomposes the predictor tensor by CP decomposition (Xu *et al.*, 2019)

$$\mathcal{X} = [\![\mathbf{A}, \mathbf{B}, \mathbf{C}]\!], \tag{4.102}$$

where $\mathcal{X} \in \mathbb{R}^{S \times T \times d}$ represents the predictor tensor, $S$ is the total number of locations, $T$ is the timestamps, $d$ is the number of predictor variables for each location $s$ at time point $t$. $\mathbf{A} \in \mathbb{R}^{S \times R}$, $\mathbf{B} \in \mathbb{R}^{T \times R}$, and $\mathbf{C} \in \mathbb{R}^{d \times R}$ are the corresponding latent factors for its spatial, temporal and feature dimensions, respectively.

Then an ensemble model is built on the latent factors directly as follows

$$\begin{aligned} &\min_{\mathscr{T}=\{\mathbf{W},\mathbf{V},\mathbf{A},\mathbf{B},\mathbf{C}\}} \mathfrak{L}(\mathscr{T}; \mathcal{X}, \mathbf{Y}) \\ =\ & \frac{1}{2}\sum_{s}^{S}\sum_{t}^{T}(\mathbf{x}_{s,t}^{\mathrm{T}}(\mathbf{W}^{\mathrm{T}}\mathbf{A}(s,:) + \mathbf{V}^{\mathrm{T}}\mathbf{B}(t,:)) - y_{s,t})^2 \\ +\ & \frac{\lambda}{2}\|\mathcal{X} - [\![\mathbf{A}, \mathbf{B}, \mathbf{C}]\!]\|_{\mathrm{F}}^2 + \beta\|\mathscr{T}\|_1, \end{aligned} \tag{4.103}$$

where $\mathbf{Y} \in \mathbb{R}^{S \times T}$ is the matrix response, $\mathbf{x}_{s,t} = \mathcal{X}(s,t,:)$ is the input vector for the corresponding response $y_{s,t} = \mathbf{Y}(s,t)$. $\mathbf{W}$ and $\mathbf{V}$ represents the spatial and temporal dependencies on the latent factors $\mathbf{A}$ and $\mathbf{B}$, respectively. The set $\mathscr{T}$ is a collection of the model parameters, $\|\mathscr{T}\|_1$ means applying $\ell_1$ norm on every model parameter separately as follows

$$\|\mathscr{T}\|_1 = \|\mathbf{W}\|_1 + \|\mathbf{V}\|_1 + \|\mathbf{A}\|_1 + \|\mathbf{B}\|_1 + \|\mathbf{C}\|_1. \tag{4.104}$$

Based on alternating minimization, each factor can be updated iteratively while the proximal gradient descent method is used to tackle each subproblem due to the non-differentiability of the $\ell_1$ norm. In this way, the feature extraction stage will be linked with the outcomes and make the important features in predicting outcomes well extracted in the first stage.

### 4.6.2 Tensor partial least squares regression

Another strand of the tensor regression model is based on the partial least squares (PLS), which is shown to be efficient for the analysis of highly collinear data. PLS based methods attempt to evaluate the relationship between the different data forms of the same subject through an assumption that the input and output are expected to contain common or highly relevant latent factors under a given factorization form. So the key problem for PLS based models is to find a set of common latent factors of the predictor and the response through projecting them onto subspaces with constraints over the latent factors. It can be seen as an integration of the tensor principal component analysis, canonical correlation analysis, and multilinear regression analysis. Here we will review two commonly used linear generations of PLS algorithm to tensors, $N$-way PLS based on CP (Bro, 1996) and HOPLS based on block term decomposition (Zhao *et al.*, 2013a).

### $N$-way PLS

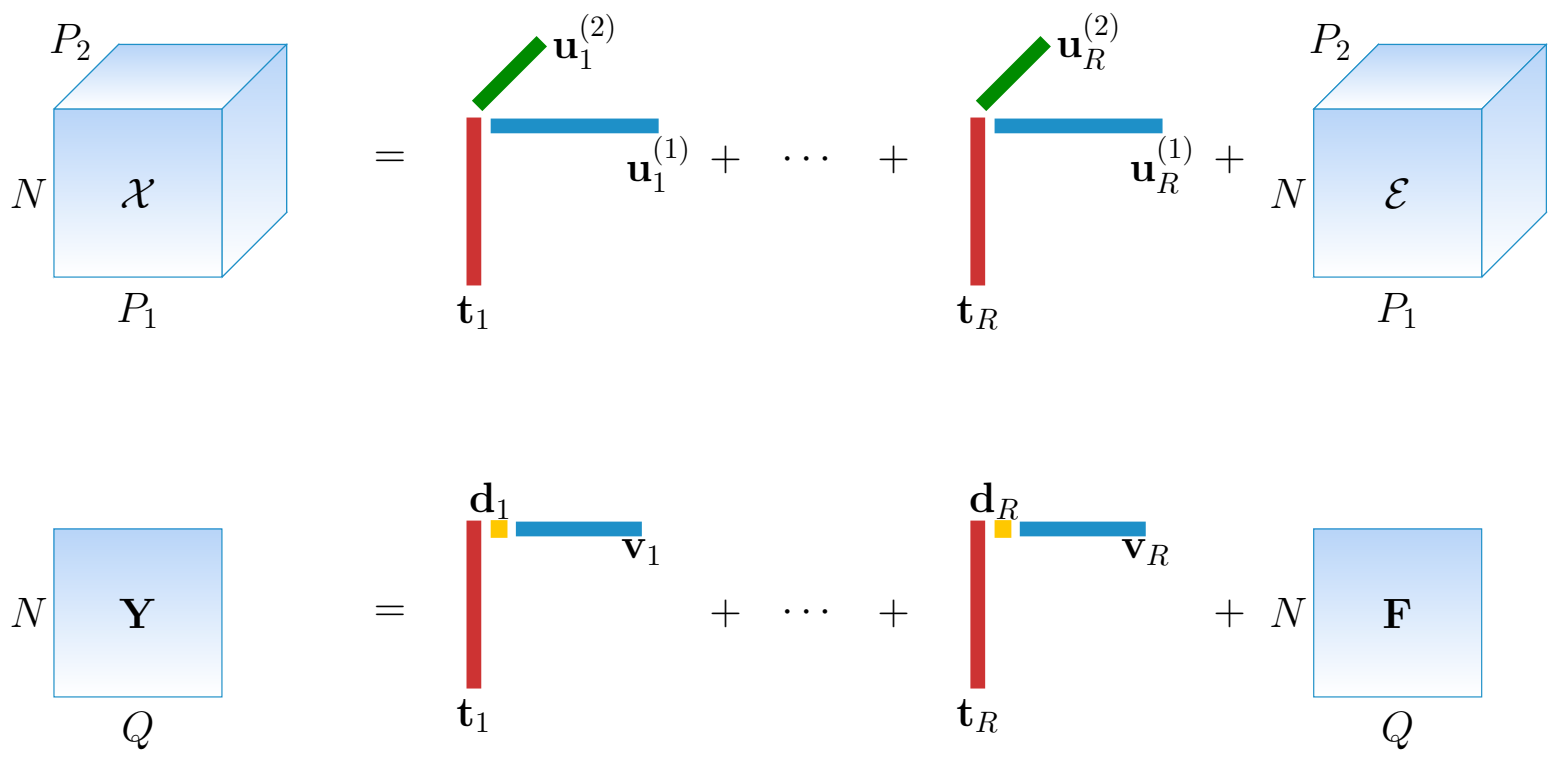

**Figure 4.7:** The diagram representation of $N$-way PLS model.

The multi-way partial least squares (PLS), namely $N$-way PLS, decomposes the $N$-th order input tensor through the CP decomposition, searches the common latent factors of the predictor and the response, and finally achieves the prediction of the response variables. It is firstly proposed in Bro (1996) and further commented by Smilde (1997) and Jong (1998).

For example, the 3-way PLS decomposes the predictor $\mathcal{X} \in \mathbb{R}^{N\times P_1\times P_2}$ and the response $\mathbf{Y} \in \mathbb{R}^{N\times Q}$ simultaneously as follows

$$\begin{aligned} \mathcal{X} &= \sum_{r=1}^{R} \mathbf{t}_r \circ \mathbf{u}_r^{(1)} \circ \mathbf{u}_r^{(2)} + \mathcal{E}, \\ \mathbf{Y} &= \sum_{r=1}^{R} d_r \mathbf{t}_r \mathbf{v}_r^{\mathrm{T}} + \mathbf{F}, \end{aligned} \tag{4.105}$$

where $\mathbf{t}_r$ is the common latent vectors of $\mathcal{X}$ and $\mathbf{Y}$, $\mathbf{u}_r^{(1)}$ and $\mathbf{u}_r^{(2)}, r = 1, \cdots, R$ are mode-1 loading vectors and mode-2 loading vectors for $\mathcal{X}$, while $\mathbf{v}_r$, $r = 1, \cdots, R$ are loading vectors of $\mathbf{Y}$, as shown in Fig. 4.7.

To figure out the common latent factors and other loading vectors, the optimization problem is given by

$$\begin{aligned} \max_{\mathbf{u}_r^{(1)},\mathbf{u}_r^{(2)},\mathbf{v}_r} \quad & \mathrm{Cov}\left(\mathbf{t}_r, \mathbf{c}_r\right) \\ \text{s.t.} \quad & \mathbf{t}_r = \mathcal{X} \times_1 \mathbf{u}_r^{(1)} \times_2 \mathbf{u}_r^{(2)}, \quad \mathbf{c}_r = \mathbf{Y}\mathbf{v}_r, \\ & \|\mathbf{u}_r^{(1)}\|_2^2 = \|\mathbf{u}_r^{(2)}\|_2^2 = \|\mathbf{v}_r\|_2^2 = 1, \end{aligned} \tag{4.106}$$

where the extracted latent factors are constrained to have maximum covariance.

Once the model parameters are obtained, the prediction of new datasets can be efficiently computed by $\mathbf{Y}' = \mathbf{X}'_{(1)}(\mathbf{U}^{(2)} \odot \mathbf{U}^{(1)})^{-1}\mathbf{D}\mathbf{V}^{\mathrm{T}}$, where $\mathbf{U}^{(l)} = [\mathbf{u}_1^{(l)}, \cdots, \mathbf{u}_R^{(l)}]$, $l = 1, 2$, $\mathbf{V} = [\mathbf{v}_1, \cdots, \mathbf{v}_R]$, $\mathbf{D} = \mathrm{diag}(d_r)$.

In addition, a Tucker-based $N$-way PLS model is proposed in Bro *et al.* (2001), where $\mathcal{X}$ is approximated by Tucker decomposition as

$$\mathcal{X} = \mathcal{G} \times_1 \mathbf{T} \times_2 \mathbf{U}^{(1)} \times_3 \mathbf{U}^{(2)} + \mathcal{E},$$

where $\mathcal{G}$ can be seen as a projection of $\mathcal{X}$ along three modes as $\mathcal{G} = \mathcal{X} \times_1 \mathbf{T}^{\mathrm{T}} \times_2 {\mathbf{U}^{(1)}}^{\mathrm{T}} \times_3 {\mathbf{U}^{(2)}}^{\mathrm{T}}$. The advantages of this new method are explained from the perspective of low-rank approximation versus subspace approximation. Meanwhile, a companion of $N$-way PLS and multiway covariates regression is given in Gurden *et al.* (2001). The results show that it is easier for multiway covariates regression to incorporate some constraints over the model coefficient tensor, but it also inevitably introduces some hyperparameters that need to be optimized.

Moreover, some theoretical analysis works of $N$-way PLS are proposed. For example, the sample-specific standard errors of prediction when applying $N$-way PLS is investigated and verified with real fluorescence spectroscopy data (Faber and Bro, 2002; Bro *et al.*, 2005). A new estimate of prediction variance from the influence function is derived in Serneels *et al.* (2005).

### HOPLS

Based on the block term decomposition, high-order PLS (HOPLS) performs constrained block term decomposition on the input tensor $\mathcal{X} \in \mathbb{R}^{N\times P_1\times P_2\times\cdots\times P_L}$ and the output tensor $\mathcal{Y} \in \mathbb{R}^{N\times Q_1\times Q_2\times\cdots\times Q_M}$ simultaneously. The HOPLS model is expressed as

$$\begin{aligned}\mathcal{X} &= \sum_{r=1}^{R} \mathcal{G}_{xr} \times_1 \mathbf{t}_r \times_2 \mathbf{U}_r^{(1)} \cdots \times_{L+1} \mathbf{U}_r^{(L)} + \mathcal{E} \qquad (4.107)\\ &= \mathcal{G}_x \times_1 \mathbf{T} \times_2 \mathbf{U}^{(1)} \cdots \times_{L+1} \mathbf{U}^{(L)} + \mathcal{E},\\ \mathcal{Y} &= \sum_{r=1}^{R} \mathcal{G}_{yr} \times_1 \mathbf{t}_r \times_2 \mathbf{V}_r^{(1)} \cdots \times_{M+1} \mathbf{V}_r^{(M)} + \mathcal{F} \qquad (4.108)\\ &= \mathcal{G}_y \times_1 \mathbf{T} \times_2 \mathbf{V}^{(1)} \cdots \times_{M+1} \mathbf{V}^{(M)} + \mathcal{F},\end{aligned}$$

where the predictor $\mathcal{X}$ is decomposed as a sum of $R$ rank-$(1, I_1, \cdots, I_L)$ subcomponents and the residuals $\mathcal{E}$, the response $\mathcal{Y}$ is decomposed as the sum of $R$ rank-$(1, K_1, \cdots, K_M)$ subcomponents and the residuals $\mathcal{F}$. The variable $\mathbf{t}_r \in \mathbb{R}^N$ is the $r$-th common latent factor between the predictor $\mathcal{X}$ and the response $\mathcal{Y}$ along the sample mode, $\{\mathbf{U}_r^{(l)} \in \mathbb{R}^{P_l\times I_l}\}_{l=1}^L$ and $\{\mathbf{V}_r^{(m)} \in \mathbb{R}^{Q_m\times K_m}\}_{m=1}^M$ are the loading matrices of $\mathcal{X}$ and $\mathcal{Y}$ along specific modes, respectively. $\mathcal{G}_{xr} \in \mathbb{R}^{1\times I_1\times\cdots\times I_L}$ and $\mathcal{G}_{yr} \in \mathbb{R}^{1\times K_1\times\cdots\times K_M}$ are the core tensors of $\mathcal{X}$ and $\mathcal{Y}$, respectively. A graphical representation for the HOPLS model is provided in Fig. 4.8, $\mathbf{T}$, $\mathbf{U}^{(l)}$, $\mathbf{V}^{(m)}$ are the collections of $\mathbf{t}_r$, $\mathbf{U}_r^{(l)}$ and $\mathbf{V}_r^{(m)}$ along $r$, $\mathcal{G}_x = \mathrm{diag}(\mathcal{G}_{x1}, \cdots, \mathcal{G}_{xR})$, $\mathcal{G}_y = \mathrm{diag}(\mathcal{G}_{y1}, \cdots, \mathcal{G}_{yR})$.

It should be noted that HOPLS will degenerate into $N$-way PLS if $I_l = 1$ and $K_m = 1$ for any $l$ and $m$. Compared with $N$-way PLS, HOPLS is able to control the model complexity by changing the size of the loading matrices with a trade-off to the modeling error. Meanwhile, it is possible for HOPLS to explore the multilinear rank rather than assuming the same rank for every mode in $N$-way PLS, which greatly improves the model flexibility and performance.

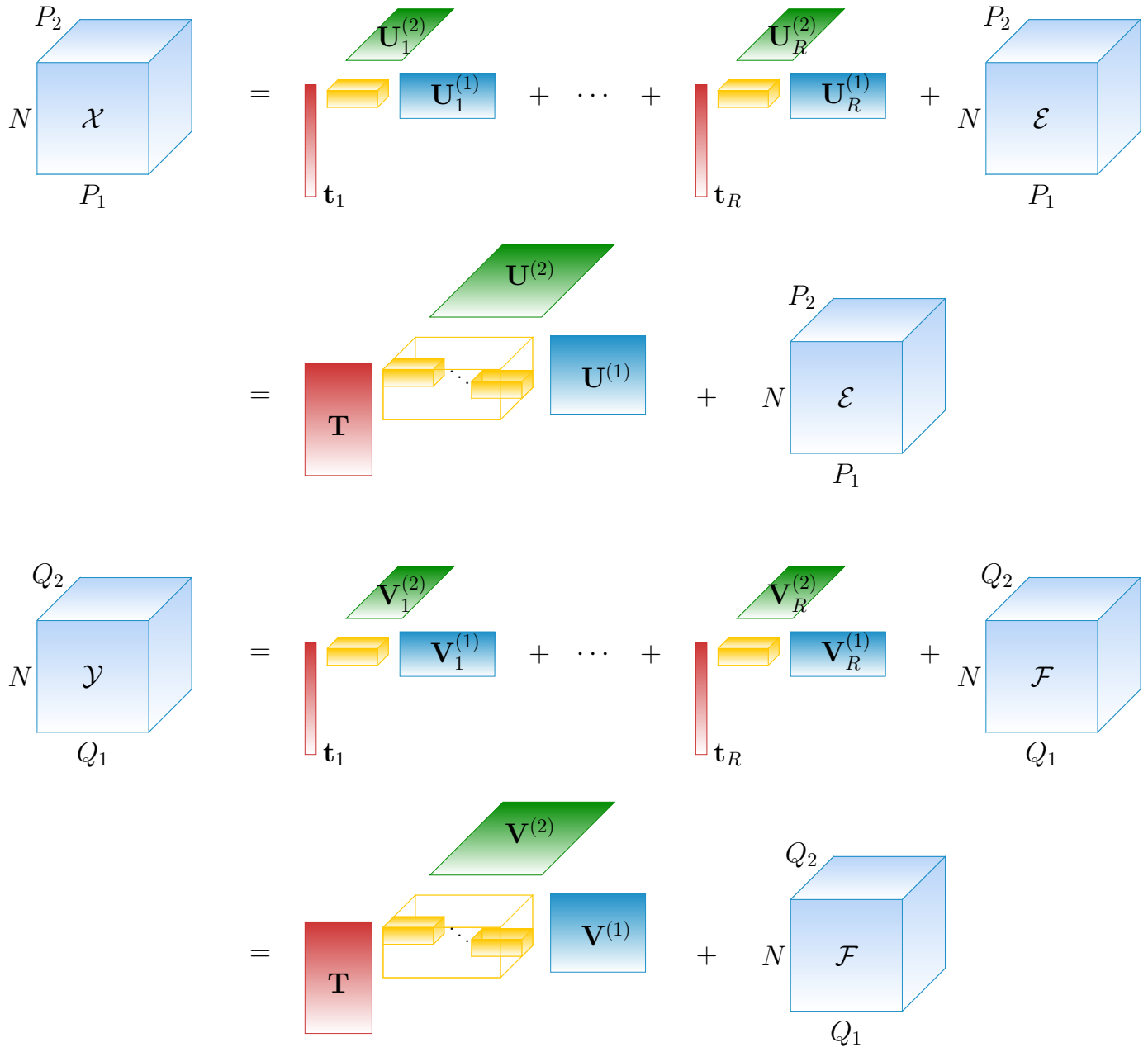

**Figure 4.8:** The diagram representation of HOPLS, where $\mathbf{t}_r \in \mathbb{R}^{N\times 1}$, $\mathbf{U}_r^{(l)} \in \mathbb{R}^{P_l\times I_l}$, $\mathbf{V}_r^{(m)} \in \mathbb{R}^{Q_m\times K_m}$, $l = 1, 2$, $m = 1, 2$.

The solving process of HOPLS is recursive, so the key problem boils down to finding the most correlated latent components in the remaining parts of $\mathcal{X}$ and $\mathcal{Y}$. For the approximation of $r$-th subtensor, the loading matrices $\{\mathbf{U}_r^{(l)}\}_{l=1}^{L}$ and $\{\mathbf{V}_r^{(m)}\}_{m=1}^{M}$ are first extracted and then the common latent factor $\mathbf{t}_r$ is estimated. The corresponding optimization problem for HOPLS is formulated as

$$\max_{\{\mathbf{U}_r^{(l)},\mathbf{V}_r^{(m)}\}} \|[\![\mathcal{Z}_r; \mathbf{U}_r^{(1)\mathrm{T}}, \cdots, \mathbf{U}_r^{(L)\mathrm{T}}, \mathbf{V}_r^{(1)\mathrm{T}}, \cdots, \mathbf{V}_r^{(M)\mathrm{T}}]\!]\|_\mathrm{F}^2 \quad (4.109)$$
$$\text{s.t.} \quad \mathbf{U}_r^{(l)\mathrm{T}}\mathbf{U}_r^{(l)} = \mathbf{I}_{I_l}, \quad \mathbf{V}_r^{(m)\mathrm{T}}\mathbf{V}_r^{(m)} = \mathbf{I}_{K_m},$$

where $\mathcal{Z}_r = \langle \overleftarrow{\mathcal{X}_r}^1, \mathcal{Y}_r\rangle_1 \in \mathbb{R}^{P_1\times\cdots\times P_L\times Q_1\times\cdots\times Q_M}$ is the tensor contraction product of $\mathcal{X}_r$ and $\mathcal{Y}_r$ along the sample dimension.

$$\mathcal{X}_r := \begin{cases} \mathcal{X}, & \text{if } r = 1 \\ \mathcal{X}_{r-1} - [\![\mathcal{G}_{x,r-1}; \mathbf{t}_{r-1}, \mathbf{U}_{r-1}^{(1)}, \ldots, \mathbf{U}_{r-1}^{(L)}]\!], & \text{otherwise ;} \end{cases}$$

$$\mathcal{Y}_r := \begin{cases} \mathcal{Y}, & \text{if } r = 1 \\ \mathcal{Y}_{r-1} - [\![\mathcal{G}_{y,r-1}; \mathbf{t}_{r-1}, \mathbf{V}_{r-1}^{(1)}, \ldots, \mathbf{V}_{r-1}^{(M)}]\!], & \text{otherwise .} \end{cases}$$

This problem can be solved by the rank-$(I_1, \cdots, I_L, K_1, \cdots, K_M)$ orthogonal Tucker decomposition of tensor $\mathcal{Z}_r$ though the higher-order orthogonal iteration algorithm (HOOI) (Kolda and Bader, 2009; De Lathauwer *et al.*, 2000).

After getting the loading matrices, the latent vector $\mathbf{t}_r$ can be easily obtained as the first leading left singular vector of matrix $(\mathcal{X}_r \times_2 \mathbf{U}_r^{(1)^{\mathrm{T}}} \cdots \times_{L+1} \mathbf{U}_r^{(L)^{\mathrm{T}}})_{(1)}$ by solving the following the optimization problem

$$\mathbf{t}_r = \arg\min_{\mathbf{t}_r} \|\mathcal{X}_r - [\![\mathcal{G}_{xr}; \mathbf{t}_r, \mathbf{U}_r^{(1)}, \cdots, \mathbf{U}_r^{(L)}]\!]\|_{\mathrm{F}}^2. \tag{4.110}$$

Finally the core tensor $\mathcal{G}_x$ and $\mathcal{G}_y$ can be computed by

$$\mathcal{G}_{xr} = \mathcal{X}_r \times_1 \mathbf{t}_r^{\mathrm{T}} \times_2 \mathbf{U}_r^{(1)^{\mathrm{T}}} \times_3 \cdots \times_{L+1} \mathbf{U}_r^{(L)^{\mathrm{T}}}, \tag{4.111}$$

$$\mathcal{G}_{yr} = \mathcal{Y}_r \times_1 \mathbf{t}_r^{\mathrm{T}} \times_2 \mathbf{V}_r^{(1)^{\mathrm{T}}} \times_3 \cdots \times_{M+1} \mathbf{V}_r^{(M)^{\mathrm{T}}}. \tag{4.112}$$

The entire solution can be achieved by repeating the above procedure until the stopping condition is reached.

In addition, variants of the tensor-based PLS are proposed, such as the penalized multiway PLS (Eliseyev and Aksenova, 2016), Bayesian multiway PLS (Camarrone and Van Hulle, 2016), the fast HOPLS (Camarrone and Van Hulle, 2018) and the tensor envelope partial least squares regression (TEPLS) (Zhang and Li, 2017).

## 4.7 Summary

In this section, we have reviewed the related works for tensor based linear regression models and algorithms. A collection of linear tensor regression models used in this work, which are grouped by families, is provided in Table 4.2.

**Table 4.2:** Summarization of linear tensor regression models, algorithms and applications.

| Family | | Algorithm | Regression Model | Low-rank tensor representation form | Optimization framework | Application |
|---|---|---|---|---|---|---|
| Simple linear regression | | MLMTL (Romera-Paredes *et al.*, 2013) | $y_t = \langle \mathbf{x}_t, \mathbf{b}_t \rangle + \varepsilon_t$ | Tucker | ADMM | Multitask learning; Spatial-temporal data analysis |
| | | Greedy (Bahadori *et al.*, 2014) | $\mathcal{Y}[:,:,m] = \mathcal{B}[:,:,m]\mathcal{X}[:,:,m] + \mathcal{E}[:,:,m]$ | CP | Geedy | Spatial-temporal data analysis |
| | | TPG (Yu and Liu, 2016) | $\mathcal{Y} = << \mathcal{X}, \mathcal{B} >> + \mathcal{E}$ | Tucker | PGD | Multitask learning; Spatial-temporal data analysis |
| | | MLTR (Hoff, 2015) | $\mathcal{Y} = \mathcal{X} \times_1 \mathbf{B}^{(1)} \cdots \times_M \mathbf{B}^{(M)} \times_{M+1} \mathbf{I}_N + \mathcal{E}$ | Tucker | ALS/ Bayesian inference | Longitudinal relational data analysis |
| Generalized linear tensor regression | | CP-GLTR (Zhou *et al.*, 2013) | $g(\mu) = \alpha + \gamma^{\mathrm{T}}\mathbf{z} + \left\langle \sum_{r=1}^{R} \mathbf{u}_r^{(1)} \circ \cdots \circ \mathbf{u}_r^{(L)}, \mathcal{X} \right\rangle$ | CP | AM | Neuroimaging analysis |
| | | Tucker-GLTR (Li *et al.*, 2013a) | $g(\mu) = \alpha + \gamma^{\mathrm{T}}\mathbf{z} + \left\langle \mathcal{G} \times_1 \mathbf{U}^{(1)} \times_2 \mathbf{U}^{(2)} \cdots \times_L \mathbf{U}^{(L)}, \mathcal{X} \right\rangle$ | Tucker | AM | Neuroimaging analysis |
| | | HT-GLTR (Hou and Chaib-Draa, 2015) | $g(\mu) = \alpha + \gamma^{\mathrm{T}}\mathbf{z} + \langle \{\mathbf{G}_1, \cdots, \mathbf{G}_N; \mathcal{B}_{1,2}, \cdots, \mathcal{B}_{N-1,N}; \mathbf{B}_{1,\cdots,N}\}, \mathcal{X} \rangle$ . | HT | AM | Neuroimaging analysis |
| | | GLTR-PGD (Chen *et al.*, 2019) | $g(\mu) = \langle \mathcal{B}, \mathcal{X} \rangle$ . | Tucker | PGD | / |
| Shrinkage methods | Ridge regression | hrTRR (Guo *et al.*, 2012) | $y_n = \left\langle \mathcal{X}_n, \sum_r^R \mathbf{u}_r^{(1)} \circ \cdots \circ \mathbf{u}_r^{(L)} \right\rangle + b$ | CP | ALS | Human age estimation; Head pose estimation; Human motion capture |
| | | TTR-CP (Lock, 2018) | $\mathcal{Y} = \langle \mathcal{X}, \mathcal{B} \rangle_L + \mathcal{E}$, $\mathcal{B} = \sum_r^R \mathbf{u}_r^{(1)} \circ \cdots \circ \mathbf{u}_r^{(L)}$. | CP | ALS | Facial imaging data analysis |
| | | TTR-TT (Liu *et al.*, 2020c) | $\mathcal{Y} = \langle \mathcal{X}, \mathcal{B} \rangle_L + \mathcal{E}$, $\mathcal{B} = [\![\mathcal{U}^{(1)}, \cdots, \mathcal{U}^{(L)}, \mathcal{V}^{(1)}, \cdots, \mathcal{V}^{(M)}]\!]$. | TT | ALS | Spatial-temporal data analysis; Human motion capture |
| | | HOLRR (Rabusseau and Kadri, 2016) | $\mathcal{Y} = \mathcal{G} \times_1 \mathbf{X}\mathbf{U}^{(0)} \cdots \times_{M+1} \mathbf{U}^{(M)} + \mathcal{E}$ | Tucker | – | Spatial-temporal data analysis; Structured ponit cloud data analysis (Yan *et al.*, 2019) |
| | Sparse regression | Remurs (Song and Lu, 2017) | $y_n = \langle \mathcal{B}, \mathcal{X}_n \rangle + \varepsilon_n$ | CP | ADMM | Neuroimaging analysis |
| | | Sturm (Li *et al.*, 2019b) | $y_n = \langle \mathcal{B}, \mathcal{X}_n \rangle + \varepsilon_n$ | t-SVD | ADMM | Neuroimaging analysis |
| | | STPG (Ahmed *et al.*, 2019) | $y_n = \langle \mathcal{B}, \mathcal{X}_n \rangle + \varepsilon_n, \mathcal{B} \in \mathscr{C}$ | Tucker | PGD | Neuroimaging analysis |
| | | SURF (He *et al.*, 2018) | $y_n = \langle \mathcal{B}, \mathcal{X}_n \rangle + \varepsilon_n$, $\mathcal{B} = \sum_r^R \mathbf{u}_r^{(1)} \circ \cdots \circ \mathbf{u}_r^{(L)}$. | CP | Greedy | Neuroimaging analysis |
| | | SMRTR (Li *et al.*, 2016) | $\mathbf{y}_n = \begin{bmatrix} \left\langle \sum_{r=1}^R \mathbf{u}_{1r}^{(1)} \circ \cdots \circ \mathbf{u}_{1r}^{(L)}, \mathcal{X}_n \right\rangle \\ \vdots \\ \left\langle \sum_{r=1}^R \mathbf{u}_{Jr}^{(1)} \circ \cdots \circ \mathbf{u}_{Jr}^{(L)}, \mathcal{X}_n \right\rangle \end{bmatrix} + \mathbf{e}_n$ | CP | ALS | Neuroimaging analysis |
| | | STORE (Sun and Li, 2017) | $\mathcal{Y}_n = \sum_r \lambda_r ((\mathbf{u}_r^{(M+1)})^{\mathrm{T}} \mathbf{x}_n) \mathbf{u}_r^{(1)} \circ \cdots \circ \mathbf{u}_r^{(M)} + \mathcal{E}_n$ | CP | ALS | Neuroimaging analysis |
| | | TETORE (Li and Zhang, 2017) | $\mathcal{Y} = \mathcal{B} \times_{M+1} \mathbf{X} + \mathcal{E}$ | – | ALS | Neuroimaging analysis |
| | | orTRR (Guo *et al.*, 2012) | $y_n = \left\langle \mathcal{X}_n, \sum_r^R \mathbf{u}_r^{(1)} \circ \cdots \circ \mathbf{u}_r^{(L)} \right\rangle + b$ | CP | ALS | Human age estimation; Head pose estimation; Human motion capture |
| | Others | GC-CP (Karahan *et al.*, 2015) | $\mathbf{Y} = \langle \mathcal{B}, \mathcal{Y} \rangle + \mathbf{E}$, $\mathcal{B} = [\![\mathbf{M}_s, \mathbf{M}_r, \mathbf{T}]\!]$ | CP | ADMM | Brain connectivity analysis |

**Table 4.2:** (Continued) Summarization of linear tensor regression models, algorithms and applications.

| Family | Algorithm | Regression Model | Low-rank tensor representation form | Optimization framework | Application |
|---|---|---|---|---|---|
| Bayesian learning | BTR (Suzuki, 2015) | $y_n = \langle \mathcal{B}, \mathcal{X}_n \rangle + \varepsilon_n$ | CP | Bayesian inference | Tensor completion |
| Bayesian learning | BTR-M-DGDP (Guhaniyogi *et al.*, 2017) | $y_n = \alpha + \mathbf{z}_n^{\mathrm{T}} \gamma + \langle \mathcal{X}_n, \mathcal{B} \rangle + \varepsilon_n$ | CP | Bayesian inference | Brain Connectome Data Analysis |
| Bayesian learning | BTDM (Billio *et al.*, 2018) | $\mathcal{Y}_t = \mathcal{B} \times_4 \mathrm{vec}(\mathcal{Y}_{t-1}) + \mathcal{E}_t$ | CP | Bayesian inference | Multilayer Dynamic Networks |
| Bayesian learning | Softer (Papadogeorgou *et al.*, 2019) | $y_n = \alpha + \mathbf{z}_n^{\mathrm{T}} \gamma + \langle \mathcal{X}_n, \mathcal{B} \rangle + \varepsilon_n$ | Soft CP | Bayesian inference | Brain Connectome Data Analysis |
| Bayesian learning | BMESTRR (Spencer *et al.*, 2019) | $\mathcal{Y}_{n,g,t} = \mathcal{B}_g x_{n,t} + d_{n,g} + \mathcal{E}_{n,g,t}$ | CP | Bayesian inference | fMRI data analysis |
| Quantile tensor regression | HDQR(Lu *et al.*, 2020) | $Q_\tau(y_n \vert \mathcal{X}_n) = \langle \mathcal{G} \times_1 \mathbf{U}^{(1)} \times_2 \cdots \times_L \mathbf{U}^{(L)}, \mathcal{X}_n \rangle + \epsilon$ | Tucker | AM/ADMM | Crowd image analysis |
| Projection methods – Tensor PCR | MPR-CP(Bro, 1997; Geladi *et al.*, 1998) | $\mathcal{X} = [\![ \mathbf{A}, \mathbf{U}^{(1)}, \mathbf{U}^{(2)} ]\!]$<br>$\mathbf{A} = \mathbf{T}\mathbf{H}^{\mathrm{T}}$<br>$\mathbf{Y} = \mathbf{T}\mathbf{v}$ | CP | AM+Greedy | Environmental analysis (Geladi *et al.*, 1998)<br>Process optimization (Bro, 1997) |
| Projection methods – Tensor PCR | MCR (Smilde and Kiers, 1999) | $\mathbf{X}_{(1)} = \mathbf{T}\mathbf{P}_x^{\mathrm{T}}$<br>$\mathbf{Y}_{(1)} = \mathbf{T}\mathbf{P}_y^{\mathrm{T}}$ | CP/Tucker | AM | Process optimization |
| Projection methods – Tensor PLS regression | *N*-way PLS (Bro, 1996) | $\mathcal{X} = \sum_{r=1}^R \mathbf{t}_r \circ \mathbf{u}_r^{(1)} \circ \mathbf{u}_r^{(2)} + \mathcal{E}$<br>$\mathbf{Y} = \sum_{r=1}^R d_r \mathbf{t}_r \mathbf{v}_r^{\mathrm{T}} + \mathbf{F}$ | CP | Greedy | chemometrics (Hasegawa *et al.*, 2000)<br>Neuroscience (Acar *et al.*, 2007) |
| Projection methods – Tensor PLS regression | HOPLS (Zhao *et al.*, 2013a) | $\mathcal{X} = \mathcal{G}_x \times_1 \mathbf{T} \times_2 \mathbf{U}^{(1)} \cdots \times_{L+1} \mathbf{U}^{(L)} + \mathcal{E}$<br>$\mathcal{Y} = \mathcal{G}_y \times_1 \mathbf{T} \times_2 \mathbf{V}^{(1)} \cdots \times_{M+1} \mathbf{V}^{(M)} + \mathcal{F}$ | BTD | Greedy | ECoG signal processing |

From the perspective of modeling, these algorithms can be classified into three groups, including vector-on-tensor regression, tensor-on-vector regression, and tensor-on-tensor regression. The vector-on-tensor regression concerns regression tasks with high dimensional predictors, including generalized linear tensor regression (Zhou *et al.*, 2013; Li *et al.*, 2013a; Hou and Chaib-Draa, 2015; Chen *et al.*, 2019), hrTRR (Guo *et al.*, 2012), sparse tensor regression (Song and Lu, 2017; Li *et al.*, 2019b; Ahmed *et al.*, 2019; He *et al.*, 2018; Li *et al.*, 2016), Bayesian based tensor regression methods (Suzuki, 2015; Guhaniyogi *et al.*, 2017; Papadogeorgou *et al.*, 2019), and HDQR (Lu *et al.*, 2020). This class of approaches is mainly designed for neuroimaging data analysis, with applications in clinical assessments estimation and disease diagnosis. The second mainly deals with the comparison tasks, in this case the response is usually a tensor and the predictor is a vector. For example, STORE (Sun and Li, 2017) and TETORE (Li and Zhang, 2017) investigate the difference of brain scans or connectivity patterns between subjects with neurological disorders and normal subjects. Yan *et al.* (2019) tries to compare the patterns of point clouds under different process variables. The last group of tensor-on-tensor regression can be seen as the generalization of the previous two, with both tensor predictors and tensor responses. The motivation for this model mainly comes from forecasting tasks, corkriging, or casual analysis for different networks in sociology, climatology, economics, geography and neuroscience; and multitask learning indexed by many dimensions. Popular solutions consist of TPG (Yu and Liu, 2016), MLTR (Hoff, 2015), TTR-CP (Lock, 2018), TTR-TT (Liu *et al.*, 2020c), BTDM (Billio *et al.*, 2018) and the projection-based methods MCR (Smilde and Kiers, 1999), $N$-way PLS (Bro, 1996) and HOPLS (Zhao *et al.*, 2013a). It is necessary to note that although these algorithms are all modeled based on both tensor input and tensor response, they are suitable for different problems and have their own pros and cons. For example, the model $\mathcal{Y} = \ll \mathcal{X}, \mathcal{B} \gg + \mathcal{E}$ is only suitable for spatial-temporal data due to the restricted data form requirement that the spatial and temporal structure of the predictor and response must be shared in common. And the order of the predictor and the response must be the same in the model $\mathcal{Y} = \mathcal{X} \times_1 \mathbf{B}^{(1)} \cdots \times_M \mathbf{B}^{(M)} \times_{M+1} \mathbf{I}_N + \mathcal{E}$.

From the perspective of regression families, as shown in Table 4.2, we divide these algorithms into six groups, including the simple linear tensor regression model, generalized linear tensor regression, tensor regression with shrinkage methods, the Bayesian based tensor regression model, quantile tensor regression, and projection-based tensor regression models. They are all counterparts in the tensor field of traditional regression methods partly listed in Table 3.1. For example, the first simple linear tensor regression model is a direct extension of simple linear regression. It regresses the predictor and the response with no additional assumptions on the coefficient tensor. Generalized linear tensor regression takes account of a large range of probability distributions over the response, such as the normal distribution in simple linear regression, the binomial distribution, the Poisson distribution, and the Gamma distribution. They all belong to the exponential family with appropriately selected functions. The simple linear tensor regression can be seen as a special form of generalized tensor linear models. Moreover, to enhance model robustness and improve model performance, different priors are considered within the shrinkage methods. As shown in Table 4.2, tensor regression models with penalty terms are widely concerned, especially tensor ridge regression and sparse tensor regression. The explanation may be that only low-rank assumptions on the coefficient tensor cannot tackle the situation that the sample number is far less than the number of model parameters.

Additionally, Bayesian learning approaches are developed for multidimensional regression problems based on the consideration of "self-calibrating", which greatly reduces the time for parameter tuning. At present, the solutions of most Bayesian approaches for tensor regression learning are based on Gibbs sampling. Although the amount of storage and computation required per iteration is very small, the convergence rate is very slow. Quantile tensor regression extends the mean regression into quantile regression, which is suitable when there are outliers or the distribution of the response is skewed. Instead of mapping the response directly over the predictor, the projection-based tensor regression models explore the relationship between the latent factors of the predictor and the response under some tensor decompositions. In this way, it can tackle some problems that could not be solved by the ordinary linear tensor regression methods, especially for the collinear datasets. Specifically, for collinear data, if the ordinary linear tensor regression method is used, the correlation of the independent variables will endanger the parameter estimation, expand the model error, and damage the stability of the model. The projection based methods use the tensor decomposition methods to extract the principal component (PCR) or most explanatory variables for the dependent variable (PLS), identifying the useful information and noise in the system, so as to better overcome the adverse effects caused by correlations between explanatory variables.

From the perspective of low-rank tensor representation methods, as shown in Table 4.2, the adopted low-rank assumptions include the traditional tensor decomposition methods CP and Tucker and the tensor networks like the tensor train. The traditional decomposition methods are widely studied in almost all kinds of tensor regression models, while there is little literature based on the tensor networks. Up to now, there has been no certain answer to which decomposition method is the best approximation for multidimensional data. The traditional tensor decomposition models are relatively simple extensions of matrix decomposition and enjoy the ease of physical interpretability. But when the data order is high, the number of model parameters required for Tucker decomposition will increase exponentially with the data order, which brings a huge amount of computation and storage. The biggest problem with CP decomposition is that it cannot explore the multilinear rank of the data. For regression problems, this restriction can easily lead to overfitting or under-fitting. The newly developed tensor networks are known for their flexible decomposition form and strong compression capability, and they will not suffer from the dimensionality curse. However, in the tensor network form, the contraction process of complicated networks in the optimization process will take a long time. This is a major difficulty in current research for tensor network-based applications.

From the perspective of optimization methods, they are mainly tackled by the classical ADMM, ALS, Greedy, PGD, and Bayesian inference methods. ADMM usually deals with the convex optimization problem with decomposable objective functions, such as the low-rankness, sparsity, or total variation constrained problems. The main concept for ADMM is to break the optimization problem into several sub-problems, which are much easier to handle or even have popular solutions. Taking the low-rankness regularized regression problems as an example, the ADMM algorithm first separates the data fitting term and the rank minimization term by introducing an auxiliary variable. Then the rank minimization subproblem can be achieved by performing the truncated SVD over the tensor unfolding matrices.

Instead of reducing the tensor rank directly, there is also another solution which first admits the original coefficient tensor into a specific decomposition form like CP with a given rank, and then the problem becomes a low-rank approximation problem. In this case, there would exist several unknown latent factors in the resulting problem, which makes it non-convex. ALS algorithm is a solution for this type of problem, which optimizes one variable with other variables fixed. Popular solutions include MLTR (Hoff, 2015), TTR-CP (Lock, 2018), TTR-TT (Liu *et al.*, 2020c), hrTRR (Guo *et al.*, 2012), and STORE (Sun and Li, 2017). But the tensor rank needs to be given in advance for these approaches. Estimating the appropriate tensor rank, especially the multilinear rank, would cost a lot of time. Therefore, some Bayesian inference methods are employed to achieve rank estimation during the training process and reduce the time for parameter tuning (Guhaniyogi *et al.*, 2017; Billio *et al.*, 2018).

Based on the Greedy learning framework, Bahadori *et al.* (2014) and He *et al.* (2018) both sequentially add a rank-1 component into the desired model, until the required loss converges within a tolerance. Since there is no need to perform SVD over large matrices as in the ADMM framework for low-rankness constrained problems, Greedy algorithms are efficient especially when the true rank of the original coefficient tensor is low. Methods based on PGD update the desired coefficient tensor using the gradient descent method and then project the estimation at each iteration onto a subspace with low rank or sparse properties, as in Yu and Liu (2016), Ahmed *et al.* (2019), and Chen *et al.* (2019). In this way, the projection step is actually a low rank approximation problem, which has many existing solutions (Cichocki *et al.*, 2016; Sidiropoulos *et al.*, 2017).

# 5

# Nonlinear tensor regression

In Chapter 4, we mainly talk about the framework and popular solvers of linear tensor regression, which assumes the dependent variable is modeled by a linear combination of the independent variables. But in practice, the relationship between the input and output of the regression system is complex and can hardly be characterized by simple linear functions. The reason for the popularity of linear regression models is due to their simplicity and ease of interpretation.

In this chapter, we will go beyond linearity and discuss some popular nonlinear tensor regression models. Fig. 5.1 gives a taxonomy of these nonlinear models, which will be introduced in the following part one by one.

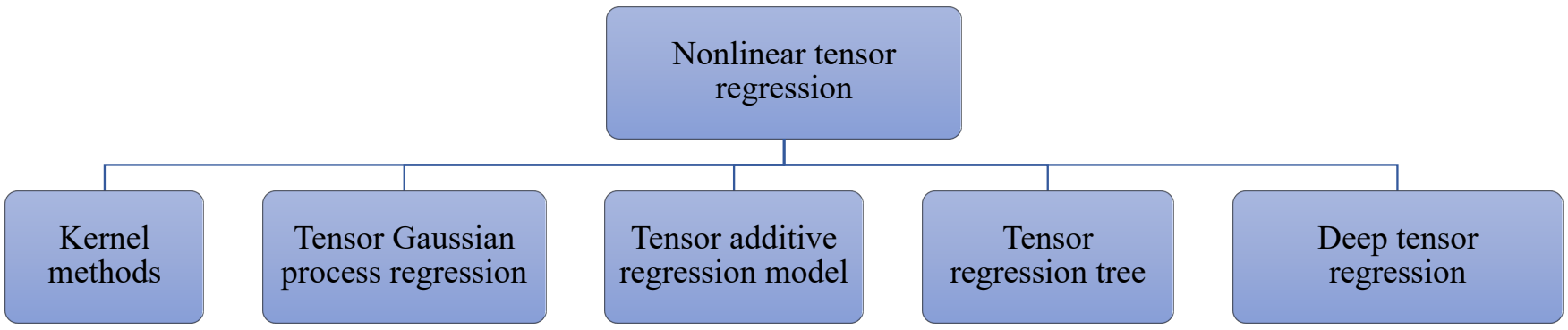

**Figure 5.1:** A taxonomy of nonlinear tensor regression models.

## 5.1 Kernel methods

In order to model the nonlinear relationship between input variables and output variables, one simple idea is to transform the input variable into a new space through a user-defined function, then the output variables can be linearly combined by the newly obtained features.

Specifically, given $N$ training samples $\{\mathcal{X}_n, \mathcal{Y}_n\}_{n=1}^N$, we transform the input variables into a new space through a function $\Phi$, then a new training dataset can be obtained as $\{\mathcal{Z}_n, \mathcal{Y}_n\}_{n=1}^N$ with $\mathcal{Z}_n = \Phi(\mathcal{X}_n)$. The regression model between the input variables and output variables is represented as

$$\mathcal{Y}_n = f(\mathcal{X}_n) + \mathcal{E}_n = \langle \mathcal{Z}_n, \mathcal{B} \rangle_L + \mathcal{E}_n, \tag{5.1}$$

where $f(\cdot)$ represents the nonlinear relationship between $\mathcal{X}_n$ and $\mathcal{Y}_n$. Mapping $\mathcal{X}_n$ into $\mathcal{Z}_n$ through $\mathcal{Z}_n = \Phi(\mathcal{X}_n)$, $\mathcal{Y}_n$ can be obtained by a linear combination of elements in $\mathcal{Z}_n$, and $\mathcal{B}$ corresponds to the weight of this linear combination. So the main problem becomes to find a suitable function $\Phi(\cdot)$ then the subsequent regression analysis can be achieved by linear models as $\mathcal{Y}_n = \langle \Phi(\mathcal{X}_n), \mathcal{B} \rangle_L + \mathcal{E}_n$.

### 5.1.1 Kernelized tensor regression

One representative work for kernelized tensor regression is the kernelized tensor partial least squares regression (KTPLS) (Zhao *et al.*, 2013b). Instead of factorizing the input and output directly, KTPLS first maps the input $\mathcal{X}_n \in \mathbb{R}^{P_1 \times \cdots \times P_L}$ and output $\mathcal{Y}_n \in \mathbb{R}^{Q_1 \times \cdots \times Q_M}$ into a new space through a nonlinear function, then finds their relationship in the newly obtained spaces. First, we need to define two mapping functions $\Phi : \mathcal{X}_n \rightarrow \Phi(\mathcal{X}_n) \in \mathbb{R}^{P'_1 \times \cdots \times P'_L}$, $\Psi : \mathcal{Y}_n \rightarrow \Psi(\mathcal{Y}_n) \in \mathbb{R}^{Q'_1 \times \cdots \times Q'_M}$. Then you will get a new dataset $\{\Phi(\mathcal{X}_n), \Psi(\mathcal{Y}_n)\}_{n=1}^N$. For the new dataset, we let $\Phi \in \mathbb{R}^{N \times P'_1 \times \cdots \times P'_L}$ with $\Phi(n, :, \cdots, :) = \Phi(\mathcal{X}_n)$, $\Psi \in \mathbb{R}^{N \times Q'_1 \times \cdots \times Q'_M}$ with $\Psi(n, :, \cdots, :) = \Psi(\mathcal{Y}_n)$. Employing the tensor partial least squares model for the new dataset, we can get

$$\begin{aligned} \Phi &= \mathcal{G}_\Phi \times_1 \mathbf{T} + \mathcal{E}_\Phi, \\ \Psi &= \mathcal{G}_\Psi \times_1 \mathbf{C} + \mathcal{E}_\Psi, \\ \mathbf{C} &= \mathbf{T}\mathbf{D} + \mathbf{E}_\mathbf{C}. \end{aligned} \tag{5.2}$$

Then the $r$-th component of latent factors $\mathbf{T}$ and $\mathbf{C}$ can be obtained by

$$\begin{aligned} \max_{\mathbf{u}_r,\mathbf{v}_r} \quad & \mathrm{Cov}(\mathbf{t}_r, \mathbf{c}_r) \\ \text{s.t.} \quad & \mathbf{t}_r = \Phi_{r(1)}\mathbf{u}_r, \mathbf{c}_r = \Psi_{r(1)}\mathbf{v}_r \text{ for } r = 1, \cdots, R, \end{aligned} \tag{5.3}$$

where $\Phi_r = \Phi - \sum_{j=1}^{r-1} \mathbf{t}_j \circ \mathbf{u}_j$, $\Psi_r = \Psi - \sum_{j=1}^{r-1} \mathbf{c}_j \circ \mathbf{v}_j$, $\Phi_{r(1)}$ and $\Psi_{r(1)}$ is mode-1 unfolding matrices of $\Phi$ and $\Psi$ respectively. The solution for Eq. (5.3) is given by

$$\begin{aligned} &\Phi_{r(1)}\Phi_{r(1)}^{\mathrm{T}}\Psi_{r(1)}\Psi_{r(1)}^{\mathrm{T}}\mathbf{t}_r = \lambda\mathbf{t}_r, \\ &\mathbf{c}_r = \Psi_{r(1)}\Psi_{r(1)}^{\mathrm{T}}\mathbf{t}_r. \end{aligned}$$

It can be seen that only inner product $\Phi_{r(1)}\Phi_{r(1)}^{\mathrm{T}}$ and $\Phi_{r(1)}\Phi_{r(1)}^{\mathrm{T}}$ sized of $N \times N$ is required to obtain the latent factors. It means that we do not need to determine the specific expression of function $\Phi(\cdot)$ and $\Psi(\cdot)$, we only need the inner product of the obtained features in the new space. That is what kernel functions do. Here, we define $\mathbf{K}_x(n, n') = k(\mathcal{X}_n, \mathcal{X}_{n'})$, $\mathbf{K}_y(n, n') = k(\mathcal{Y}_n, \mathcal{Y}_{n'})$. Then the solution is rewritten as $\mathbf{K}_x\mathbf{K}_y\mathbf{t}_r = \lambda\mathbf{t}_r$,$\mathbf{c}_r = \mathbf{K}_y\mathbf{t}_r$, which greatly reduces the computational complexity. The prediction for the newly arrived sample $\mathcal{X}'$ is

$$\mathrm{vec}(\mathcal{Y}') = \mathbf{Y}^{\mathrm{T}}\mathbf{T}(\mathbf{C}^{\mathrm{T}}\mathbf{K}_x\mathbf{T})^{-1}\mathbf{C}^{\mathrm{T}}\mathbf{k}', \tag{5.4}$$

where $\mathbf{Y} \in \mathbb{R}^{N \times Q_1 \cdots Q_M}$, $\mathbf{Y}(n, :) = \mathrm{vec}(\mathcal{Y}_n)$, $\mathbf{k}'(n) = k(\mathcal{X}', \mathcal{X}_n)$. As shown in Eq. (5.4), the prediction for newly arrived samples is in fact a linear combination of the dependent variable of observed samples, and the weight of each sample is affected by the similarity of corresponding explanatory variables of the observed one and the newly arrived one, namely $k(\mathcal{X}', \mathcal{X}_n)$. As illustrated, the selection of kernel functions will directly affect the performance of the kernelized model. Therefore, how to construct a suitable kernel function for tensor data is a prominent problem in the kernelization of tensor regression. To this end, we will explain in detail in the next subsection and do not repeat it here.

Besides the KTPLS, Rabusseau and Kadri (2016) extended the tensor-on-vector regression model into a kernelized version as

$$\mathcal{Y} = \mathcal{B} \times_1 \Phi(\mathbf{X}) + \mathcal{E},$$

where $\mathbf{X} \in \mathbb{R}^{N \times P}$, $\Phi(\mathbf{X}) \in \mathbb{R}^{N \times P'}$. Since the input variables are represented in vector form, the kernelization of this model is similar to traditional kernelized regression in the vector or matrix field. Meanwhile, by introducing alternative loss functions, support tensor machines (Cai *et al.*, 2006) for classification tasks are extended to regression problems in Gao and Wu (2012). Gao and Wu (2012) defined a non-linear mapping function for 2-order tensors as follows

$$\begin{aligned} k(\mathbf{X}_n, \mathbf{X}_{n'}) &= \Phi(\mathbf{X}_n)\Phi(\mathbf{X}_{n'})^{\mathrm{T}} \\ &= \begin{bmatrix} \Phi(\mathbf{x}_{n1}) \\ \cdots \\ \Phi(\mathbf{x}_{nP_1}) \end{bmatrix} \begin{bmatrix} \Phi(\mathbf{x}_{n'1}) \\ \cdots \\ \Phi(\mathbf{x}_{n'P_1}) \end{bmatrix}^{\mathrm{T}} \\ &= \begin{bmatrix} \Phi(\mathbf{x}_{n1})\Phi(\mathbf{x}_{n'1})^{\mathrm{T}} & \cdots & \Phi(\mathbf{x}_{n1})\Phi(\mathbf{x}_{n'P_1})^{\mathrm{T}} \\ \vdots & \ddots & \vdots \\ \Phi(\mathbf{x}_{nP_1})\Phi(\mathbf{x}_{n'1})^{\mathrm{T}} & \cdots & \Phi(\mathbf{x}_{nP_1})\Phi(\mathbf{x}_{n'P_1})^{\mathrm{T}} \end{bmatrix}, \end{aligned} \tag{5.5}$$

where $\mathbf{x}_{np_1}$ is the $p_1$-th row of $\mathbf{X}_n \in \mathbb{R}^{P_1 \times P_2}$, $\Phi(\mathbf{X}_n) \in \mathbb{R}^{P_1 \times P_2'}$. It can be seen that the kernel function defined in Eq. (5.5) results in a matrix instead of a scalar as normal settings. The corresponding regression model employed is the bilinear model as

$$y_n = \langle \mathbf{B}, \Phi(\mathbf{X}_n) \rangle + \epsilon = \langle \mathbf{u} \circ \mathbf{v}, \Phi(\mathbf{X}_n) \rangle + \epsilon = \mathbf{u}^{\mathrm{T}} \Phi(\mathbf{X}_n) \mathbf{v} + \epsilon,$$

where $\mathbf{B}$ is assumed to be rank-1 tensor as $\mathbf{B} = \mathbf{u} \circ \mathbf{v}$, $\mathbf{u} \in \mathbb{R}^{P_1}$ is the left projection vector, $\mathbf{v} \in \mathbb{R}^{P_2'}$ is the right projection vector.

### 5.1.2 Kernel functions for tensorial data

As discussed in the last subsection, there is no need to know the explicit expression of function $\Phi(\cdot)$ for prediction in kernelized tensor regression. Only the inner product in feature space $\langle \Phi(\mathcal{X}), \Phi(\mathcal{X}') \rangle$ is required. The key idea in kernel methods is to replace the inner product in feature space by a function over the input variables as $k(\mathcal{X}, \mathcal{X}')$, which actually measures the similarity of two input variables in higher-dimensional space.

**Definition 5.1 (Kernel functions).** The kernel functions of given tensors $\mathcal{X}$ and $\mathcal{Z}$ in same size $I_1 \times \cdots \times I_d \times \cdots \times I_D$ can be denoted as

$$k(\mathcal{X}, \mathcal{Z}) = \langle \Phi(\mathcal{X}), \Phi(\mathcal{Z}) \rangle, \tag{5.6}$$

where $\Phi$ is a mapping function

$$\Phi : \mathcal{X} \rightarrow \Phi(\mathcal{X}) \in \mathbb{R}^{I_1' \times \cdots \times I_D'}.$$

The simplest idea to construct kernel functions for tensorial data is to reshape the tensor into a vector and then use the existing kernel functions, such as the linear kernel and Gaussian-RBF kernel.

$$\text{linear kernel:} \qquad k(\mathcal{X}, \mathcal{X}') = \langle \text{vec}(\mathcal{X}), \text{vec}(\mathcal{X}') \rangle, \tag{5.7}$$

$$\text{Gaussian kernel:} \qquad k(\mathcal{X}, \mathcal{X}') = \exp(-\frac{1}{2\sigma^2} \|\mathcal{X} - \mathcal{X}'\|_{\text{F}}^2). \tag{5.8}$$

However, it cannot exploit the multidimensional structure of high dimensional data. Based on the recent development of tensor decompositions, Signoretto *et al.* (2011) considered measuring the similarity of tensorial data through measuring the similarity of their unfolding matrices

$$k(\mathcal{X}, \mathcal{X}') = \sum_{d=1}^{D} k(\mathbf{X}_{(d)}, \mathbf{X}'_{(d)}), \tag{5.9}$$

where $\mathbf{X}_{(d)}$ is mode-$d$ unfolding matrix of $\mathcal{X}$. Meanwhile, the probabilistic product kernels are proposed in Zhao *et al.* (2013b),

$$k(\mathcal{X}, \mathcal{X}') = \alpha^2 \sum_{d=1}^{D} \exp(-\frac{1}{2\beta_d^2} \, \text{S}_d(\mathcal{X} \| \mathcal{X}')), \tag{5.10}$$

where $\text{S}_d(\mathcal{X}\|\mathcal{X}')$ represents the similarity of $\mathcal{X}$ and $\mathcal{X}'$ in mode-$d$, which is measured by the information divergence between the distributions of $\mathcal{X}$ and $\mathcal{X}'$, $\alpha$ is a magnitude parameter and $\beta_d$ denotes length-scales parameter.

Besides, instead of using unfolding matrices, some works employ the latent factors. Signoretto *et al.* (2011) first applied the SVD over the mode-$d$ unfolding matrices of $\mathcal{X}$ as

$$\begin{aligned} \mathbf{X}_{(d)} &= \mathbf{U}_{\mathcal{X}}^{(d)} \Sigma_{\mathcal{X}}^{(d)} (\mathbf{V}_{\mathcal{X}}^{(d)})^{\text{T}}, \\ \mathbf{X}'_{(d)} &= \mathbf{U}_{\mathcal{X}'}^{(d)} \Sigma_{\mathcal{X}'}^{(d)} (\mathbf{V}_{\mathcal{X}'}^{(d)})^{\text{T}}, \end{aligned}$$

then measures the similarity of $\mathbf{V}_{\mathcal{X}}^{(d)}$ and $\mathbf{V}_{\mathcal{X}'}^{(d)}$ over all modes. The mathematical expression based on Chordal distance is as follows

$$k(\mathcal{X}, \mathcal{X}') = \sum_{d=1}^{D} \exp(-\frac{1}{2\beta_d^2} \|\mathbf{V}_{\mathcal{X}}^{(d)} (\mathbf{V}_{\mathcal{X}}^{(d)})^{\text{T}} - \mathbf{V}_{\mathcal{X}'}^{(d)} (\mathbf{V}_{\mathcal{X}'}^{(d)})^{\text{T}} \|_{\text{F}}^2). \tag{5.11}$$

Moreover, the latent factors after TT factorization is also considered (Chen *et al.*, 2020)

$$k(\mathcal{X},\mathcal{X}') = \sum_{d=1}^{D}\sum_{r_d=1}^{R_d}\sum_{r_{d+1}=1}^{R_{d+1}} k(\mathcal{G}_{\mathcal{X}}^{(d)}(r_d,:,r_{d+1}),\mathcal{G}_{\mathcal{X}'}^{(d)}(r_d,:,r_{d+1})), \quad (5.12)$$

where $\mathcal{G}_{\mathcal{X}}^{(d)}$ and $\mathcal{G}_{\mathcal{X}'}^{(d)}$ are the $d$-th core factor of tensor $\mathcal{X}$ and $\mathcal{X}'$ respectively, $\{R_1,\cdots,R_{D+1}\}$ is the tensor train rank. Furthermore, Ragusa *et al.* (2019) developed a similarity measure function between two tensors based on eigenvalues of aligned basis constituted by the extremal energy directions. Specifically, for tensors $\mathcal{X}$ and $\mathcal{X}'$, we first decompose the $\mathcal{X}$ to get its orthogonal bases

$$\mathcal{X} = \tilde{\mathcal{X}} \times_1 \mathbf{U}^{(1)} \times_2 \cdots \times_D \mathbf{U}^{(D)}.$$

Then a pseudo tensor of $\mathcal{X}'$ is obtained as $\tilde{\mathcal{X}}' = \mathcal{X}' \times_1 {\mathbf{U}^{(1)}}^{\mathrm{T}} \times_2 \cdots \times_D {\mathbf{U}^{(D)}}^{\mathrm{T}}$. In this way, the similarity function $k(\mathcal{X},\mathcal{X}')$ is expressed as

$$k(\mathcal{X},\mathcal{X}') = \frac{2}{1+\sigma} - 1 \quad (5.13)$$

with

$$\sigma = \sqrt{\sum_{d=1}^{D}\sum_{i_d=1}^{I_d-\zeta_d} \frac{(\sigma_{i_d}-\sigma'_{i_d})^2}{\sigma_{i_d}\sigma'_{i_d}}},$$

where $\sigma_{i_d} = \|\tilde{\mathbf{X}}_{(d)}(i_d,:)\|$, $\sigma'_{i_d} = \|\tilde{\mathbf{X}}'_{(d)}(i_d,:)\|$, $\tilde{\mathbf{X}}_{(d)}$ and $\tilde{\mathbf{X}}'_{(d)}$ are mode-$d$ unfolding matrices of $\tilde{\mathcal{X}}$ and $\tilde{\mathcal{X}}'$, respectively. The parameters $\zeta_d$ implies that the decomposition process of $\mathcal{X}$ can be implemented with reduced dimensionality and will reduce the computational complexity of the similarity function.

In fact, in addition to being widely used in tensor regression, tensor kernel functions can also be used for many other tasks, such as solving $\ell_p$ norms regularization (Salzo *et al.*, 2018) and multi-view classification (Houthuys and Suykens, 2021). Those who are interested in these topics can go to the specific paper for a detailed illustration.

## 5.2 Tensor Gaussian process regression

Gaussian process regression is a nonparametric regression model which conducts regression analysis through estimation of Gaussian process models. The main assumption in Gaussian process regression is the Gaussian process prior over the regression functions. Specifically, given samples $\{\mathcal{X}_n, y_n\}_{n=1}^N$, the nonparametric regression model can be generalized as

$$y_n = f(\mathcal{X}_n) + \epsilon_n. \tag{5.14}$$

Assuming $\epsilon_n \sim \mathcal{N}(0, \sigma)$, the distribution of the observations is as

$$\mathbf{y} \sim \prod_{n=1}^{N} \mathcal{N}(y_n | f(\mathcal{X}_n), \sigma), \tag{5.15}$$

where $\mathbf{y} = [y_1, \cdots, y_N]$. In Zhao *et al.* (2014), the Gaussian process prior is imposed over the nonlinear function $f(\cdot)$ as

$$f(\mathcal{X}_n) \sim \mathrm{GP}(m(\mathcal{X}_n), k(\mathcal{X}_n, \mathcal{X}_{n'})|\theta), \tag{5.16}$$

where $m(\mathcal{X}_n)$ is the mean function, $k(\mathcal{X}_n, \mathcal{X}_{n'})$ is the covariance function, $\theta$ is a hyperparameter. The covariance function $k(\mathcal{X}_n, \mathcal{X}_{n'})$ is the same as the kernel functions discussed in last section. Different selections of $k(\mathcal{X}_n, \mathcal{X}_{n'})$ will effect the performance of subsequent regression analyses.

Under this assumption, when a new sample $\mathcal{X}'$ arrives, the prediction function $f(\mathcal{X}')$ obeys a Gaussian distribution as

$$f(\mathcal{X}') \sim \mathcal{N}(m', \sigma'^2), \tag{5.17}$$

where

$$m' = \mathbf{k}'^{\mathrm{T}}(\mathbf{K} + \sigma\mathbf{I})^{-1}\mathbf{y},$$
$$\sigma'^2 = k'' - \mathbf{k}'^{\mathrm{T}}(\mathbf{K} + \sigma\mathbf{I})^{-1}\mathbf{k}',$$

with $\mathbf{k}'(n) = k(\mathcal{X}', \mathcal{X}_n)$, $\mathbf{K}(n, n') = k(\mathcal{X}_n, \mathcal{X}_{n'})$, $k'' = k(\mathcal{X}', \mathcal{X}')$, $n = 1, \cdots, N$. As shown in Eq. (5.17), Gaussian process regression not only provides the expectation of the prediction, but also provides the uncertainty of the prediction.

In addition, an online extension of this method is proposed in Hou *et al.* (2015), namely online local Gaussian process for tensor-variate regression (OLGP). At first, it classifies the training data samples into $K$ local small-sized experts through measuring the similarity of the input variables, namely the kernel function $k(\mathcal{X}_n, \mathcal{X}_{n'})$. When a new input variable $\mathcal{X}'$ arrives, OLGP computes its similarity with the center of each local expert $k(\mathcal{X}', \mathcal{C}_k)$ and selects the nearest expert for subsequent prediction, where $\mathcal{C}_k$ is the center of $k$-th local expert. In this way, only several small-sized local experts are employed for the prediction of the newly arrived sample, which naturally reduces the computational complexity and benefits the processing of large-scale datasets.

Moreover, Yu *et al.* (2018) discussed the relationship between the low rank tensor regression model and the Gaussian process and developed a probabilistic counterpart of low rank tensor regression, namely multilinear Gaussian process (MLGP). It proves that the optimizer of low rank tensor regression and the estimator of the multilinear Gaussian process in fact pursue the same latent feature representation and result in the same prediction if the employed kernel enjoys low rankness. Specifically, considering from weight-space,

$$y_n = \langle \text{vec}(\mathcal{B}), \Phi(\mathcal{X}_n) \rangle + \epsilon_n,$$

where $\mathcal{B}$ is assumed to be low Tucker rank as $\mathcal{B} = \mathcal{G} \times_1 \mathbf{U}^{(1)} \times_2 \mathbf{U}^{(2)} \times_3 \mathbf{U}^{(3)}$, here the order of $\mathcal{B}$ is assumed to be 3. If $\text{vec}(\mathcal{G}) \sim \mathcal{N}(0, \sigma \mathbf{I})$, then

$$\text{vec}(\mathcal{B}) = \mathbf{U}^{(3)} \otimes \mathbf{U}^{(2)} \otimes \mathbf{U}^{(1)} \text{vec}(\mathcal{G}) \sim \mathcal{N}(0, \sigma \otimes_{d=1}^{3} \mathbf{U}^{(d)} \mathbf{U}^{(d)^{\mathrm{T}}}).$$

Then

$$\begin{aligned} f(\mathcal{X}_n) = \quad & \langle \text{vec}(\mathcal{B}), \Phi(\mathcal{X}_n) \rangle \\ & \sim \mathcal{N}(0, \sigma \Phi(\mathcal{X}_n) \otimes_{d=1}^{3} \mathbf{U}^{(d)} \mathbf{U}^{(d)^{\mathrm{T}}} \Phi(\mathcal{X}_{n'})^{\mathrm{T}}), \end{aligned} \quad (5.18)$$

where $\otimes_{d=1}^{3} \mathbf{U}^{(d)} \mathbf{U}^{(d)^{\mathrm{T}}} = \mathbf{U}^{(1)} \mathbf{U}^{(1)^{\mathrm{T}}} \otimes \mathbf{U}^{(2)} \mathbf{U}^{(2)^{\mathrm{T}}} \otimes \mathbf{U}^{(3)} \mathbf{U}^{(3)^{\mathrm{T}}}$. If starting from the function-space, we can get

$$y_n = f(\mathcal{X}_n) + \epsilon_n,$$

with

$$f(\mathcal{X}_n) \sim \text{GP}(0, \mathbf{K}(n, n')). \quad (5.19)$$

Comparing Eq. (5.18) and Eq. (5.19), defining the kernel function in Eq. (5.19) as

$$\mathbf{K} = \Phi(\mathcal{X}) \otimes_{d=1}^{3} \mathbf{K}_d \Phi(\mathcal{X})^{\mathrm{T}},$$

with $\mathbf{K}_d = \mathbf{U}^{(d)} \mathbf{U}^{(d)^{\mathrm{T}}}$ for $d = 1, 2, 3$, we can get the same result as Eq. (5.18). In this way, through estimating the hyperparameters in the multilinear Gaussian process with low rank kernels, we can achieve an approximation of the low-rank tensor regression optimization with a confidence interval. It has also proven the oracle inequality. With a defined space for the tensor of functionals $\mathcal{B}$ as

$$\begin{aligned}\mathscr{C}_N = \{\mathcal{B} : \quad & \mathcal{B} = \mathcal{G} \times_1 \mathbf{U}^{(1)} \times_2 \mathbf{U}^{(2)} \times_3 \mathbf{U}^{(3)}, \\ & \|\mathcal{G}_{(1)}\|_* = O(\frac{N}{P_2 P_3 + \log(P_1 P_2 P_3)})^{1/4}\},\end{aligned}$$

letting the $\hat{\mathcal{B}}$ be the estimator achieved by minimizing the empirical risk $\mathfrak{L}(f(\mathcal{X}_n, \mathcal{B}), y_n)$ over the space $\mathcal{B} \in \mathscr{C}_N$, then the excess risk satisfies

$$\mathfrak{L}(\hat{\mathcal{B}}) - \inf_{\mathcal{B} \in \mathscr{C}_N} (\mathfrak{L}(\mathcal{B})) \xrightarrow{P} 0. \tag{5.20}$$

The asymptotic results indicate that the excess risk will approach 0 with a large sample number $N$ and an assumption over $\mathcal{G}$. In addition, the derived non-asymptotic learning curve for the MLGP model under both full rank and low rank scenarios has confirmed the dependency of performance of MLGP over the covariance function along different tasks or features and also the number of samples in each task.

Furthermore, the Gaussian process is also employed for higher-dimensional output regression models in Zhe *et al.* (2019). Given samples $\{\mathbf{x}_n, \mathcal{Y}_n\}_{n=1}^N$, the high order Gaussian process regression (HOGP) model is

$$\mathcal{Y}_n = \mathcal{G} \times_1 \mathbf{U}^{(1)} \times_2 \cdots \times_D \mathbf{U}^{(D)} \times_{D+1} \Phi(\mathbf{x}_n)^{\mathrm{T}} + \mathcal{E}.$$

Similarly, if $\mathrm{vec}(\mathcal{G}) \sim \mathcal{N}(0, \mathbf{I})$,

$$\mathrm{vec}(\mathcal{Y}) \sim \mathcal{N}(0, \Sigma) \tag{5.21}$$

with $\Sigma = \otimes_{d=1}^D \mathbf{U}^{(d)} \mathbf{U}^{(d)\mathrm{T}} \otimes \Phi(\mathbf{X})\Phi(\mathbf{X})^{\mathrm{T}}$, $\mathcal{Y}(:, \cdots, :, n) = \mathcal{Y}_n$, $\mathbf{X}(n, :) = \mathbf{x}_n$ for $n = 1, \cdots, n$. Then the parameters of the derived Gaussian process model can be estimated through Bayesian inference.

An extension to the Bayesian mixed effect model is analyzed in Kia *et al.* (2018). It considers the model

$$\mathcal{Y}_n = \mathcal{A} \times_{D+1} \mathbf{x}_n^{\mathrm{T}} + \mathcal{Z} + \mathcal{E},$$

where $\mathcal{A}$ represents the fixed effect along different subjects, $\mathcal{Z}$ denotes the random effect across different subjects, $\mathcal{E}$ is the noise. The estimation of $\mathcal{A}$ is obtained through tackling the linear equations, namely $\text{vec}(\mathbf{Y}_n) = \mathbf{A}_{<D>}\mathbf{x}_n$ for $n = 1.\cdots, N$. After getting the coefficient tensor $\mathcal{A}$, the MLGP prior with low rank kernels is imposed over the random effect tensor $\mathcal{Z}$ and noise $\mathcal{E}$ for efficient estimation. Experiments on a clinical neuroimaging dataset for Neuropsychiatric Phenomics show the efficiency of the multitask Gaussian process tensor regression model in processing whole-brain neuroimaging data. Predicting the multidimensional output in a mixed effect model also benefits the exploration of variations along different modes.

## 5.3 Tensor additive models

Additive models are a group of nonparametric models that assume the effect of each explanatory variable over the response is independent and either linear or nonlinear. Specifically, for training dataset $\{\mathbf{x}_n \in \mathbb{R}^P, y_n\}$, the additive modes can be represented as

$$y_n = \alpha + f_1(x_{n,1}) + \cdots + f_P(x_{n,P}), \tag{5.22}$$

where $f_p$ for $p = 1, \cdots, P$ are nonparametric smooth functions, $\alpha$ is the bias. Actually, if we define $f_p(x_{n,p}) = \beta_p x_{n,p}$, the model degenerates into the classical linear regression model. However, if additive models relax the linear constraints, the employed functions $f_p$ for $p = 1, \cdots, P$ can be either linear or nonlinear.

When it comes to tensorial data, Kanagawa *et al.* (2016) first started with the rank one tensor, namely $\mathcal{X}_n = \mathbf{x}_n^{(1)} \circ \cdots \circ \mathbf{x}_n^{(L)}$ for $n = 1, \cdots, N$. The corresponding response is expected to be generated by the additive model as

$$y_n = \sum_{r=1}^{R} \prod_{l=1}^{L} {f^*}_r^l(\mathbf{x}_n^{(l)}) + \epsilon_n, \tag{5.23}$$

where $\epsilon_n$ is the independently identically distributed zero-mean noise, $f^*(\mathbf{x}^{(1)}, \cdots, \mathbf{x}^{(L)}) = \sum_{r=1}^{R} \prod_{l=1}^{L} {f^*}_r^l(\mathbf{x}^{(l)})$ is the true function needed to be persued which represents the relationship between the variables $\{\mathbf{x}_n^{(1)}, \cdots, \mathbf{x}_n^{(L)}\}_{n=1}^N$ and the responses $\{y_n\}_{n=1}^N$.

Extending it into rank $M$ tensors $\mathcal{X}_n = \sum_{m=1}^{M} \mathbf{x}_{n,m}^{(1)} \circ \cdots \circ \mathbf{x}_{n,m}^{(L)}$, Imaizumi and Hayashi (2016) proposed the additive-multiplicative non-parametric regression model (AMNR) as

$$y_n = \sum_{m=1}^{M} \sum_{r=1}^{R} \prod_{l=1}^{L} f^{*l}_{r}(\mathbf{x}^{(l)}_{n,m}) + \epsilon_n. \tag{5.24}$$

As shown in Eq. (5.24), the desired function space and the multidimensional input space are simultaneously decomposed, which greatly reduces the model complexity and effect of noise. In order to figure out the appropriate local functions over the latent factors decomposed from the input space, the Bayesian estimator with Gaussian process prior is established. Instead of imposing the GP prior over the function $f(\cdot)$ in Eq. (5.19), additive models assume that each local function $f^{*l}_{r}$ admits a zero-mean multivariate Gaussian distribution

$$f^{*l}_{r} \sim \mathcal{N}(\mathbf{0}, \mathbf{K}^{(l)}_{r}), \tag{5.25}$$

then the prior distribution of $f^{*}(\mathbf{x}^{(1)}, \cdots, \mathbf{x}^{(L)})$ is

$$p(f^{*}) = \prod_{r=1}^{R} \prod_{l=1}^{L} p(f^{*l}_{r}). \tag{5.26}$$

In this way, the posterior distribution can be determined by the likelihood function and the prior through Bayes' rule. But in practice, the support of the assumed Gaussian process prior in an infinite dimensional setting is commonly bigger than the reproducing kernel Hilbert space (RKHS) of the covariance function of input variables. Thus, Kanagawa *et al.* (2016) proposed a Gaussian process prior with scaled kernel function $\mathbf{K}^{(l)}_{r} = \mathbf{K}^{(l)}_{r}/\lambda_{r,k}$. Meanwhile, Kanagawa *et al.* (2016) also imposed a prior distribution over the rank of the estimated function, while Imaizumi and Hayashi (2016) simply used a reduced CP rank for approximation. An alternating minimization procedure for fitting Eq. (5.23) is proposed in Kanagawa *et al.* (2016). The corresponding optimization problem is

$$\min_{\{f^{*l}_{r}\}_{r,l}} \frac{1}{N} \sum_{n=1}^{N} (y_n - \sum_{r=1}^{R} \prod_{l=1}^{L} f^{*l}_{r}(\mathbf{x}^{(l)}_{n}))^2 + C_n \sum_{r=1}^{R} \sum_{l=1}^{L} \Omega(f^{*l}_{r}), \tag{5.27}$$

where $\Omega(f^{*l}_{r})$ is the regularization term. Based on alternating minimization, each local function can be obtained by tracking a variant of the kernel ridge regression model while fixing other local functions.

For nonparametric models, if the desired function belongs to a Sobolev space, namely $f^* \in \mathcal{W}^\beta$, the general optimal convergence rate is $O(N^{-\beta/(2\beta+\prod_{l=1}^{L} P_l)})$, where $\beta$ represents the smoothness of the desired true function, $\prod_{l=1}^{L} P_l$ denotes the total number of entries in each input tensor $\mathcal{X}_n$. Imaizumi and Hayashi (2016) proved the Gaussian process based tensor additive model AMNR can achieve a faster convergence rate as $O(N^{-\beta/(2\beta+\max_l P_l)})$, which is $\prod_{l' \neq l} P_{l'}$ times better. Compared with Bayesian estimation, alternating minimization procedure requires a stronger assumption to satisfy the minimax optimality, but it is computational efficient which avoids the time-consuming posterior sampling (Suzuki *et al.*, 2016).

As described above, for methods in Imaizumi and Hayashi (2016), Kanagawa *et al.* (2016), and Suzuki *et al.* (2016), the kernel function is employed with no requirement to determine the specific expressions of the feature functions. However, for large datasets which possess numerous samples, the computation of kernel functions will be very time-consuming. Thus, some works use basis functions to approximate the true functions using weighting factors. In other words, the true function which needs to be estimated can be represented by basis functions, such as finite multidimensional Fourier series (Wahls *et al.*, 2014), polynomial splines (Hao *et al.*, 2021), and B-splines (Zhou *et al.*, 2020b). Specifically,

$$y_n = f^*(\mathcal{X}_n) + \epsilon_n = \sum_{p_1,\cdots,p_L,h} b_{p_1,\cdots,p_L,h} \Psi_h(x_{n,p_1,\cdots,p_L}) + \epsilon_n, \qquad (5.28)$$

where $x_{n,p_1,\cdots,p_L}$ is $p_1,\cdots,p_L$-th entry of $\mathcal{X}_n \in \mathbb{R}^{P_1\times\cdots\times P_L}$, $\{\Phi_h(\cdot)\}_{h=1}^H$ is a group of basis functions. Concatenating the weighting factors and local functions, we can get

$$y_n = f^*(\mathcal{X}_n) + \epsilon_n = \sum_{h=1}^{H} \langle \mathcal{B}_h, \mathbf{F}^*_{n,h} \rangle + \epsilon_n, \qquad (5.29)$$

where $\mathcal{B}_h(p_1,\cdots,p_L) = b_{p_1,\cdots,p_L,h}$, $\mathbf{F}^*_{n,h}(p_1,\cdots,p_L) = \Psi_h(x_{n,p_1,\cdots,p_L})$. In this way, the additive model in Eq. (5.29) boils down to a simple multilinear tensor regression model with parameter tensors $\{\mathcal{B}_h\}_{h=1}^H$. Then the true function can be estimated through fitting the linear tensor regression model with certain constraints. For example, Hao *et al.* (2021) and Zhou *et al.* (2020b) both assumed the coefficient tensor to be low CP rank, while low rank tensor train approximation is employed in Wahls *et al.* (2014). In addition, the sparsity constraints are enforced over the latent factors to select the important correlated subregions for the response, such as the group sparsity constraint (Hao *et al.*, 2021), and the elastic-net penalty (Zhou *et al.*, 2020b). The alternating minimization method is used to update the latent factors. Wahls *et al.* (2014) also provided statistical analysis for the penalized alternating minimization and derived a non-asymptotic error bound for the estimator from each iteration.

## 5.4 Random forest based tensor regression

Random forest is a bagging algorithm, which belongs to a method of ensemble learning. The general idea of ensemble learning is to train multiple weak models and pack them into a strong model. The performance of the resulting strong model is usually much better than that of a single weak model. Random forest is a collection of multiple decision trees, which can be used for tackling classification and regression tasks. However, one limitation of traditional random forest regression is the lack of reasoning ability. That is to say, the prediction of the test sample is usually limited by the labels of existing samples. In order to tackle this issue, regression models are imported for prediction at each leaf node.

Defining a forest $T = \{T_h\}_{h=1}^{H}$, the training of random forest is achieved by fitting each decision tree $T_h$ with a set of patches that are randomly sampled from the training samples. The construction of each decision tree is a process which finds the best way to split the datasets into subgroups and reach the final decision. For tensorial data $\{\mathcal{X}_n \in \mathbb{R}^{P_1 \times \cdots \times P_L}, \mathcal{Y}_n \in \mathbb{R}^{Q_1 \times \cdots \times Q_M}\}_{n=1}^{N}$, the decision tree is constructed by selecting an important feature and splitting the dataset by a threshold $S$

$$\mathbb{R}_{\text{left}}(\mathbf{p}, S) = \{\mathcal{X}_n | \mathcal{X}_{n,p_1,\cdots,p_L} \leq S\}, \mathbb{R}_{\text{right}}(\mathbf{p}, S) = \{\mathcal{X}_n | \mathcal{X}_{n,p_1,\cdots,p_L} > S\}, \tag{5.30}$$

where $\mathbf{p} = [p_1, \cdots, p_L]$, $p_l = 1, \cdots, P_l$, $l = 1, \cdots, L$. To find the key

features and corresponding best splitting point, Kaymak and Patras (2014) employ the metric information gain, namely the difference between the differential entropy of the set $\mathbb{R}$ at the internal node and the weighted sum of differential entropies at the left child node $\mathbb{R}_{\text{left}}(\mathbf{p}, S)$ and the right child node $\mathbb{R}_{\text{right}}(\mathbf{p}, S)$ after splitting. Mathematically,

$$J = H(\mathbb{R}) - (w_{\text{left}} H(\mathbb{R}_{\text{left}}(\mathbf{p}, S)) + w_{\text{right}} H(\mathbb{R}_{\text{right}}(\mathbf{p}, S))), \tag{5.31}$$

where $\mathbb{R} = \{\mathbb{R}_{\text{left}}(\mathbf{p}, S), \mathbb{R}_{\text{right}}(\mathbf{p}, S)\}$. When the predefined depth of the decision tree is reached or the number of patches in a node is less than a threshold, the splitting process will stop. A tensor regression model is employed to fit the patches in each leaf node

$$\mathcal{Y}_n = \langle \mathcal{X}_n, \mathcal{B} \rangle_L + \mathcal{E}_n, \tag{5.32}$$

where $\mathcal{B}$ admits a low CP rank approximation $\mathcal{B} = \sum_{r=1}^{R} \mathbf{u}^{(1)} \circ \cdots \circ \mathbf{u}^{(L)} \circ \mathbf{v}^{(1)} \circ \cdots \circ \mathbf{v}^{(M)}$, $\mathcal{E}_n$ is the zero-mean Gaussian noise. If the number of leaf nodes in decision tree $T_h$ is $C_h$, you need to fit $\sum_h C_h$ regression models in total to get the final random forest for prediction. A graphical illustration for this tensor based random forest is given in Fig. 5.2.

A similar tensor basis random forest is employed for modeling Reynolds Averaged Navier-Stokes (RANS) simulations. The main difference is that Kaandorp and Dwight (2020) use mean square error to both split the dataset and fit the patches reached in the leaf nodes. It considers the regression model

$$\mathbf{Y}_n = \mathcal{X}_n \times_3 \mathbf{b} + \mathbf{E}_n, \tag{5.33}$$

where $\mathbf{Y}_n \in \mathbb{R}^{3\times 3}$, $\mathcal{X}_n \in \mathbb{R}^{3\times 3\times 10}$, $\mathbf{b} \in \mathbb{R}^{10}$ is the coefficient vector. The cost function of each splitting is represented as

$$J = \sum_{\mathcal{X}_n \in \mathbb{R}_{\text{left}}(\mathbf{p},S)} \|\mathbf{Y}_n - \mathcal{X}_n \times_3 \mathbf{b}_{\text{left}}\|_F^2 + \sum_{\mathcal{X}_n \in \mathbb{R}_{\text{right}}(\mathbf{p},S)} \|\mathbf{Y}_n - \mathcal{X}_n \times_3 \mathbf{b}_{\text{right}}\|_F^2. \tag{5.34}$$

Minimizing the cost function, we can get the weighting parameters $\mathbf{b}_{\text{left}}$ and $\mathbf{b}_{\text{right}}$ through solving two least square problems and get the best splitting channel and point.

In fact, the construction of random forest for tensor regression is time consuming, but suitable for large scale datasets. It reduces the risk of overfitting, is flexible, and it is easy to determine the feature importance.

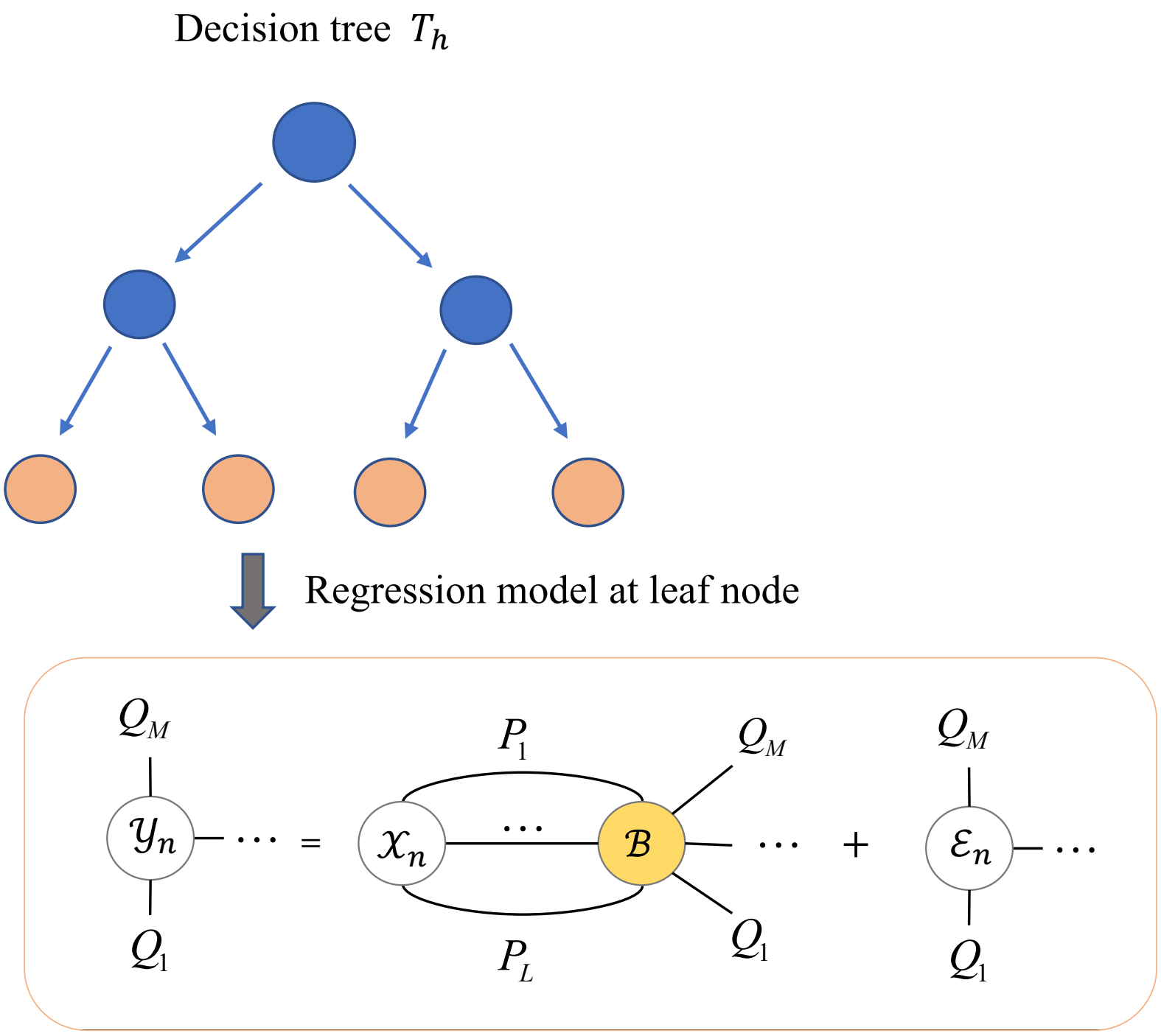

**Figure 5.2:** The diagram description of the random forest based tensor regression.

## 5.5 Deep tensor regression

Deep learning is a new and promising research direction in the field of machine learning. Deep neural networks have powerful representation capabilities and can fit any complex function. It is dedicated to learning the inherent laws of data and mining features useful for specific tasks. Deep learning is a complex machine learning algorithm that has achieved impressive results in related fields such as recommender systems, image and video classification, natural language processing, and object recognition or tracking. However, in addition to its powerful capabilities, numerous network parameters, huge computing requirements, and black box characteristics have also attracted a lot of attention.

To tackle these issues, tensor methods have been employed recently for deep learning networks, such as tensor factorization for network compression (Lebedev *et al.*, 2015; Chen *et al.*, 2018; Kossaifi *et al.*, 2019a; Kossaifi *et al.*, 2020), higher order extensions of existing deep neural networks through exploring multilinear correlations (Chien and Bao, 2017; Su *et al.*, 2020; Blagoveschensky and Phan, 2020; Kolbeinsson *et al.*, 2021; Houthuys and Suykens, 2021), training deep neural networks using tensor learning (Xu *et al.*, 2020; Idelbayev and Carreira-Perpinán, 2020; Hawkins and Zhang, 2021), and understanding and analyzing deep learning using tensor representations (Cohen *et al.*, 2017). Here we only focus on tensor regression networks and higher order extensions of existing deep neural networks.

### 5.5.1 Tensor regression networks

A key aspect of deep learning is its efficiency. To employ deep neural networks in resourced-constricted devices, a lot of works aim to reduce the network parameters while not sacrificing too much performance of trained networks. It is possible to find a compact model to approximate the deeper neural networks due to redundancy (Ba and Caruana, 2013).

Tensor, as a natural presentation of multidimensional data, provides a natural way to represent either the input variables or model weighting parameters through multiple indices. Specifically, the fully connected layer (FCL) is expressed as

$$\mathbf{y} = \mathbf{A}^{\mathrm{T}}\mathbf{x}, \tag{5.35}$$

where $\mathbf{x} \in \mathbb{R}^I$, $\mathbf{y} \in \mathbb{R}^O$ and $\mathbf{A} \in \mathbb{R}^{I \times O}$ is the input, output and parameter matrix of the fully connected layer. Representing in tensor form, it can be rewritten as

$$\mathcal{Y}(o_1, \cdots, o_M) = \sum_{i_1, \cdots, i_L} \mathcal{X}(i_1, \cdots, i_L)\mathcal{A}(i_1, \cdots, i_L, o_1, \cdots, o_M), \tag{5.36}$$

where $\mathcal{X} \in \mathbb{R}^{I_1 \times \cdots \times I_L}$, $\mathcal{Y} \in \mathbb{R}^{O_1 \times \cdots \times O_M}$, and $\mathcal{A} \in \mathbb{R}^{I_1 \times \cdots \times I_L \times O_1 \times \cdots \times O_M}$ with $I = \prod_l I_l$, $O = \prod_m O_m$. In fact, it can be seen as a tensor regression model if a low rank tensor approximation is performed over the parameter tensor $\mathcal{A}$.

Similarly, for convolutional layers (CL), the output tensor $\mathcal{Y} \in \mathbb{R}^{H \times W \times O}$ is generated by a convolutional operation between the input tensor $\mathcal{X} \in \mathbb{R}^{H \times W \times I}$ and a kernel tensor $\mathcal{K} \in \mathbb{R}^{D \times D \times I \times O}$

$$\mathcal{Y}(h, w, o) = \sum_{d_1, d_2, i} \mathcal{X}(h', w', i)\mathcal{K}(d_1, d_2, i, o), \tag{5.37}$$

with $h' = (h-1)s + d_1 - p$, $w' = (w-1)s + d_2 - p$, where $s$ is stride size, $p$ is the zero-padding size. If $I$ and $O$ are large, we can reshape $\mathcal{K}$ into a higher order tensor $\mathcal{K} \in \mathbb{R}^{D \times D \times I_1 \times \cdots \times I_L \times O_1 \times \cdots \times O_M}$ and then decompose it into several latent factors.

Different tensor factorization methods employed for the parameters $\mathcal{A}$ or $\mathcal{K}$ will result in different network forms and present different compression ratios in both parameters and computation (Lebedev *et al.*, 2015; Novikov *et al.*, 2015; Garipov *et al.*, 2016; Yang *et al.*, 2017; Wang *et al.*, 2018; Kossaifi *et al.*, 2020). Fig. 5.3 gives a comparison of the fully connected layer or convolutional layer based on different tensor factorization methods, where all the multilinear ranks are assumed to equal to $R$. Defining $I_s = \sum_{l=1}^{L} I_l$,$O_s = \sum_{m=1}^{M} O_m$, the compression factor in terms of parameters is presented in Table 5.1.

As shown in Fig. 5.3, the parameters that need to be trained become the latent factors instead of the high order coefficient tensor $\mathcal{A}$ in FCL or $\mathcal{K}$ in CL. One issue for tensor regression networks is the partial gradients with respect to each latent factor, which is necessary for implementing

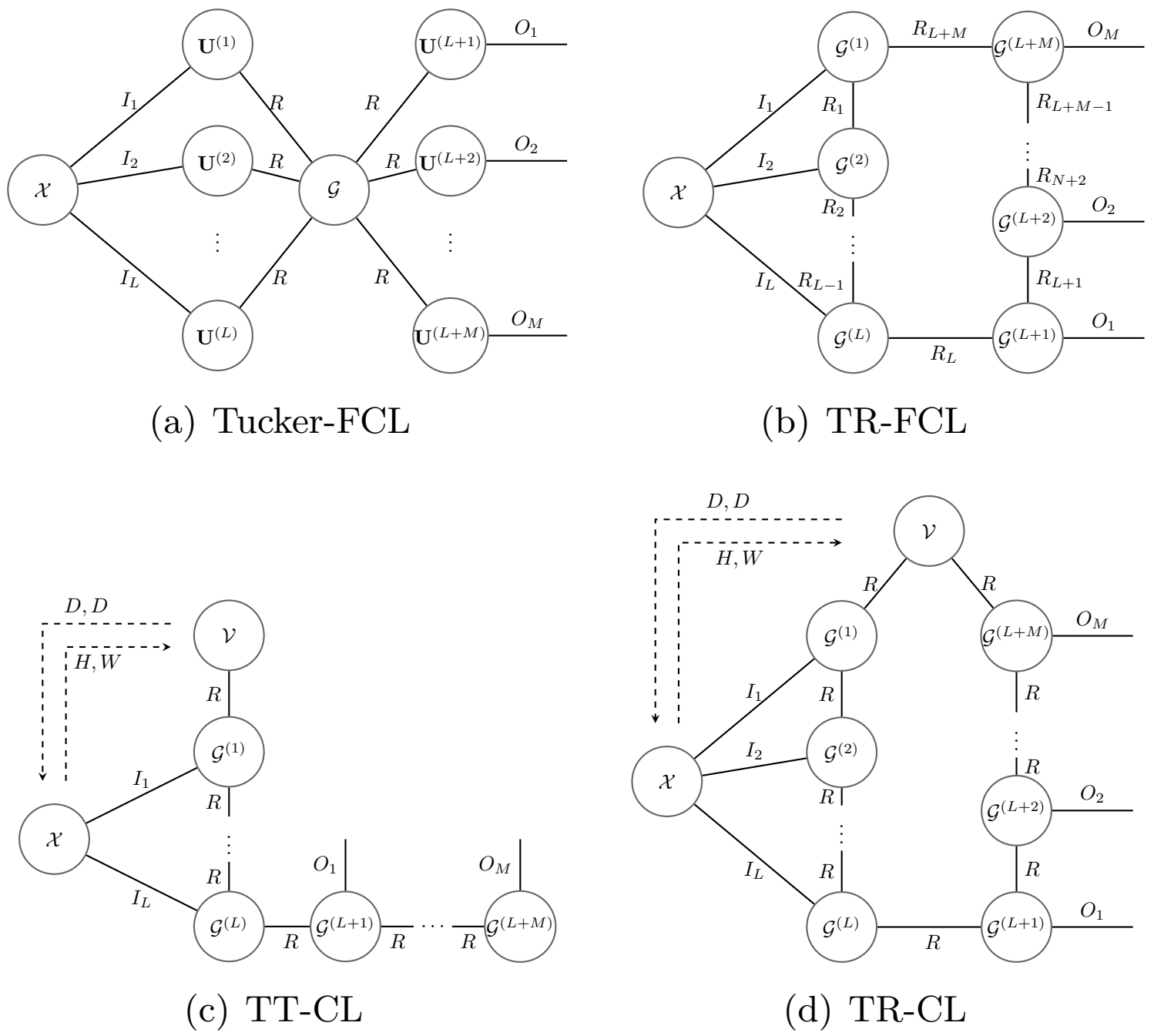

**Figure 5.3:** A comparison of the fully connected layer or convolutional layer based on different tensor factorization methods.

the backpropagation algorithm. Taking the Tucker form for FCL as an example, $\mathcal{A} = \mathcal{G} \times_1 \mathbf{U}^{(1)} \times_2 \cdots \times_{L+M} \mathbf{U}^{(L+M)}$, Eq. (5.36) can be rewritten as

$$\mathrm{vec}(\mathcal{Y}) = f(\mathcal{X}) = \mathbf{U}^{(L+M)} \otimes \cdots \otimes \mathbf{U}^{(L+1)} \mathbf{G}_{<L>}^{\mathrm{T}} \mathbf{W}^{\mathrm{T}} \mathrm{vec}(\mathcal{X}), \quad (5.38)$$

where $\mathbf{G}_{<L>} \in \mathbb{R}^{R_1 \cdots R_L \times R_{L+1} \cdots R_{L+M}}$ is $k$-unfolding matrix of core tensor $\mathcal{G}$, $\mathbf{W} = \mathbf{U}^{(L)} \otimes \cdots \otimes \mathbf{U}^{(1)}$. The partial derivative with respect to $\mathbf{U}^{(l)}$ for $l = 1, \cdots, L$ is

$$\frac{\partial f(\mathcal{X})}{\partial \mathbf{U}^{(l)}}_{(1)} = ([\![\mathcal{G}; \mathbf{U}^{(1)}, \cdots, \mathbf{I}_{R_l}, \cdots, \mathbf{U}^{L+M}]\!])_{<L>}^{\mathrm{T}} (\mathbf{X}_{(l)} \otimes \mathbf{I}_{R_l})^{\mathrm{T}}, \quad (5.39)$$

where $\frac{\partial f(\mathcal{X})}{\partial \mathbf{U}^{(l)}}_{(1)}$ means the mode-1 unfolding matrix of $\frac{\partial f(\mathcal{X})}{\partial \mathbf{U}^{(l)}}$. When $m = 1, \cdots, M$, the partial derivative of $f(\mathcal{X})$ with respect to $\mathbf{U}^{(L+m)}$ is

$$\frac{\partial f(\mathcal{X})}{\partial \mathbf{U}^{(L+m)}}_{(1)} = \mathbf{C}_{(m)}^{\mathrm{T}} \otimes \mathbf{I}_{O_m} \quad (5.40)$$

with

$$\mathcal{C} = \langle \mathcal{X}, [\![\mathcal{G}; \mathbf{U}^{(1)} \cdots, \mathbf{U}^{(L+m-1)}, \mathbf{I}_{R_{L+m}}, \mathbf{U}^{(L+m+1)}, \cdots, \mathbf{U}^{(L+M)}]\!] \rangle_L,$$

where $\mathcal{C} \in \mathbb{R}^{O_1 \times \cdots \times O_{m-1} \times R_{L+m} \times O_{m+1} \times \cdots O_M}$, $\mathbf{C}_{(m)}$ is mode-$m$ unfolding of $\mathcal{C}$. And the partial derivative of $f(\mathcal{X})$ with respect to $\mathcal{G}$ is

$$\frac{\partial f(\mathcal{X})}{\partial \mathcal{G}}_{(1)} = (\mathbf{U}^{(L+M)} \otimes \cdots \otimes \mathbf{U}^{(L+1)}) \otimes (\text{vec}(\mathcal{X})^{\mathrm{T}} \mathbf{W}). \qquad (5.41)$$

In this way, the desired parameters can be learned end-to-end using backpropagation algorithms.

Besides the factorization listed in Table 5.1, there have also been some other attempts at compressing FCL, such as block term decomposition (Ye *et al.*, 2020), hierarchical Tucker (Yin *et al.*, 2020), and randomized CPD or Tucker decomposition (Kolbeinsson *et al.*, 2021). The randomized CPD and Tucker decompositions are achieved by the nonlinear tensor dropout over the low rank subspace, which achieves a stochastic reduction of the rank during the training process and thus avoids overfitting and improves robust learning.

**Table 5.1:** The compression factor (CF) of tensor based FCL and CL.

| CF | CP | Tucker | TT | TR |
|---|---|---|---|---|
| FCL | $\frac{IO}{R(I_s+O_s)}$ | $\frac{IO}{R(I_s+O_s)+R^{L+M}}$ | $\frac{IO}{RI_1+RO_M+R^2(\sum_{l=2}^{L} I_l+\sum_{m=1}^{M-1} O_m)}$ | $\frac{IO}{R^2(I_s+O_s)}$ |
| CL | $\frac{D^2IO}{R(I_s+O_s+2D)}$ | / | $\frac{D^2IO}{RD^2+RO_M+R^2(I_s+\sum_{m=1}^{M-1} O_m)}$ | $\frac{D^2IO}{R^2(I_s+O_s+D^2)}$ |

Moreover, Li *et al.* (2019a) analyzed the influence of spatial similarity over the low rank structure of kernels in convolutional layers. The analysis shows that the kernels can be effectively compressed even when they are randomly shuffled, which implies that the spatial structure is not the key factor in why the kernels can be compressed. Furthermore, some complexity analysis both in the space and time domain is proposed in Wang *et al.* (2018) and Yin *et al.* (2020), and the theoretical analysis in Cao and Rabusseau (2017) shows that the representation power of the decomposed networks is upper-bounded by the ranks of the corresponding tensor approximations. Besides the compression capability, Kossaifi *et al.* (2020) verified the ability of tensor regression networks to retain the multi-modal structure throughout the network, particularly crucial for prediction tasks from MRI data. This is the first paper discussing the ability of multidimensional structural information retention in tensor regression networks.

### 5.5.2 Higher order extensions of recurrent neural networks

One vital reason why we use tensor presentation for regression models is its capability to model multidimensional relatedness. Driven by this, Soltani and Jiang (2016), Zhang and Woodland (2018), and Yu *et al.* (2017) extended the recurrent neural networks into higher order versions through taking the long-term dependency over time into account. Specifically, the recurrent neural network is a type of neural network for modeling sequential data or time series data, which is widely used in social or climate data analysis, natural language processing, and speech recognition. It models dynamical systems using the first order Markovian models,

$$\mathbf{h}_t = f(\mathbf{h}_{t-1}, \mathbf{x}_t, \theta_f) = \phi(\mathbf{W}^{hx}\mathbf{x}_t + \mathbf{W}^{hh}\mathbf{h}_{t-1} + \epsilon^h), \quad (5.42)$$

$$y_t = g(\mathbf{h}_t, \theta_g) = \phi(\mathbf{W}^{yh}\mathbf{h}_t + \epsilon^y), \quad (5.43)$$

where $f$ is the state transition function, $g$ is the output function, $\theta_f$ and $\theta_g$ are the corresponding parameters, $\phi$ is nonlinear activation function, $\mathbf{W}^{hx}, \mathbf{W}^{hh}, \mathbf{W}^{yh}$ are the weighting matrices, $\epsilon^h, \epsilon^y$ are the biases. It can be seen that the current state $\mathbf{h}_t$ is only affected by the last state $\mathbf{h}_{t-1}$ and the input variable $\mathbf{x}_t$.

Extending it into $L$-order Markov processes, Soltani and Jiang (2016) represented the long-term dependency between the current state $\mathbf{h}_t$ and $L$ previous states $\{\mathbf{h}_{t-1}, \cdots, \mathbf{h}_{t-L}\}$ through linear combinations

$$\mathbf{h}_t = \phi(\mathbf{W}^{hx}\mathbf{x}_t + \sum_{l=1}^{L} \mathbf{W}_l \mathbf{h}_{t-l}). \quad (5.44)$$

But this improvement inevitably brings an increase in the number of parameters that need to be estimated in each RNN layer. Thus, Zhang and Woodland (2018) proposed to project the hidden state vectors, namely $\{\mathbf{h}_{t-1}, \cdots, \mathbf{h}_{t-L}\}$, into a small-sized vector with a projection matrix $\mathbf{P}$. Mathematically, the Eq. (5.44) is transformed as

$$\mathbf{h}_t = \phi(\mathbf{W}^{hx}\mathbf{x}_t + \sum_{l=1}^{L} \mathbf{W}_l'\mathbf{P}\mathbf{h}_{t-l}). \tag{5.45}$$

The size of the parameters $\mathbf{W}_l'$ will be reduced in this way. Besides the linear combinations, Yu *et al.* (2017) represented the dependency between the current state and previous states through degree-P polynomial interactions

$$\mathbf{h}_t(j) = \phi(\mathbf{W}^{hx}(j,:)\mathbf{x}_t + \langle \mathcal{W}_j, \underbrace{\mathbf{s}_{t-1} \otimes \cdots \otimes \mathbf{s}_{t-1}}_{P} \rangle), \tag{5.46}$$

where $\mathbf{s}_{t-1}^{\mathrm{T}} = [1, \mathbf{h}_{t-1}^{\mathrm{T}}, \cdots, \mathbf{h}_{t-L}^{\mathrm{T}}]$ is a collection of all the $L$ historical states which are assumed to be related with the current state, $\mathbf{s}_{t-1} \otimes \cdots \otimes \mathbf{s}_{t-1}$ includes all polynomial expansions of hidden states up to order $P$, $\mathcal{W}_j$ indicates the relationship between the $j$-th entry of current state $\mathbf{h}_t$ and the polynomial expansions of hidden states. Fig. 5.4 gives a comparison of the classical RNNs and the higher-order tensor RNNs (Hot-RNN). For dimensionality reduction, the tensor train decomposition is performed over the parameter tensor $\mathcal{W}$ with $\mathcal{W}(j,:,\cdots,:) = \mathcal{W}_j$, reducing the required parameters from $(JL+1)^P$ to $(JL+1)R^2P$ with $R$ being the maximum value of all the tensor train ranks, where $J$ is the size of the hidden states. Based on this, the approximation guarantees and the variance bounds of proposed higher-order RNN with sequential inputs are also established.

In addition, a higher-order extension for a variant of RNN, named convolutional LSTMs, is proposed in Su *et al.* (2020) with a newly defined convolutional tensor train decomposition method to parameterize the convolution operations. The defined higher-order convolutional LSTM is computationally tractable and can explore both the local spatial and long-term temporal dependencies.

## 5.6 Summary

The relevant studies for tensor based nonlinear regression models and algorithms, such as kernelized tensor methods, tensor Gaussian process, tensor additive models, tensor based random forest, and deep based tensor regression models, have been covered in this chapter. They are either higher order extensions of classic nonlinear regression models or extensions of linear tensor regression models.

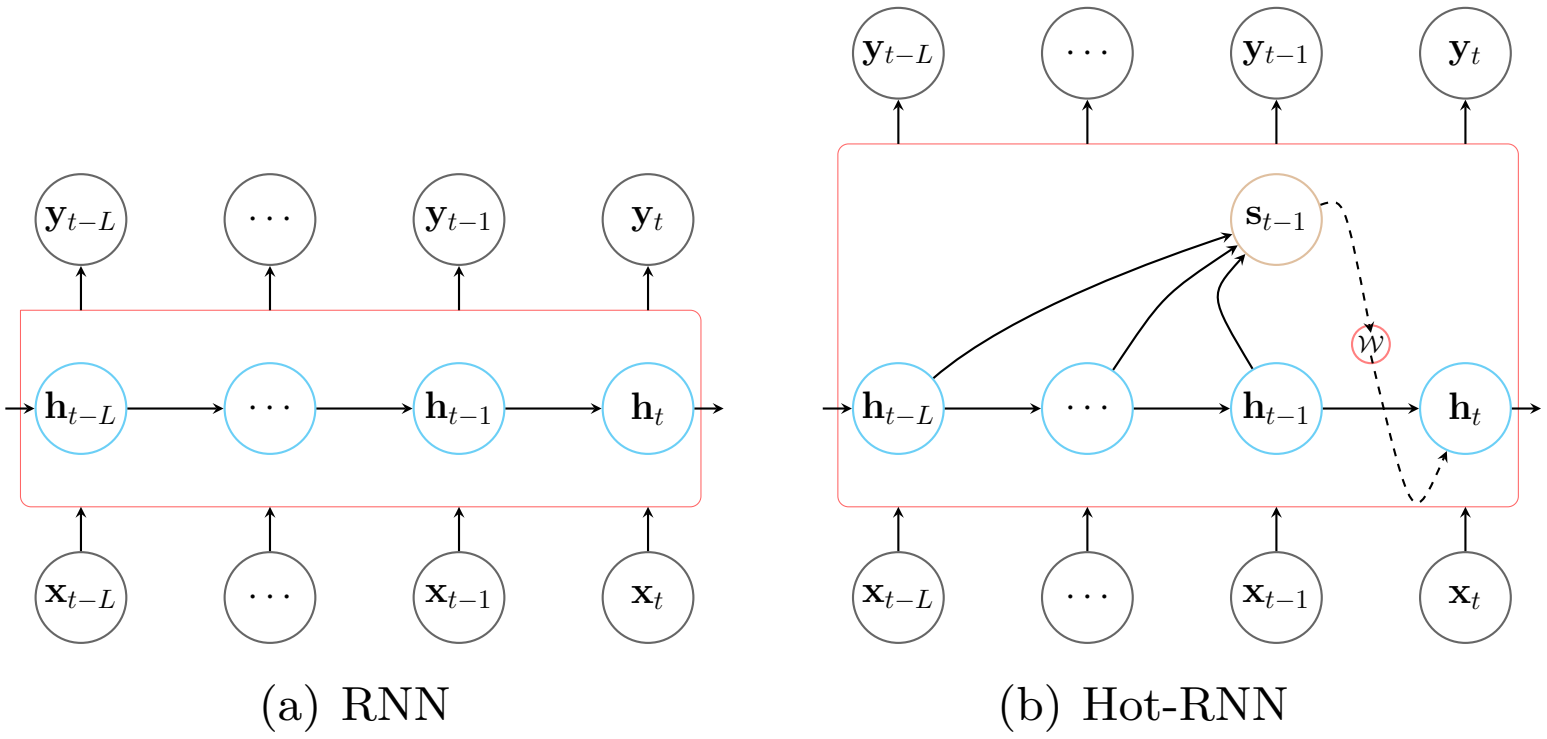

**Figure 5.4:** A comparison of the classical RNNs and the higher-order RNNs.

Kernelized tensor regression and tensor Gaussian process regression are both nonlinear extensions to simple linear models. They both consider the nonlinear function $f$ to be the weighted sum of a finite set of multiple basis functions, namely $f(\mathcal{X}_n) = \langle \mathcal{B}, \Phi(\mathcal{X}_n) \rangle$, where $\Phi(\mathcal{X}_n)$ is the collection of all the basis functions performed over $\mathcal{X}_n$, and $\mathcal{B}$ is the weight of each basis function. $\Phi(\mathcal{X}_n)$ can also be thought of as a mapping function from the input space into another high dimensional space. The key issue with these two methods is figuring out which tensor kernels are suited for specific applications. Once the kernel function has been created, only a few hyperparameters need to be tuned to fit the training data. As a result, the performance of these two methods highly depends on the kernel functions employed and will be difficult to apply to large datasets because the computation of kernel functions and their inverse requires $O(N^2)$ memory and $O(N^3)$ computation. Fitting models in subsets (Das *et al.*, 2018) or finding the solution using doubly stochastic functional gradients (Yan *et al.*, 2015) are some of the solutions for this problem.

The additivity hypothesis of simple linear models is maintained in additive models, which assumes that the impact of each explanatory variable or subsets of explanatory factors on the output is independent and nonlinear. Tensor-based additive models investigate multidirectional correlations by factoring the input tensor into latent factors and employing latent factors in an additive model, or by approximating the function $f$ using basis functions through a low rank coefficient tensor. The fundamental issue with additive models is model selection and overfitting; one must determine which type of basis functions to use, as well as the smoothing or regularization term and the weighting factor. Although additive models simplify function approximation on multidimensional data, they inevitably overlook data collinearity. It is likely that the multidimensional explanatory variable's included features have correlations and have similar effects on the response in reality.

Random forest and neural networks are methods arising from the field of machine learning, which can be used in both classification and regression. Random forest is very robust and stable since it incorporates multiple decision trees for ensemble learning. In addition, each decision tree in the forest is trained using samples randomly selected from the training datasets, reducing the correlation between distinct trees and thereby improving the random forest's prediction performance. However, building a random forest for tensor regression takes a long time. Although neural networks have demonstrated impressive performance in computer vision applications, they may confront difficulties when dealing with huge video collections. Exploring multidirectional relatedness in multidimensional datasets, such as long-term dependency in sequential data, can help improve the performance of existing neural networks. However, in the case of neural networks, the network architecture design has a significant impact on the network's capacity to model specific tasks. The performance of neural networks is influenced by hyperparameters such as network depth, width, and tensor rank, and the training process is time-consuming. This is also why a number of network compression, network acceleration, and rank determination techniques are present.

# 6

# Strategies for efficient implementation

## 6.1 Sketching

Sketching is a dimensionality reduction approach that uses a well-designed randomized matrix to compress huge matrices (Woodruff *et al.*, 2014). Sketching has found successful applications in least squares regression, low-rank approximation, clustering, etc.

The following is the most basic application of sketching in least squares regression

$$\min_{\mathbf{B}} \|\mathbf{Y} - \mathbf{X}\mathbf{B}\|_{\mathrm{F}}^2 \quad \longrightarrow \quad \min_{\mathbf{B}} \|(\mathbf{S}\mathbf{Y}) - (\mathbf{S}\mathbf{X})\mathbf{B}\|_{\mathrm{F}}^2, \tag{6.1}$$

where $\mathbf{X} \in \mathbb{R}^{N \times P}$ is the predictor, $\mathbf{Y} \in \mathbb{R}^{N \times Q}$ is the response, $\mathbf{B}$ is the coefficient matrix that needs to be estimated, $\mathbf{S} \in \mathbb{R}^{R \times N}$ is a randomized matrix. If $R \ll N$, the memory and computational time required will be greatly reduced. And for well-designed $\mathbf{S}$, the recovery accuracy of the original system is guaranteed.

For the low-rank approximation problem, given a matrix $\mathbf{A}$, one wishes to compute $\tilde{\mathbf{A}}_R$ such that (Clarkson and Woodruff, 2017)

$$\|\mathbf{A} - \tilde{\mathbf{A}}_R\|_{\mathrm{F}} \leq (1 + \epsilon)\|\mathbf{A} - \mathbf{A}_R\|_{\mathrm{F}}, \tag{6.2}$$

where $\mathbf{A}_R$ is the best rank-$R$ approximation of $\mathbf{A}$ obtained from trun-cated SVD. Time-optimal algorithms for computing $\tilde{\mathbf{A}}_R$ are present in Sarlos (2006) and Clarkson and Woodruff (2017).

For the clustering problem, suppose a data matrix $\mathbf{A} \in \mathbb{R}^{N \times D}$ is given, where the $n$-th row represents the $n$-th data point with dimension $D$. The simplest $K$-means clustering computes $K$ clusters $C_1$, $C_2$, $\cdots$, $C_K$ such that the cost function

$$\sum_{k=1}^{K} \sum_{n \in C_i} \|\mathbf{A}(n,:)^{\mathrm{T}} - \mathbf{c}_k\|_{\mathrm{F}}^2 \tag{6.3}$$

is minimized, where $\mathbf{c}_k \in \mathbb{R}^D$ is the clustering center of $C_k$, $\mathbf{a}_n$ is the $n$-th row of $\mathbf{A}$. It can be expressed in matrix form as

$$\|\mathbf{A} - \mathbf{BC}\|_{\mathrm{F}}^2, \tag{6.4}$$

where $\mathbf{B} \in \mathbb{R}^{N \times K}$ represents the cluster indication matrix, and $\mathbf{B}(n,k) = 1$ means $\mathbf{A}(n,:)$ belongs to $k$-th cluster, $\mathbf{C} \in \mathbb{R}^{K \times D}$ represents the centroids of the clusters as $\mathbf{C} = [\mathbf{c}_1^{\mathrm{T}}; \cdots; \mathbf{c}_K^{\mathrm{T}}]$. By applying sketching, $\mathbf{A}$ is right-multipled by a count sketch matrix $\mathbf{S}$. The count sketch matrix $\mathbf{S} \in \mathbb{R}^{D \times M}$ is constructed by uniformly picking a row $d$ and assigning $\{1, -1\}$ to the element $\mathbf{S}(d,m)$ for each column $m$. Then clustering is done on $\mathbf{AS}$ with reduced dimension. It is shown in Cohen *et al.* (2015) that for any given $K$-means algorithm, the cost function on $\mathbf{AS}$ gives a $(1+\epsilon)$-approximation of the cost function on $\mathbf{A}$, if $\mathbf{S}$ has $O(K^2/\epsilon^2)$ rows.

Sketching also serves as an efficient tool in tensor algebra. For example, tensor regression is done in Yu and Liu (2016) by employing the count sketch as a subsampling step before the estimation of the regression coefficient tensors. The sketching is performed over the tensor predictor and response along the first mode as $\tilde{\mathcal{X}} = \mathcal{X} \times_1 \mathbf{S}$, $\tilde{\mathcal{Y}} = \mathcal{Y} \times_1 \mathbf{S}$. In this way, the number of samples included in the subsequent regression analysis is greatly reduced, and the huge computational complexity and memory requirements are relaxed. Besides, a similar trick is used in Li *et al.* (2017) for general high dimensional vector-on-tensor regression models with applications in neuroimaging analysis. But there are mainly two disadvantages for both Yu and Liu (2016) and Li *et al.* (2017).

One is that they sample the sketch matrix from predefined distributions, with no consideration of the data instinct structures. To address this issue, Zhang *et al.* (2020) proposed the concept of importance sketching, which first finds the important sketching directions through factorization of the covariates tensor before performing regression analysis using the sketched factors. In this way, the important sketches which contribute to the model parameters will be reserved in the sketching process, thus compressing the predictor tensor most and improving the performance of subsequent regression analysis in terms of accuracy, computational complexity and memory requirements. In recent years, learning-based methods have also been proposed to enhance the performance of random sketching by adapting the structure of sketching matrices to specific tasks and data (Indyk *et al.*, 2019; Hsu *et al.*, 2019; Liu *et al.*, 2020b).

The other is that Yu and Liu (2016) and Li *et al.* (2017) still used the matrix based count sketching methods, ignoring correlations between other modes. In recent years, multidimensional tensor sketching methods have been proposed to exploit the complete multidimensional structures and achieve further compression. For example, in Shi and Anandkumar (2020), a higher-order count sketch is proposed to sketch a high dimensional tensor into a low-dimensional one with the same order, therefore the multidimensional structure information within tensors can be well exploited.

The applications of multidimensional tensor sketching methods include tensor decomposition (Wang *et al.*, 2015; Malik and Becker, 2018), and network approximation (Kasiviswanathan *et al.*, 2018). Specifically, Wang *et al.* (2015) suggested using the tensor sketching methodology to speed up two CP decomposition algorithms: robust tensor power method (RTPM) and alternating least squares (ALS). In Malik and Becker (2018), tensor sketching is applied in accelerating the higher-order orthogonal iteration (HOOI) algorithm for Tucker decomposition. For network approximation, Kasiviswanathan *et al.* (2018) applied tensor sketching to approximate the convolutional layer and fully-connected layer of the convolutional neural networks (CNNs). Using the tensor sketching approaches, the network parameters can be considerably reduced with guarantees of approximation accuracy.

## 6.2 Online learning

Large-scale data streams may fluctuate with the change of time or space domain in real data analysis, especially in sociology or climatology, posing challenges to batch learning. For real-time processing and analysis, online learning that can dynamically perform model estimation and prediction is critical. Online learning not only avoids the massive computing load caused by large batch processing, but also allows for real-time data prediction, reducing the model's feedback time significantly. The most common strategy for achieving online learning is to fit fresh data samples while maintaining the estimated model as close to the prior model as possible. Mathematically, the model for online learning is formulated in Xu *et al.* (2019) as

$$\min_{\tilde{\mathscr{T}}} \mathfrak{Q}(\mathscr{T}, \tilde{\mathscr{T}}) = \mathfrak{L}\left(\tilde{\mathscr{T}}; \mathcal{X}_{\text{new}}, \mathbf{Y}_{\text{new}}\right) + \mathfrak{H}(\mathscr{T}, \tilde{\mathscr{T}}), \tag{6.5}$$

where $\mathscr{T}$ is the set of the old model parameters, $\tilde{\mathscr{T}}$ is the new set of model parameters that need to be estimated. This is, in fact, the central principle of online learning. The first term $\mathfrak{L}\left(\tilde{\mathscr{T}}; \mathcal{X}_{\text{new}}, \mathbf{Y}_{\text{new}}\right)$ minimizes the fitting error on the new samples, while the second term $\mathfrak{H}(\mathscr{T}, \tilde{\mathscr{T}})$ assesses the similarity of the new model and the previous model.

The collection of model parameters and the solution procedure for different regression models can differ. As an incremental learning algorithm for the regression model (4.103), WISDOM (Xu *et al.*, 2019) jointly minimizes the objective function in model (4.103) for the newly arrived dataset and the regularization term

$$\mathfrak{H}(\mathscr{T}, \tilde{\mathscr{T}}) = \|\mathbf{W} - \tilde{\mathbf{W}}\|_{\text{F}}^2 + \|\mathbf{V} - \tilde{\mathbf{V}}\|_{\text{F}}^2 + \|\mathbf{A} - \tilde{\mathbf{A}}\|_{\text{F}}^2 + \|\mathbf{B} - \tilde{\mathbf{B}}\|_{\text{F}}^2 + \|\mathbf{C} - \tilde{\mathbf{C}}\|_{\text{F}}^2, \tag{6.6}$$

where $\{\mathbf{W}, \mathbf{V}, \mathbf{A}, \mathbf{B}, \mathbf{C}\}$ are old model parameters as defined in (4.103), $\{\tilde{\mathbf{W}}, \tilde{\mathbf{V}}, \tilde{\mathbf{A}}, \tilde{\mathbf{B}}, \tilde{\mathbf{C}}\}$ are new parameters that we need to estimate when a new dataset arrives. The estimated new parameters can be made similar to the prior ones by minimizing $\mathfrak{H}(\mathscr{T}, \tilde{\mathscr{T}})$.

In Yu *et al.* (2015), an incremental learning scheme for the regression model (4.17) is proposed, namely accelerated low-rank tensor online learning (ALTO). It first updates $\mathcal{B}$ when a new mini-batch arrives at

time $T$ as follows

$$\mathcal{B}^k[:,:,m] = (1-\alpha)\mathcal{B}^{k-1}[:,:,m] + \alpha\mathcal{Y}_{\text{new}}[:,:,m](\mathcal{X}_{\text{new}}[:,:,m])^{\dagger}, \quad (6.7)$$

which can be considered as a solution for the model (6.5) when old estimation and new data are taken into account, where $\alpha$ is a weighing factor. To enforce the estimated coefficient tensor to be low rank, a projection procedure is adopted efficiently through making use of the previous projected factors.

Moreover, the PLS based methods $N$-way PLS and HOPLS are also extended into incremental versions as recursive $N$-way PLS (RNPLS) (Eliseyev and Aksenova, 2013), online incremental higher-order partial least squares (IHOPLS) (Hou and Chaib-draa, 2016), and recursive higher-order partial least squares (RHOPLS) (Hou and Chaib-draa, 2017). The primary idea behind these frameworks is to append fresh arriving data to existing data, extract new factors from the combined data, and lastly compress all model components to produce a new set of factors.

But RNPLS and IHOPLS directly add the new tensor data into the old tensor data, while RHOPLS first approximates the new tensor data using HOPLS and adds the latent factors into the old factors. This eliminates the requirement to run the iterative approximation technique on the joint tensor as in RNPLS or IHOPLS, which saves a significant amount of time. Due to the superiority of Tucker over CP decomposition, RHOPLS can yield more accurate predictions and converges faster than RNPLS. According to IHOPLS, it is designed to deal with infinite time-dependent tensor steams. The main difference between IHOPLS and RHOPLS is twofold. One is the concatenating method of the newly arrived data stream and the old data as described above. The second point is that IHOPLS uses clustering to find the old factors that are closest to the incremental factors and absorbs them into specific latent space, while RHOPLS directly uses the matrix factorization method to compress the factor matrix.

In addition to sketching methods and online learning, there also exist methods for scalable computation, such as gradient descent based methods and distributed algorithms. At the moment, the majority of works on distributed algorithms and gradient descent-based methods focus on tensor decomposition (Beutel *et al.*, 2014; Papalexakis *et al.*, 2012; De Almeida and Kibangou, 2014; Wang *et al.*, 2019) and tensor completion (Papastergiou and Megalooikonomou, 2017; Ge *et al.*, 2018; Yuan *et al.*, 2019; Song *et al.*, 2019) because for regression tasks, usually model parameters cannot be sampled unless sparsity assumption is imposed. Furthermore, many recent tensor regression tasks have a limited sample size and a large number of model parameters, making this method ineffective. The stochastic gradient approach is well suited to large sample sizes, and the deep network is one of the highlights. As in Kossaifi *et al.* (2020), the computational efficiency of gradient descent based Tucker regression is discussed and verified on MRI prediction tasks. The stochastic gradient descent algorithm only considers one predictor-response sample at a time. Mini-batch gradient descent is between the stochastic gradient algorithm and the batch gradient algorithm, which employs $b$ samples at each iteration with $b \ll N$. In this way, the computational burden can be reduced, and faster iterations can be achieved with a slower convergence speed.

Meanwhile, for the tensor approximation problem, the choice of rank has always been a problem under exploration. Tensor rank estimation can be done using a variety of ways. (1) One simple way is to set a higher bound on the true rank and strive to lower it during training. For example, Liu *et al.* (2020c) use the tensor training rounding algorithm to reduce the predefined tensor train rank with acceptable approximation error. Deep rank estimation networks are also proposed to give an approximation of the original tensor in factorization form with a lower rank (Zhou *et al.*, 2019). They are all dedicated to solving the optimization problem

$$\min_{\mathcal{B}} \operatorname{rank}(\hat{\mathcal{B}}) \quad \text{s.t.} \|\hat{\mathcal{B}} - \mathcal{B}\|_{\mathrm{F}} \leq \epsilon \|\mathcal{B}\|_{\mathrm{F}} \tag{6.8}$$

but with different methods. (2) Transferring the rank minimization problem into a sparse coding problem, such as the group sparsity constraints developed for rank adaptivity in Guo *et al.* (2012) and Liu *et al.* (2020a), can reduce the unnecessary components of the tensor decomposition, as described in Section 4.3.2. (3) Considering from the perspective of probability theory, incorporating the sparsity with a prior distribution over the given upper bound of tensor rank (Suzuki, 2015;

Guhaniyogi *et al.*, 2017; Papadogeorgou *et al.*, 2019; Spencer *et al.*, 2019) is also a promising direction, as descried in Section 4.4.

# 7

# Applications and available datasets

Tensor regression has been investigated in a wide range of areas, including MLMTL, spatio-temporal data analysis in climatology, sociology and economics, human motion analysis, facial image analysis, neuroimaging analysis, and chemometrics. In the following, we will go through each of these application categories in detail, as well as provide links to some publicly available datasets.

## 7.1 Performance evaluation

We initially offer several metrics to evaluate the prediction performance of learnt models before illustrating the associated applications or instances of tensor regression. One popular strategy is to apply the learnt models to a new test dataset to make predictions. Let $\mathcal{Y}$ and $\hat{\mathcal{Y}}$ be the ground truth and predicted result of the test predictor $\mathcal{X}$, then the most used evaluation indicators are formulated in the following.

- The mean squares error (MSE)

$$\text{MSE} = \frac{\|\mathcal{Y} - \hat{\mathcal{Y}}\|_{\text{F}}^2}{N_{\text{test}} Q_1 \cdots Q_M}, \tag{7.1}$$

where $N_{\text{test}}Q_1 \cdots Q_M$ is the total size of $\mathcal{Y}$, which equals to the total number of values in $\text{vec}(\mathcal{Y})$.

- The root mean squares error (RMSE)

$$\text{RMSE} = \sqrt{\frac{\|\mathcal{Y} - \hat{\mathcal{Y}}\|_{\text{F}}^2}{N_{\text{test}}Q_1 \cdots Q_M}}. \tag{7.2}$$

- The mean absolute error (MAE)

$$\text{MAE} = \frac{\sum |\mathcal{Y}(n_{\text{text}}, q_1, \cdots, q_M) - \hat{\mathcal{Y}}(n_{\text{text}}, q_1, \cdots, q_M)|}{N_{\text{test}}Q_1 \cdots Q_M}. \tag{7.3}$$

- $Q^2$ (Zhao *et al.*, 2013a)

$$Q^2 = 1 - \|\mathcal{Y} - \hat{\mathcal{Y}}\|_{\text{F}}^2 / \|\mathcal{Y}\|_{\text{F}}^2. \tag{7.4}$$

- Correlation

$$\text{Corr}(\mathcal{Y}, \hat{\mathcal{Y}}) = \langle \text{vec}(\mathcal{Y}), \text{vec}(\hat{\mathcal{Y}}) \rangle. \tag{7.5}$$

  It should be noted that correlation needs to be calculated under the assumption that $\mathcal{Y}$ and $\hat{\mathcal{Y}}$ are mean-centered.

Aside from that, model complexity is an essential metric. As previously said, it is not the case that the more complicated the model, the better the performance. Our goal is to choose the simpler model as much as feasible in order to avoid modeling errors. This is one of the reasons why linear models are so popular. Although it has certain modeling flaws, its simplicity makes it simple to adopt and execute. BIC (Bayesian Information Criterion) is a model selection index that is defined as follows:

$$\text{BIC} = k \log(N) - 2 \log(\mathfrak{L}(\mathscr{T})), \tag{7.6}$$

where $k$ is the number of required model parameters, $N$ is the number of observed samples, $\mathscr{T}$ is a set of model parameters, $\mathfrak{L}(\mathscr{T})$ is the likelihood of the estimated model. Statistically, the model with the lowest BIC is considered to be the best of several alternative models.

## 7.2 Multitask learning

Multitask learning (MTL) is a promising field in machine learning, which aims to exploit the useful or shared information contained in multiple learning tasks in order to acquire more accurate learners for each task. With the assumption that all tasks (at least some of them) are related, learning multiple tasks jointly can get better performance than learning them individually. But as described before, traditional MTL methods can not exploit the inherent structure within the multi-dimensional data. To address this issue, Romera-Paredes *et al.* (2013) proposed an approach based on the convex relation of the tensor rank, namely MLMTL. MLMTL is better suited for datasets specified by many indices since it can preserve multidimensional indices and study correlations across different modes. Other linear solutions include the efficient TPG algorithm in Yu and Liu (2016) and a newly proposed regression model based on latent factors with batch learning and online learning approaches in Xu *et al.* (2019). It should be noted that although Xu *et al.* (2019) uses tensor-based learning, it only explores the multidimensional information within the input side, the correlation between the multiple tasks is not included in the modeling. Some nonlinear methods are proposed for performance improvement, including the multilinear Gaussian process regression (Yu *et al.*, 2018) and tensor additive models (Kanagawa *et al.*, 2016; Suzuki *et al.*, 2016).

The three benchmark datasets for model evaluation in MTL include the Restaurant & Consumer Dataset, Shoulder Pain Dataset and School Dataset with descriptions in the following:

- **Restaurant & Consumer Dataset** (Vargas-Govea *et al.*, 2011): A dataset to predict rating values for specific restaurants and variables based on corresponding descriptive attributes. The dataset records the users' rating values of different restaurants according to different variables, including food quality, service quality and overall quality, and 44 descriptive attributes of the restaurants are consisted.

- **Shoulder Pain Dataset** (Lucey *et al.*, 2011): A dataset to infer the Action units (AU) intensity level to different AUs and different

patients. The dataset consists of video clips of facial expressions of people who suffer from shoulder pain when performing activities. The predictor variables for each task are formed by 132 attributes representing facial expressions. The response variables consist of the intensity level (0-5) of different AUs defined by the facial action coding system.

- **School Dataset** (Nuttall *et al.*, 1989): A dataset to predict examination scores for students based on student-related or school-related input variables. The dataset includes the examination records of 31623 students at age 16 from 1985 to 1987, which is taken from the Inner London Education Authority with 140 secondary schools.

In addition, a climate dataset extracted from the Global Surface Summary of the Day (GSOD)[1] website and coarse-scale gridded climate data[2], are used for climate prediction in Xu *et al.* (2019). The output variables are extracted from monthly climate data with respect to the total precipitation, and the maximum, minimum and average temperature across over 30000 sites over the world from 1942 to the present in the GSOD. The input variables are extracted from hundreds of variables for corresponding sites and timepoints in the GSOD.

$N$-way PLS is used to simulate the relationship between the average features of the global climate from the General Circulation Models (GCM) and the local climate characteristics, so as to predict the GCM characteristics of a specific local region (Bergant and Kajfež-Bogataj, 2005). The predictor comes from the simulation results of five different GCMs in the NCEP/NCAR reanalysis dataset (Kistler *et al.*, 2001). It contains the monthly average near-ground air temperature, sea-level pressure and near-ground specific humidity fields throughout Central and Western Europe. The dependent variables are the monthly average near-ground air temperature, dew temperature and daily precipitation amount of four locations in different climate zones in Slovenia.

[1] https://data.noaa.gov/dataset/dataset/global-surface-summary-of-the-day-gsod

[2] http://www.esrl.noaa.gov/psd/data/gridded/data.ncep.reanalysis.derived.html

Moreover, Kanagawa *et al.* (2016) and Suzuki *et al.* (2016) applied the tensor Gaussian process models to an online shopping dataset to predict online shopping sales. The dataset is collected from the purchase histories on Yahoo! Japan shopping, recording how many times each product is purchased by each consumer. The objective is to predict the quantity of a certain product purchased by each user in a particular shop, given the characteristics of each user such as age, gender, and industry type of occupation. With different types of shops and products, it will yield a multitask learning problem.

## 7.3 Spatio-temporal analysis

There are primarily two sorts of tasks for spatio-temporal data: forecasting and cokriging. Forecasting tasks attempt to anticipate future trends using past data, whereas cokriging attempts to interpolate likely values for certain variables in a new unknown area using observations from nearby sites or similar patterns in other known locations. In other words, forecasting is used to make predictions in the temporal mode using an autoregression model, whereas cokriging is used to make predictions in the spatial mode using local consistency and similar patterns in known areas. In Bahadori *et al.* (2014), the unified tensor regression framework Greedy is developed which can perform both two types of tasks. Later, some advanced methods are proposed for extensions to Tucker-based learning (Yu and Liu, 2016; Rabusseau and Kadri, 2016), TT-based learning (Liu *et al.*, 2020c), Bayesian based learning (Billio *et al.*, 2018), multilinear Gaussian process regression (Yu *et al.*, 2018), and real time analysis (Yu *et al.*, 2015). Deep tensor regression models, namely higher order extensions of RNNs (Yu *et al.*, 2017; Su *et al.*, 2020), have been used to tackle long-term forecasting and challenging video tasks. The main application fields of spatio-temporal analysis include, but are not limited to, climate & weather, social networks, transportation, and human mobility, with some benchmark datasets as follows:

1) Climate & weather

- **USHCN**: The U.S. historical climatology network contains two datasets. One consists of the daily measurements of five climate

variables including temperature max, temperature min, precipitation, snow fall and snow depth over 100 years[3]. The other is the monthly records of the climatological variables maximum, minimum and average temperatures over a month and total monthly precipitations from 1915 to 2014[4].

- **CCDS**[5]: The comprehensive climate dataset is a collection of measurements of 17 climatological variables in North America from 1990 to 2001 across 125 locations.
- **Meteo-UK**: The dataset is collected from the meteorological office of the UK.[6] It contains monthly measurements of five variables in 16 stations across the UK from 1960 to 2000.

2) Social networks

- **Foursquare check-in**: The Foursquare check-in dataset (Long *et al.*, 2012) records the number of check-ins in the Pittsburgh area from Feb 24 to May 23, 2012. The data is categorized by different venue types such as Art & Entertainment and College & University. The users' friendship network is also provided.
- **Yelp**: The Yelp dataset[7] includes the rating values (1-5) of users to 22 categories of business over 500 time intervals on Yelp and the corresponding social graph for the active 137 users.

3) Transportation

- **PeMS**: The dataset is collected from the California Department of Transportation[8] for traffic flow prediction. PEMS contains loop detector data for more than 8,100 highway locations in California over 10 years. Due to the huge amount of data, the data is usually aggregated into a 15-minute or 20-minute period. Usually, some time periods, roads, and sensors are selected for research.

[3] https://cdiac.ess-dive.lbl.gov/ftp/ushcn_daily/
[4] https://cdiac.ess-dive.lbl.gov/ftp/ushcn_v2.5_monthly/
[5] https://viterbi-web.usc.edu/~liu32/data.html
[6] http://www.metoffice.gov.uk/public/weather/climate-historic/
[7] http://www.yelp.com/dataset_chellege
[8] https://pems.dot.ca.gov/

4) Human mobility

- **KTH action dataset** (Schuldt *et al.*, 2004): The KTH dataset is one of the most standard datasets for human actions. The database contains 6 actions (walking, jogging, running, boxing, waving and clapping) performed by 25 people in 4 different scenarios, with a total of 2391 video samples. For each action of each actor, the scene will vary with changes in proportions, clothing and lighting.

- **Moving-MNIST-2-dataset** (Srivastava *et al.*, 2015): The dataset contains 10000 sequences of moving digits with a resolution of $64 \times 64$. Each video sequence contains 20 images. Usually, the first 10 images are tread as inputs and the latter 10 images as outputs.

- **Something-Something V2 dataset** (Goyal *et al.*, 2017): The dataset verifies RGB-based action recognition related algorithms. The Something-Something V2 dataset is a large labeled dataset that records actions between humans and some objects in daily life. There are 174 categories of actions. It contains 220,847 videos, including 168,913 in the training set, 24,777 in the validation set, and 27,157 in the test set.

Taking the Foursquare dataset as an example, we give a comparison between the tensor regression models for spatio-temporal data forecasting. We first extract the hourly check-in records of 121 users in 15 different venues over 1200 hours. The classical autoregression model with 3 lags is used and each time sequence is normalized for the selected subset. For a fair comparison, we use 5-fold cross validation to select the hyper-parameters during the training process and report the mean and standard deviation of several metrics like RMSE, $Q^2$, Correlation as well as the consuming time with 10 repetitions on a machine with 3.40GHz CPU and 16GB RAM, as shown in Table. 7.1. The maximum value of the rank is set to be 10.

**Table 7.1:** A comparison of tensor regression methods for forecasting task on Foursquare.

| Model | Parameters | RMSE | $Q^2$ | Correltion | CPU time |
|---|---|---|---|---|---|
| Greedy | - | 0.1212 | 0.0285 | 0.1696 | 611.27 ± 4.17 |
| TPG | $R = 6$ | 0.1216 | 0.0227 | 0.1530 | 33.47±0.07 |
| HOPLS | $R = 8, \lambda = 10$ | 0.1195 | 0.0566 | 0.2378 | 51.74±0.11 |
| HOLRR | $R = [10, 10, 10], \lambda = 10^2$ | 0.1201 | 0.0462 | 0.2148 | 31.58±3.93 |
| TTR-CP | $R = 10, \lambda = 10^2$ | 0.1204 | 0.0416 | 0.2038 | 1.647 $\times 10^3$±199.58 |
| TTR-TT | $R = (1, 10, 7, 7, 10, 1), \lambda = 10^2$ | 0.1190 | 0.0644 | 0.2538 | 4.48±0.37 |

## 7.4 Human motion analysis

Human motion analysis, including pose estimation, human motion trajectory reconstruction and object tracking, plays an important role in computer vision and is widely applied in many fields, such as animation production, gait analysis, biomechanics, ergonomics, human-computer interaction, intelligent surveillance, and vehicle navigation. Since image or video sequences are intrinsically multidimensional arrays, tensor-based learning can better explore the spatial and temporal correlations within the multiway data and greatly improve the algorithm performance. For example, Guo *et al.* (2012) employ the CP decomposition to model the relationship between the image sequences and corresponding head pose or map the image sequences of people performing specific activities into the human motion trajectories defined by the 3D position of body joints. But the correlations in the output side are ignored in Guo *et al.* (2012) since all the responses are learned individually. Later, tensor-on-tensor regression models such as HOPLS (Hou and Chaib-draa, 2016; Hou and Chaib-draa, 2017) and TTR-TT (Liu *et al.*, 2020c) are used to consider not only the spatial and temporal dependencies within the input image sequences but also the shared parameters among different tasks. Moreover, Kaymak and Patras (2014) proposed the multimodal random forest based tensor regression for head pose estimation through integrating both the RGB images and the depth data. Furthermore, instead of individual analysis, Zhao *et al.* (2014) use a tensor Gaussian process regression model to establish a good model for crowd behavior patterns. A 4D object tracking method is proposed based on the HOPLS in Zhong *et al.* (2016). Several benchmark datasets here are shown in the following:

- **IDIAP dataset** (Ba and Odobez, 2005): A dataset consisting of 23 video sequences of people in natural activities. To use the dataset for head pose estimation, a face detector needs to be used to extract the bounding box of each face in video frames. The input variables are two types of features extracted from the obtained facial images while the output variables are the head pose Euler angles calculated from the rotation matrix of the head configuration.

- **Boston University dataset** (La Cascia *et al.*, 2000): A dataset containing 45 video sequences with 5 subjects performing 9 different motions in a standard office. The preprocessing procedure is the same as IDIAP.

- **Pointing'04 dataset** (Gourier *et al.*, 2004): A dataset involving a set of facial images of 15 subjects in different poses. These subjects are in different skin colors, with glasses or not. The pose or head orientation, which forms the response of the head pose estimation task, varies from $-90^\circ$ to $+90^\circ$ in horizontal and vertical.

- **HumanEva-I dataset** (Sigal *et al.*, 2010): A dataset for human pose and motion estimation. HUMANEVA-I contains the video and motion information of 4 subjects performing a set of 6 predefined actions, such as walk, jog, gestures and box.

- **UMPM** (Van der Aa *et al.*, 2011): A dataset to explore the relationship between the video sequences recording human motions in daily life activities, including multi-person motion interaction and the human motion trajectories data collected by the motion capture system.

- **Biwi Kinect head pose database** (Fanelli *et al.*, 2011): A dataset for head pose estimation which contains both depth data and RGB images of the upper body region of 20 people (14 males and 6 females). All images are marked with the head center position and head rotation angle, which constitute the output variables of the head estimation task. The head rotation range is approximately between ±75° for yaw, ±60° for pitch, and ±50° for roll.

- **ICT-3DHP database** (Baltrušaitis *et al.*, 2012): A dataset for head pose estimation which contains 14000 framers (both depth data and RGB images), which are divided into 10 video sequences. All images are marked with the head center position and head rotation angle using Polhemus FASTRAK flock of birds tracker.
- **Pedestrian traffic dataset**[9]: A dataset for crowd counting which contains an hour of video captured at 30 fps with resolution $740 \times 480$. The video contains a total of 49,885 pedestrian instances and is later downsampled into $238 \times 158$ and 10fps.

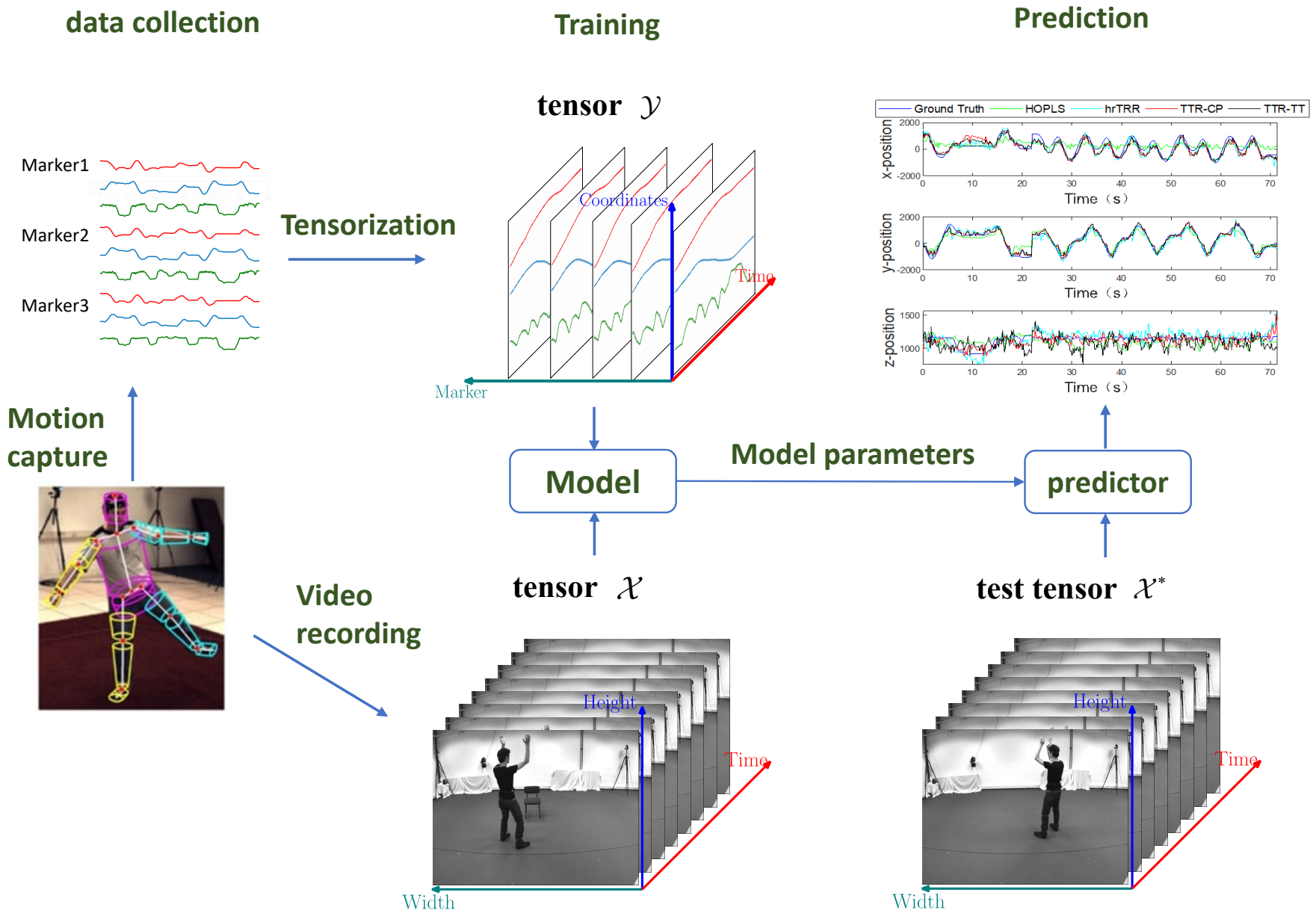

**Figure 7.1:** The human motion trajectories reconstruction on UMPM dataset.

[9] http://www.svcl.ucsd.edu/projects/peoplecnt

For example, Fig. 7.1 gives a description of the experimental process of the human motion trajectories reconstruction based on the UMPM dataset. First, the input variables are the downsampled image sequences in the size of $frames \times 32 \times 24$, and the output is the corresponding 3D positions of 37 markers in the size of $frames \times 3 \times 37$. Splitting the collected dataset into a training set and a test set, we can obtain the learned model parameters and perform a prediction on the new test tensor. A comparison of the ground truth and predicted trajectories of one specific marker by the popular solutions hrTRR, HOPLS, TTR-CP, TTR-TT is shown in the upper right corner of Fig. 7.1.

## 7.5 Facial image analysis

Facial image analysis is a specialized field in image analysis and provides a large amount of valuable information on each individual, such as age, gender, expression, and identity, which is useful for realizing identity authentication, human-computer interaction and even smart cities. With the diversification of feature extraction methods, the extracted features from the facial images appear multidimensional. However, traditional regression methods cannot reasonably utilize the correlation information between the extracted features. This is why Guo *et al.* (2012) proposed using tensor learning to estimate one's age from features derived from facial images and saw a significant improvement. In addition, a tensor-on-tensor regression model is proposed to predict multiple descriptive attributes rather than only age from the facial images in Lock (2018). The two datasets used are as follows:

- **FG-NET dataset**[10]: A dataset to build the human age estimation system based on one's facial images or the features extracted from the facial images. The FG-NET comprises 1002 facial images of 82 individuals aging from 0 to 69.
- **LFW dataset** (Learned-Miller *et al.*, 2016): The Labeled Faces in the Wild dataset contains over 13000 facial images taken from the Internet. For each individual, there are multiple images with different lighting or angles. The describable attributes of each image, such as age, gender, smiling and make-up, can be obtained by the popular attribute classifier (Kumar *et al.*, 2009) and used as the response of the attributes estimation task.

[10] https://yanweifu.github.io/FG_NET_data/index.html

## 7.6 Neuroimaging analysis

### 7.6.1 Disease diagnosis

With the development of computer technology, medical images play an increasingly important role in assisting doctors in diagnosis and treatment. Exploring the intrinsic relationship between medical images and neuropsychiatric diseases or neurodegenerative diseases is one of the intriguing scientific issues and an urgent problem to be solved in the medical field. If this problem can be solved, then we can predict and control the onset of some malignant diseases. Studies on tensor regression models for medical diagnosis include the GLTR, STPG for attention deficit hyperactivity disorder (ADHD) (Zhou *et al.*, 2013; Li *et al.*, 2013a; Hou and Chaib-Draa, 2015; Ahmed *et al.*, 2019), Sturm for autism spectrum disorder (ASD) (Li *et al.*, 2019b), SMRTR for Alzheimer's disease (AD) (Li *et al.*, 2016), SURF for Parkinson's disease (He *et al.*, 2018), scalable multitask Gaussian process tensor regression (Kia *et al.*, 2018) for psychiatric disorders, and BTR-M-DGDP, Softer for predicting human traits based on brain connectome data (Guhaniyogi *et al.*, 2017; Papadogeorgou *et al.*, 2019; Kossaifi *et al.*, 2020; Kolbeinsson *et al.*, 2021). Instead of simply labeling the subjects as normal controls or patients (Zhou *et al.*, 2013; Li *et al.*, 2013a; Hou and Chaib-Draa, 2015; Li *et al.*, 2019b), SMRTR, SURF and tensor regression networks (Kossaifi *et al.*, 2020; Kolbeinsson *et al.*, 2021) employ continuous clinical assessment index as responses in order to obtain richer information on one's health status. The detailed description of datasets used for disease diagnosis is given in the following.

- **ADHD-200**[11]: The ADHD-200 dataset consists of the fMRI data for subjects with ADHD and several normal controls. The raw

[11] http://neurobureau.projects.nitrc.org/ADHD200/Data.html

fMRI data has been preprocessed and can be directly used for medical diagnosis.

- **ABIDE**[12]: The Autism Brain Imaging Data Exchange (ABIDE) consists of the fMRI data for subjects with ASD and several normal controls. The raw fMRI data has been preprocessed and made available online.

- **ADNI**[13]: This dataset is used for learning the association between image scans and multiple clinical outcomes, including Mini-Mental State Examination (MMSE) and Alzheimer's Disease Assessment Scale-Cognitive Subscale (ADAS-Cog). The dataset includes the MRI data and clinical assessment measures for 93 subjects with mild AD and 101 normal controls.

- **PPMI**[14]: The Parkinson's progression markers initiative (PPMI) dataset includes both the DTI and MRI images for 656 human subjects while the Montreal Cognitive Assessment (MoCA) scores are recorded for each subject.

- **UK Biobank** (Sudlow *et al.*, 2015): UK Biobank is a very large dataset which includes over 500,000 participants aged 40-69 years when recruited in 2006-2010. It was established for research about the genetic and non-genetic factors of the diseases of middle and old age. Available data includes brain images, health-related outcomes, lifestyle factors and others.

Here, we perform a group of disease diagnosis experiments on the ADHD and ABIDE datasets using the popular methods, including GLTR-CP, GLTR-Tucker, GLTR-HT, Remurs, Sturm, SMRTR and SURF. For ABIDE, two largest subsets by New York University (NYU) and the University of Michigan (UM) are selected. The detailed information of the selected dataset is shown in Table 7.2. The input fMRI image of ADHD is resized in advance to reduce the computational complexity as in Zhou *et al.* (2013). For each classification task, ten-fold cross validation is used. The performance comparison is provided in Table 7.3 both in terms of classification accuracy and sparsity.

[12] http://fcon_1000.projects.nitrc.org/indi/abide
[13] http://adni.loni.usc.edu/
[14] https://www.ppmi-info.org/data/

**Table 7.2:** Summary of ADHD and ABIDE datasets.

| Data set | Positive samples | Negative samples | Input data size |
|---|---|---|---|
| ADHD | 356 | 575 | $11 \times 10 \times 11$ |
| ABIDE | 128 | 158 | $61 \times 73 \times 61$ |

**Table 7.3:** Performance comparison of disease diagnosis experiments on the ADHD and ABIDE datasets.

| Data set | Method | Classification accuracy (%) | Sparsity |
|---|---|---|---|
| ADHD | Sparse GLTR-CP | 63.37±13.16 | 0.99±0.00 |
| | Sparse GLTR-Tucker | 64.13±11.31 | 0.00±0.00 |
| | Remurs | 59.18±11.87 | 0.66±0.25 |
| | Sturm | 58.86± 12.01 | 0.71± 0.21 |
| | SMRTR | 42.63±18.96 | 0.00±0.00 |
| | SURF | 38.34±16.37 | 0.95±0.01 |
| ABIDE | Sparse GLTR-CP | 54.22 ±8.78 | 0.99±0.00 |
| | Sparse GLTR-Tucker | 50.03±6.56 | 0.16±0.02 |
| | Remurs | 55.2±5.62 | 0.64±0.01 |
| | Sturm | 57.27±24.71 | 0.85± 0.18 |
| | SMRTR | 46.76±8.43 | 0.08±0.02 |
| | SURF | 46.18±3.31 | 0.99±0.00 |

From Table 7.3, the performance of the GLTR-based methods, including the sparse GLTR-CP and sparse GLTR-Tucker, is much better than other methods for ADHD but sightly worse than Remurs and Sturm for ABIDE. This can be explained by the extra inclusion of other important vector features like age or handiness. But there is no difference between the included predictors in the experiments over the ABIDE dataset.

Besides, the application of sparse GLTR-Tucker and SMRTR on this classification task is not promising in terms of the model complexity. In practice, it is difficult to find an appropriate hyper-parameter for sparse GLTR-Tucker and SMRTR to control the sparsity of the regression coefficient tensor. This is probably because a slight change over the hyper-parameter of sparse GLTR-Tucker and SMRTR would lead to a sharp variation of the number of required model parameters. The performance of sparse GLTR-Tucker and SMRTR is sensitive to its hyperparameters especially in terms of model sparsity.

Overall, there is a balance between model complexity and prediction accuracy. It is difficult to specify how many features are significant, but penalized models can help you figure out which feature is more important for predicting specific outcomes.

### 7.6.2 Neural decoding

Brain-Computer Interface (BCI) refers to the direct connection created between the human or animal brain and external devices to exchange information between the brain and external devices. The main task is to analyze brain neural activity and convert it into control signals of external devices. It provides a way for people with severe movement disorders to interact with the outside world to improve their quality of life. In recent decades, several methods have been developed to deal with neural decoding problems.

For example, for binary outputs, $N$-way PLS is employed for calibration of binary self-paced BCI systems in Eliseyev *et al.* (2011), which predicts the movement intention from neural activities. In Song and Lu (2017), Remurs maps the variety in the fMRI data into the binary information whether the subject has a certain disease. For multiple outputs, a sparse variant called L1-Penalized $N$-way PLS is used to predict the pressing intentions of non-human primates on four different pedals under simple reward-oriented tasks from electroencephalogram (ECoG) signals (Eliseyev *et al.*, 2012). As for continuous outputs, Eliseyev and Aksenova (2013) and Zhao *et al.* (2013a) map the brain activity of monkeys expressed by the electrocorticogram (ECoG) signals into the corresponding motion trajectories in the real world using the $N$-way PLS and HOPLS algorithms respectively, as shown in Fig. 7.2. Later, some nonlinear models such as kernelized HOPLS (Zhao *et al.*, 2013b), online tensor Gaussian process regression models (Hou *et al.*, 2015), penalized $N$-way PLS (Eliseyev and Aksenova, 2016), online adaptive group-wise sparse $N$-way PLS (Moly *et al.*, 2020) and broadcasted nonparametric tensor regression (Zhou *et al.*, 2020b), are also used to decode ECoG signals. It is worth noting that most of the ECOG signal decoding works mentioned above are based on experimental cases of primates, while Moly *et al.* (2020) decodes the ECoG recordings of a tetraplegic patient.

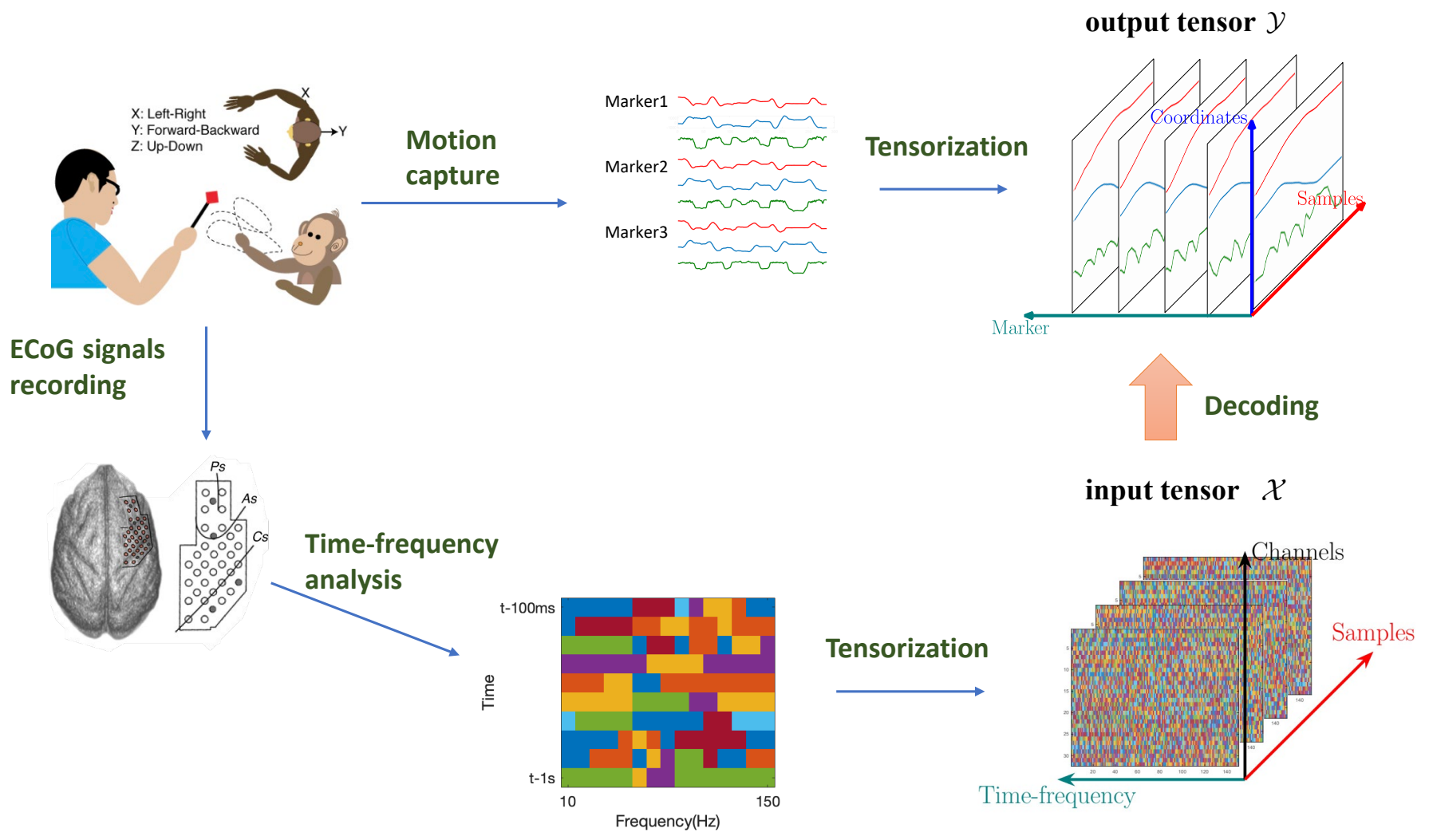

**Figure 7.2:** Decoding of ECoG signals.

Using the developed algorithms for neural decoding, predictions can be made based on brain activities even before an event occurs. In other words, neural decoding can be seen as a process of “mind reading”. The two datasets used are as follows:

- **CMU2008 dataset**[15]: The CMU2008 dataset (Mitchell *et al.*, n.d.) includes the 3D fMRI data in the size of $51 \times 61 \times 23$ for 11 subjects viewing different kinds of words or figures, such as tools, animals and buildings.
- **Food tracking task**[16]: This dataset can be used for decoding ECoG signals. The dataset includes a set of records of ECoG

[15] http://www.cs.cmu.edu/afs/cs/project/theo-73/www/science2008/data.html
[16] http://neurotycho.org

signals and the temporally synchronized motion data of a monkey during the food tacking task.

These two datasets are task-based signals, which means the brain signals or images are obtained when the subject is performing specific tasks, such as thinking or looking at figures. This kind of dataset is usually used for decoding brain neurons, in order to better understand the function of each neuron and the brain activation under specific behaviors. Furthermore, this technology is expected to help doctors understand the requirements of fantasy patients and cognitively impaired patients, and to help judges or police officers determine whether criminals are lying.

### 7.6.3 Brain activation and connectivity analysis

As an inverse problem for disease diagnosis or neural decoding, understanding how the brain processes information when performing specific tasks, or discovering the difference of brain activation or brain connectivity patterns between patients and normal controls, is also an attractive research field. As in Sun and Li (2017), Li and Zhang (2017), and Kia *et al.* (2018), tensor response regression models are proposed for comparisons of brain activation between subjects with ADHD, ASD, schizophrenia or suffering from alcoholism and the normal controls. Despite the comparison of rest-state fMRI data between different subjects, Spencer *et al.* (2019) analyze the task-based fMRI data of individuals performing risk-taking behavior. The main goal of these methods is to understand the functional connectivity of brain neurons, which means to explore regions of the brain that are spatially different but similar in function and will be activated jointly under certain behaviors. Another type of connectivity in the brain is called effective connectivity, which tends to find causal relationships between brain regions. Specifically, activation of one region may affect or cause activation of another region. These regions may not be activated at the same time, but the influence of one region over another exists, namely the Granger Causality (GC). In Karahan *et al.* (2015), the GC is formulated as a tensor autoregression model and applied into an fMRI data study. Besides the ADHD

200 and the ABIDE datasets, there are three other datasets used for brain connectivity analysis as follows:

- **EEG data for an alcoholism study**[17]: The dataset includes the EEG images of 64 channels sampled at 256 Hz for 77 subjects suffering from alcoholism and 44 normal controls.
- **LA5c Study**[18]: This is a clinical neuroimaging dataset from the UCLA Consortium for Neuropsychiatric Phenomics. This study collects both the fMRI scans and neuropsychological tests of 138 normal controls and individuals with schizophrenia (59), bipolar disorder (49) and ADHD (45).
- **Balloon Analog Risk-Taking Task**[19]: This study collects the fMRI data of individuals undergoing a risk-taking behavior and tries to associate the brain responses with the change of the degree of risk. In this study, real-world risk behaviors are simulated by inflating a balloon to make it larger without exploding to get more points. Subjects need to balance the potential for reward versus loss.

## 7.7 Chemometrics

Chemometrics is an emerging branch of chemistry, which is an interdisciplinary combination of mathematics, statistics, computer technology, and chemistry. Its main purpose is to extract useful information from the collected data to help process monitoring, optimization of experimental parameters, experimental design, and optimization.

Chemometrics involves a lot of multidimensional and enormous data. For example, fluorescence spectroscopy is a highly redundant data type used for multiplex analysis. It is usually obtained by measuring the autofluorescence of each sample at multiple excitation wavelengths and multiple emission wavelengths. The data obtained can be represented by a third-order tensor $\mathcal{X} \in \mathbb{R}^{I \times J \times K}$ with $I$ samples, $J$ denoting the emission wavelength, and $K$ indicating the excitation wavelength. In addition, for batch production of chemical products, $J$ process variables are measured at $K$ time points for each batch process, which yields a third order tensor sized $I \times J \times K$ with $I$ batches used in process optimization. How to extract useful chemical information from a large amount of data to solve related practical problems is a difficult problem faced by analytical chemists. This is also why tensor representation, the natural representation of multidimensional data, is widely used in chemometrics.

[17] https://archive.ics.uci.edu/ml/datasets/EEG+Database
[18] https://openfmri.org/dataset/da000030
[19] https://openneuro.org/datasets/ds000001/versions/00006

Typical tasks in chemometrics include calibration, process optimization, fault diagnosis and detection. Calibration is a field aimed at exploring the relationship between chemical properties related to substance concentration and instrument response. It is one of the most important branches in chemometrics. Multiway data analysis methods are commonly used in second-order sensor calibration since the response of the second-order sensor is a matrix. For example, Kiers and Smilde (1998) and Smilde *et al.* (1994) considered the calibration of a chemical sensor for chlorinated hydrocarbons, which is commonly used for environmental analysis. $N$-way PLS is used to detect several microorganisms that may cause ventilator-associated pneumonia using the electronic nose (Serneels *et al.*, 2004; Serneels and Van Espen, 2005), determination of the mixture components (Pettersson and Karlberg, 1997; Trevisan and Poppi, 2003; Sena and Poppi, 2004; Amigo *et al.*, 2005; Chow *et al.*, 2006; Sikorska *et al.*, 2008; Boltia *et al.*, 2019), toxicity identification of fresh and weathered crude oils (Mostafapour and Parastar, 2015), screening for dioxin contamination in fish oil (Pedersen *et al.*, 2002), and quantitative structure-activity relationship (QSAR) (Nilsson *et al.*, 1997; Hasegawa *et al.*, 1999; Hasegawa *et al.*, 2000; Goodarzi and De Freitas, 2010).

Process optimization, used for the development, decision-making and subsequent improvement of chemical processes, aims to maximize process performance while minimizing processing costs. The establishment of a relationship model from process variables to product quality can help find better process parameters and design scientific experimental procedures. For example, Smilde and Kiers (1999) used the multiway covariate regression model to a fat-hardening batch process; (Nomikos and MacGregor, 1995; Boqué and Smilde, 1999) for a simulated styrene-butadiene rubber (SBR) manufacturing process monitoring; (Bro, 1999) for sugar manufacturing process; (Lopes and Menezes, 2003) for the industrial fermentation process; (Kai *et al.*, 2004) for internal rubber mixing process; (Gao *et al.*, 2019), and for the time series metabolomic process. In addition, Yan *et al.* (2019) proposed a Tucker-based tensor regression model for process analysis. It uses a point cloud to represent the structure of the manufactured part and models it as a function of process parameters. Hansen *et al.* (2002) applied $N$-way PLS to predict grain yield and protein content by repeating seasonal spectral measurements over the time range from early stem elongation to heading.

Another important task in process monitoring is fault diagnosis and detection. Early fault diagnosis can help to avoid serious equipment abnormalities and adjust related parameters so that the processing process runs normally. In contrast to the above-mentioned process data analysis, fault diagnosis and detection investigates variation in a process caused by special issues rather than common reasons. For example, $N$-way PLS was used in problematic wine fermentations (Emparán *et al.*, 2012). A dynamic $N$-way PLS was employed for online monitoring of the batch process (Chen and Yen, 2003). Idé (2019) proposed a probabilistic tensor regression model for change detection and analysis and applied it to a semiconductor manufacturing process for fault diagnosis.

Some example datasets employed are listed as:

- **Fat hardening** (Smilde and Kiers, 1999): The dataset is taken from the manufacturing batch process of catalytic hydrogenation of soybean oil. This dataset records the measured values of three process variables, flow, pressure and temperature on 20 batches at 101 time points. In addition, the response related to the fat melting curve measured by NMR after the manufacturing is used to evaluate the product quality.

- **SBR** (Nomikos and MacGregor, 1994): A simulated dataset of the production of SBR. Nine process parameters were measured over 50 batches at 200 time points and 5 quality parameters were measured on the final product.

- **Fluorescence** (Bro, 1999)[20]: A dataset for predicting the quality of refined sugar from fluorescence data. It records the emission spectra of 268 sugar samples at seven excitation wavelengths. To reduce the size of the dataset, the resolution of the emission spectrum is usually down-sampled and the quality-related parameters of each sample are recorded, including color and ash content.

- **Sensor data**[21]: A bread dataset to predict the salt content from the assessments from different individuals. It collects the assessments of 8 people in terms of 11 different attributes and also the salt content of each sample of bread.

Note that only part of the related works are listed here, and interested readers can refer to the review papers (Smilde *et al.*, 2005; Bro, 2006; Escandar *et al.*, 2007; Gómez and Callao, 2008; Rendall *et al.*, 2019) for more details on multiway analysis in chemistry.

## 7.8 Others

There are many application fields covered by tensor regression, including but not limited to some applications and datasets mentioned above. Here we give a brief description of some others.

**Online advertising** In Hao *et al.* (2021), the sparse tensor additive model is employed for click-through rate (CTR) prediction, which is a very important indicator to evaluate the performance of online advertising. CTR is defined as the ratio of the number of clicks to the number of views, which influences online revenue. The click-through rate prediction system can predict the CTR based on some features of the user, the page, and the ad itself before actually displaying it. In this way, ranking and placing the ads according to the predicted CTR can help improve the real CTR.

**Language modeling and Speech recognition** Language modeling is a fundamental model in Natural Language Processing (NLP), and it is commonly used in speech recognition and machine translation. The long-term dependency between natural languages is a key aspect of language modeling tasks. Thus, Soltani and Jiang (2016) proposed higher-order RNNs to explore the long-term dependency through tracking more preceding RNN states. The datasets used include the standard Penn Treebank (PTB) corpus which contains 1M words and the English text8 corpora which contains about 17M words.

[20] http://www.models.kvl.dk/Sugar_Process
[21] http://www.models.kvl.dk/Sensory_Bread

Similarly, speech is also a complex time signal with long-term dependency between frames. Usually, several words before and after will have an impact on the word we want to say. Considering the long-term correlation of speech signals, the natural idea is to use a neural network model with long-term modeling capabilities, like the high order RNN in Zhang and Woodland (2018). The British English multi-genre broadcast (MGB3) dataset from the MGB3 speech recognition challenge task[22] (Bell *et al.*, 2015) is used for algorithm performance evaluation.

**Physics** Zhe *et al.* (2019) employed the scalable high order Gaussian process regression (SHOGP) model to find the best topological structure needed to withstand specific external forces. One small benchmark dataset is the Cantilever which collects the material structures that have maximum stiffness when bearing specific forces from the right side. Each input variable represents the features of a specific force, and the corresponding output is a vector in the size of 100 describing the structure in a $10 \times 10$ domain.

The Navier-Stokes equation is the core of fluid flow modeling. Solving these equations under specific boundary conditions (such as inlets, outlets, and walls) can predict the fluid velocity and pressure in a given geometry. It is very difficult to solve the incompressible Navier-Stokes equations under large Reynolds numbers. Fine-grained grids are usually required to ensure the convergence of the numerical solver, which also means that the dimensionality of the output field is very high. The specific task considered in Zhe *et al.* (2019) is to predict a pressure field of the lid-driven cavity flow, whose output is organized as a $100 \times 100 \times 100$ tensor. But due to computational cost, usually only hundreds of training samples are collected, resulting in a typical "small $N$, Large $p$" problem. The experimental results have verified the linear scalability of the SHOGP model to large output dimensions.

[22] http://www.mgb-challebge.org

A tensor random forest model is proposed in Kaandorp and Dwight (2020) as a data-driven turbulence modeling tool for Reynolds Average Navier-Stokes (RANS) simulations. The tensor random forest model is trained based on several reference DNS/LES datasets and used to predict the Reynolds stress anisotropy tensor of new flows. It helps to build a better closure model for specific types of flows and improve the current RANS models.

Table 7.4 summarizes the datasets for the tensor regression described above for easy reference and divides them into three groups from the representation form of predictor and response, including vector-on-tensor regression, tensor-on-vector regression and tensor-on-tensor regression.

**Table 7.4:** A summary of applications for tensor regression models.

| Predictor | Response | Applications | Dataset |
|---|---|---|---|
| Tensor | Vector | Disease diagnosis | ADHD-200 (Zhou *et al.*, 2013; Li *et al.*, 2013a; Hou and Chaib-Draa, 2015; Ahmed *et al.*, 2019);<br>ABIDE (Li *et al.*, 2019b);<br>ADNI (Li *et al.*, 2016);<br>PPMI (He *et al.*, 2018);<br>UK Biobank (Kossaifi *et al.*, 2020) |
| | | Human age estimation | FG-NET (Guo *et al.*, 2012)<br>UK Biobank (Kossaifi *et al.*, 2020; Kolbeinsson *et al.*, 2021) |
| | | Human attributes prediction | LFW (Lock, 2018) |
| | | Head pose estimation | IDIAP dataset (Guo *et al.*, 2012);<br>Boston University dataset (Guo *et al.*, 2012);<br>Pointing'04 dataset (Guo *et al.*, 2012)<br>Biwi Kinect head pose dataset (Kaymak and Patras, 2014);<br>ICT-3DHP database (Kaymak and Patras, 2014); |
| | | Crowd counting | Pedestrian traffic dataset (Zhao *et al.*, 2014) |
| | | Neuro decoding | CMU2008 dataset (Song and Lu, 2017) |
| | | Process optimization | Fat hardening (Smilde and Kiers, 1999);<br>SBR (Nomikos and MacGregor, 1995; Boqué and Smilde, 1999);<br>Fluorescene (Bro, 1999);<br>Sensor data (Smilde *et al.*, 2005) |
| Vector | Tensor | Brain activation or connectivity | ADHD(Li and Zhang, 2017)<br>ABIDE (Sun and Li, 2017)<br>EEG data for an alcoholism study (Li and Zhang, 2017)<br>LA5c Study (Kia *et al.*, 2018)<br>Balloon Analog Risk-Taking Task (Spencer *et al.*, 2019) |
| Tensor | Tensor | Multitask learning | Restaurant & Consumer Dataset (Romera-Paredes *et al.*, 2013; Yu and Liu, 2016; Yu *et al.*, 2018; Suzuki *et al.*, 2016; Kanagawa *et al.*, 2016);<br>Shoulder Pain dataset (Romera-Paredes *et al.*, 2013);<br>School dataset (Yu *et al.*, 2018; Kanagawa *et al.*, 2016) |
| | | Spatial-temporal analysis | USHCN (Bahadori *et al.*, 2014; Yu *et al.*, 2018; Yang *et al.*, 2017);<br>CCDS (Bahadori *et al.*, 2014; Yu *et al.*, 2018);<br>Meteo-UK (Bahadori *et al.*, 2014);<br>Foursquare check-ins (Bahadori *et al.*, 2014; Yu *et al.*, 2018);<br>Yelp (Bahadori *et al.*, 2014; Yu *et al.*, 2018);<br>PeMS (Yang *et al.*, 2017);<br>KTH action dataset (Su *et al.*, 2020);<br>moving-MNIST-2-dataset (Su *et al.*, 2020);<br>something-something V2 dataset (Su *et al.*, 2020); |
| | | Human motion reconstruction | HumanEva-I dataset(Guo *et al.*, 2012);<br>UMPM (Liu *et al.*, 2020c); |
| | | Neuro decoding | Food-tracking task (Zhao *et al.*, 2013a; Zhao *et al.*, 2013b; Hou *et al.*, 2015; Zhou *et al.*, 2020b) |

# 8

## Open-source software frameworks

For tensor-based analysis, there are several available software libraries for tensor decompositions and tensor networks. In the following, we introduce some commonly used ones with a brief illustration.

- *tensorlab* (Vervliet *et al.*, 2016) (http://www.tensorlab.net/) A Matlab toolbox for tensor operations, CP and Tucker decompositions and optimization methods. It also supports the sparse and incomplete tensor and includes several practical examples for structured data fusion, independent component analysis.

- *tensor-toolbox* (Kolda and Bader, 2006) (http://www.tensortoolbox.com/) An open source which defines the classes and functions of dense, sparse, symmetric or structured tensors like CP and Tucker format.

- *ht-toolbox* (Kressner and Tobler, 2012) (https://www.epfl.ch/labs/anchp/index-html/software/htucker/) A MATLAB toolbox for the construction and manipulation of tensors in the Hierarchical Tucker (H-Tucker) format.

- *tt-toolbox* (Oseledets and Dolgov, 2012) in MATLAB (https://github.com/oseledets/TT-Toolbox) and in Python (http://github.

com/oseledets/ttpy). A MATLAB and Python implementation of basic operations with tensors in TT-format.

- *Tensorly* (Kossaifi *et al.*, 2019b) (https://github.com/tensorly/tensorly). A Python implementation of common tensor decompositions, including CPD, Tucker decomposition and TT decomposition, and corresponding low rank tensor regression models.

- *tntorch* (https://github.com/rballester/tntorch) A PyTorch powered modeling and learning library based on tensor networks, such as CP, Tucker, TT and Hybrids like CP-Tucker and TT-Tucker. It supports basic tensor operations, tensor decompositions, tensor reconstruction and sensitivity analysis.

- *Scikit- tensor* (https://github.com/tensorly/tensorly). A Python module for multilinear algebra, basic tensor operations and tensor decompositions like CP and Tucker.

- *hottbox* (https://github.com/hottbox/hottbox) A Python module for tensor decompositions, visualization and nonlinear classification methods. It constructs an ecosystem of tensor class, which represents different forms of tensor, such as folding, unfolding and different decomposition forms, using meta information in order to more conveniently visualize the current state of the data.

These implementations bring great convenience to tensor-based analysis, and also benefit tensor regression. There are also some tensor libraries developed on C++, such as *mptensor* for basic tensor operations (https://github.com/smorita/mptensor) and *itensor* for tensor network calculations (Fishman *et al.*, 2020). We will not go into detail here because they are complicated and currently mainly used for efficient implementation of tensor calculations.

In the following, we give a brief summary of the available implementations of some tensor regression algorithms in the order in which we introduce these algorithms in Chapter 4, Chapter 5 and Chapter 6.

### Simple linear tensor regression models

The *Greedy* package (http://roseyu.com/code.html) provides an efficient greedy solution for the simple linear tensor regression model (Bahadori *et al.*, 2014). Tensorly (https://github.com/tensorly/tensorly) provides an implementation of the CP- and Tucker-based vector-on-tensor regression model using ALS and gradient descent methods. The implementation of both ALS-based and Bayesian approaches for Tucker-based tensor-on-tensor regression model (Hoff, 2015) can be found in (http://www2.stat.duke.edu/~pdh10/Code/hoff_2014/).

### Generalized linear tensor regression models

The CP and Tucker extensions of the GLM, namely GLTR-CP (Zhou *et al.*, 2013) and GLTR-Tucker (Li *et al.*, 2013a), are implemented in the *TensorReg* software (https://hua-zhou.github.io/TensorReg/). In addition, the sparsity constrained versions can also be performed with use of the *SparseReg* software (https://github.com/Hua-Zhou/SparseReg).

### Penalized tensor regression methods

The *tensor learning for regression* software (http://www.eecs.qmul.ac.uk/~ioannisp/source.htm) provides the possibility of learning the low rank approximations of the regression coefficient tensor with the Frobenius norm or group sparsity constraints (Guo *et al.*, 2012). Additional implementations of penalized tensor regression models include tensor ridge regression models such as the *HOLRR* package (https://github.com/grwip/HOLRR) developed for tensor response regression tasks in (Rabusseau and Kadri, 2016), the *MultiwayRegression* R package (https://github.com/lockEF/MultiwayRegression) and the *TTR* package (https://github.com/liujiani0216/TTR) for predicting a multiway tensor from a multiway tensor developed in Lock (2018) and Liu *et al.* (2020c) respectively, sparse ones like the *Remurs* package (http://staffwww.dcs.shef.ac.uk/people/H.Lu/) with embedded feature selection, the *SURF* package (https://github.com/LifangHe/NeurIPS18_SURF) as a tensor extension of stagewide regres-sion developed in He *et al.* (2018), the *SparseTensorRegression* package (https://github.com/xmlyqing00/SparseTensorRegression) for multi-response regression problems in (Li *et al.*, 2016) and the *STORE*, *TensorEnvelope* packages (http://lexinli.biostat.berkeley.edu/software.html) for sparse tensor response regression models in Sun and Li (2017) and Li and Zhang (2017). Besides, the open source of other constraints can refer to the *tensor-regression-granger-causality* package (https://github.com/esinkarahan/tensor_regression_granger-causality) for brain connectivity Granger causality analysis (Karahan *et al.*, 2015) and *TensorRegression* (https://github.com/hyan46/TensorRegression) for structured point cloud data analysis (Yan *et al.*, 2019).

### Bayesian tensor learning

An implementation of tensor-based tensor learning can be found in the *Bayesian-Tensor-Regression* package (https://github.com/rajguhaniyogi/Bayesian-Tensor-Regression) in Guhaniyogi *et al.* (2017).

### Quantile tensor regression

The implementation of quantile tensor regression can be found in (https://github.com/WenqiLu/QuantileTensorReg), which includes an alternating update algorithm based on Tucker decomposition and an ADMM algorithm for sparse version.

### Projection-based tensor regression

The PLS-based tensor regression methods, such as $N$-way PLS (Bro, 1996) and HOPLS (Zhao *et al.*, 2013a), could be implemented with the *Nway toolbox* (https://ww2.mathworks.cn/matlabcentral/fileexchange/1088-the-n-way-toolbox?requestedDomain=zh) and the *HOPLS-tools* in MATLAB (https://github.com/sqsun/HOPLSC_TFBSs/tree/master/matlab%20code/HOPLS_Tools) and in Python (https://github.com/arthurdehgan/HOPLS) respectively. An extension based on tensor envelope can refer to the *TEPLS* package (Zhang and Li, 2017) (http://lexinli.biostat.berkeley.edu/software.html).

### Kernel methods

The kernelized version of HOLRR can refer to the HOLRR package. For tensor kernels, one can also refer to the kernelized support tensor train machines for classification in Chen *et al.* (2020) with available software in Matlab (https://github.com/git2cchen/KSTTM).

### Tensor Gaussian process regression

The implementation of the multilinear Gaussian process regression (Yu *et al.*, 2018) can be found in (https://github.com/yuqirose/MultilinearGP), which takes the high dimensional arrays as predictor and the vector as response, namely the vector-on-tensor regression model. For multidimensional output regression tasks, one can refer to the HOGP package (Zhe *et al.*, 2019) (https://www.cs.utah.edu/~zhe/) and the implementation of scalable multitask Gaussian process tensor regression (Kia *et al.*, 2018) based on the Bayesian mixed effect model in (https://github.com/smkia/MTNorm).

### Tensor additive models

The implementation of a high dimensional multivariate additive model using Fourier series (Wahls *et al.*, 2014) is available online at (http://bitbucket.com/wahls/mdfourier), which treats the coefficients as a tensor and borrows tensor train decomposition to reduce the number of parameters. Meanwhile, a tensor additive model with B-splines (Zhou *et al.*, 2020b) is implemented in (https://github.com/yazhou2019/BNTR), which employs the CP decomposition to approximate the coefficient tensor.

### Random forest based tensor regression

The implementation of tensor basis random forest regression for RANS turbulence modeling (Kaandorp and Dwight, 2020) can be found in (https://github.com/mikaelk/TBRF).

### Deep tensor regression

*TensorLy-Torch* (https://github.com/tensorly/torch), built on top of TensorLy and PyTorch, provides a platform to use tensor methods in deep learning, including tensor regression layers, tensor contraction layers, factorized convolutional layers and some hooks like tensor dropout and rank selection. It includes many methods proposed for network compression and some hooks for performance enhancement, which enables interested users to better understand and compare various deep tensor based methods.

The application of tensor factorization in network compression can also be found in:

- *TT-RNN* package (https://github.com/Tuyki/TT-RNN) which employs the tensor train decomposition for factorizing the input-to-hidden weight matrix in RNNs (Yang *et al.*, 2017).
- *caffe-cp-decomposition* package (https://github.com/ddfabbro/caffe-cp-decomposition) which uses the CP decomposition to compress the kernel of convolutional layer (Lebedev *et al.*, 2015).
- *TensorNet-TF* package (https://github.com/timgaripov/TensorNet-TF) which supports the tensor train fully connected (TT-FC) layer (Novikov *et al.*, 2015) and tensor train convolutional (TT-conv) layer (Garipov *et al.*, 2016).
- *pytorch-tensor-deccompositions* (https://github.com/jacobgil/pytorch-tensor-deccompositions) which implements the Tucker and CP based convolutional layers (Lebedev *et al.*, 2015; Kim *et al.*, 2015) based on *Tensorly*.
- *LowRankTRN* (https://github.com/xwcao/LowRankTRN) which implements the CP regression layer, Tucker regression layer and tensor train regression layer (Cao and Rabusseau, 2017).

For high-order extensions of RNNs, one can refer to the *tensor-train-RNN* package ( https://github.com/yuqirose/tensor-train-RNN) implemented in Tensorflow and the *conv-tt-lstm* package (https://github.com/NVlabs/conv-tt-lstm) implemented in Pytorch.

In addition, the application of online learning strategies can be found in the *ALTO* package (Yu *et al.*, 2015) (http://roseyu.com/code.html) and *SMART* package (Xu *et al.*, 2019) (https://github.com/Jianpeng-Xu/TKDE-SMART) for spatio-temporal analysis. Tensor based sketch methods can be found in *tensorsketch* package (Pagh, 2013) (https://github.com/RuiShu/tensorsketch) and *tucker-tensorsketcch* package (Malik and Becker, 2018) (https://github.com/OsmanMalik/tucker-tensorsketch). Gradient descent algorithm for efficient implementation of simple linear regression can be referred to the *Tensoly* package.

# 9

## Conclusions and discussions

We provide a detailed overview of tensor regression models, techniques, and applications in this paper. We demonstrate the motivation and necessity of tensor regression research by starting with classic regression models. Then, for existing tensor regression problems, recent advances and popular approaches are provided in detail, with comparison, summary, and discussion from several perspectives. Some strategies for efficient implementation are introduced to speed up the training process and address the large-scale datasets, while model evaluation metrics are summarized for assessing the performance of learnt models. Finally, we provide an overview of related applications of tensor regression in various fields with several examples illustrating the usefulness and effectiveness of some popular solutions.

Several concerns need to be addressed in future investigations. To begin with, determining the optimum approximation for a high dimensional complex regression system is always a research hotspot. The low rank approximation method has been shown to be useful in reducing model parameters while preserving considerable information in the regression coefficient array. However, how to more efficiently and compactly represent high-order data, as well as the performance evaluation and theoretical analysis of tensor decomposition algorithms, remain hot subjects in tensor-related research. For high dimensional data, there is currently no definitive answer as to which decomposition approach provides the optimal low rank approximation. Besides, developing efficient algorithms for constrained / regularized tensor regression with different structural features (sparsity, smoothness) along different modes, particularly for high dimensional regression problems with a limited sample size, is also an important topic for reducing model parameters, stabilizing estimates, and improving model interpretability.

Secondly, developing a robust tensor regression model to limit the detrimental impacts of outliers or sparse noise would be a significant step forward. Most recent studies on tensor regression use a Gaussian distribution to characterize measurement noise, however, if the data is skewed by outliers or sparse noise, the algorithm's performance suffers significantly. Reasonable system noise modeling, as well as the use of prior information to aid modeling, can increase the model's robustness and lessen the negative effects of outliers, sparse noise, structured noise, or mixed noise.

Thirdly, multi-view or multimodal information exploitation and fusion is also a major focus of current research. As the types of sensors increase, we can usually obtain data from different views of the same object. How to combine multi-view data to help better prediction is a problem that many researchers are concerned about, such as disease diagnosis based on multimodal imaging data (Acar *et al.*, 2017; Tang *et al.*, 2020). Meanwhile, tensor regression analysis with partly observed entries (Feng *et al.*, 2019; Zhou *et al.*, 2020a) is also an area that needs to be filled.

Finally, when it comes to large-scale data analysis, computational complexity is always a crucial consideration when evaluating established algorithms. Computing SVD for the unfolding matrices or contracting the remaining elements in the tensor regression model would be computationally very expensive, as we previously indicated. As a result, solutions for efficient implementation are considered, including sketching, online learning, and distributed algorithms for tensor operations. Stochastic gradient descent-based and deep-based tensor regression, in particular, could be interesting research subjects.